\documentclass[10pt,twocolumn,letterpaper]{article}

 \usepackage[pagenumbers]{cvpr} 

\usepackage{colortbl} 
\usepackage{enumitem}

\definecolor{yellow}{rgb}{1, 1, 0.7}
\definecolor{orange}{rgb}{1, 0.85, 0.7}
\definecolor{tablered}{rgb}{1, 0.7, 0.7}
\definecolor{red}{rgb}{1, 0, 0}

\definecolor{tablethree}{rgb}{0.7, 1, 1}
\definecolor{tabletwo}{rgb}{0.7, 0.85, 1}
\definecolor{tableone}{rgb}{0.7, 0.7, 1}

\newcommand{\sbest}{\cellcolor{orange}}
\newcommand{\tbest}{\cellcolor{yellow}}
\newcommand{\best}[1]{\cellcolor{tablered} \textbf{#1}}

\usepackage{array, makecell} %
\usepackage{soul}
\usepackage{listings}
\definecolor{codegray}{gray}{0.95}

\renewcommand{\paragraph}[1]{\vspace{.2em}\noindent\textbf{#1.}}

\definecolor{cvprblue}{rgb}{0.21,0.49,0.74}
\usepackage[pagebackref,breaklinks,colorlinks,allcolors=cvprblue]{hyperref}
\usepackage{mathtools}
\def\paperID{} 
\def\confName{CVPR}
\def\confYear{2026}

\title{Illustrator's Depth: Monocular Layer Index Prediction for Image Decomposition}

\author{
Nissim Maruani$^*$\\[-0.5mm]
\small{Inria, UCA}\\[-1mm]
\and
Peiying Zhang\\[-0.5mm]
\small{CityUHK}\\[-1mm]
\and
Siddhartha Chaudhuri\\[-0.5mm]
\small{Adobe Research}\\[-1mm]
\and
Matthew Fisher\\[-.5mm]
\small{Adobe Research}\\[-1mm]
\and
Nanxuan Zhao\\[-0.5mm]
\small{Adobe Research}\\[-1mm]
\and
Vladimir G. Kim\\[-0.5mm]
\small{Adobe Research}\\[-1mm]
\and
Pierre Alliez\\[-.5mm]
\small{Inria, UCA}\\[-1mm]
\and
Mathieu Desbrun\\[-.5mm]
\small{Inria/X, IP Paris}\\[-1mm]
\and
Wang Yifan\\[-.5mm]
\small{Adobe Research}\\[-1mm]
}

\begin{document}
\maketitle

\renewcommand{\thefootnote}{\fnsymbol{footnote}}
\addtocounter{footnote}{1}
\footnotetext{Work performed during an internship at Adobe Research}
\renewcommand{\thefootnote}{\arabic{footnote}}

\begin{abstract}

We introduce Illustrator’s Depth, a novel definition of depth that addresses a key challenge in digital content creation: decomposing flat images into editable, ordered layers. Inspired by an artist’s compositional process, illustrator’s depth infers a layer index for each pixel, forming an interpretable image decomposition through a discrete, globally consistent ordering of elements optimized for editability. 
We also propose and train a neural network using a curated dataset of layered vector graphics to predict layering directly from raster inputs. Our layer index inference unlocks a range of powerful downstream applications. In particular, it significantly outperforms state-of-the-art baselines for image vectorization while also enabling high-fidelity text-to-vector-graphics generation, automatic 3D relief generation from 2D images, and intuitive depth-aware editing. By reframing depth from a physical quantity to a creative abstraction, illustrator's depth prediction offers a new foundation for editable image decomposition.
\vspace*{-3mm}
\end{abstract}
    
\section{Introduction}
\label{sec:intro}



\begin{figure}[!t]\vspace*{3mm}
    \centering
    \includegraphics[width=0.99\linewidth]{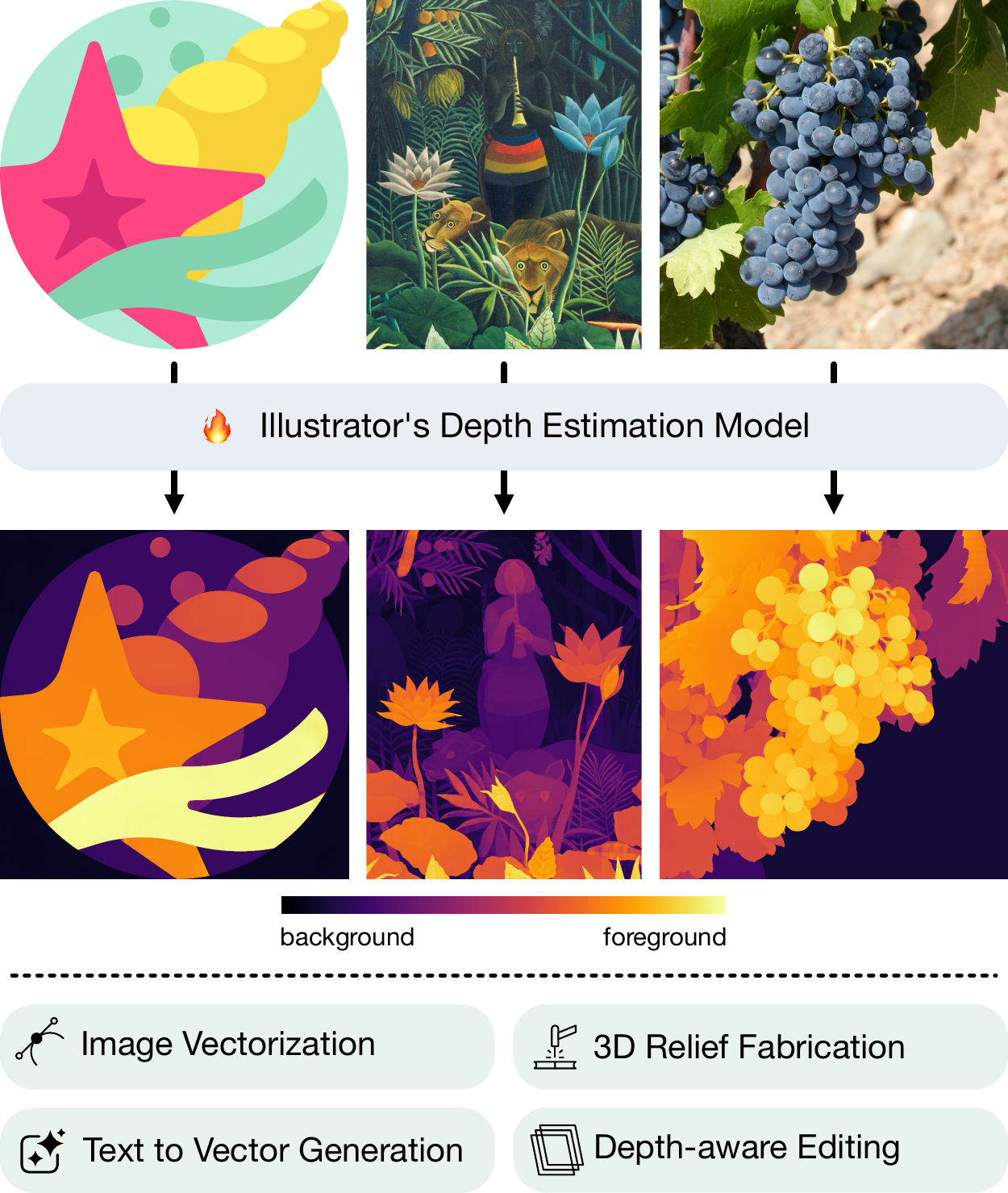}
    \vspace*{-2mm}
    \caption{\textbf{Overview.} Given an input image, our model predicts \emph{Illustrator’s Depth}, a learned ordering of compositional layers that reflects how an artist might have structured the image layout. This representation, applicable broadly to illustrations (left), paintings (middle), or even some realistic images (right), enables multiple downstream applications such as vectorization, intuitive editing, text-to-vector generation, and 3D relief fabrication.
    \vspace*{-4mm}
    }
    \label{fig:intro_overview}
\end{figure}

The organization of a digital artwork into a stack of layers is a fundamental concept in creative software. This paradigm, common to both vector-based and raster graphics tools, is central to the creative process as it allows for the independent manipulation and editing of individual compositional elements. This layering is also inherently related to the physical depth of objects within a scene, in that closer elements obscure those that are farther away.
\smallskip

\begin{figure}[b]\vspace*{-3mm}
    \centering
    \captionsetup[subfigure]{justification=centering, singlelinecheck=false}

    \begin{subfigure}[b]{0.325\linewidth}
        \centering
        \includegraphics[width=\linewidth]{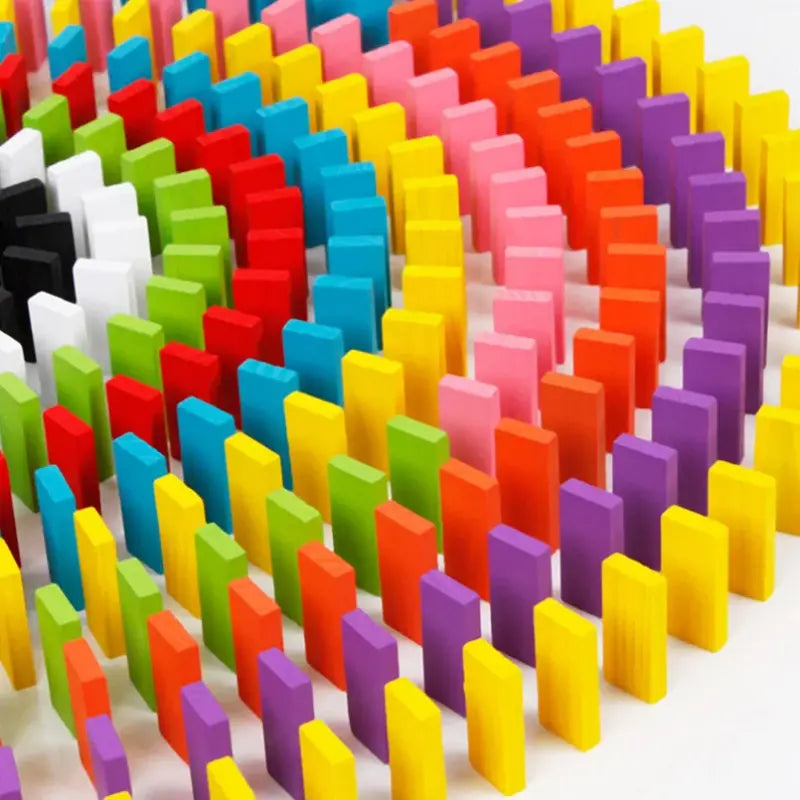}
    \end{subfigure}
    \hfill
    \begin{subfigure}[b]{0.325\linewidth}
        \centering
        \includegraphics[width=\linewidth]{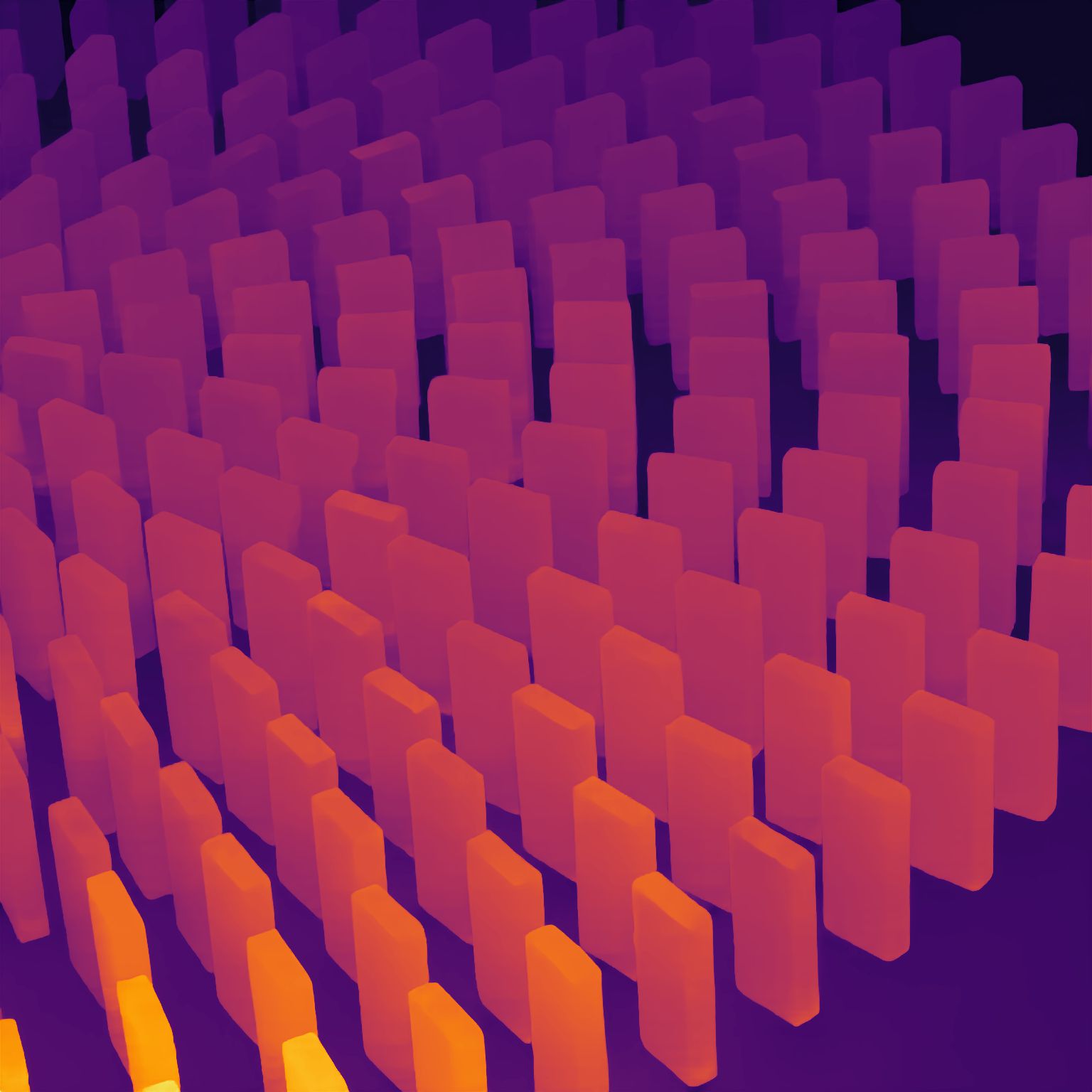}
    \end{subfigure}
    \hfill
    \begin{subfigure}[b]{0.325\linewidth}
        \centering
        \includegraphics[width=\linewidth]{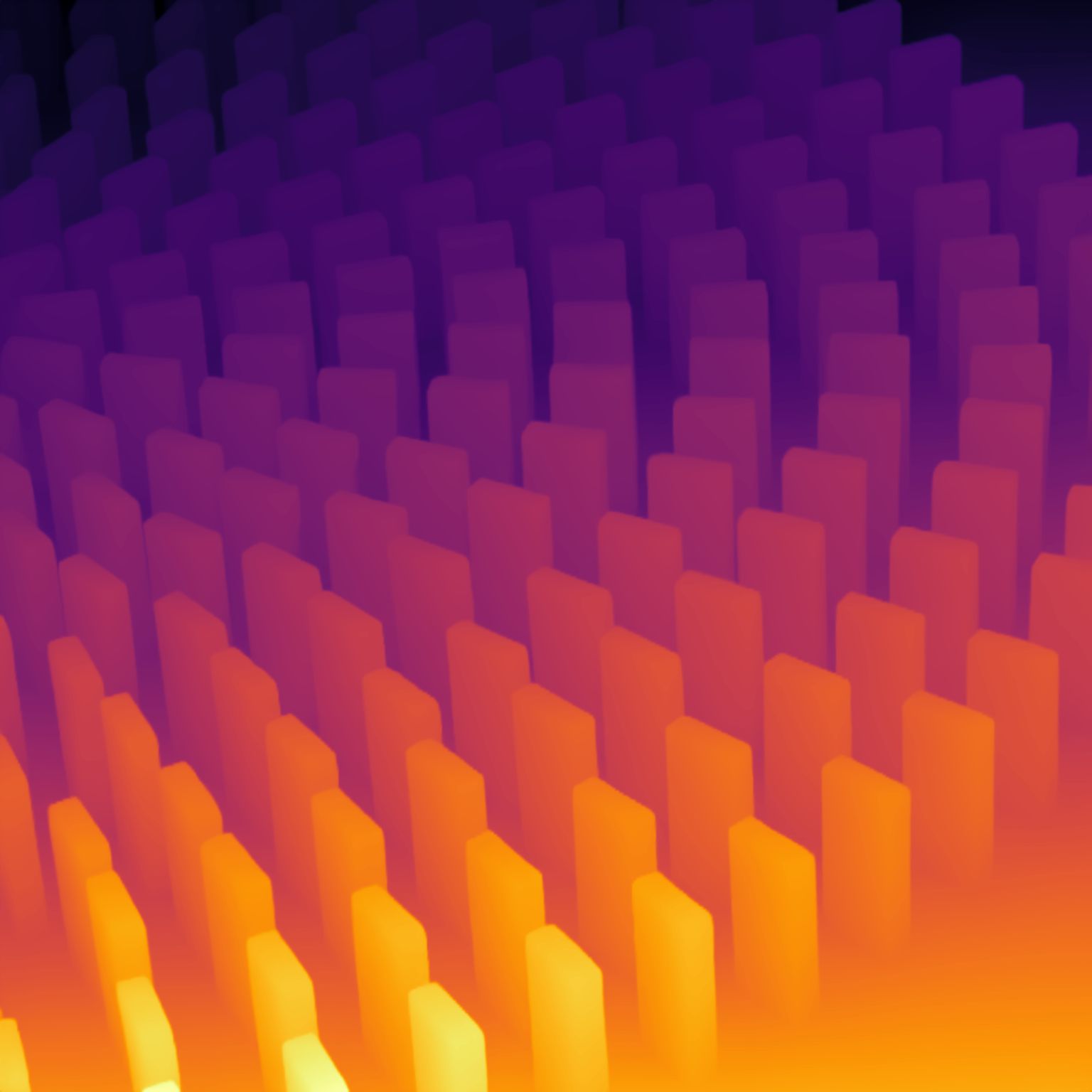}
    \end{subfigure}  
    
    \begin{subfigure}[b]{0.325\linewidth}
        \centering
        \includegraphics[width=\linewidth]{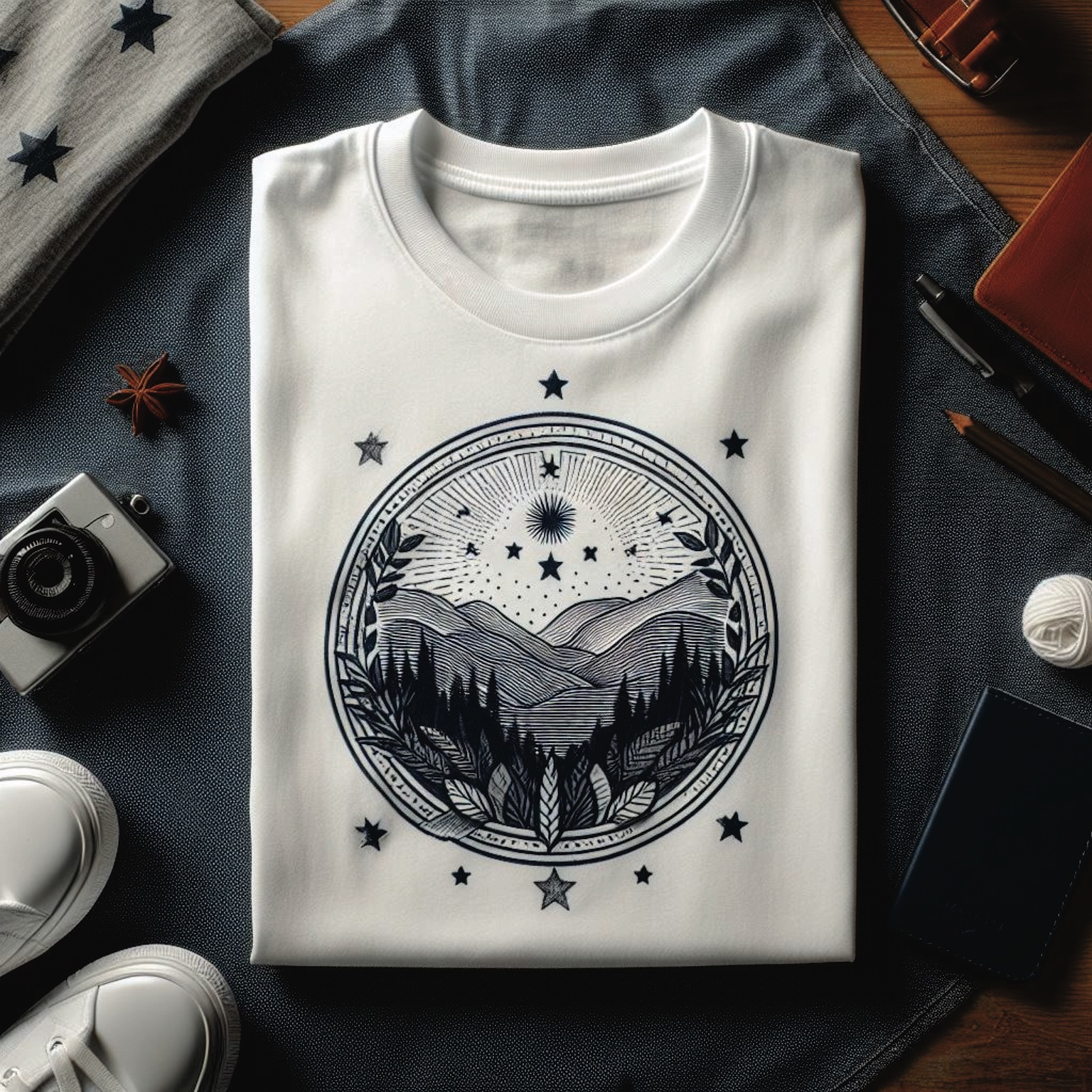}
    \end{subfigure}
    \hfill
    \begin{subfigure}[b]{0.325\linewidth}
        \centering
        \includegraphics[width=\linewidth]{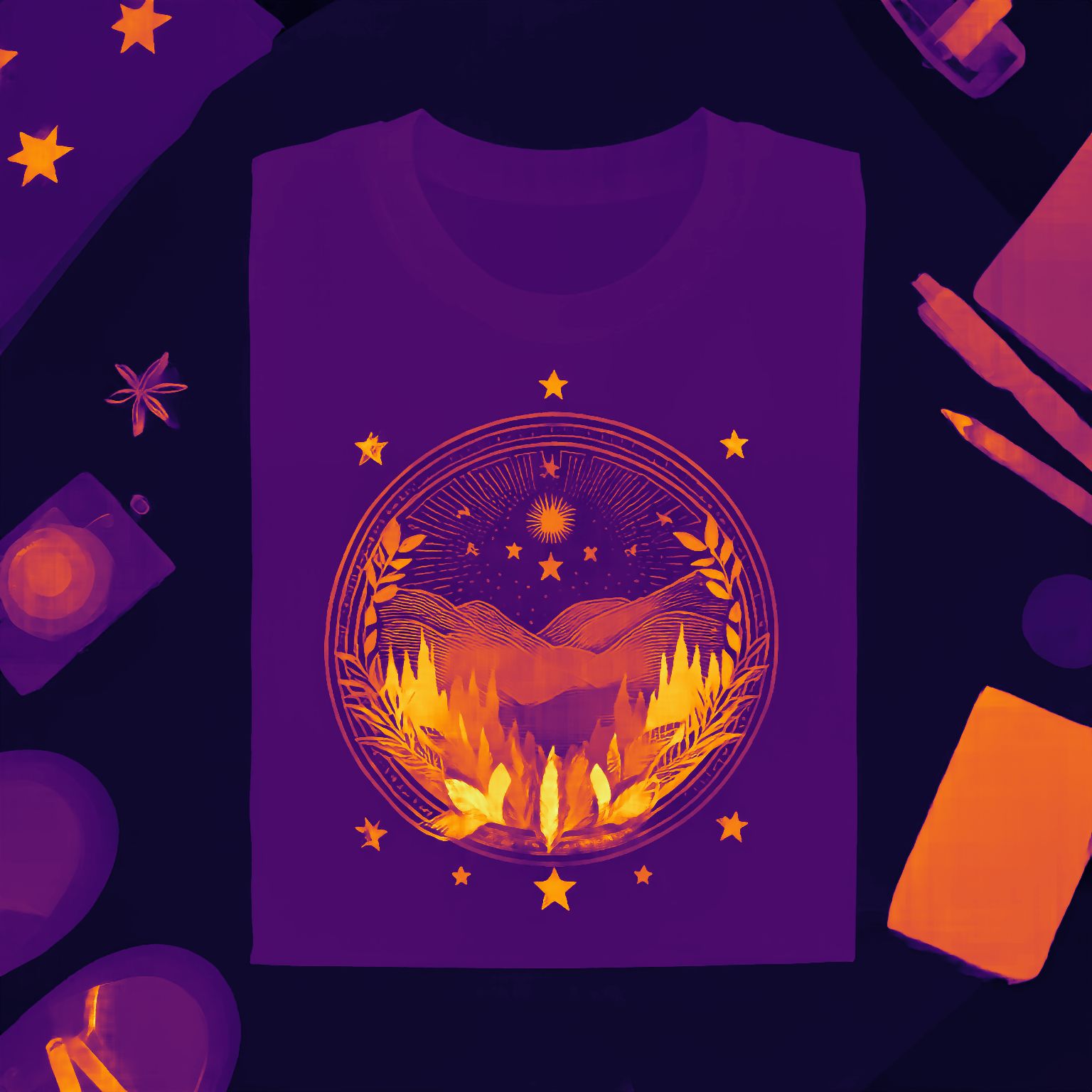}
    \end{subfigure}
    \hfill
    \begin{subfigure}[b]{0.325\linewidth}
        \centering
        \includegraphics[width=\linewidth]{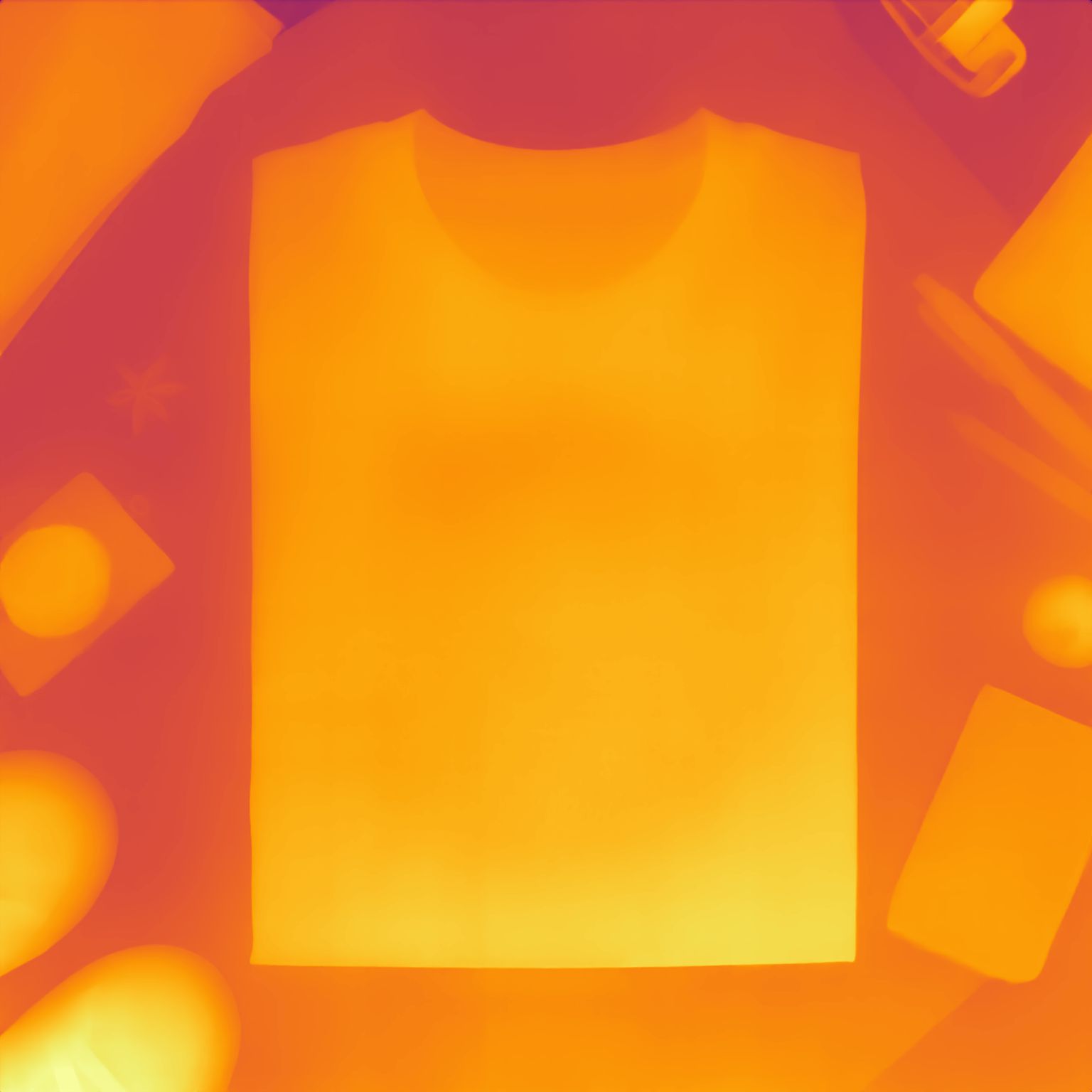}
    \end{subfigure}    
    \begin{subfigure}[b]{0.325\linewidth}
        \centering
        \includegraphics[width=\linewidth]{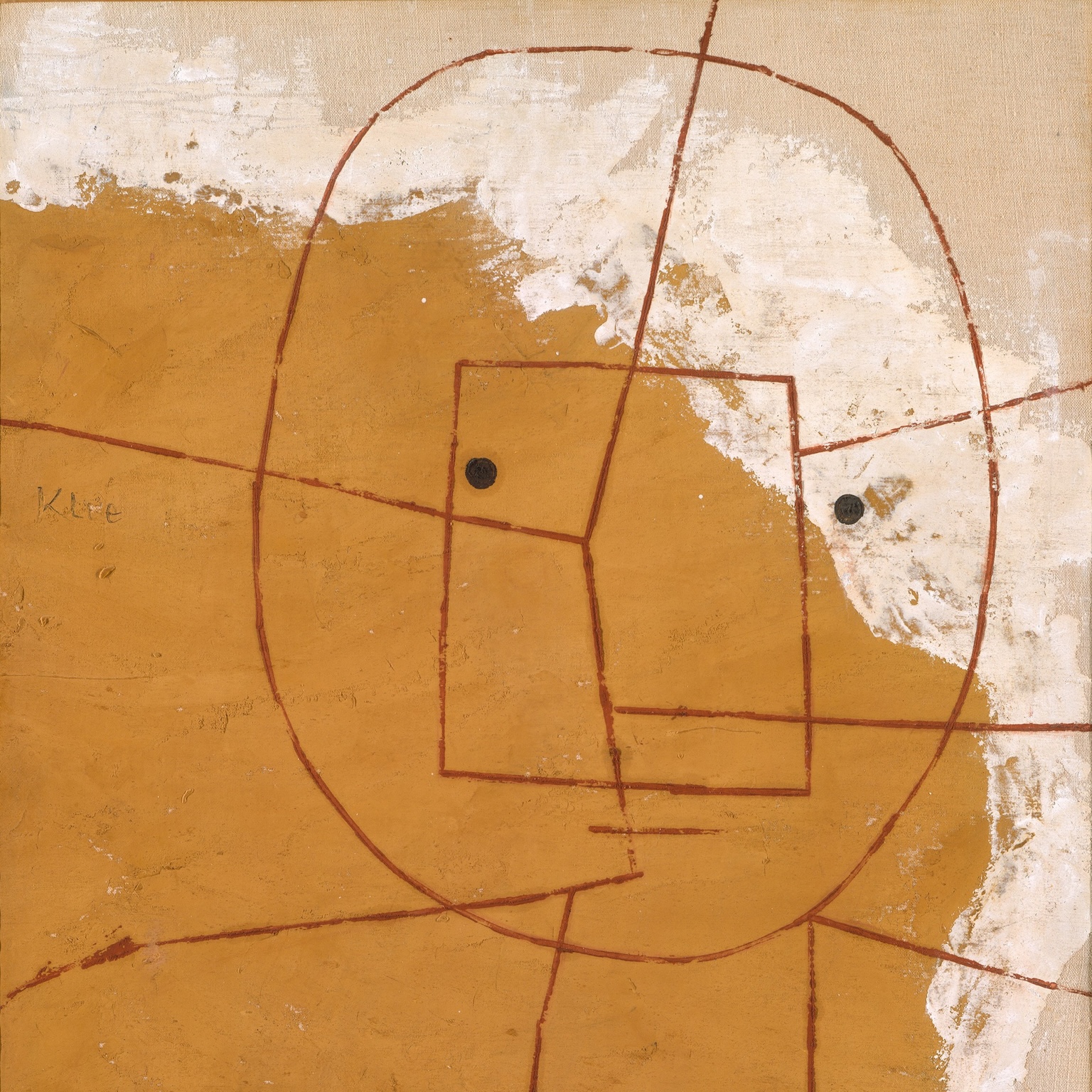}
        \caption{\centering Input Image\\ \phantom{empty}}
    \end{subfigure}
    \hfill
    \begin{subfigure}[b]{0.325\linewidth}
        \centering
        \includegraphics[width=\linewidth]{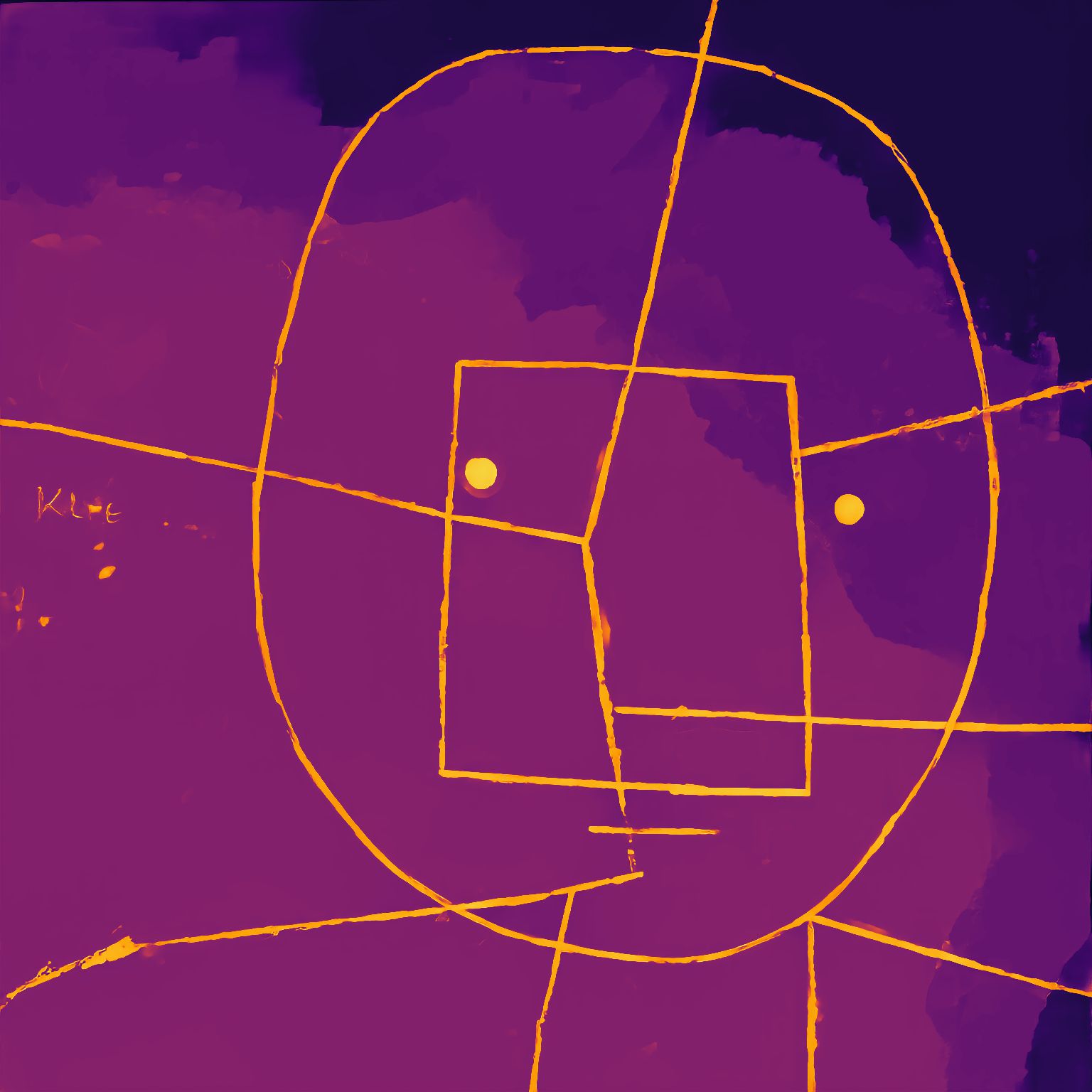}
        \caption{Illustrator's Depth \\ \scriptsize (Ours)}
    \end{subfigure}
    \hfill
    \begin{subfigure}[b]{0.325\linewidth}
        \centering
        \includegraphics[width=\linewidth]{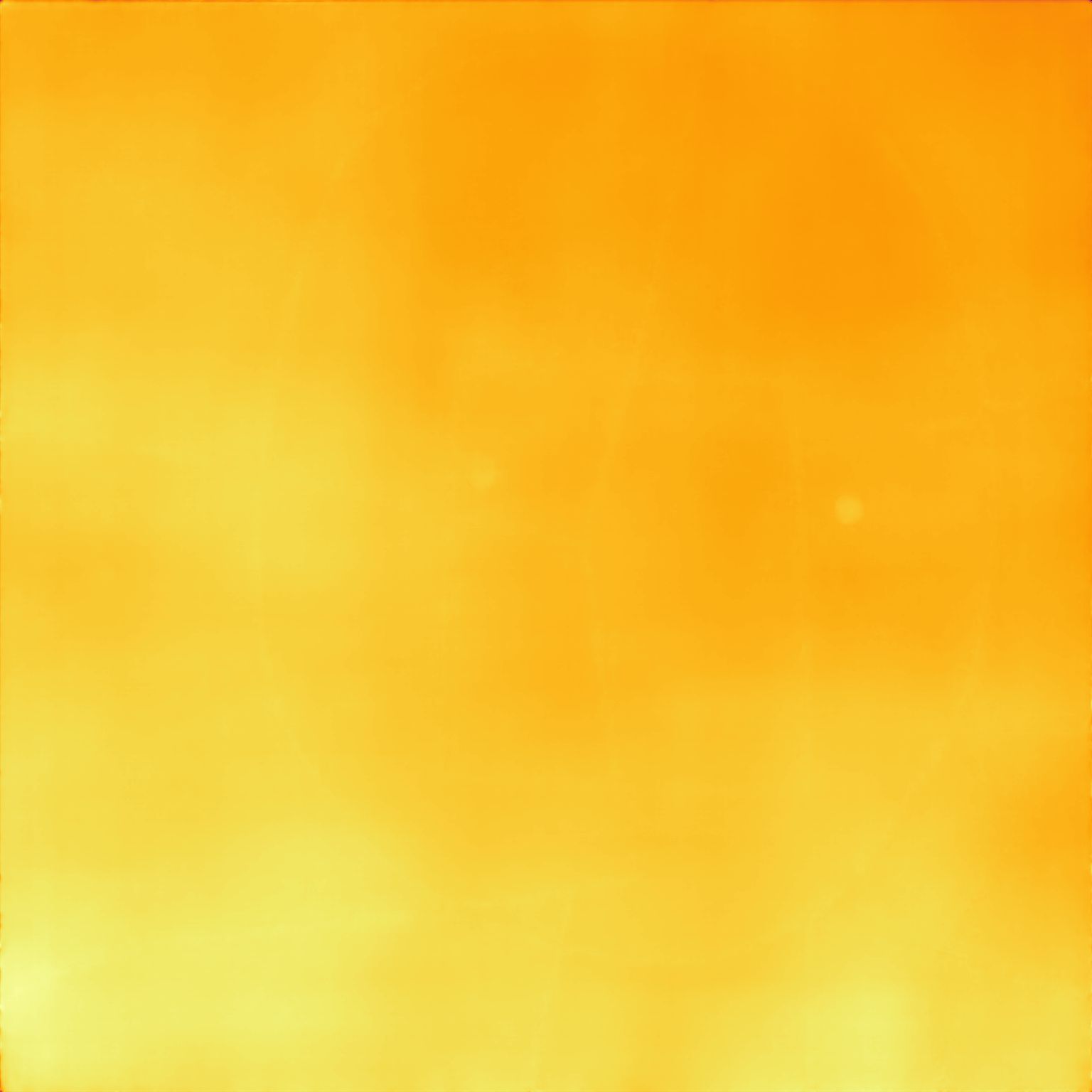}
        \caption{\centering Physical Depth\\ \scriptsize (Depth Pro~\cite{bochkovskii_depth_2025})}
    \end{subfigure}
    \vspace*{-7mm}
    \caption{\textbf{Physical~\emph{vs.}~Illustrator's Depth.}~Unlike monocular depth estimation, illustrator’s depth (middle, in false colors) produces piecewise-flat regions corresponding to layers and preserves compositional ordering even for printed or flat elements (e.g., shadows, drawings, or textures) that lack real-world depth (right).
    }
    \label{fig:depth_overview}
\end{figure}

While recent neural architectures can efficiently and accurately predict monocular depth from images~\cite{yang_depth_2024,bochkovskii_depth_2025} or compute panoptic segmentations~\cite{kirillov_panoptic_2019,ravi_sam_2025}, they are unable to decompose input illustrations or images into useful, ordered layers for three main reasons.
First, illustrative layers differ fundamentally from physical depths: important visual elements such as shadows may be placed \emph{above} the objects on which they are cast, and non-orthogonal flat surfaces with overlapping physical \emph{depth gradients} may nevertheless be mapped to discrete, sortable layers (see dominoes in \cref{fig:depth_overview}). 
Second, because illustrations typically appear on flat media (i.e., book pages, posters, or paintings), monocular depth estimation models are explicitly trained to \emph{ignore} them (see t-shirt in \cref{fig:depth_overview}).
Third, illustrative layering is conceptually independent from standard panoptic segmentation: contemporary segmentation models~\cite{ravi_sam_2025} generate predictions without encoding any relative \emph{ordering}, addressing fundamentally distinct aspects of scene understanding.
Layer depths derive, instead, from a subtle mix of segmentation and depth ordering to facilitate both design and editing. 

Although rarely acknowledged as such, layer inference is a core challenge in vectorization that impacts numerous downstream applications by offering an intuitive layer decomposition enhancing editing capabilities. Existing state-of-the-art methods, however, remain limited in scope, handling only simple inputs~\cite{wu_layerpeeler_2025, song_layertracer_2025} or relying on brittle heuristics~\cite{pun_vtracer_2025, zhao_less_2025, law_image_2025, ma_towards_2022, zhou_segmentation-guided_2025, hirschorn_optimize_2024} which do not consistently yield useful results. In the raster domain, several approaches have explored transparent layer extraction or generation~\cite{zhang_transparent_2024, leonardis_objectdrop_2024, pu_art_2025, lee_generative_2025, yang_generative_2025, tang_instance_2025}, yet these operate exclusively at the object level. 
To the best of our knowledge, no existing technique achieves fine-grained image layer decomposition.\smallskip

We introduce Illustrator’s Depth, a novel concept 
for representing the structural layering of vector graphics
Specifically, we define the illustrator’s depth of an image as the inverse mapping from each pixel to its corresponding layer index in its digital mockup, effectively capturing the spatial and compositional ordering of the artwork. 
We infer illustrators’ depth from arbitrary images automatically by leveraging a Depth Pro based neural network~\cite{bochkovskii_depth_2025} trained on a large, curated SVG dataset.
Our model operates in a feed-forward manner to predict pixel-level layer indices, enabling a wide range of applications such as 
image editing and depth-aware vector graphics manipulation.
More specifically, we present a number of contributions: 
\begin{itemize}[labelsep=0.4ex,leftmargin=2ex]
    \item We introduce the notion of \emph{Illustrator's Depth} and train a network to predict it, enabling fast layer decomposition;
    \item We show that incorporating our model into standard vectorization pipelines yields consistently layered SVGs with state-of-the-art visual fidelity;
    \item We propose a novel method for evaluating layer quality in vector graphics by rasterizing the predicted illustrator's depth and assessing its consistency with the ground truth;
    \item We demonstrate that coupling our pipeline with Text2Img models substantially enhances the generation of high-quality, editable vector illustrations from text;
    \item Finally, we showcase other applications of illustrator's depth in layer-based segmentation, depth-aware object insertion, tactile graphics creation, and artwork analysis. 
\end{itemize}

\section{Related Work}
\label{sec:related}


\paragraph{Monocular depth estimation (MDE)}~Classical learning approaches for depth estimation from images~\cite{saxena_make3d_2009,zhang_monocular_2015,hoiem_recovering_2007,rezaeirowshan_monocular_2016,eigen_depth_2014,fu_deep_2018} have evolved into strong backbones trained on diverse data~\cite{ranftl_towards_2020,farooq_bhat_adabins_2021,yang_depth_2024,bochkovskii_depth_2025}.
They yield finely detailed and \emph{continuous} (relative or metric) depth maps that serve as robust physical priors. Yet, they remain blind to content without true volume, like printed posters or patterns on clothing.
In contrast, our objective is to produce a new kind of depth \emph{prioritizing user editability over metric prediction}.

\paragraph{Layered depth for view synthesis}~Layered Depth Images store multiple depth samples per ray to model occlusions~\cite{shade_ldi_1998,dhamo2019peeking}, while Multiplane Images approximate scenes by many fronto-parallel planes for novel-view rendering~\cite{zhou_stereomag_2018,mildenhall_llff_2019}. These abstractions excel at detecting disocclusions and synthesizing new views, but they produce \emph{multi-sample} or \emph{multi-plane depths}, not a single discrete index per pixel that a designer can restack. Furthermore, they focus on physical depth like MDE, unlike our illustrator's depth which focuses on layer index prediction. 

\paragraph{Amodal / instance / panoptic segmentation}~Moving from geometry to semantics, segmentation families group regions by categories, but do not encode geometric ordering. Standard instance and panoptic methods provide high-quality visible masks~\cite{he_maskrcnn_2017,kirillov_panoptic_2019,kirillov_panoptic_fpn_2019,cheng_panopticdeeplab_2020,cheng_mask2former_2022,kirillov_sam_2023,ravi2025sam} without global per-pixel depth ordering. Amodal instance and amodal-panoptic formulations extend masks to occluded regions (for countable “thing” categories, typically), while “stuff” categories remain modal; representative datasets and models include~\cite{zhu2017semantic,xiao2021amodal,qi_kins_2019,mohan2022amodal}. Occlusion-aware and amodal transformers refine completion and boundary reasoning~\cite{lee2022instance,tran_aisformer_2022,ke_bilayer_2021,dhamo2019peeking}, yet supervision and metrics remain instance-centric or pairwise. None imposes a single, transitive ordering across all pixels, which is the target of our globally-consistent ordinal layer map.

\paragraph{Generative decompositions for editing}
Inspired by traditional approaches~\cite{richardt2014vectorising, tan2016decomposing, favreau2017photo2clipart}, editing-focused decompositions produce per-subject RGBA layers to facilitate local edits. Examples include real-time human matting~\cite{lin_real-time_2021}, generative pipelines that output editable layers for subjects and effects~\cite{lee_generative_2025,yang_generative_2025}, and atlas-based video methods that unwrap scenes into a few textures with an alpha channel for temporal consistency~\cite{lopes_learned_2019,law_image_2025}. These layers are effective for targeted edits but are \emph{independent} and not constrained to a global, per-pixel depth order. Instead, we seek a single ``illustrator's depth'' map that provides an coherent ordering of \emph{all} pixels in order to facilitate further editing.

\paragraph{Layering in vectorization}
An obvious application of our layer index estimation is vectorization: given our per-pixel ordinal map, standard raster-to-vector pipelines can group paths by layer and export edit-ready stacks. Existing systems based on heuristics or optimization~\cite{ma_towards_2022, hirschorn_optimize_2024, pun_vtracer_2025, law_image_2025, zhou_segmentation-guided_2025} often fail to infer a clean, useful layering. Learning-based approaches~\cite{lopes_learned_2019, reddy_im2vec_2021, rodriguez_starvector_2025, rodriguez_rendering-aware_2025, yang_omnisvg_2025} can, in principle, learn layer order from examples, but their training often compounds all the steps of the vectorization process (including Bézier control points), resulting in frequent reconstruction failures on complex inputs. Very recent works explore explicit layer predictions for better editing~\cite{wu_layerpeeler_2025,song_layertracer_2025}, but remain limited in the amount of paths and details they generate. Instead, our layer index prediction provides a supervised signal for \emph{ordering itself}, allowing traditional vectorizers to assemble SVGs in a manner most useful for further editing.

\section{Method}
\label{sec:method}


We now introduce our notion of illustrator's depth in Sec.~\ref{sec:IDepth}, before describing our dataset curation in~\cref{sec:method_depth_data}, and finally presenting our neural network implementation and training in~\cref{sec:NN+T}.
Evaluation tests and ablation studies will be presented and discussed at length in Sec.~\ref{sec:experiments}.

\begin{figure}[!t]
    \centering
    \includegraphics[width=0.99\linewidth]{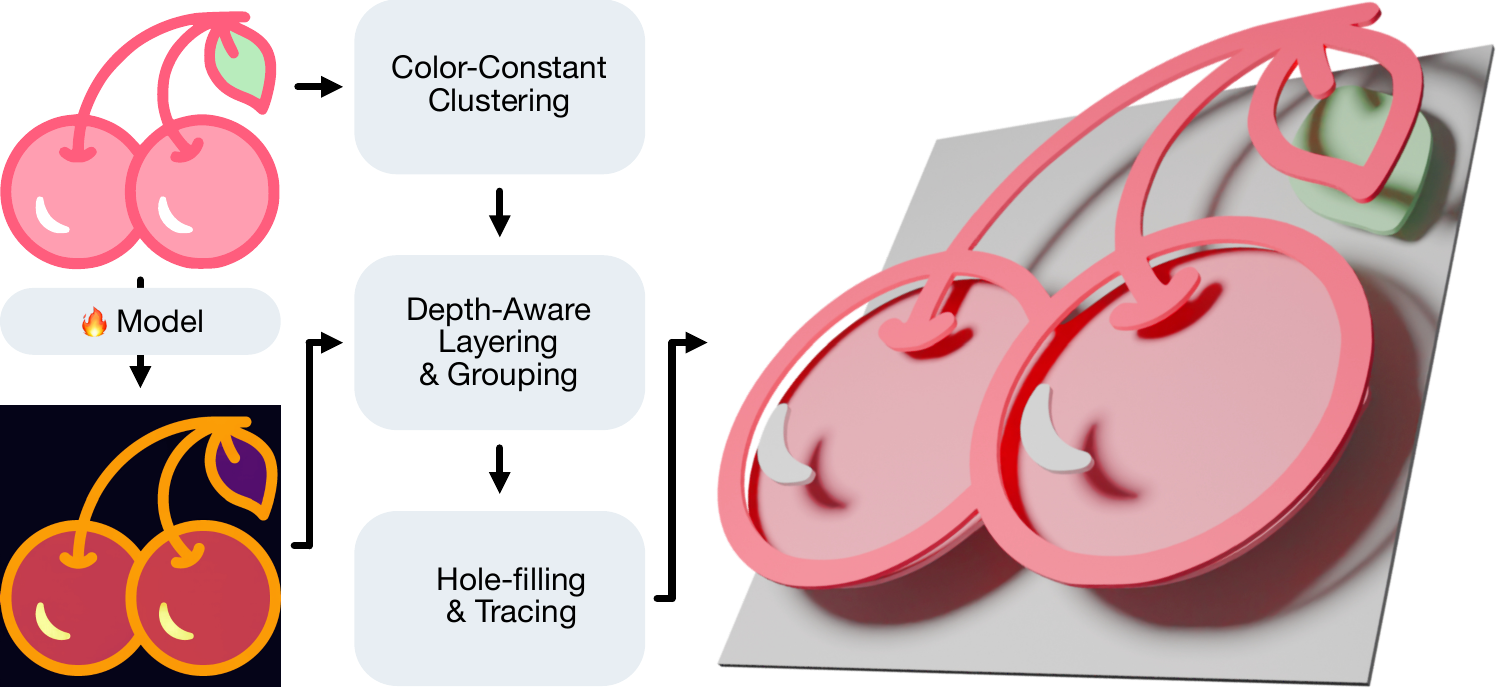}
    \vspace*{-1mm}
     \caption{\textbf{Depth-aware Image Vectorization.} Our predicted illustrator’s depth map (bottom left) can be integrated in traditional vectorization pipelines to produce well-layered SVG images (right, in 3D for clarity). On this example, our model allows the grouping of two disconnected white clusters to form a single background layer, while accurately separating the white highlights.
    \vspace*{-3mm}
    }
    \label{fig:layer_decomp}
\end{figure}

\subsection{Illustrator's Depth}
\label{sec:IDepth}
From an input \emph{illustration} $I$, represented as a raster $H \!\times\! W$ RGB image, we refer to its \emph{illustrator’s depth} as the mapping from each image pixel of the input to a layer index $i \!\in\! \{1 ... N\}$.
Conceptually, this map represents how an artist might have structured the image as a composition of $N$ separate layers (see \cref{fig:layer_decomp}), each corresponding to a different element or object drawn at a particular depth. Thus, 
illustrator’s depth provides a per-pixel layer assignment that captures an interpretable notion of structural depth implicit in the artist’s compositional workflow, which can then be directly leveraged for editing purposes.
This paper proposes predicting this mapping from an image using a neural network and a curated training set of layered compositions, yielding an illustrator's depth image $D_\theta(I) \!\in\! \mathbb{R}^{H \!\times\! W}$ where depth is treated as a continuous value rather than a discrete layer index as it still captures relative ordering while allowing straightforward binning into discrete values if necessary.


\subsection{Curating a Training Dataset}
\label{sec:method_depth_data}
Training our network to predict Illustrator's Depth requires a large-scale dataset of images paired with their ground-truth layer structure. Scalable Vector Graphics (SVG) files are an ideal source for this data, as they are inherently composed of layered vector paths that define the stacking order of a composition. We leverage this property by developing a three-stage data preparation pipeline: first, we source a suitable dataset of layered SVGs; second, we curate it to reduce ambiguity; and finally, we rasterize the vector files into corresponding image and depth map pairs for training.

\paragraph{Data sourcing}~While SVGs provide a structural foundation, the quality of their layering is crucial. Many SVG datasets, while visually correct when rendered, contain disorganized or programmatically generated layers that do not reflect an artist's intent. Yet, effective learning depends on a dataset with intuitively and consistently structured compositions. After reviewing existing options, we selected the MMSVG-Illustration dataset~\cite{yang_omnisvg_2025}, which features SVGs where elements are layered in a consistent and meaningful way, with layers systematically organized from the lowest index for the background to the highest index for the foreground, and outline strokes always placed above their corresponding color fills for instance. 

\paragraph{Data curation}~Even a high-quality dataset like MMSVG contains inherent ambiguities that can hinder learning. Artistic layering is often subjective; for instance, multiple distinct objects might logically share the same depth level, and different artists may have different layering habits. This variability can create a noisy training signal. To normalize these variations and create a more consistent ground truth, we perform two curation steps. First, we merge consecutive layers that share the same RGB color to simplify the structure. Second, we exclude ambiguous cases where non-consecutive layers of the same color overlap in the final rendered image, as this significantly improves training stability.

\begin{figure}[t] 
    \centering
    \captionsetup[subfigure]{justification=centering, singlelinecheck=false}
    \begin{subfigure}[b]{0.19\linewidth}
        \centering
        \includegraphics[width=\linewidth]{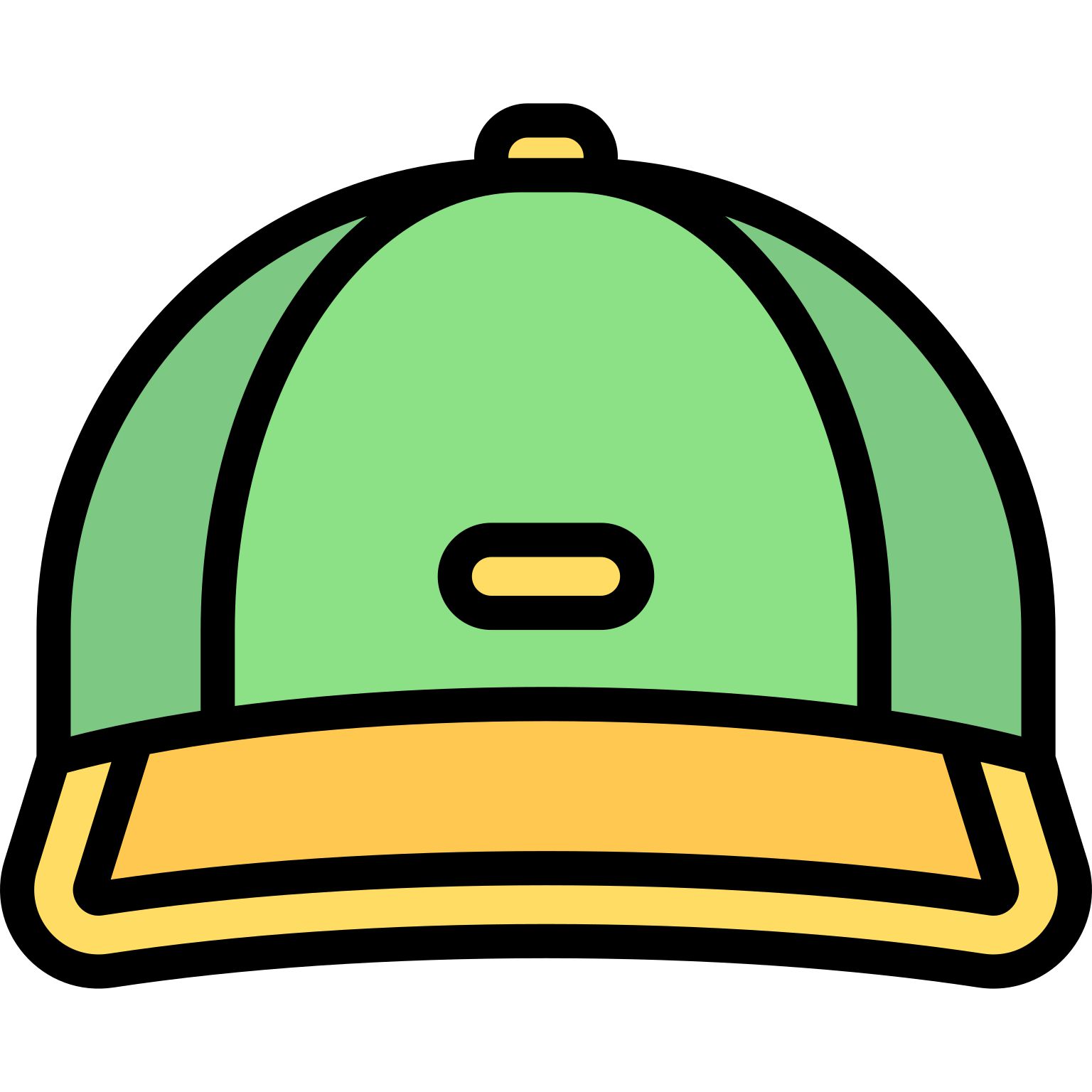}
    \end{subfigure}
    \hfill
    \begin{subfigure}[b]{0.19\linewidth}
        \centering
        \includegraphics[width=\linewidth]{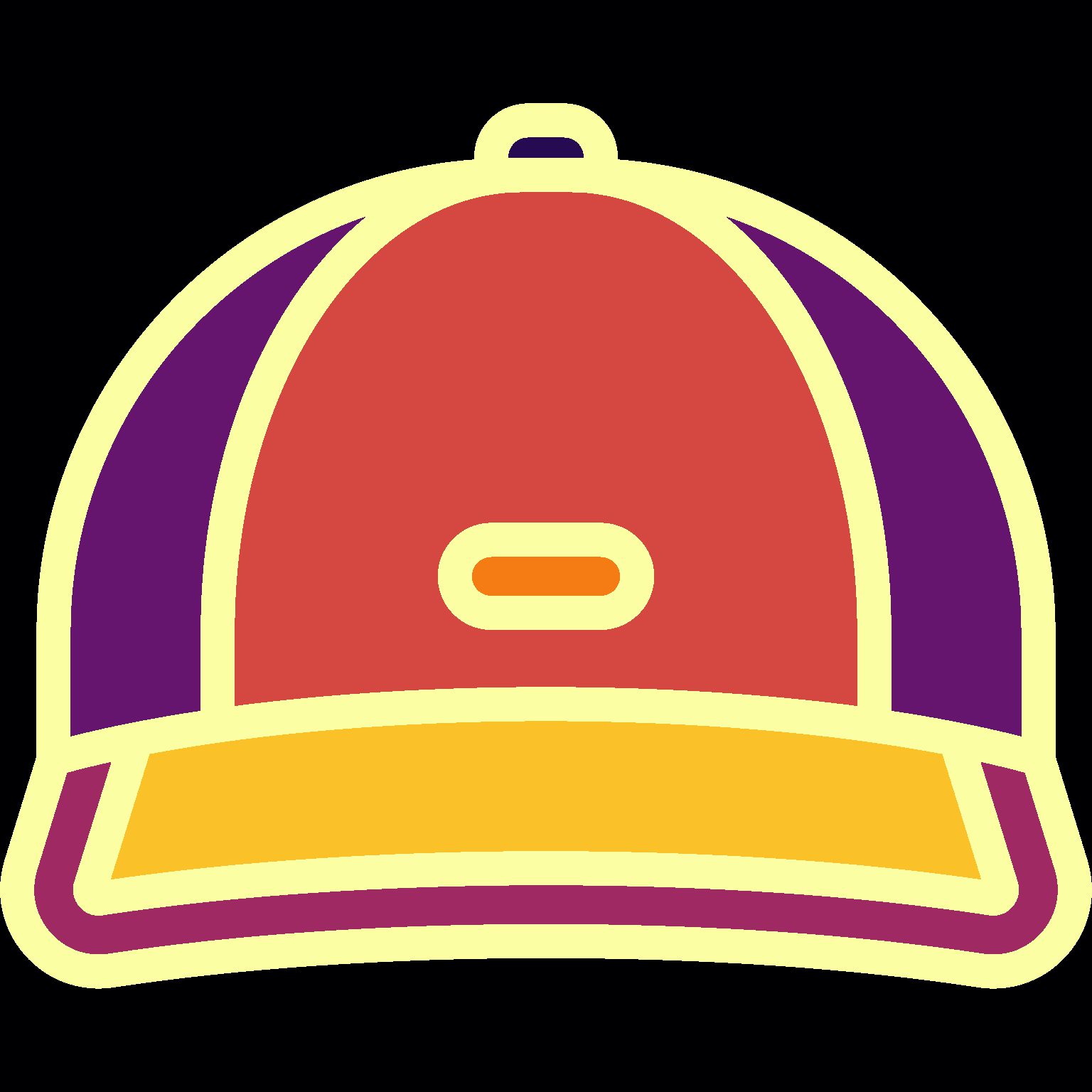}
    \end{subfigure}
    \hfill
    \begin{subfigure}[b]{0.19\linewidth}
        \centering
        \includegraphics[width=\linewidth]{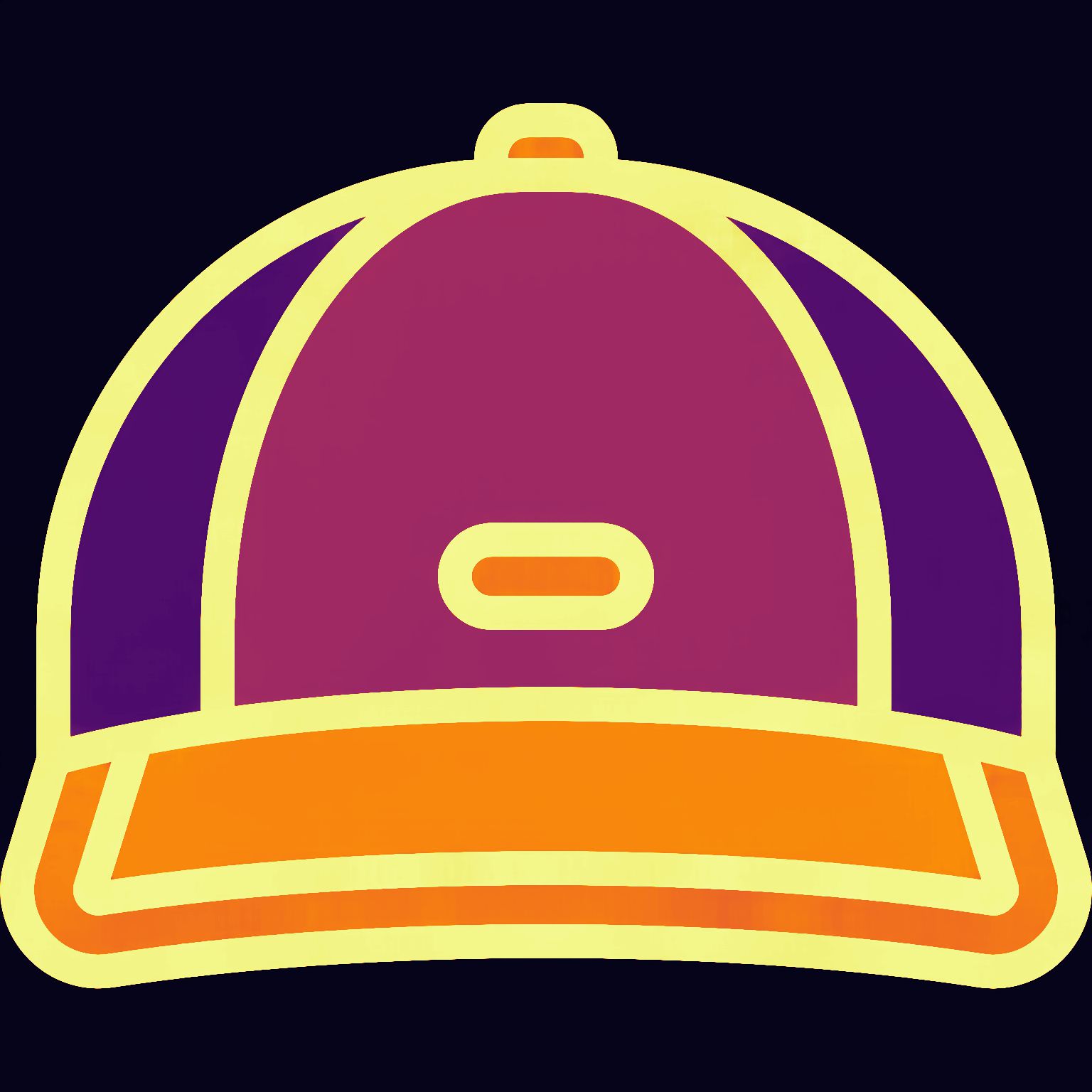}
    \end{subfigure}
    \hfill
    \begin{subfigure}[b]{0.19\linewidth}
        \centering
        \includegraphics[width=\linewidth]{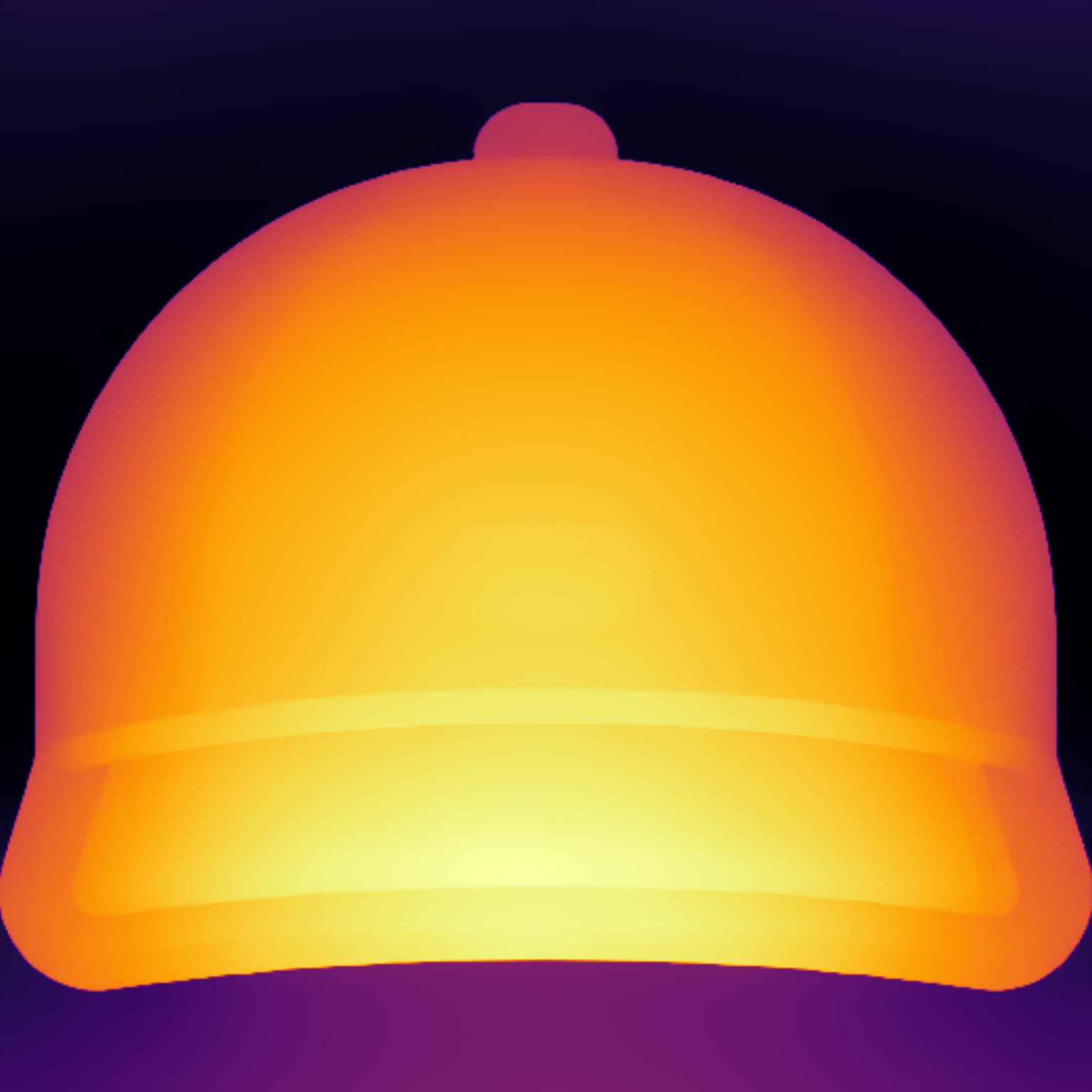}
    \end{subfigure}
    \hfill
    \begin{subfigure}[b]{0.19\linewidth}
        \centering
        \includegraphics[width=\linewidth]{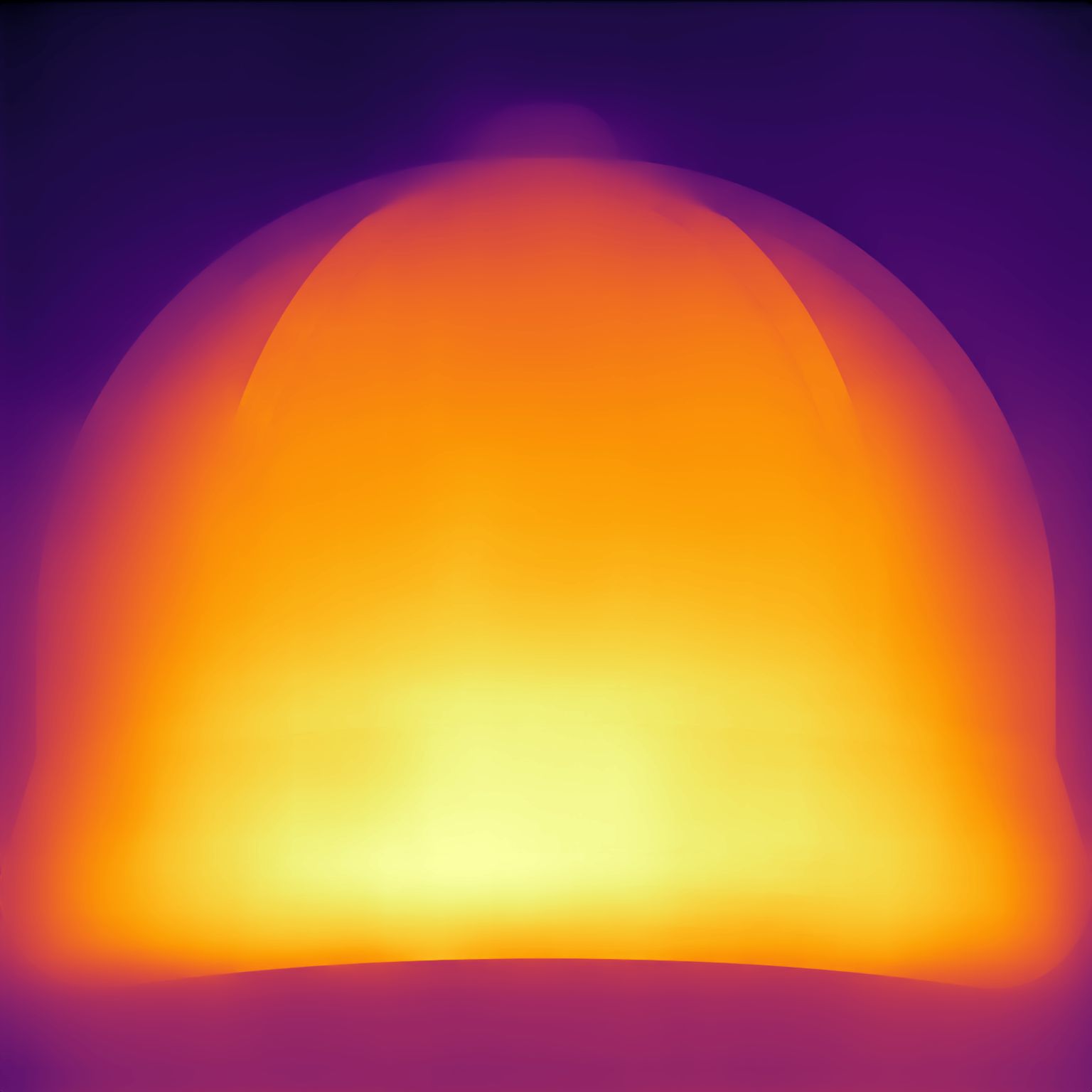}
    \end{subfigure}    

    \centering
    \begin{subfigure}[b]{0.19\linewidth}
        \centering
        \includegraphics[width=\linewidth]{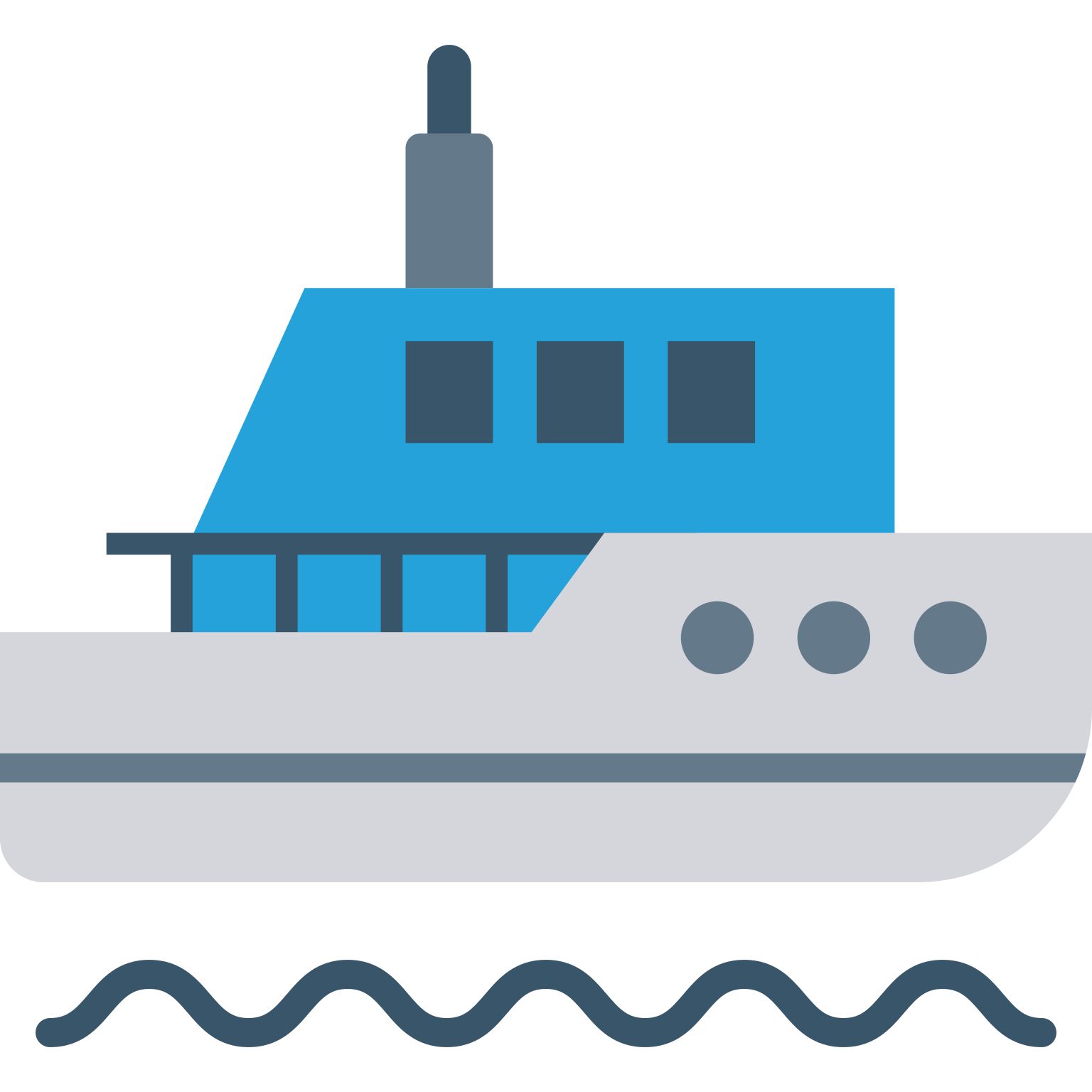}
        \caption{Input}
    \end{subfigure}
    \hfill
    \begin{subfigure}[b]{0.19\linewidth}
        \centering
        \includegraphics[width=\linewidth]{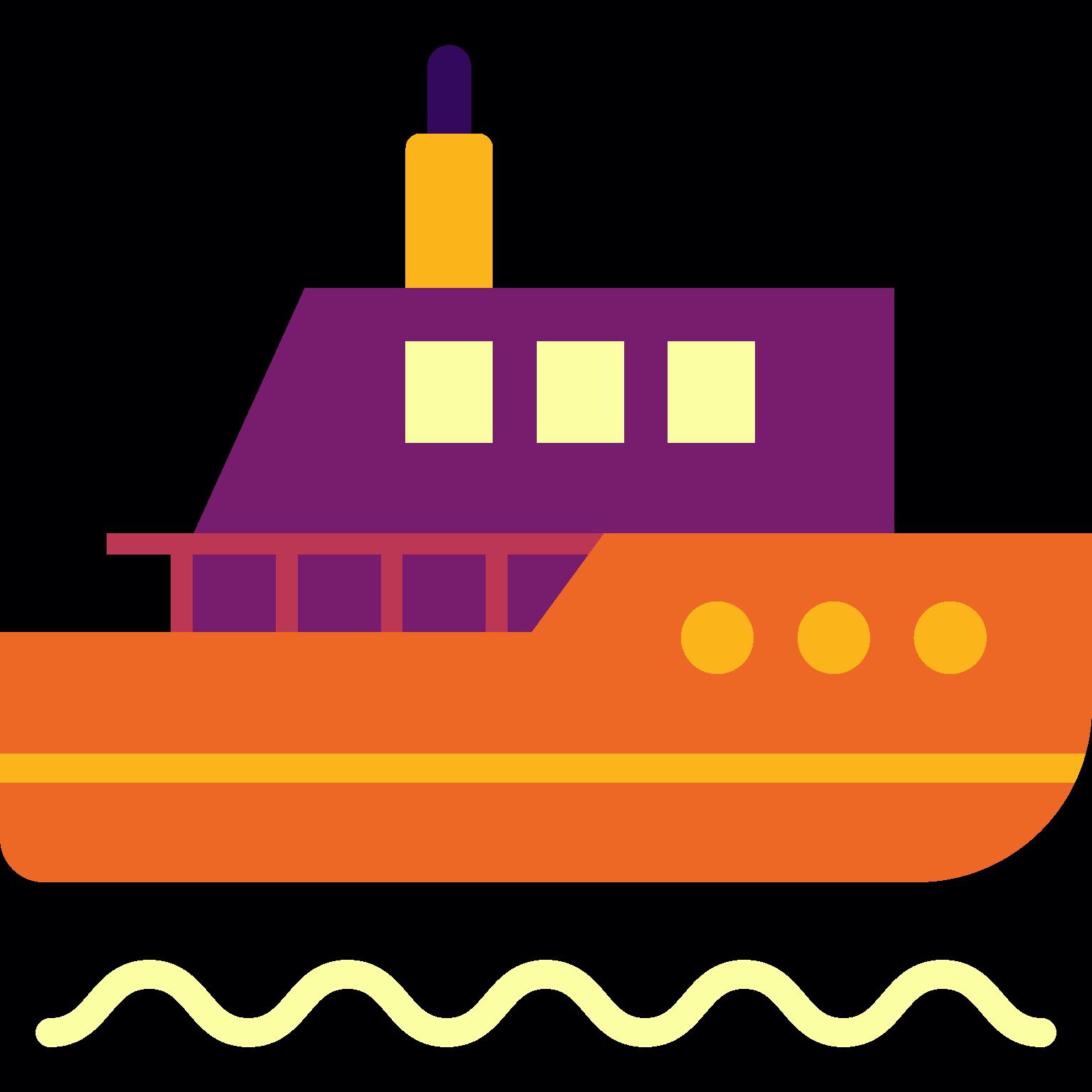}
        \caption{GT}
    \end{subfigure}
    \hfill
    \begin{subfigure}[b]{0.19\linewidth}
        \centering
        \includegraphics[width=\linewidth]{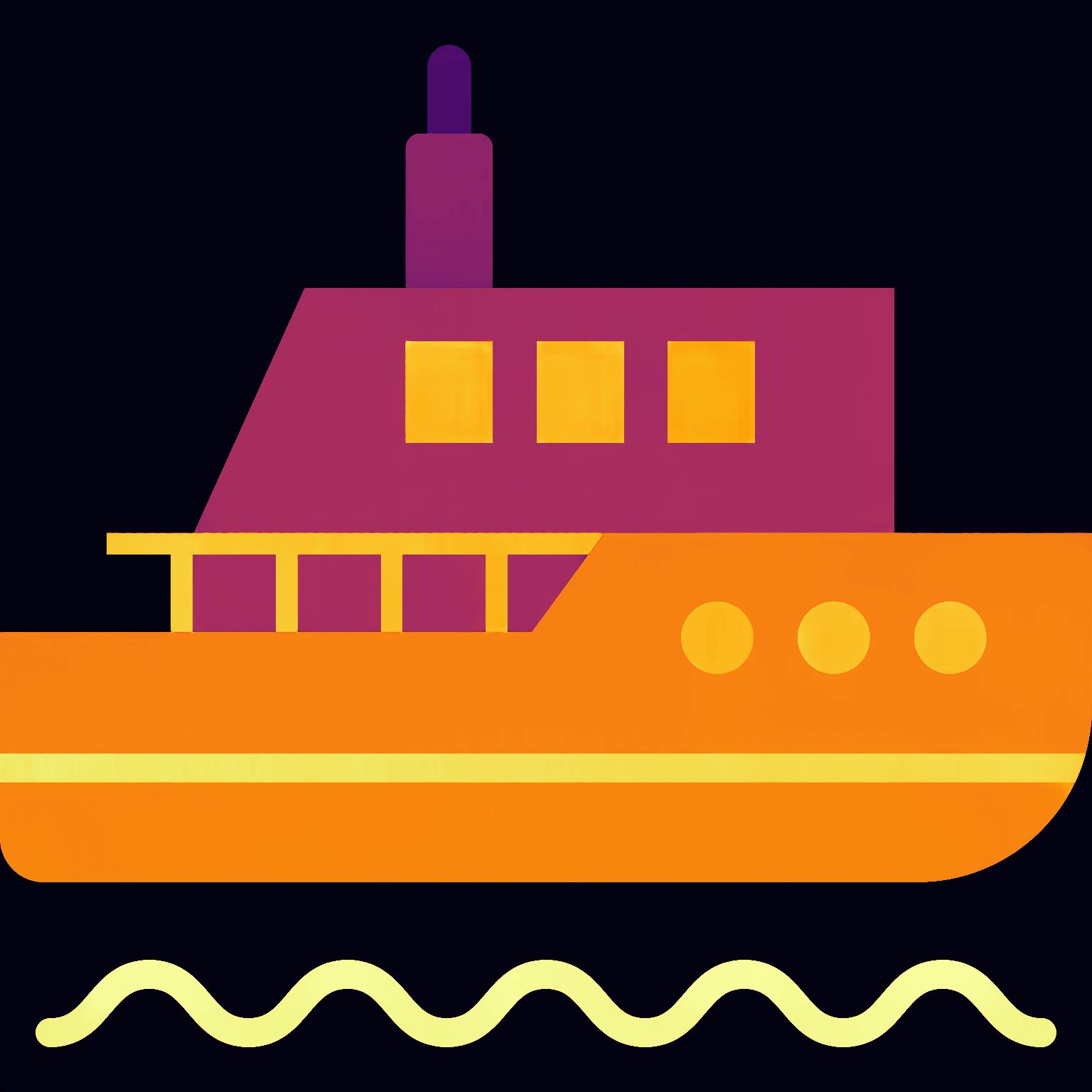}
        \caption{Ours}
    \end{subfigure}
    \hfill
    \begin{subfigure}[b]{0.19\linewidth}
        \centering
        \includegraphics[width=\linewidth]{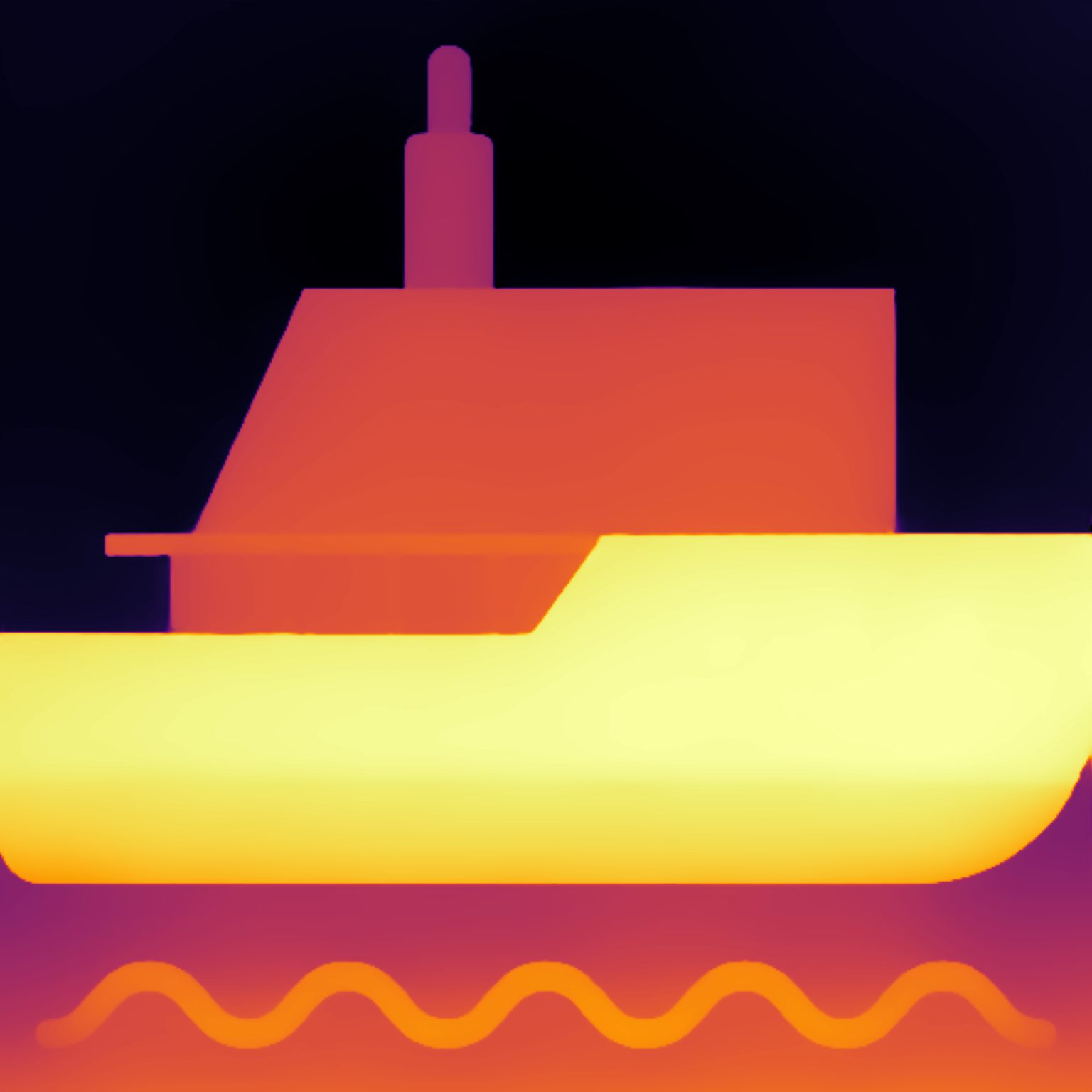}
    \caption{Dep.A.-v2}
    \end{subfigure}
    \hfill
    \begin{subfigure}[b]{0.19\linewidth}
        \centering
        \includegraphics[width=\linewidth]{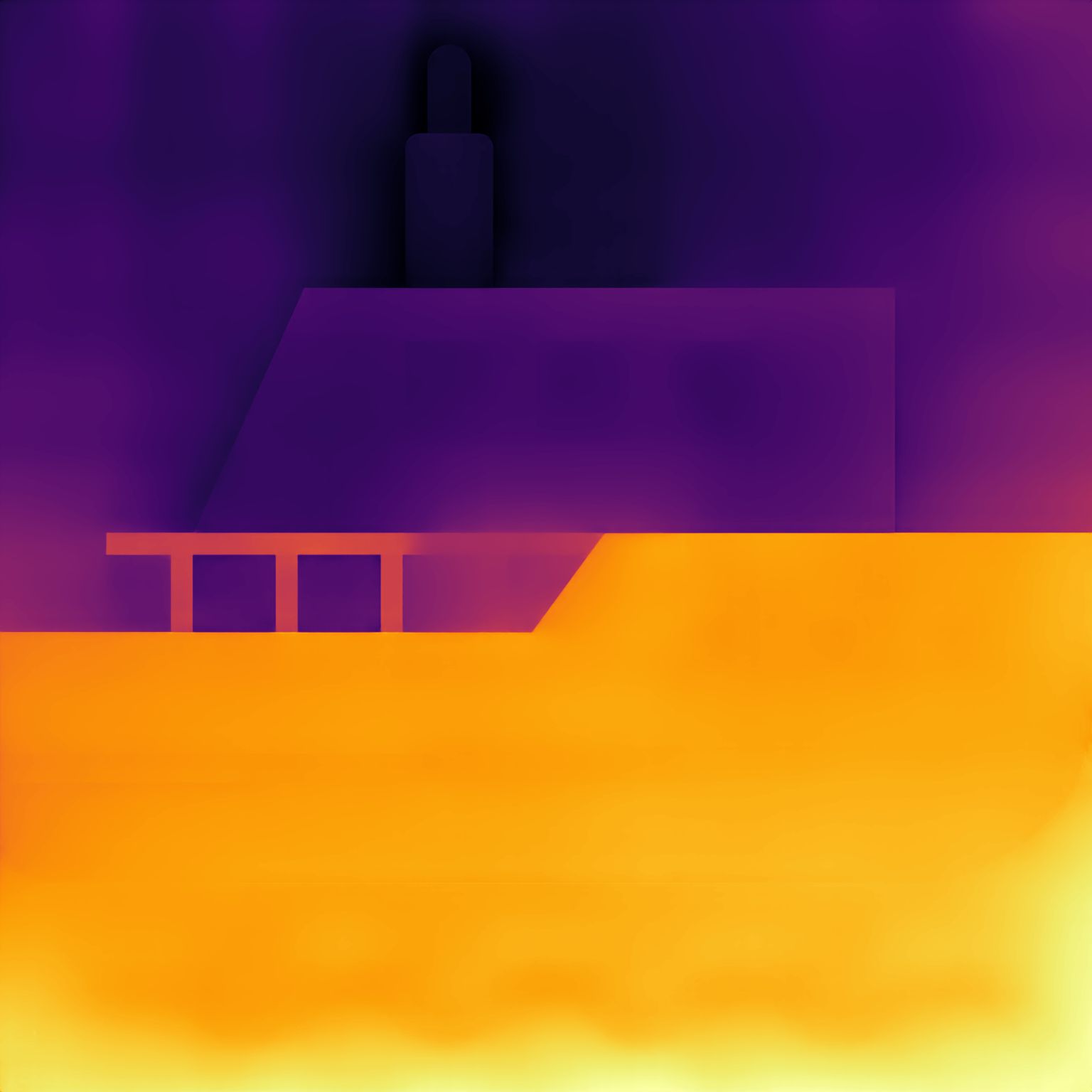}
        \caption{DepthPro}
    \end{subfigure}    \vspace*{-2mm}
    \caption{\textbf{Predicted illustrator's depth evaluation.} 
    Conventional monocular depth models (DepthAnything-v2~\cite{yang_depth_2024} (d), DepthPro~\cite{bochkovskii_depth_2025} (e)) predict physical depth; in contrast, our model (c) accurately infers layer indices suitable for illustration decomposition.
    \vspace*{-3mm}}
    \label{fig:mde_comp}
\end{figure}

\paragraph{Ground-truth rasterization}~Once the SVG dataset is curated, the final step is to generate the rasterized image-depth pairs for training. For each curated SVG file, we generate its corresponding RGB input image $I$ and ground-truth illustrator's depth map $D(I)$ of size $H \!\times\! W$ through a custom rasterization process.
First, we create a temporary version of the SVG where each layer's original color is replaced by a unique color representing its layer index $i$ in base 256: the index is thus encoded across the RGB channels via 
\vspace*{-2.25mm}
\[\left(i \bmod 256, \lfloor i / 256 \rfloor \bmod 256, \lfloor i / 256^2 \rfloor \bmod 256\right).
\vspace*{-2.25mm}
\]
We then rasterize this modified SVG; the resulting ``false color" image is converted back into a per-pixel integer depth map using the formula $D\!\left(I\right) \!=\! R+256\!\cdot\!G + 256^2\!\cdot\!B.$
This encoding strategy allows us to efficiently represent a large number of layers with virtually no additional data loading overhead. 
All the resulting pairs \(\{I_k, D(I_k)\}_k\) of images and their illustrator's depths form our training dataset. 
\begin{figure*}[h]
\centering
    \rotatebox{90}{\hspace{0.15cm} \footnotesize{Rasterized RGB}}
    \begin{subfigure}[b]{0.135\linewidth}
        \centering
        \includegraphics[width=\linewidth]{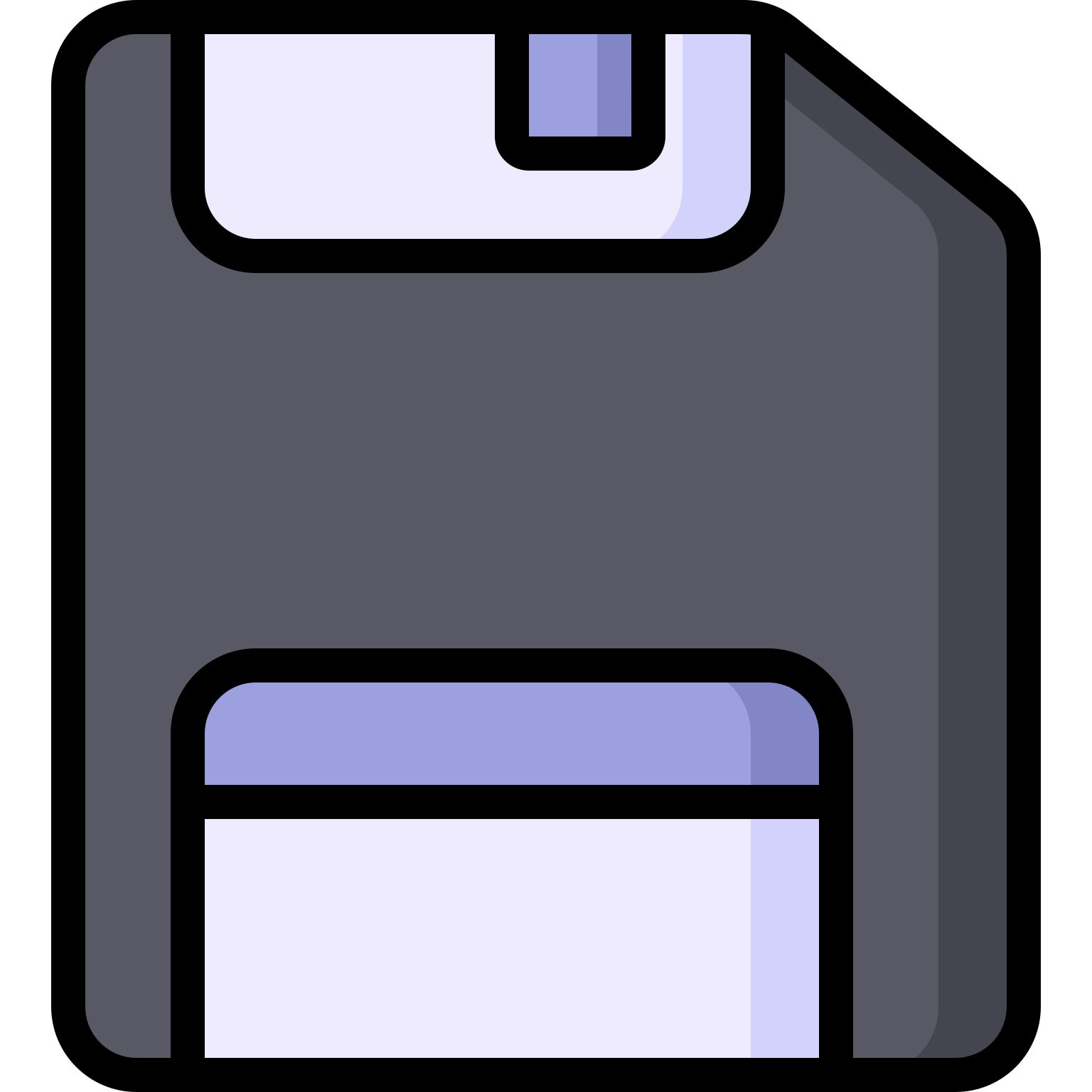}
    \end{subfigure}
    \hfill
    \begin{subfigure}[b]{0.135\linewidth}
        \centering
        \includegraphics[width=\linewidth]{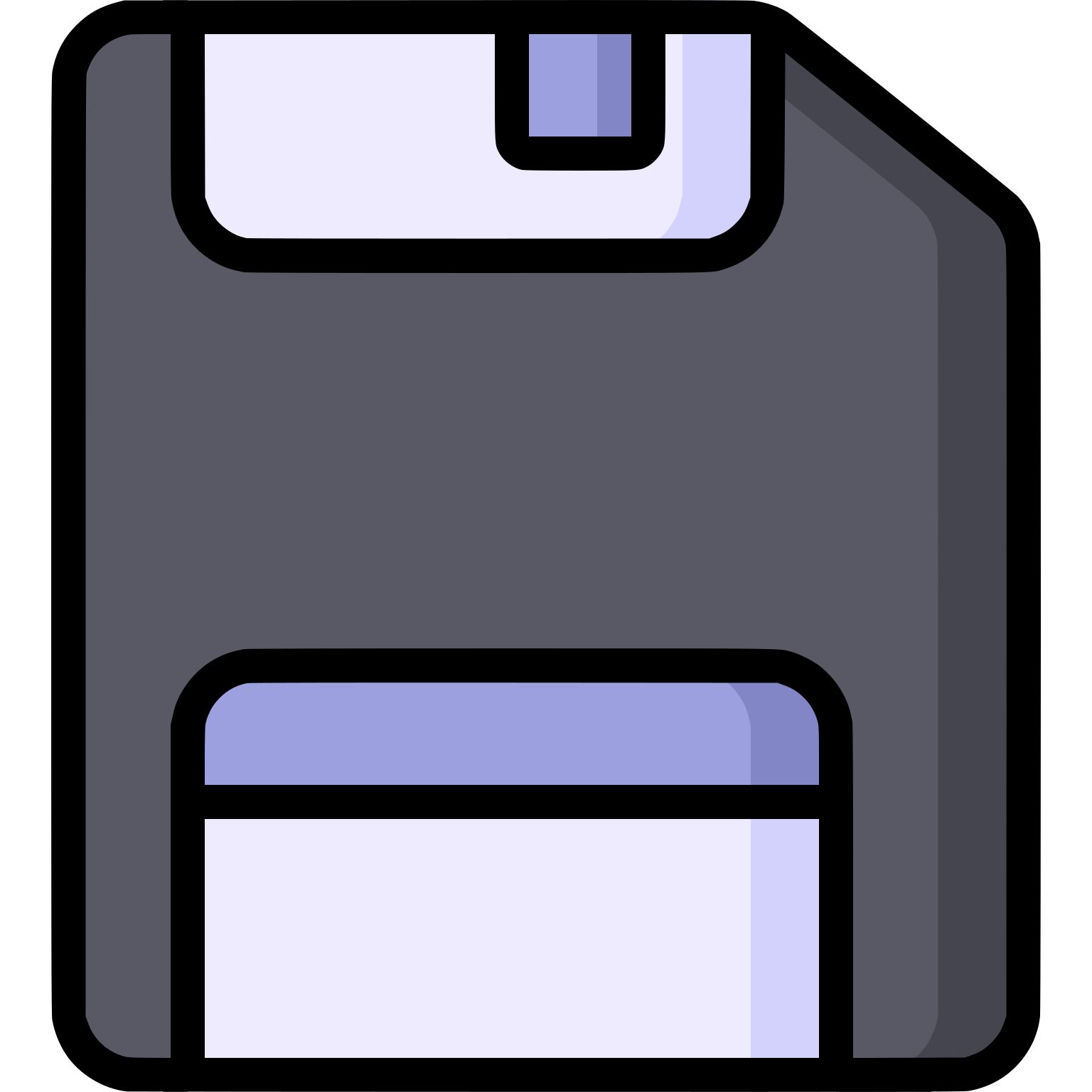
        }
    \end{subfigure}
    \hfill
    \begin{subfigure}[b]{0.135\linewidth}
        \centering
        \includegraphics[width=\linewidth]{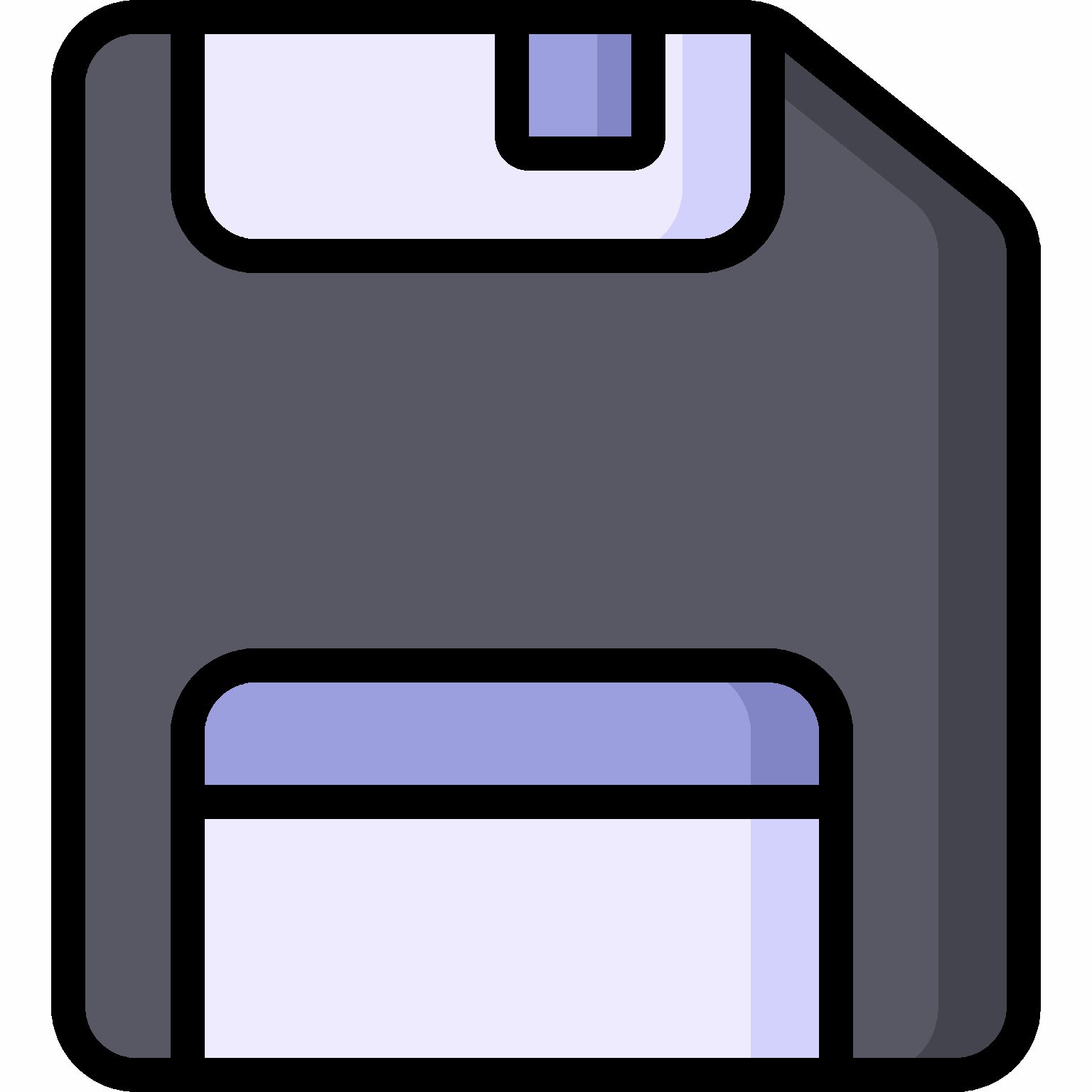}
    \end{subfigure}
    \hfill
    \begin{subfigure}[b]{0.135\linewidth}
        \centering
        \includegraphics[width=\linewidth]{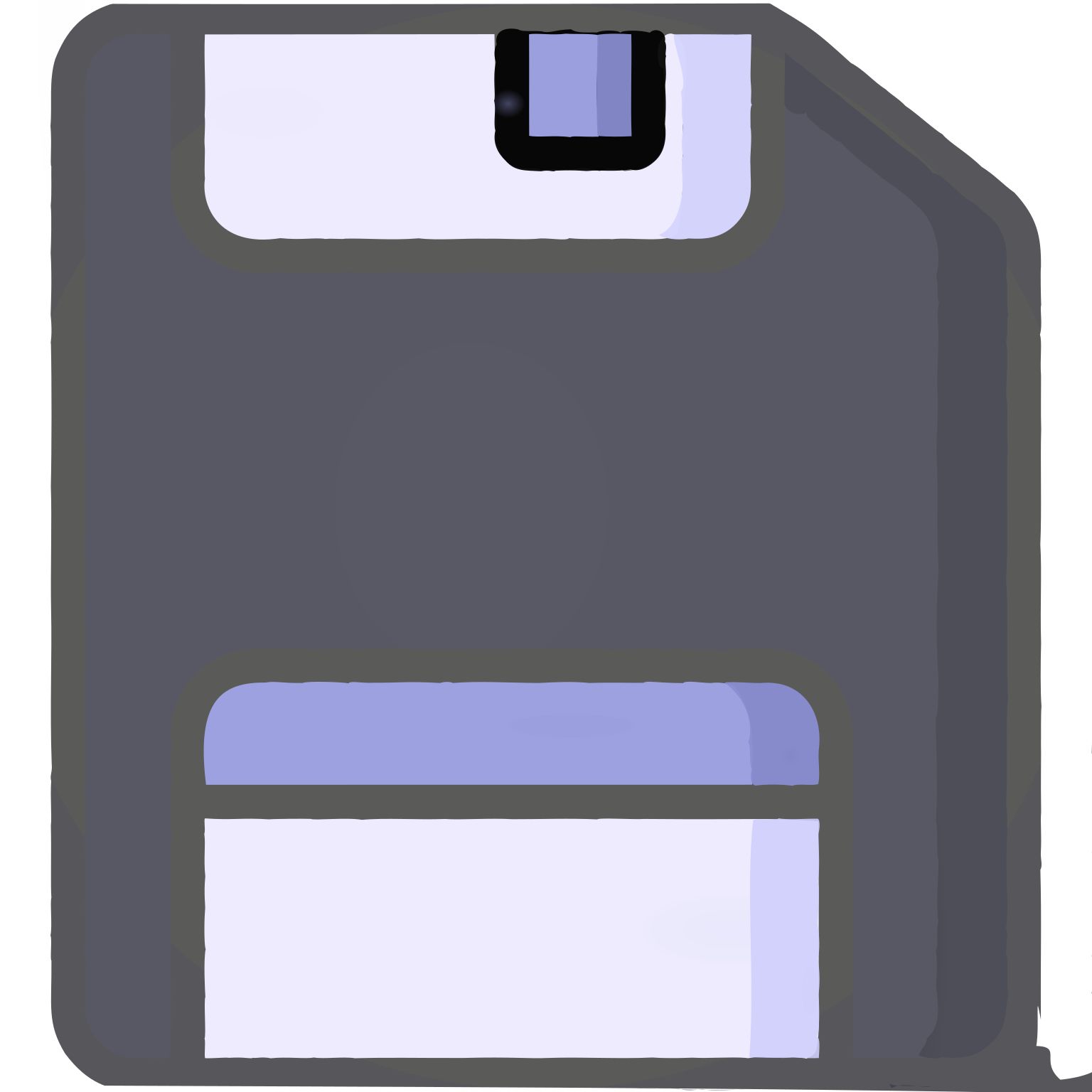}
    \end{subfigure}
    \hfill
    \begin{subfigure}[b]{0.135\linewidth}
        \centering
        \includegraphics[width=\linewidth]{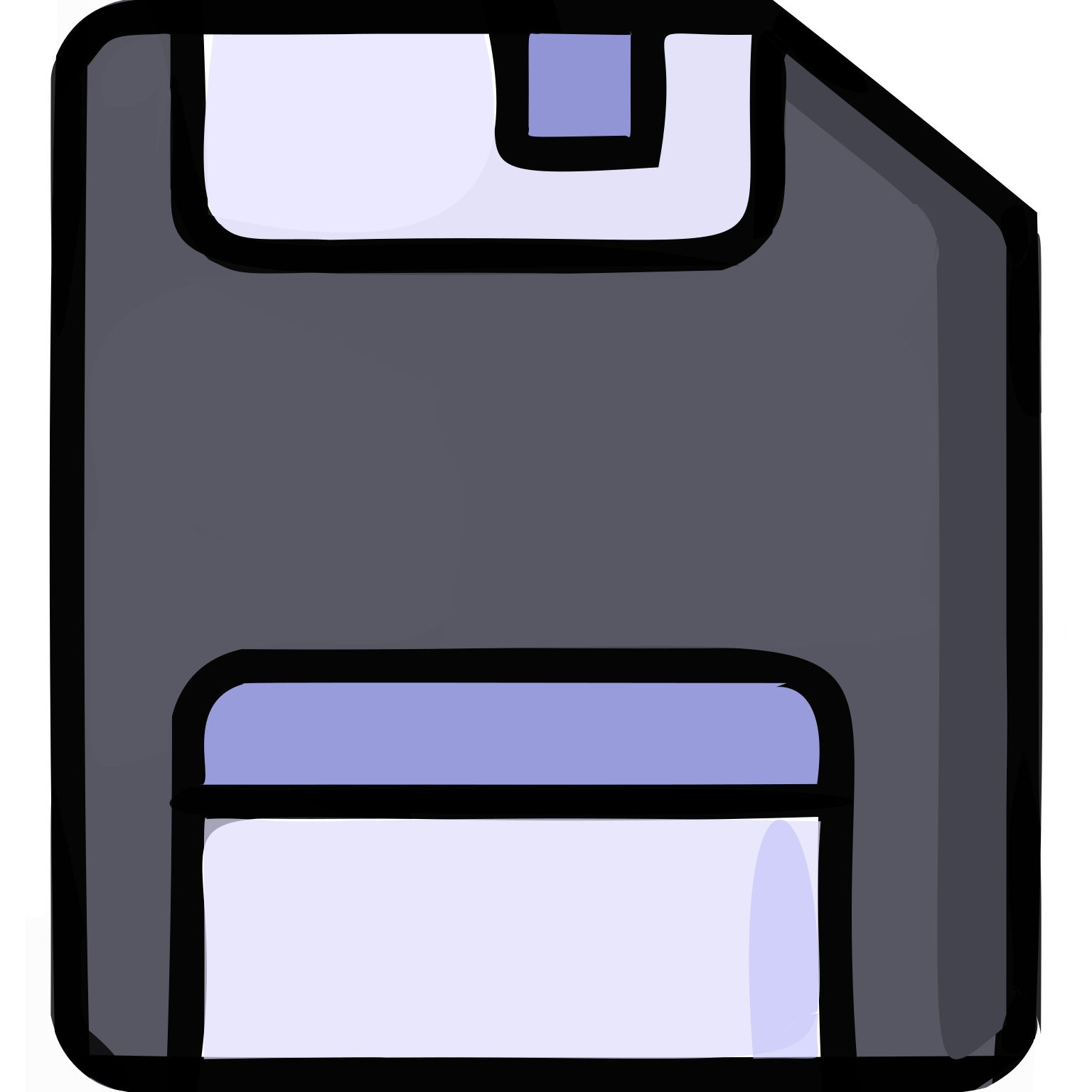}
    \end{subfigure}
    \hfill
    \begin{subfigure}[b]{0.135\linewidth}
        \centering
        \includegraphics[width=\linewidth]{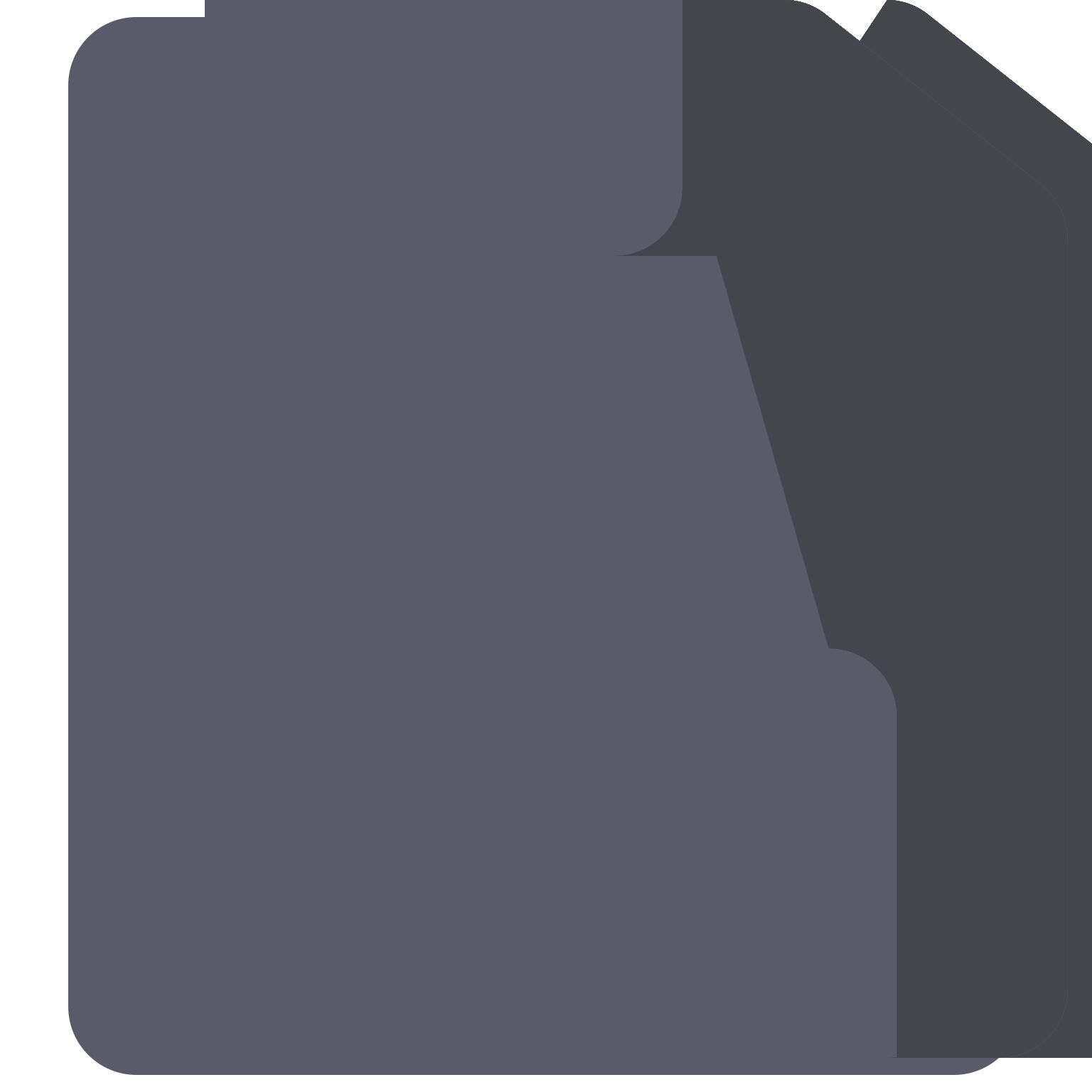}
    \end{subfigure}
    \hfill
    \begin{subfigure}[b]{0.135\linewidth}
        \centering
        \includegraphics[width=\linewidth]{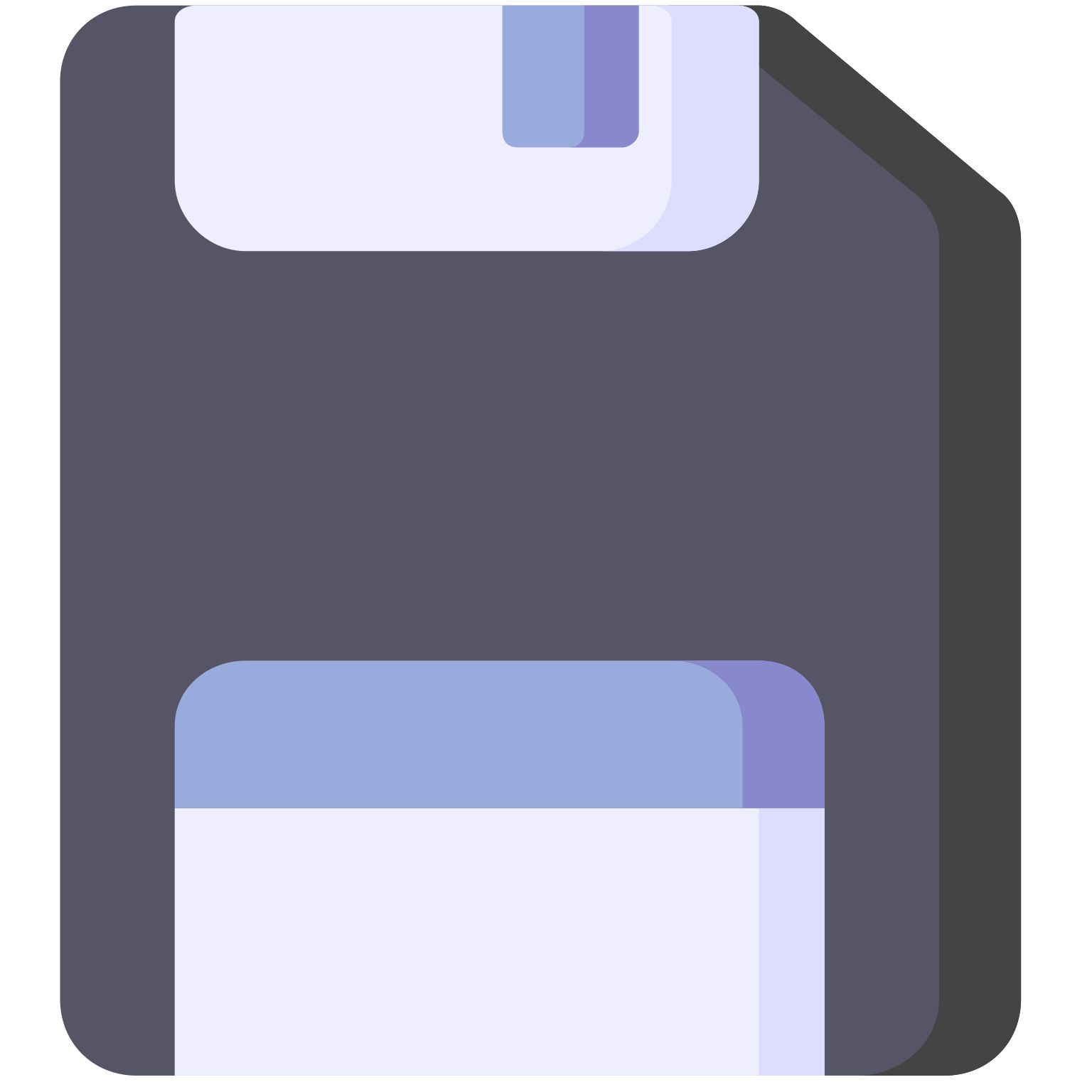}
    \end{subfigure}

     \rotatebox{90}{\hspace{0.15cm} \footnotesize{Rasterized Depth}}
    \begin{subfigure}[b]{0.135\linewidth}
        \centering
        \includegraphics[width=\linewidth]{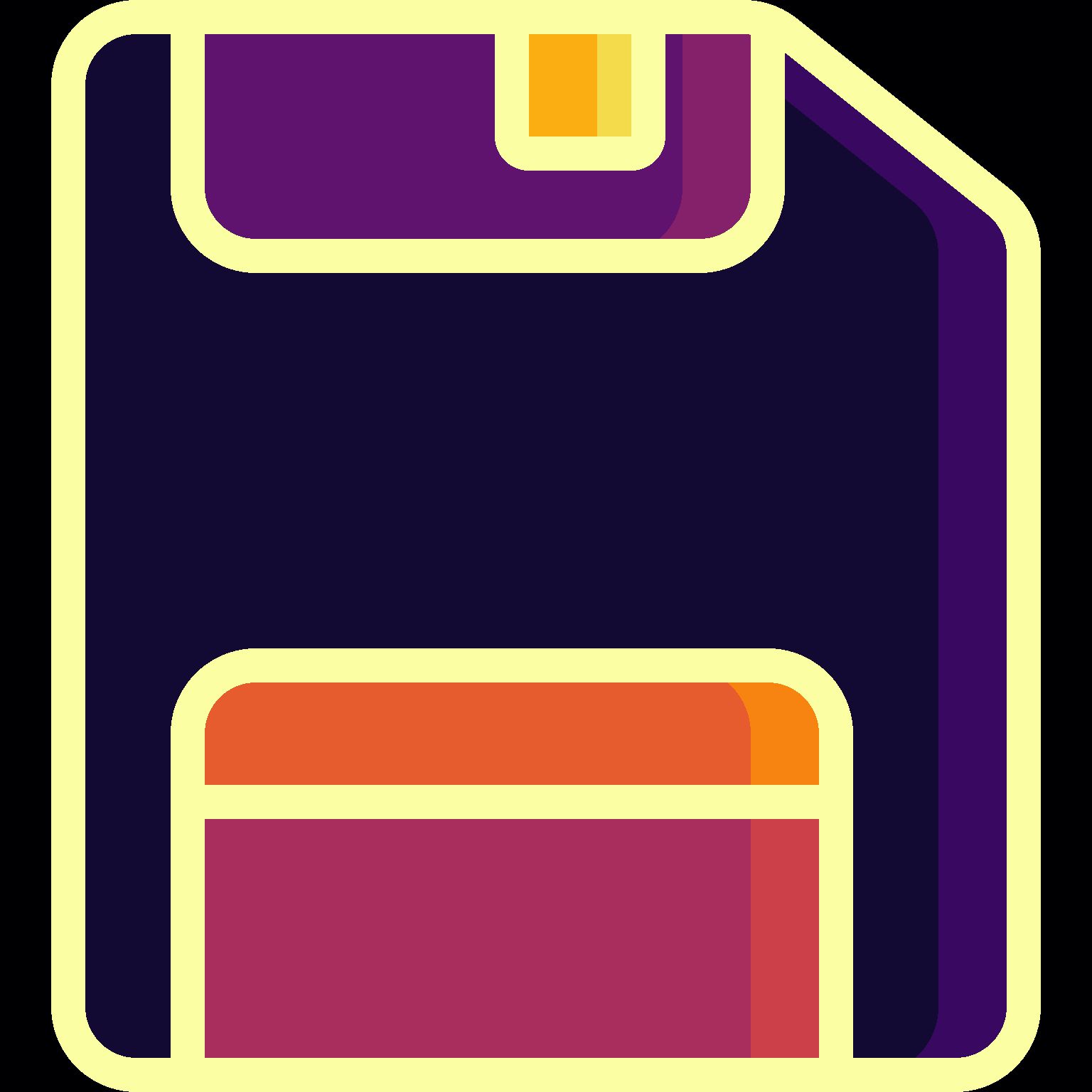}
    \end{subfigure}
    \hfill
    \begin{subfigure}[b]{0.135\linewidth}
        \centering
        \includegraphics[width=\linewidth]{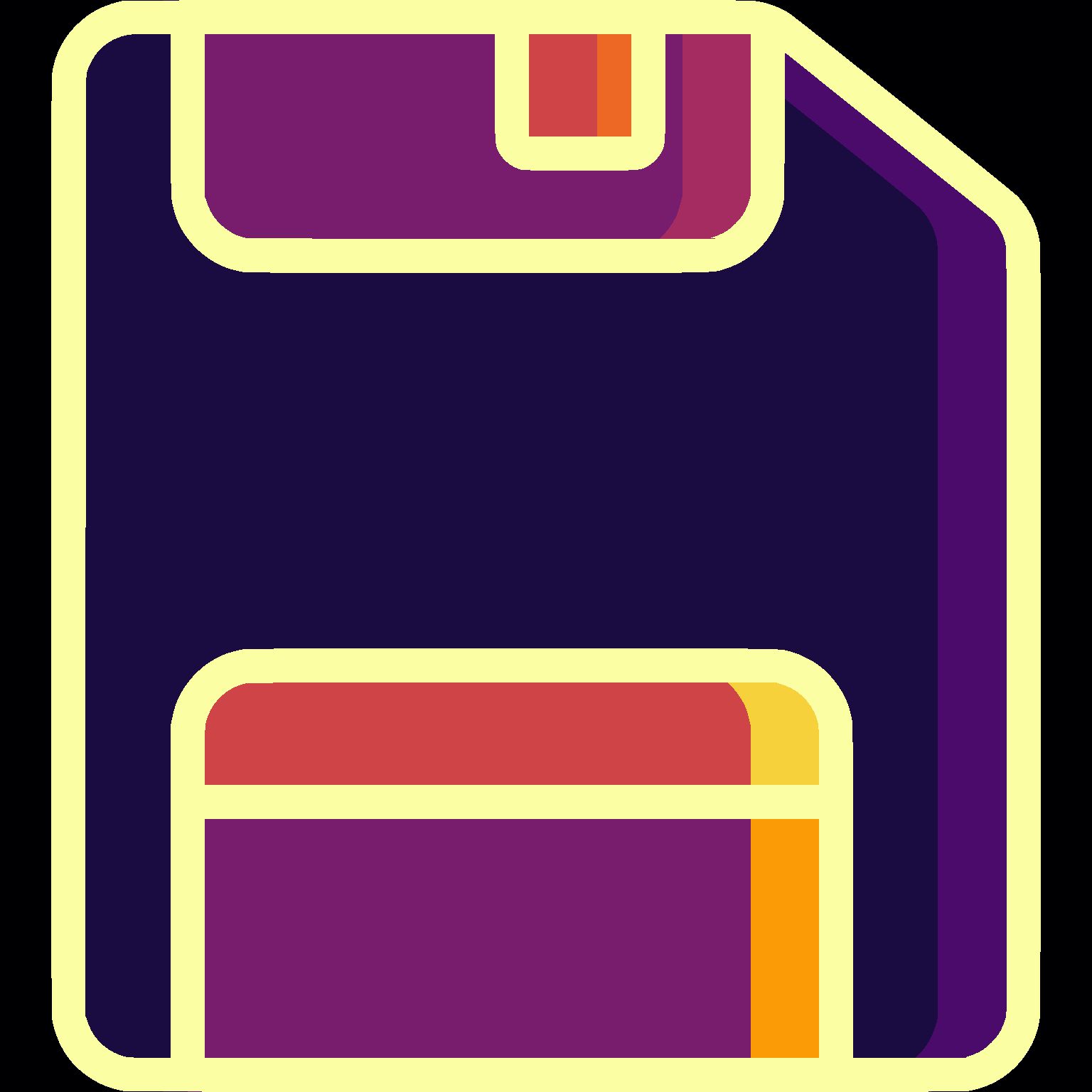}
    \end{subfigure}
    \hfill
    \begin{subfigure}[b]{0.135\linewidth}
        \centering
        \includegraphics[width=\linewidth]{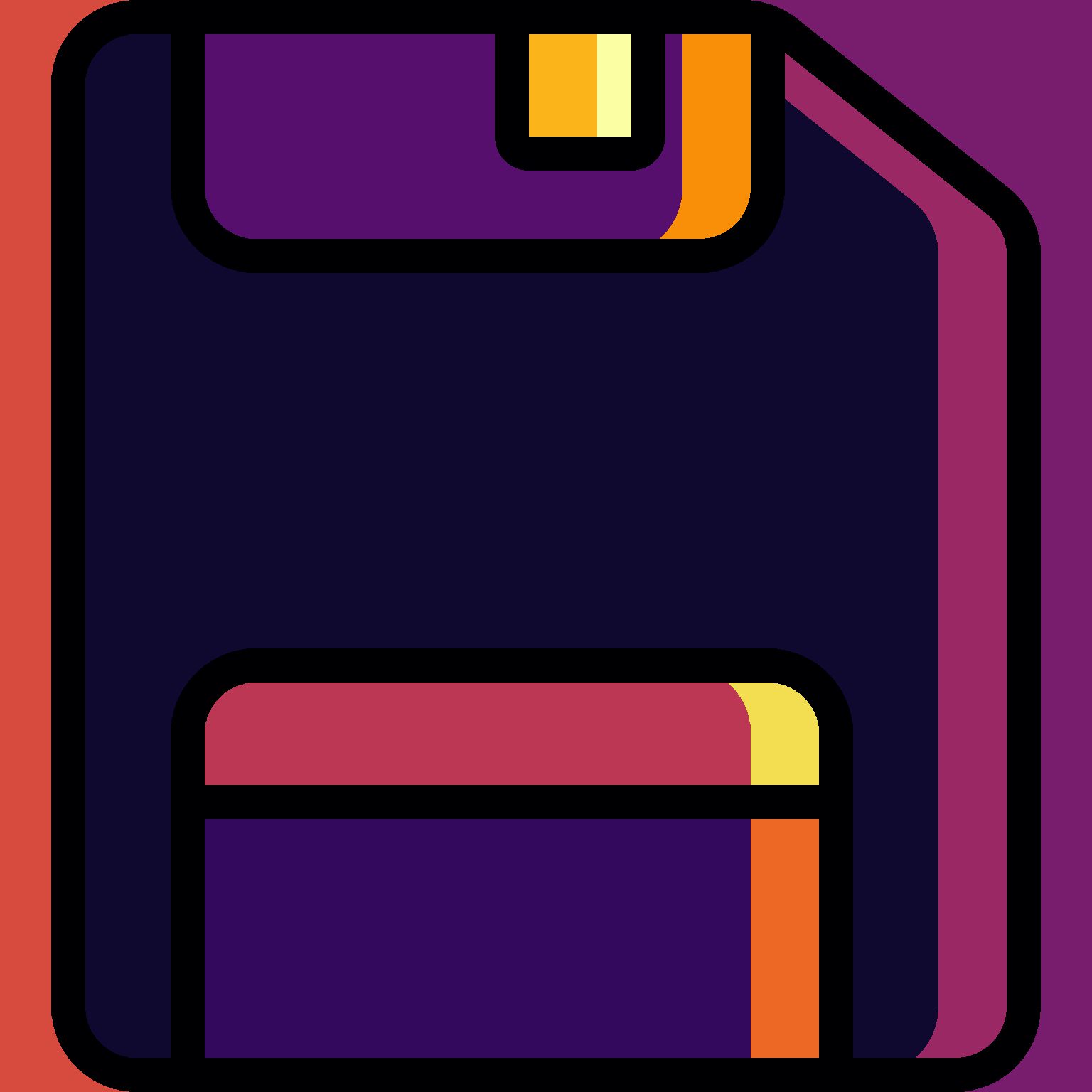}
    \end{subfigure}
    \hfill
    \begin{subfigure}[b]{0.135\linewidth}
        \centering
        \includegraphics[width=\linewidth]{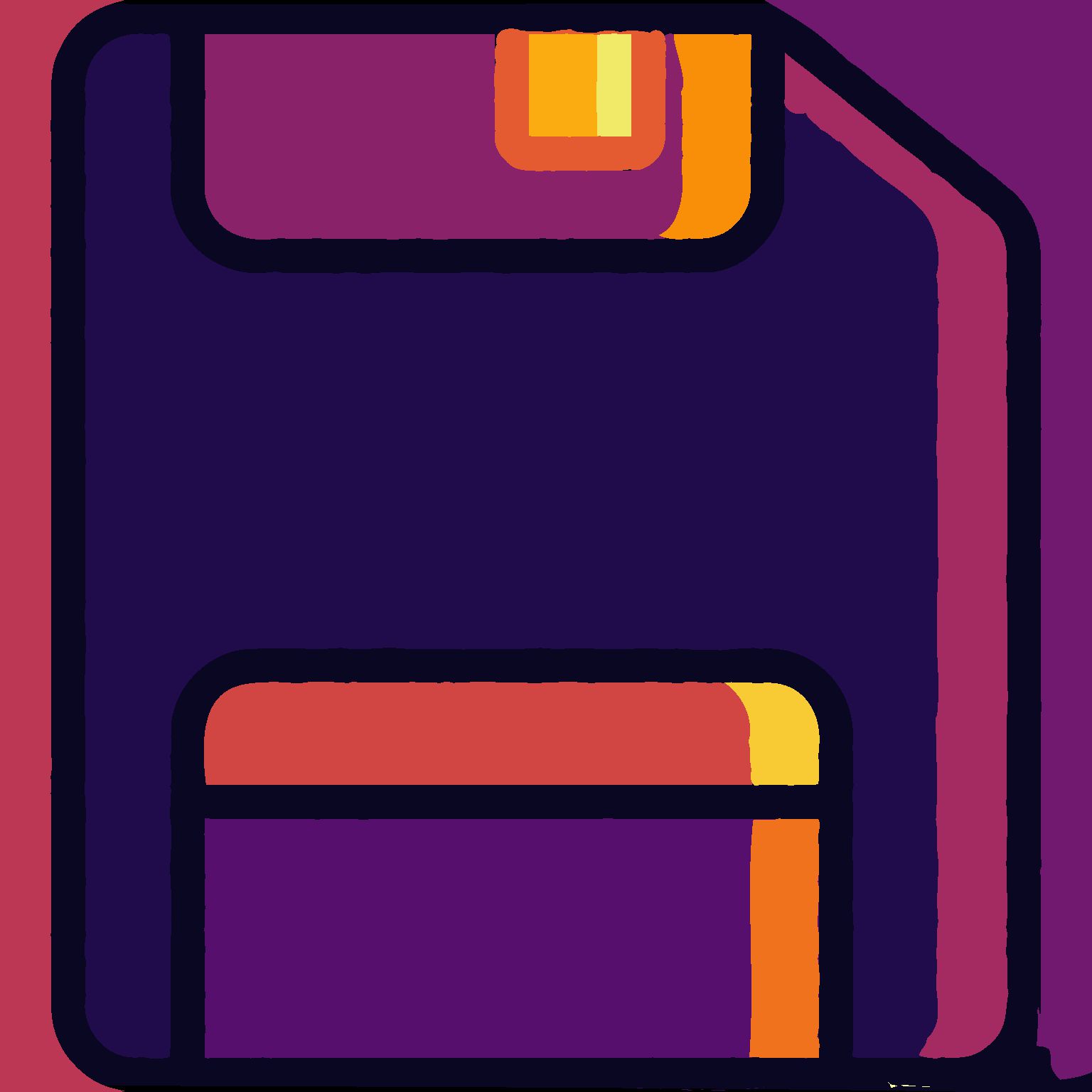}
    \end{subfigure}
    \hfill
    \begin{subfigure}[b]{0.135\linewidth}
        \centering
        \includegraphics[width=\linewidth]{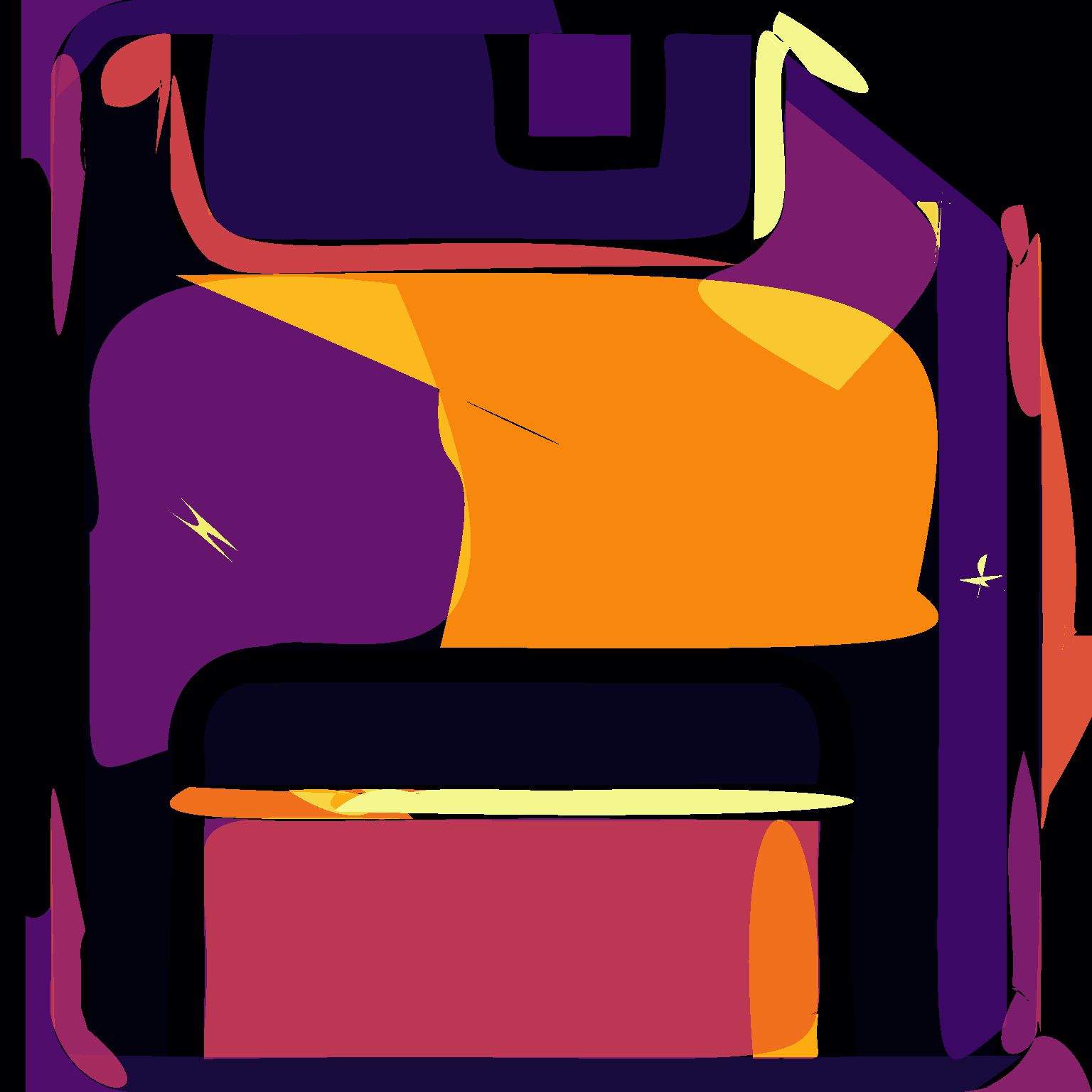}
    \end{subfigure}
    \hfill
    \begin{subfigure}[b]{0.135\linewidth}
        \centering
        \includegraphics[width=\linewidth]{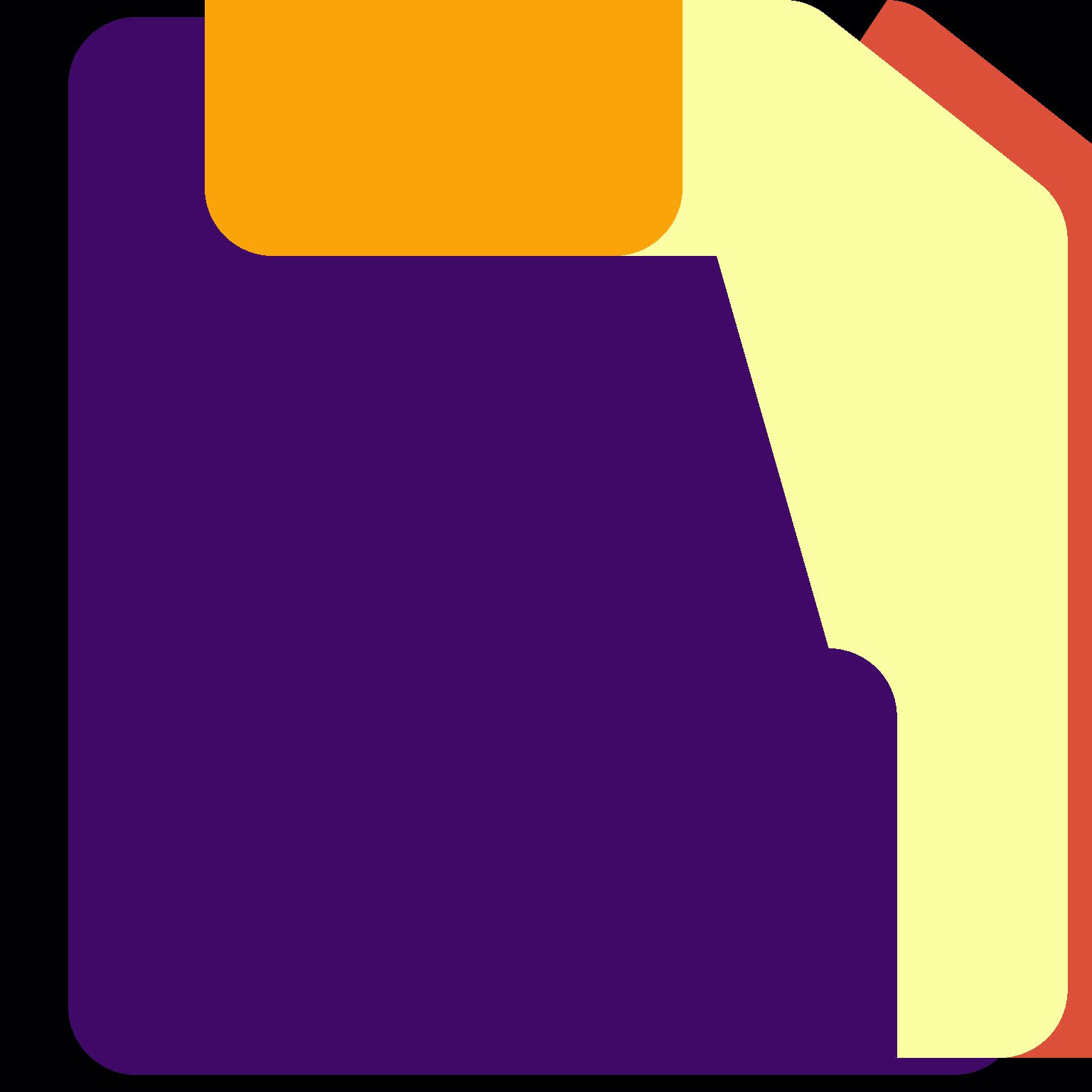}
    \end{subfigure}
    \hfill
    \begin{subfigure}[b]{0.135\linewidth}
        \centering
        \includegraphics[width=\linewidth]{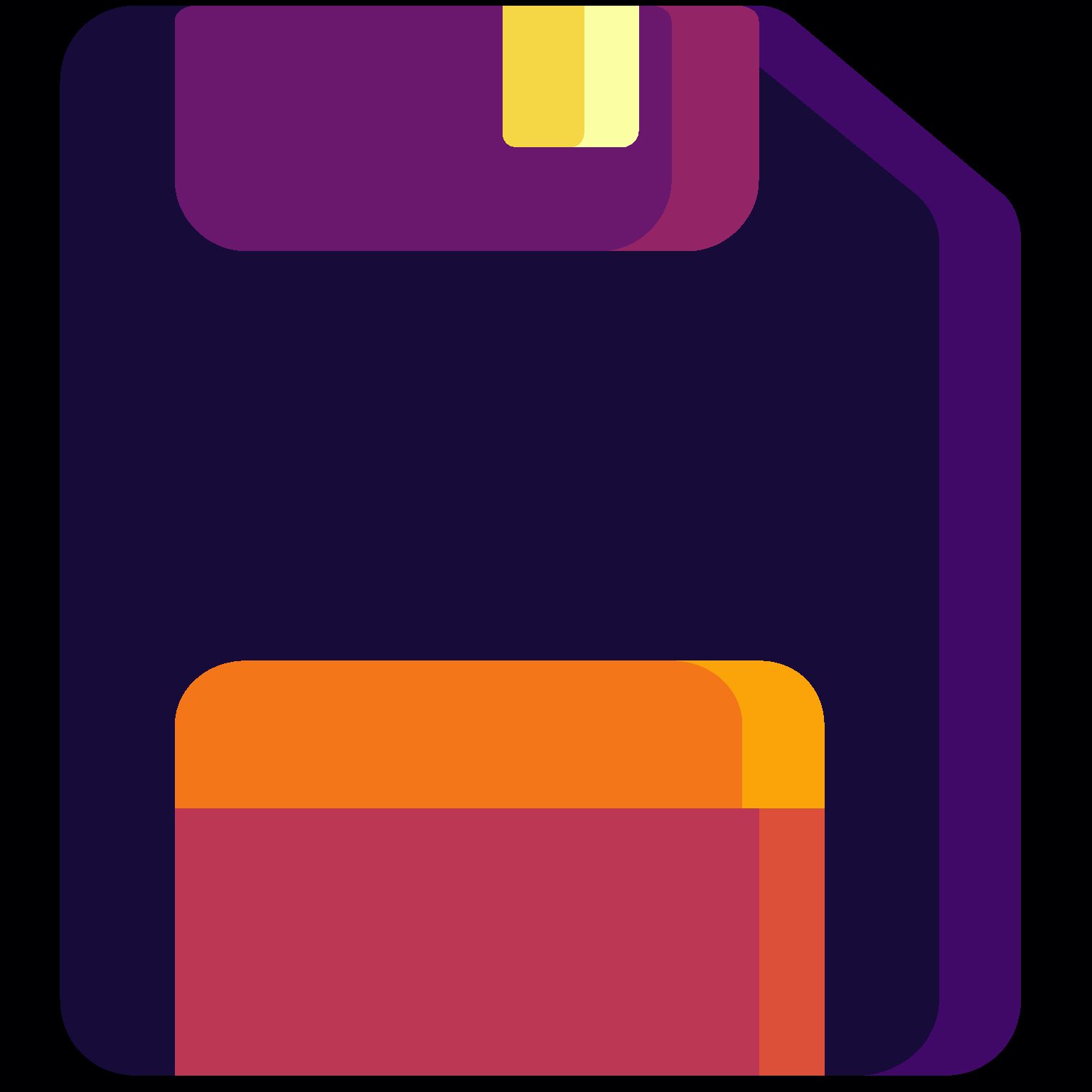}
    \end{subfigure}

    \rotatebox{90}{\hspace{0.3cm}\footnotesize{Rasterized RGB}}
    \begin{subfigure}[b]{0.135\linewidth}
        \centering
        \includegraphics[width=\linewidth]{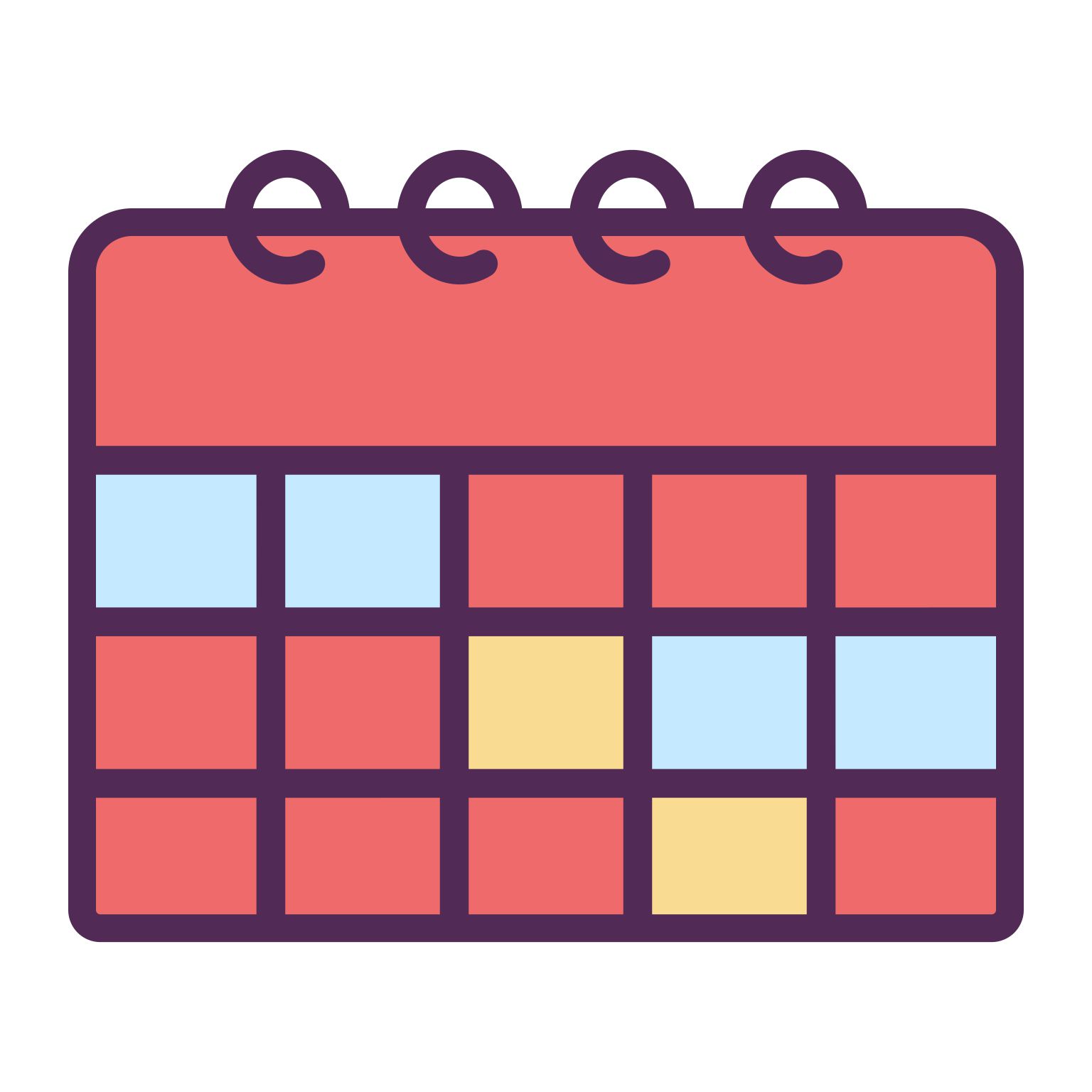}
    \end{subfigure}
    \hfill
    \begin{subfigure}[b]{0.135\linewidth}
        \centering
        \includegraphics[width=\linewidth]{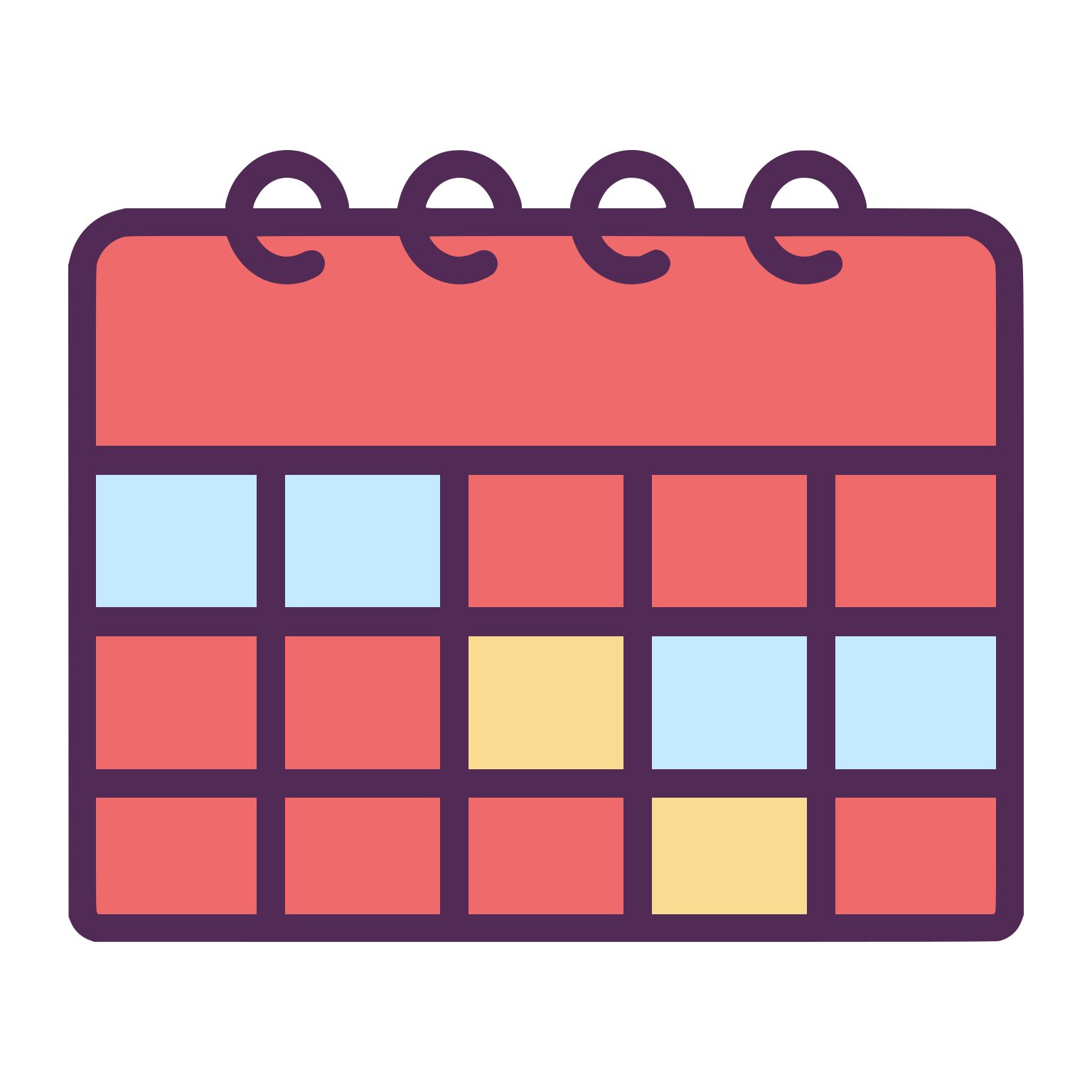}
    \end{subfigure}
    \hfill
    \begin{subfigure}[b]{0.135\linewidth}
        \centering
        \includegraphics[width=\linewidth]{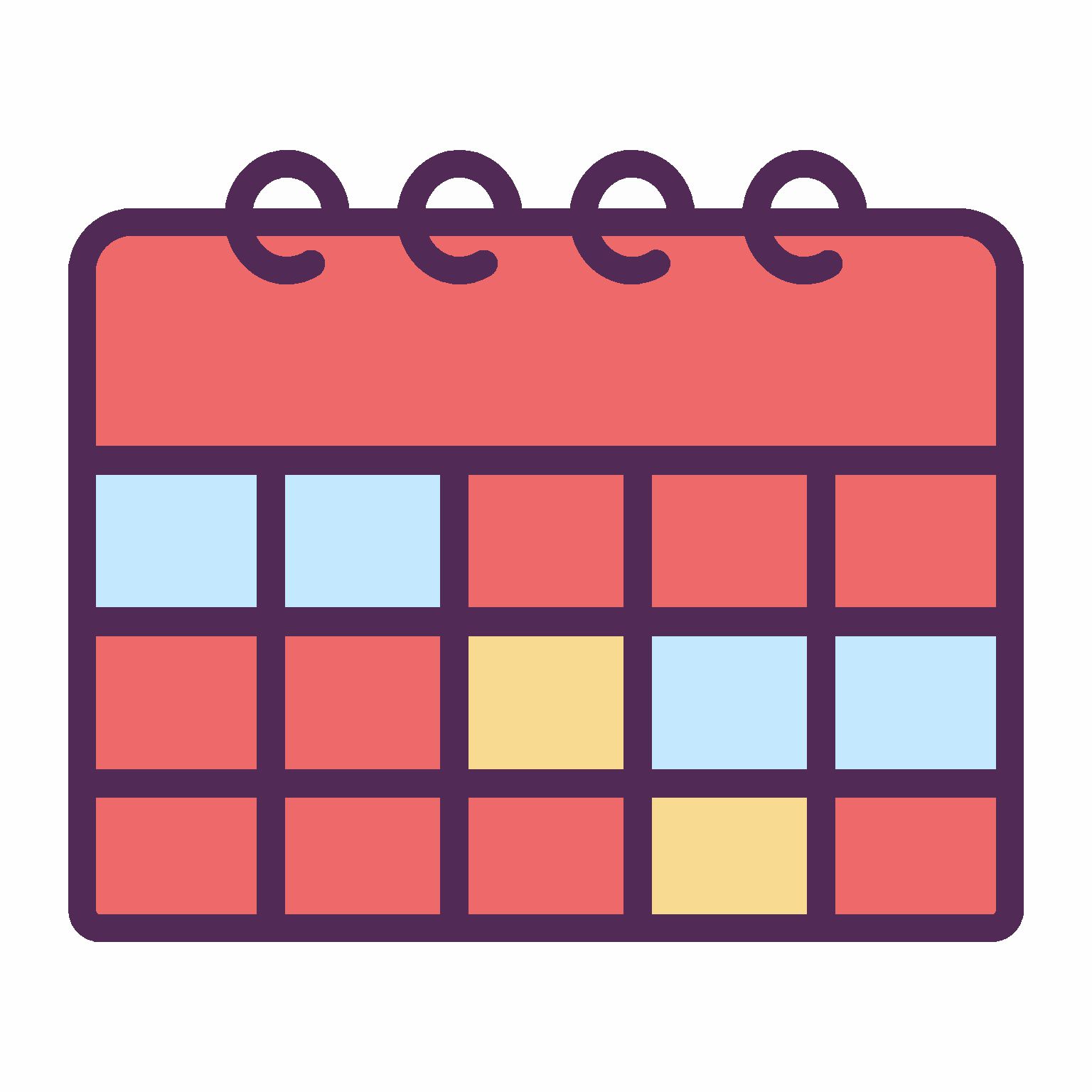}
    \end{subfigure}
    \hfill
    \begin{subfigure}[b]{0.135\linewidth}
        \centering
        \includegraphics[width=\linewidth]{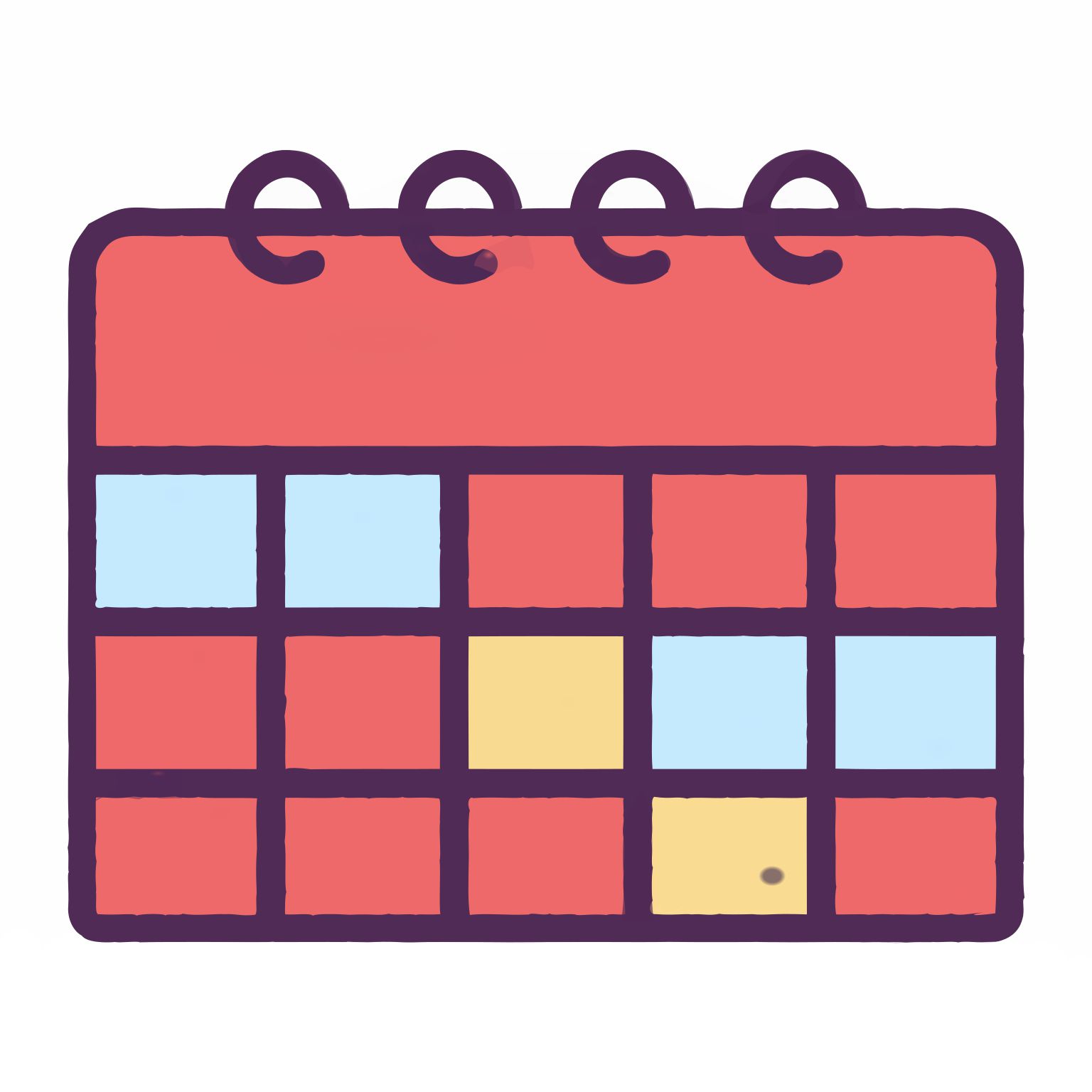}
    \end{subfigure}
    \hfill
    \begin{subfigure}[b]{0.135\linewidth}
        \centering
        \includegraphics[width=\linewidth]{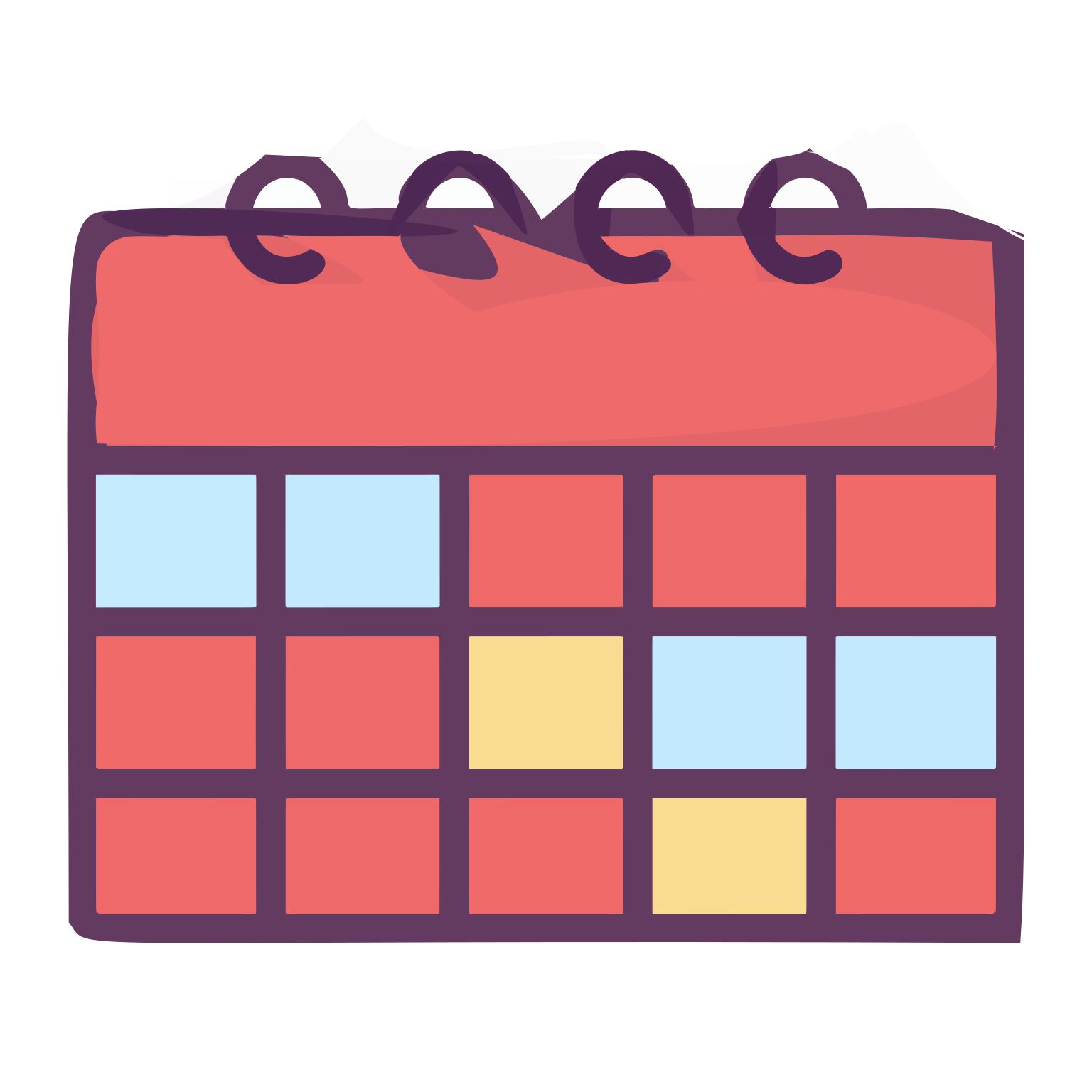}
    \end{subfigure}
    \hfill
    \begin{subfigure}[b]{0.135\linewidth}
        \centering
        \includegraphics[width=\linewidth]{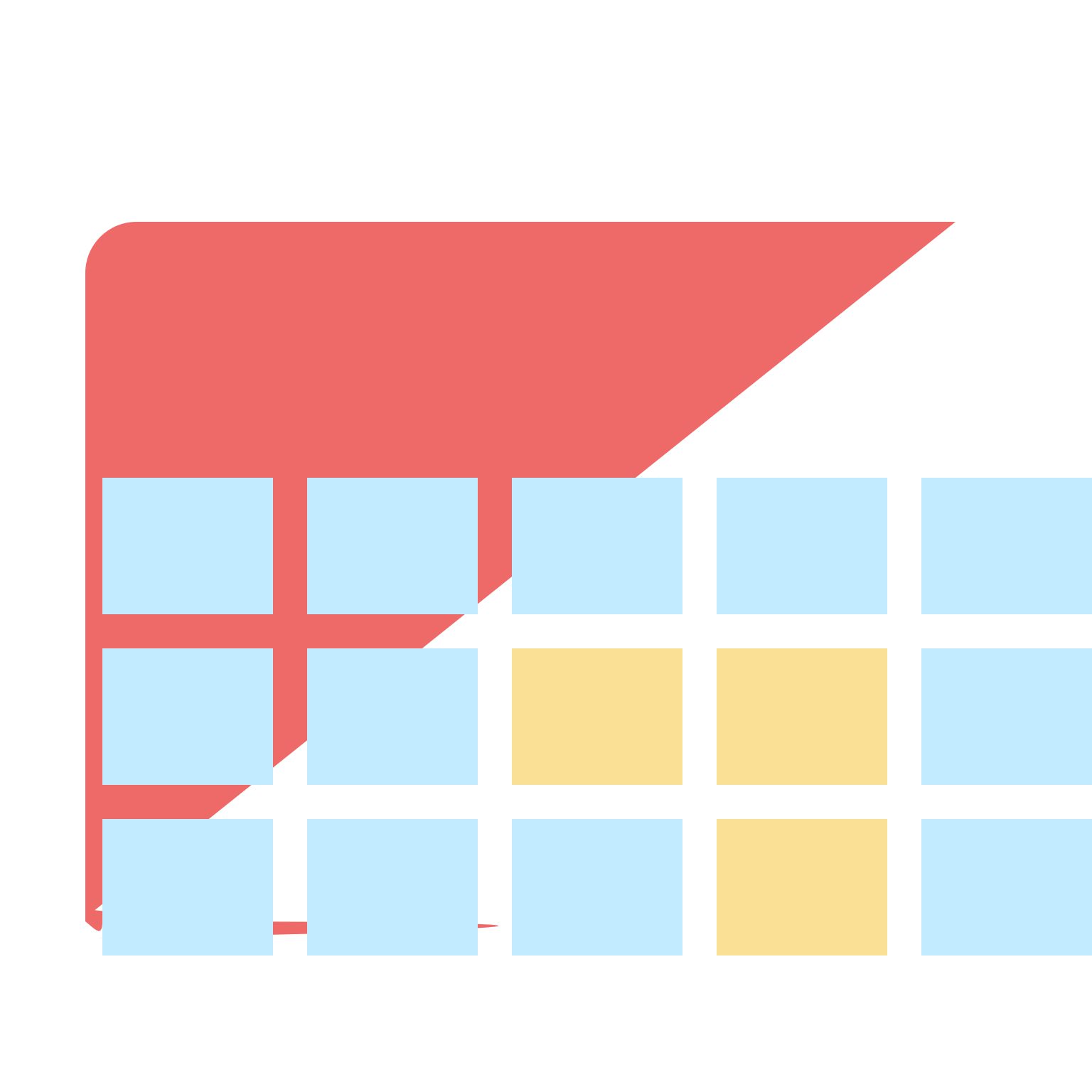}
    \end{subfigure}
    \hfill
    \begin{subfigure}[b]{0.135\linewidth}
        \centering
        \includegraphics[width=\linewidth]{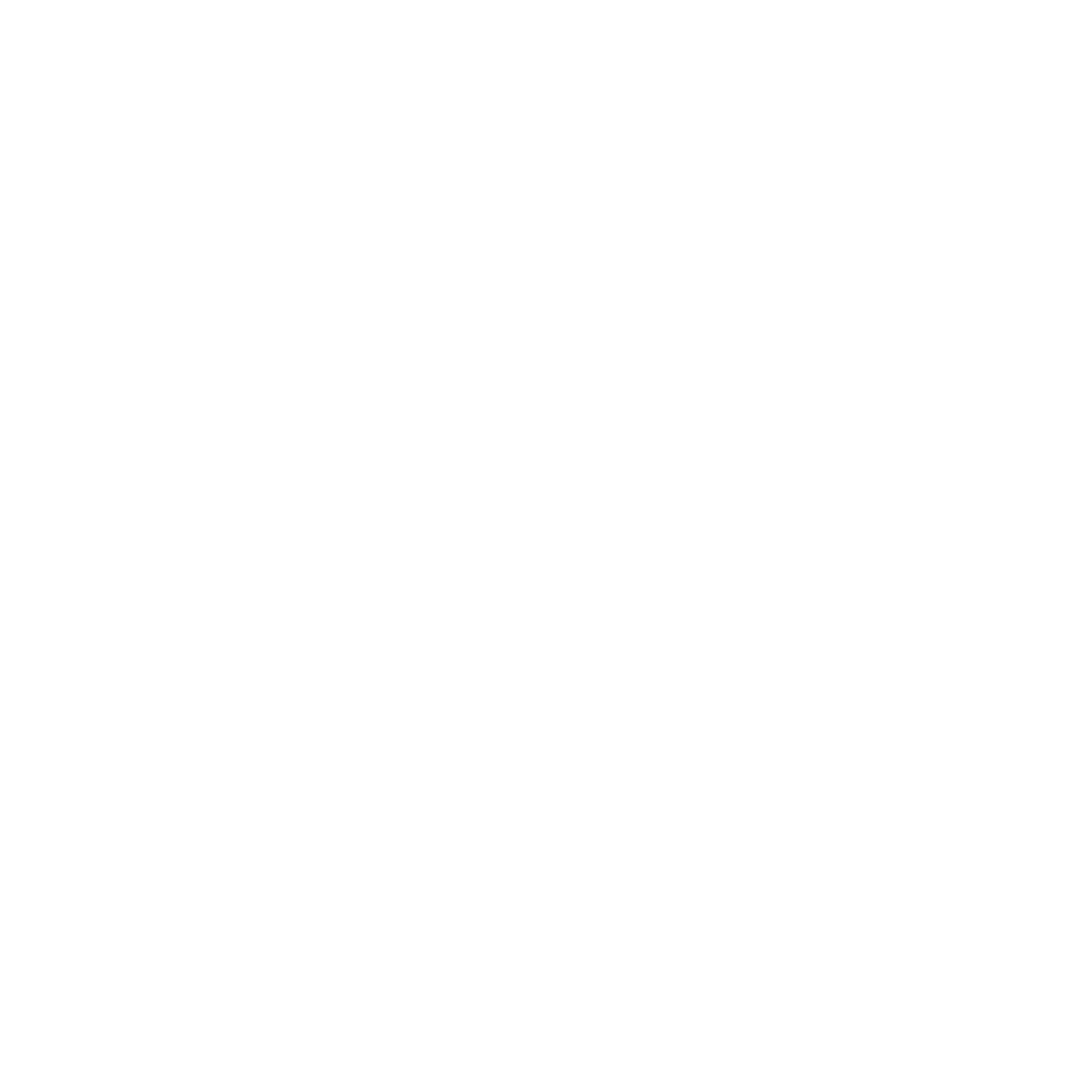}
    \end{subfigure}
    
    \rotatebox{90}{\hspace{0.55cm} \footnotesize{Rasterized Depth}}
    \begin{subfigure}[b]{0.135\linewidth}
        \centering
        \includegraphics[width=\linewidth]{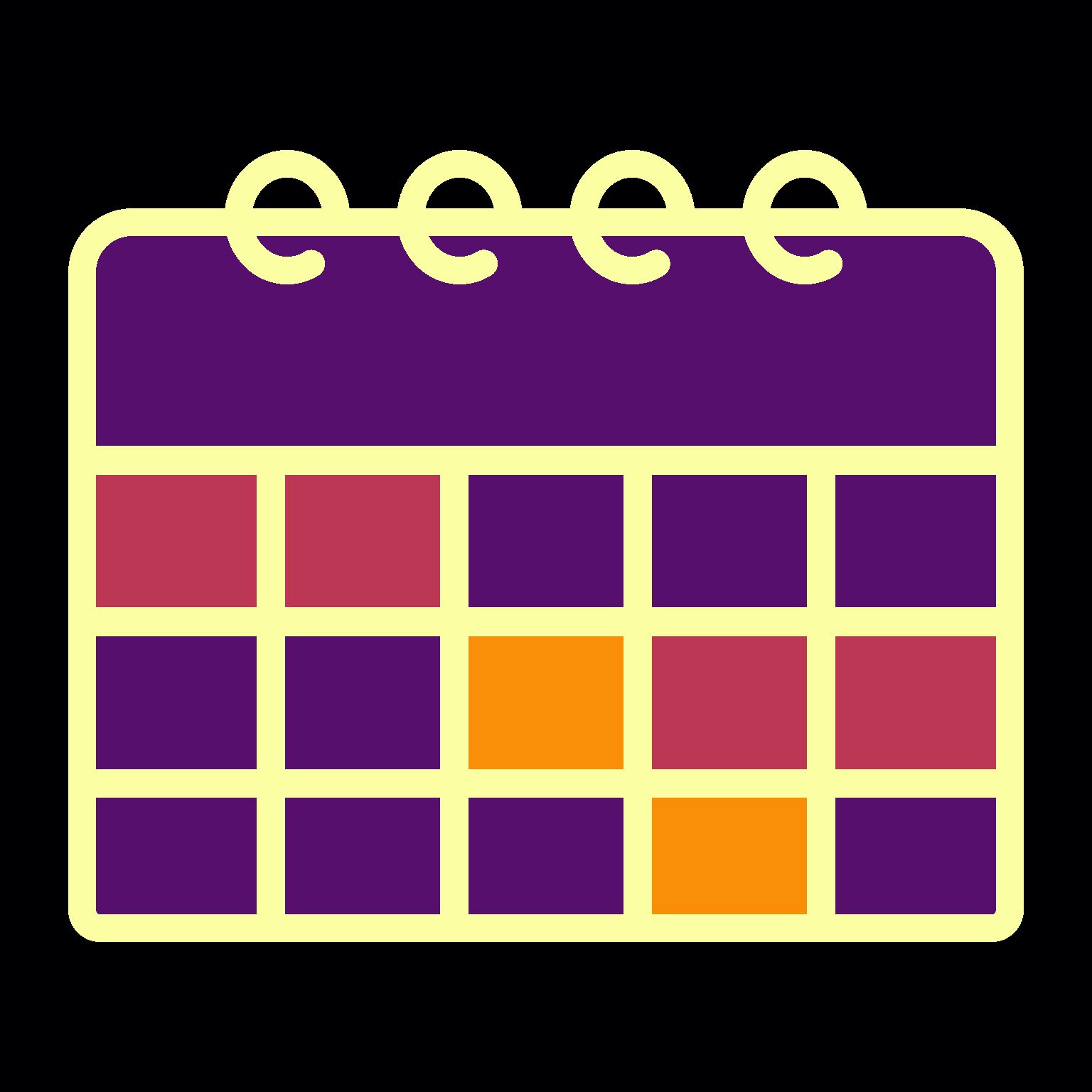}
        \caption{GT}
    \end{subfigure}
    \hfill
    \begin{subfigure}[b]{0.135\linewidth}
        \centering
        \includegraphics[width=\linewidth]{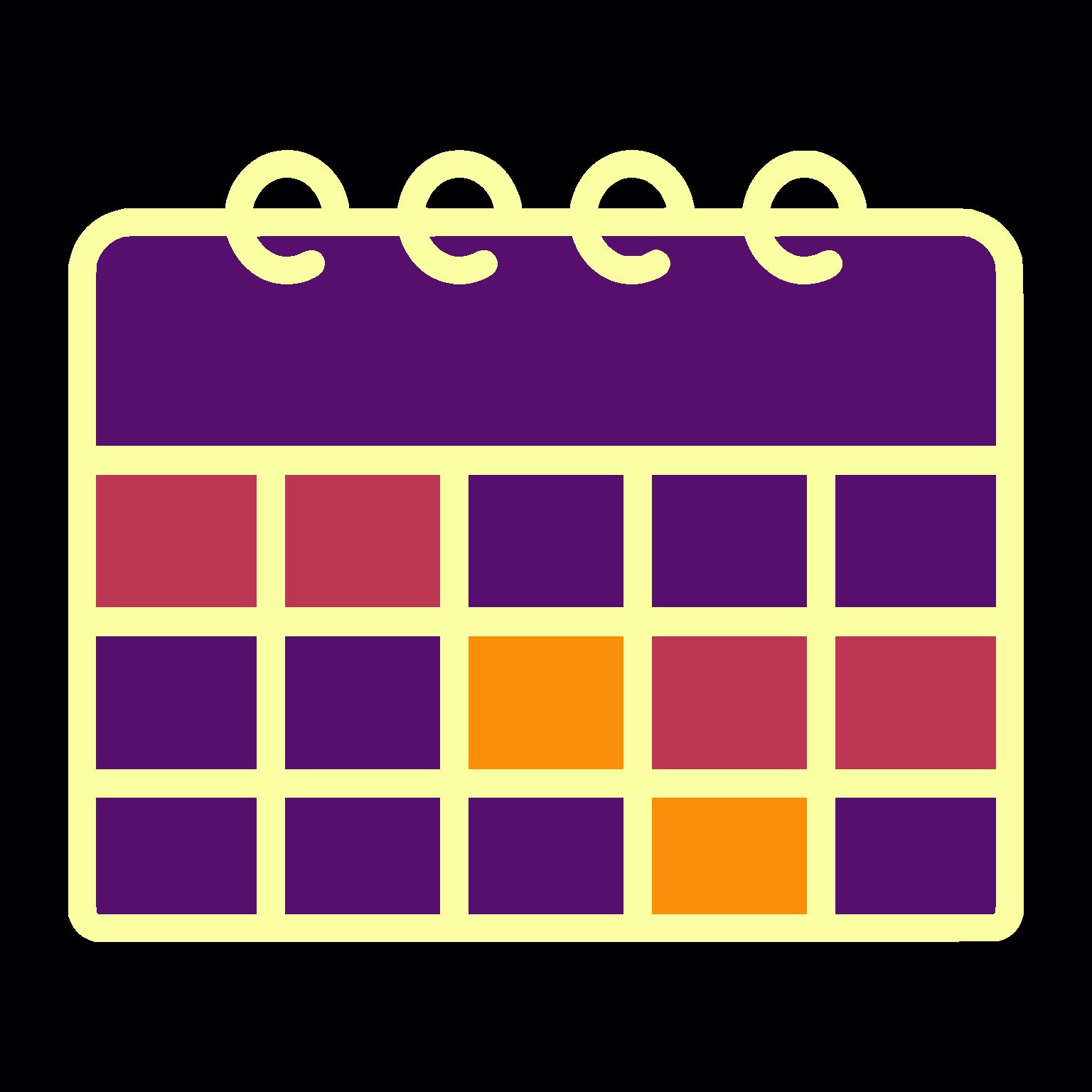}
        \caption{Ours +~\cite{selinger_potrace_2003, pun_vtracer_2025}}
    \end{subfigure}
    \hfill
    \begin{subfigure}[b]{0.135\linewidth}
        \centering
        \includegraphics[width=\linewidth]{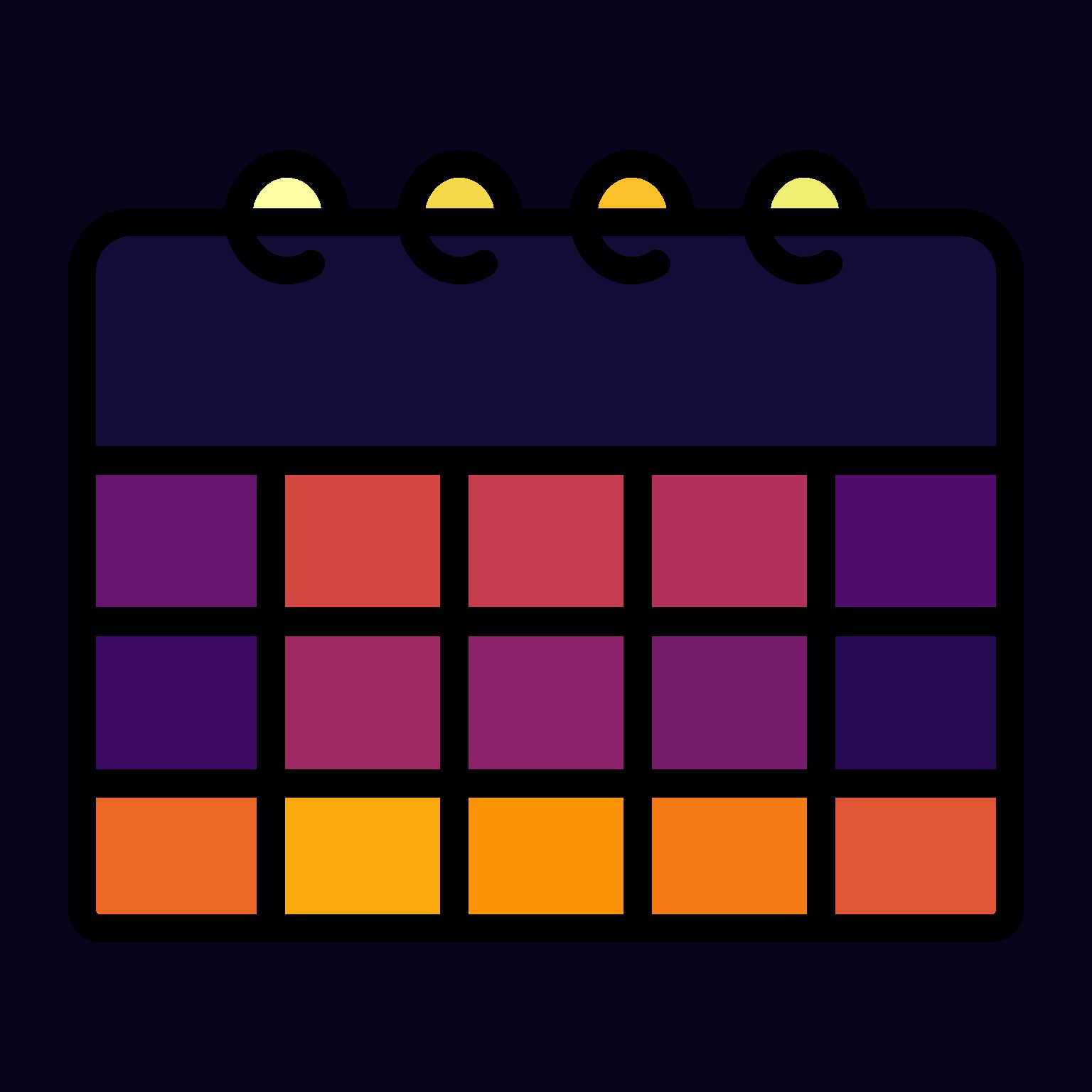}
        \caption{Vtracer~\cite{pun_vtracer_2025}}
    \end{subfigure}
    \hfill
    \begin{subfigure}[b]{0.135\linewidth}
        \centering
        \includegraphics[width=\linewidth]{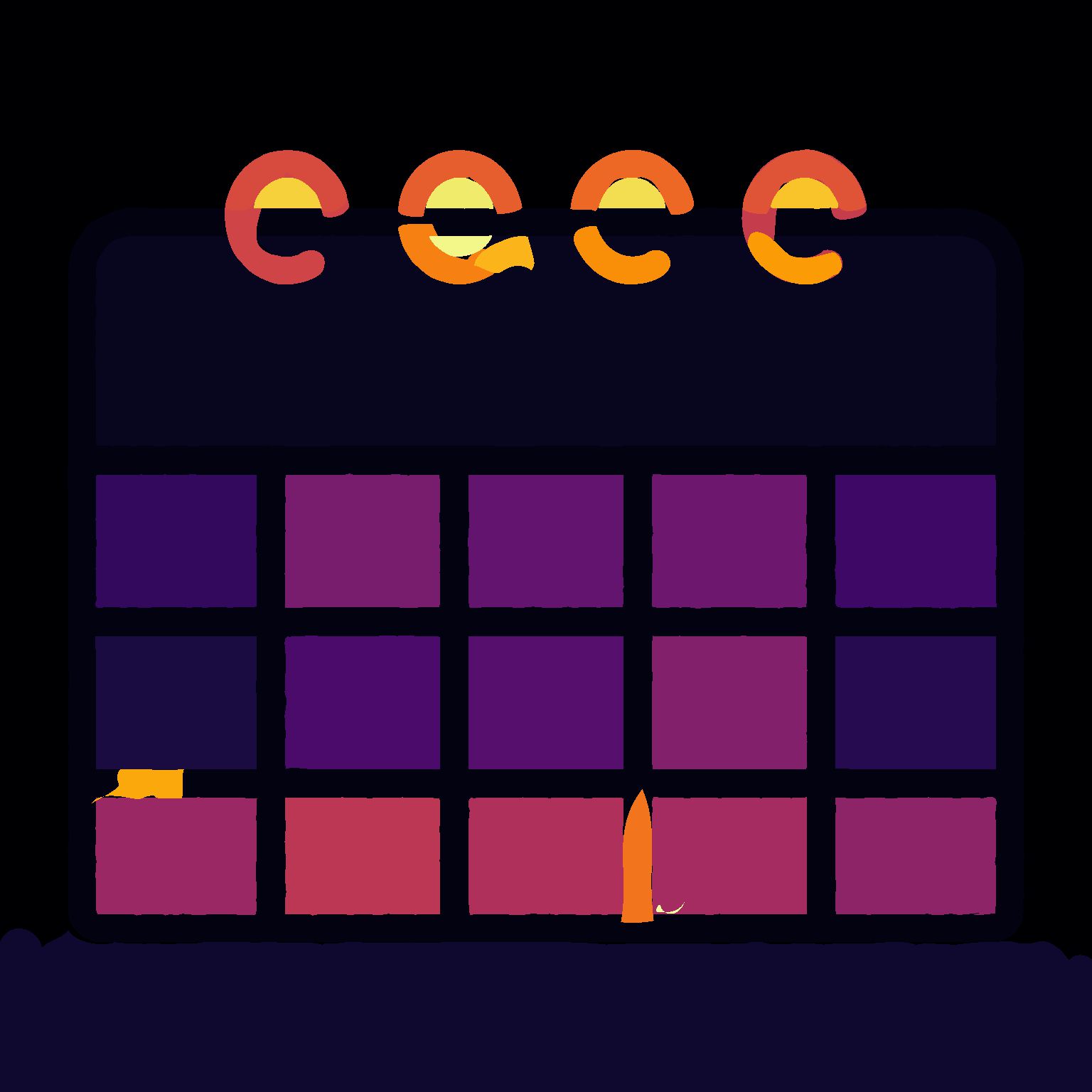}
        \caption{L.I.M.~\cite{zhao_less_2025}}
    \end{subfigure}
    \hfill
    \begin{subfigure}[b]{0.135\linewidth}
        \centering
        \includegraphics[width=\linewidth]{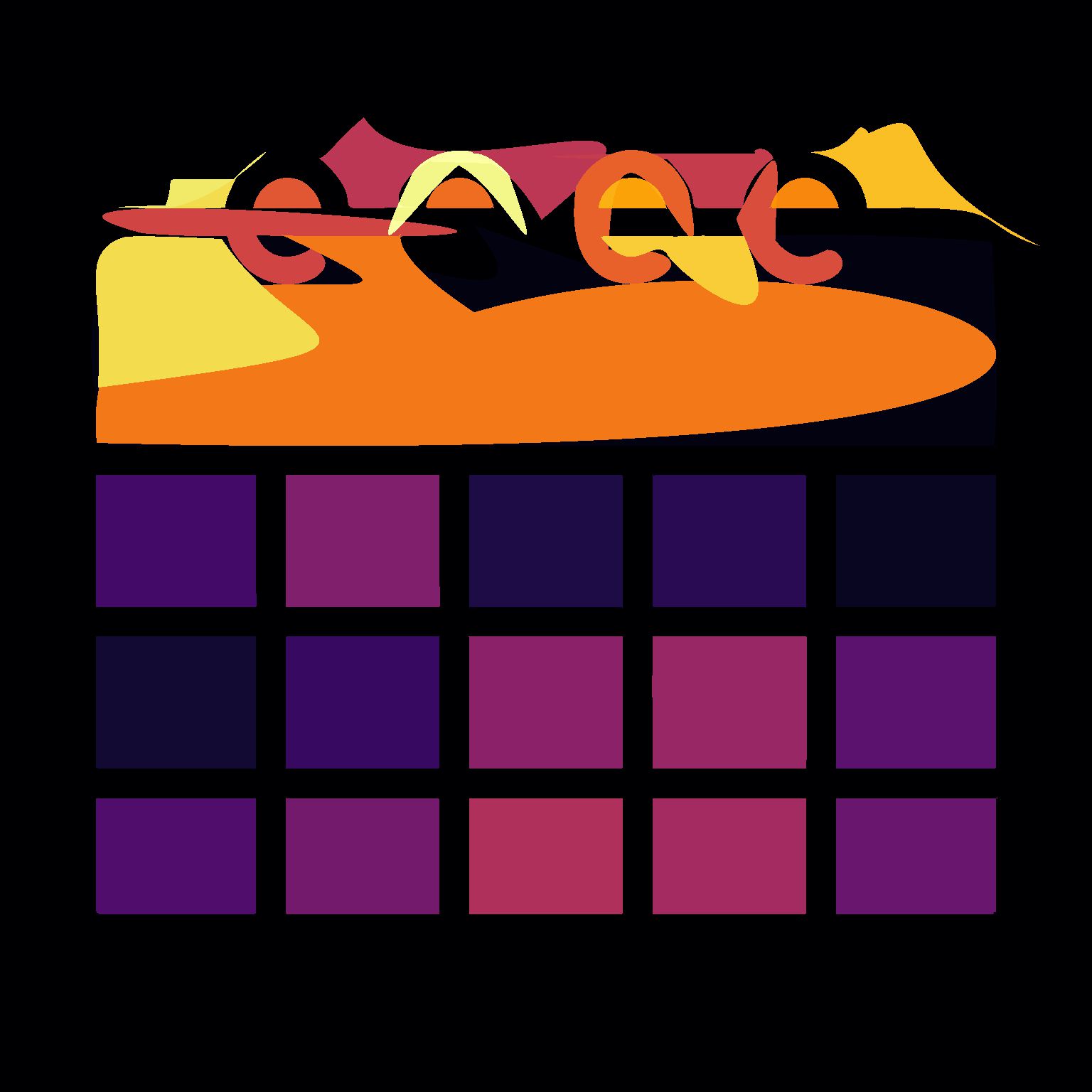}
        \caption{LIVE~\cite{ma_towards_2022}}
    \end{subfigure}
    \hfill
    \begin{subfigure}[b]{0.135\linewidth}
        \centering
        \includegraphics[width=\linewidth]{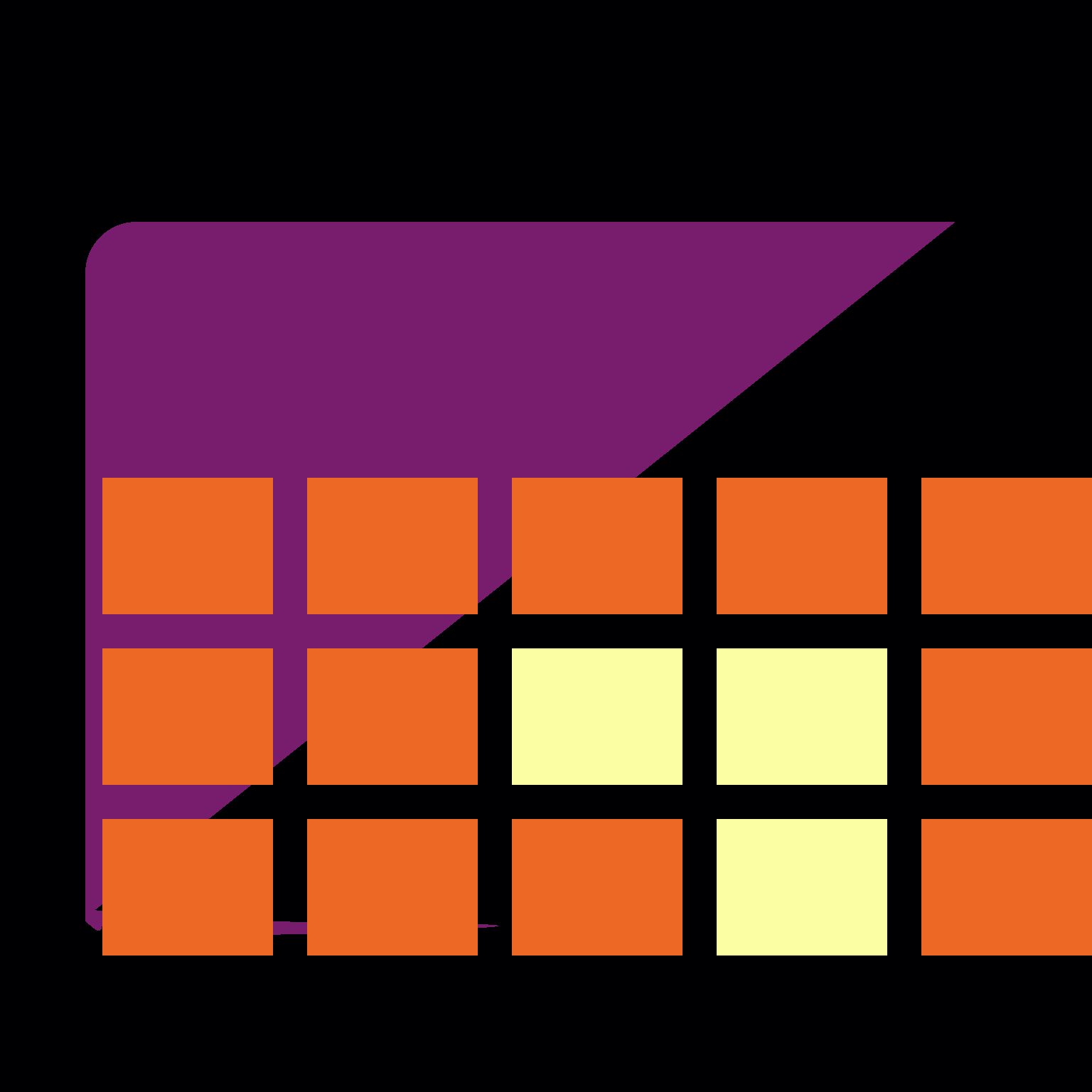}
        \caption{Starvector~\cite{rodriguez_starvector_2025}}
    \end{subfigure}
    \hfill
    \begin{subfigure}[b]{0.135\linewidth}
        \centering
        \includegraphics[width=\linewidth]{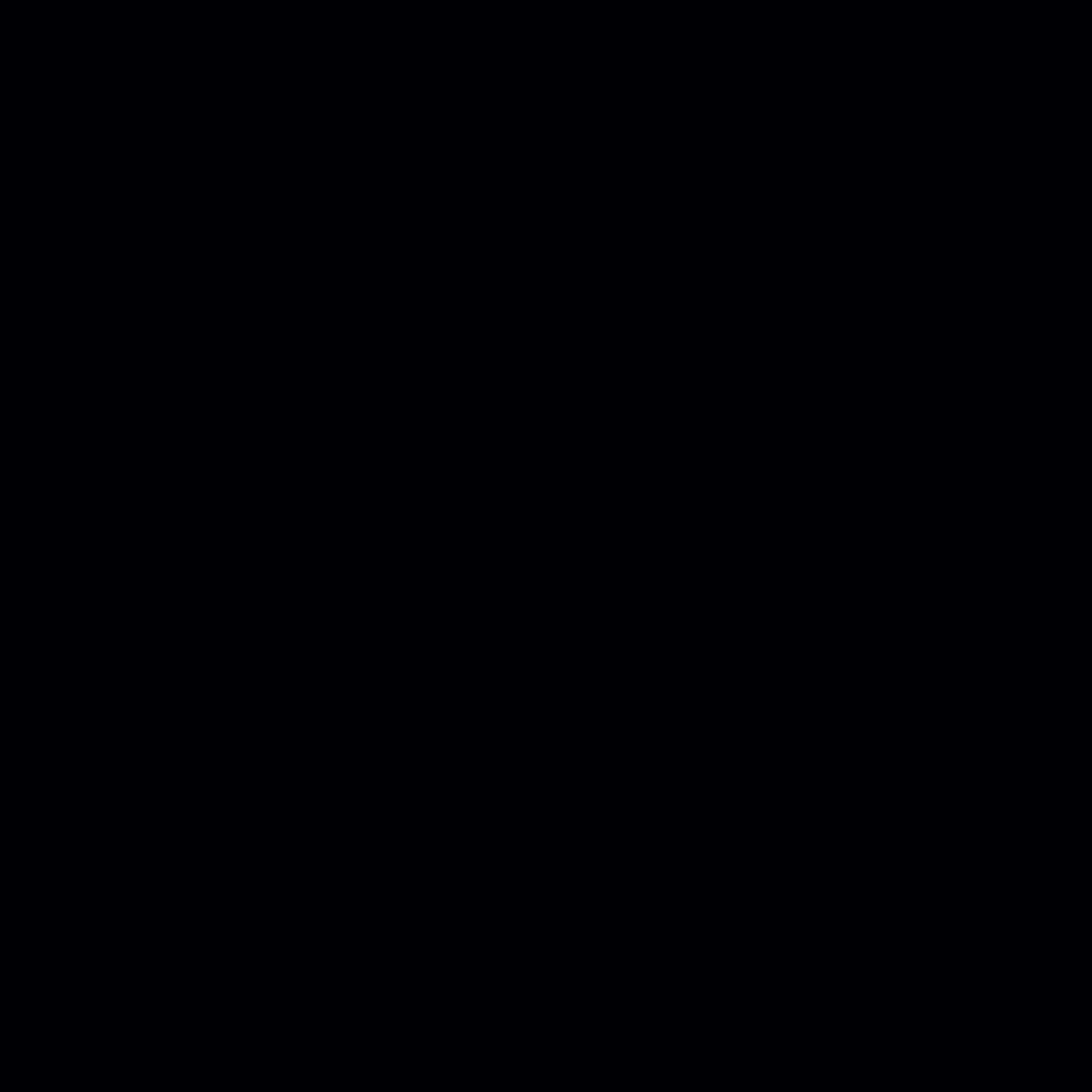}
        \caption{O-SVG~\cite{yang_omnisvg_2025}}
    \end{subfigure} \vspace*{-3mm}
    \caption{\textbf{Image vectorization with illustrator's depth.} 
    Paired with standard vectorization pipelines, our method produces editable, depth-ordered SVGs that closely preserve the structure of the input image. Compared to heuristic, optimization-driven, or learning-based baselines, our approach systematically yields much cleaner layering and higher visual fidelity.
    \vspace*{-3mm}}
    \label{fig:svg_depth}
\end{figure*}

\subsection{Neural Network \& Training}
\label{sec:NN+T}

\paragraph{Model}~Predicting illustrator's depth requires reasoning about object boundaries, occlusion, and grouping.
While distinct from physical depth estimation, this task benefits immensely from the powerful priors learned by state-of-the-art monocular depth estimation (MDE) models. In particular, we find that Depth Pro~\cite{bochkovskii_depth_2025}, built on Dino-v2~\cite{oquab_dinov2_2024} and equipped with a multi-scale encoder, provides a robust feature extractor that allows our model to generalize well from our training set of simple vector graphics to complex, artistic images, as we will demonstrate in~\cref{sec:experiments}. We initialize our model with Depth Pro's pre-trained weights, leveraging its learned understanding of geometry and occlusion as a crucial prior for our task to enable broad generalization.

\paragraph{Scale-invariant loss function}~In natural images, distant objects correspond to large physical depth values, which are inherently more challenging to estimate accurately. Therefore, most MDE models~\cite{yang_depth_2024, bochkovskii_depth_2025, ranftl_towards_2020} learn \emph{inverse} depth values $1/d$, prioritizing the accuracy of foreground objects over distant ones. In contrast, illustrations are composed in a structured, layer-wise manner from background to foreground, where depth values typically range from $1$ to $N$. In this setting, estimating the illustrator’s depth is not inherently harder for background layers than for foreground ones. Instead of learning in disparity space, we thus train our model to predict discrete ground-truth layer indices ($1,...,N$) directly, assigning \emph{equal} importance to all image layers (see ablation studies in~\cref{sec:monocular_exp}). Our primary objective, however, is to recover the correct \emph{relative} ordering of these layers rather than their absolute index values.  To focus the training on this relative structure, and remain robust to the potentially large range of $N$, we adopt a scale-invariant normalization scheme similar to MiDaS~\cite{ranftl_towards_2020}. For any depth map $D$, we compute its median $m$ and mean absolute deviation $s$, and normalize each depth value $d$ as $\hat{d} \!\coloneqq\! {(d\!-\!m)}/{s}$. We then train the network using a Mean Absolute Error (MAE) loss on these normalized maps, \ie, using the loss:\vspace*{-2mm}
\begin{equation}
L_{\scriptscriptstyle\text{MAE}} (D(I), D_\theta(I)) = \overline{| \hat{D}(I) - \hat{D}_\theta(I)|}. \vspace*{-1.5mm} 
\end{equation}

\paragraph{Training}~The network is trained on our SVG dataset using standard training practices, including data augmentation (color jitter, random inversion, random blur) and a cosine learning rate schedule. Additionally, we follow~\cite{bochkovskii_depth_2025} by emplying two distinct learning rates for the encoder (DINO-v2 \cite{oquab_dinov2_2024}) and the CNN-based decoder. Details are provided in Sec.~\ref{sec:experiments}, the Supplementary Material, and the code.

\paragraph{Post-processing}~As our network outputs pixel-wise illustrator's depth estimates, \emph{optional} post-processing can be applied to derive discrete layer indices. Depending on the target application, two common strategies are advisable: (1) direct segmentation of depth values using binning or thresholding, and (2) clustering in RGB space followed by assigning each cluster its median depth. We typically adopt the first strategy for raster image processing (\cref{sec:beyond}), whereas the second is better suited for vectorization tasks (Secs.~\ref{sec:vectorization} and \ref{sec:texttovector}) where inputs typically exhibit color-consistent regions. In the latter case, clusters with similar colors and depths can be further merged to simplify the resulting SVG paths (see \cref{fig:svg_depth}). Notably, even without post-processing, our predicted illustrator's depth maps are visually coherent and structurally clean, see ~\cref{fig:intro_overview,fig:depth_overview,fig:mde_comp,fig:tactile}.

\section{Experiments and Applications}
\label{sec:experiments}
In this section, we outline our training setup in \cref{sec:monocular_exp} and benchmark our model against state-of-the-art monocular depth estimators.  
Then we demonstrate a variety of applications of illustrator's depth. 
First, we embed our trained model into a vectorization pipeline (\cref{sec:vectorization}), which outperforms state-of-the-art methods, and show how it enables a creative, fully editable workflow when paired with generative image models (\cref{sec:texttovector}).
We then showcase diverse raster-based editing tools enhanced by our predicted illustrator's depths (\cref{sec:beyond}), including relief generation for tactile graphics and 
layer-wise decomposition.

\subsection{Predicting Illustrator's Depth}
\label{sec:monocular_exp}

\begin{table}[b]\vspace*{-4mm}
    \centering
    \caption{\textbf{Evaluation of illustrator's depth on MMSVG.}~While models trained for physical depth perform poorly, our method achieves near-perfect layer ordering and low 
    layering error.
    \vspace*{-2mm}}
    \label{tab:mde_comp}
    \resizebox{0.99\linewidth}{!}{
    \begin{tabular}{lccc}
    \toprule
     & Order $\uparrow$ & MAE $\downarrow$ & MSE $\downarrow$ \\
     \midrule
    Depth Pro~\cite{bochkovskii_depth_2025} 
    & \tbest{0.636} & \tbest{1.44} &  \tbest{4.76}
    \\
    Depth Anything-v2~\cite{yang_depth_2024} 
   	& \sbest{0.791} & \sbest{1.16} &  \sbest{3.58}
    \\
    Ours 
    & \best{0.987}  & \best{0.12} & \best{0.26}
    \\
     \bottomrule
    \end{tabular}}
\end{table}

\begin{table}[h]
    \centering
    \caption{\textbf{Impact of key components on illustrator’s depth.}~
    Removing depth prior, data cleaning, or training in disparity space degrades layer-order consistency and accuracy, at times significantly, confirming the contribution of each component to the overall performance of our layer index predictions.
    \vspace*{-2mm}}
    \label{tab:depth_ablation}
    \resizebox{0.99\linewidth}{!}{
    \begin{tabular}{ccc ccc}
    \toprule
       \makecell{Depth prior \\ initialization} &  \makecell{Data \\ cleaning} &  \makecell{Direct index\\ training}  & Order $\uparrow$ & MAE $\downarrow$ & MSE $\downarrow$ \\
     \midrule
     & \color{black}\checkmark & \color{black}\checkmark
    & 0.903 & \tbest{0.51}  & \sbest{1.17}
    \\
    \color{black}\checkmark &  & \color{black}\checkmark
    & \tbest{0.905} & 0.53  & \tbest{1.21}
    \\
    \color{black}\checkmark &  \color{black}\checkmark & 
    & \sbest{0.980} & \sbest{0.50}  & 1.88 
    \\
     \color{black}\checkmark  &  \color{black}\checkmark & \color{black}\checkmark 
    & \best{0.981} & \best{0.16} & \best{0.29}
    \\
     \bottomrule
    \end{tabular}}\vspace*{-4mm}
\end{table}

\paragraph{Training}~As detailed in~\cref{sec:method}, our model is trained on the MMSVG-Illustration dataset~\cite{yang_omnisvg_2025}. Following data cleaning and rasterization to a resolution of $1536\!\times\!1536$, the dataset comprises approximately 100K consistently layered SVG images, with $80\%$ allocated for training and $20\%$ reserved for evaluation. In line with~\cite{zhao_less_2025}, we randomly select 100 images for quantitative analysis --- see Supplementary Material for results on the SVGX-Core dataset and nearest neighbors analysis. Training is done for 40 epochs on 8 Nvidia\textsuperscript{\textregistered\!} A100 GPUs, with a cosine learning rate schedule, a max learning rate of $5\cdot\!10^{-6}$, and a batch size of 8. 

\begin{table*}[ht]
    \centering
    \caption{\textbf{Evaluation of the vectorization.} 
    Here, we test different vectorization methods --- grouped by layering strategies --- on the validation set of the MMSVG dataset.
    Our approach achieves the best combination of layering accuracy, path compactness, and reconstruction fidelity, systematically outperforming heuristic, optimization-based, and data-driven baselines.
    \vspace*{-2mm}
    }
    \label{tab:svg_depth}
    \resizebox{0.99\linewidth}{!}{
    \begin{tabular}{lcccccccc}
    \toprule
      & & \multicolumn{4}{c}{\textit{Layering Quality}} & \multicolumn{3}{c}{\textit{Visual Fidelity}} \\
      Method & Layering Prior  & Order $\uparrow$ & MAE $\downarrow$ & MSE $\downarrow$ & Path Number $\downarrow$ & MSE ($\times10^{-2}$) $\downarrow$ & SSIM $\uparrow$ & LPIPS $\downarrow$ \\
     \midrule
    Vtracer~\cite{pun_vtracer_2025}+Potrace~\cite{selinger_potrace_2003} & Heuristics
    & 0.694 & 1.67 & \tbest{8.68}  & 0.83 & \sbest{0.019} & \best{0.997} & \best{0.005}\\
    Less Is More~\cite{zhao_less_2025} & Heuristics
    & 0.746 & 2.43 & 21.10  & 5.54  & 0.663 & \sbest{0.961} & \sbest{0.043}
    \\
    LIVE~\cite{ma_towards_2022}  & Optimization-based
    & 0.838 & 4.88 & 96.91  & 8.62 & \tbest{0.297} & \tbest{0.946} & \tbest{0.053}
    \\
    Starvector~\cite{rodriguez_starvector_2025} & Data-driven
    & \tbest{0.918} & \tbest{1.52} & 9.75  & \sbest{0.53} & 9.123 & 0.858 & 0.302
    \\
    OmniSVG~\cite{yang_omnisvg_2025} & Data-driven
    & \sbest{0.925} & \sbest{1.31} & \sbest{8.08}  & \tbest{0.54} & 9.997 & 0.830 & 0.317
    \\
    Ours + ~\cite{pun_vtracer_2025,selinger_potrace_2003} & Data-driven
    &\best{0.987} & \best{0.46} & \best{2.09}  & \best{0.16} & \best{0.018} & \best{0.997} & \best{0.005}
    \\
     \bottomrule
    \end{tabular}
   }\vspace*{-3mm}
\end{table*}

\paragraph{Baselines}~We compare our approach with two state-of-the-art monocular depth estimation (MDE) methods, Depth Pro~\cite{bochkovskii_depth_2025} and Depth Anything-v2~\cite{yang_depth_2024}.

\paragraph{Metrics}~We evaluate performance by rendering illustrator's depth maps from ground-truth SVGs as described in \cref{sec:method_depth_data}. 
Since each method produces depth estimates in its own scale, we first normalize all predicted depth maps using the procedure described in \cref{sec:NN+T} prior to computing Mean Squared Error (MSE) and Mean Absolute Error (MAE). While both MSE and MAE assess pixel-wise depth accuracy, many of our target applications  require a \emph{globally consistent layer ordering} rather than \emph{precise} depth values. Therefore, following Zhang et al.~\cite{zhang_monocular_2015}, we further evaluate \emph{depth ordering consistency} by randomly sampling pixel pairs from the ground truth and predictions, and checking whether their relative depth order is preserved (see Supplementary Material for details). The resulting \emph{depth ordering consistency}  metric (abbreviated as Order in Tabs.~\ref{tab:mde_comp}-\ref{tab:svg_depth}) measures the percentage of correctly ordered pixel pairs, providing a complementary measure of global depth consistency.


\paragraph{Results}~While related, physical depth and illustrator’s depth do capture fundamentally different concepts (\cref{fig:depth_overview}). Standard MDE models, trained to predict real-world geometry, struggle to recover correct layer ordering in illustrations (\cref{fig:mde_comp}); our model, purposely trained to infer layer indices, achieves markedly better results, outperforming all baselines by a wide margin (\cref{tab:mde_comp}). Inference takes less than one second on current GPUs as reported in~\cite{bochkovskii_depth_2025}. 

\paragraph{Ablation studies}~We conduct a series of ablation studies to validate our design choices discussed in \cref{sec:method}. As detailed in~\cref{tab:depth_ablation}, both Depth Pro initialization (leveraging a physical depth prior from weights learned on millions of images) and data cleaning (removing inconsistencies and ambiguities in ground-truth layers) boost the depth ordering consistency
quite sharply. 
Although training directly with layer indices $(1,..., N)$ instead of disparity space $(1/d)$ yields comparable global ordering scores, it facilitates a more balanced optimization between foreground and background layers: this results in better depth transitions and a clear advantage across \emph{all} evaluation metrics; see the Supplementary Material for additional qualitative evaluations.


\subsection{Vectorization}
\label{sec:vectorization}
Image vectorization, which consists in converting raster images to vector graphics, is a particularly straightforward application of illustrator's depth.

\paragraph{Pipeline}~Our model integrates seamlessly into existing vectorization pipelines such as VTracer~\cite{pun_vtracer_2025}, where we replace area-based sorting heuristics with our predicted illustrator's depth. We first compute color clusters, sort them using our layer index prediction, inpaint layers to fill holes and bridge gaps (with, e.g., Scikit-Image~\cite{van2014scikit}), before vectorizing each layer with Potrace~\cite{selinger_potrace_2003}.
The whole process, including our illustrator's depth prediction, only takes seconds. 

\paragraph{Baselines}~We benchmark our pipeline against key state-of-the-art approaches, based on simple area heuristics (VTracer~\cite{pun_vtracer_2025} combined with Potrace~\cite{selinger_potrace_2003} including the same cleanups
we use in our method for fairness) or more advanced cluster-sorting strategies (Less Is More~\cite{zhao_less_2025}), optimization methods (LIVE~\cite{ma_towards_2022}), and LLM-based tools (StarVector~\cite{rodriguez_starvector_2025}, OmniSVG~\cite{yang_omnisvg_2025}). 

\paragraph{Metrics}~Vectorization demands both compactness and accuracy for best editability. We thus measure \emph{layering quality} using the depth ordering consistency (Order), mean squared error (MSE), and mean absolute error (MAE), as well as path count errors $|N\!-\!\tilde{N}|/N$ to compare the number of paths in ground-truth ($N$) vs. reconstructed ($\tilde{N}$) SVGs.
We then evaluate \emph{visual fidelity} by measuring the rasterized output compared the input using MSE in RGB space, Structural Similarity Index Measure (SSIM)~\cite{zhou_wang_image_2004}, and Learned Perceptual Image Patch Similarity (LPIPS)~\cite{zhang_unreasonable_2018}.


\paragraph{Results}~Although most vectorization methods produce outputs that look quite close to the input raster images, visualizing their layer indices in false colors 
reveals substantial differences in layering quality (\cref{fig:svg_depth}). 
Methods relying on heuristics such as VTracer~\cite{pun_vtracer_2025} and Less is More~\cite{zhao_less_2025} frequently misorder layers; for instance,  spiral binding holes in the calendar in \cref{fig:svg_depth} are incorrectly positioned on top despite belonging to the background. Optimization-based LIVE~\cite{ma_towards_2022} introduces spurious layers and shapes, while LLM-based approaches~\cite{rodriguez_starvector_2025, yang_omnisvg_2025} often fail (sometimes, spectacularly) to achieve full reconstruction. 
In contrast, our pipeline is able to faithfully reconstruct the input while producing layer indices close to the ground truth.
Additionally, quantitative results from~\cref{tab:svg_depth} confirm these observations: 
our method matches VTracer's reconstruction fidelity while outperforming all SOTA competitors in layer-index accuracy. 
Interestingly, our layering evaluation reveals a clear divide between methods excelling at reconstruction but weak in layering (VTracer, Less is More) and those with opposite strengths (Starvector, OmniSVG).
Our approach thus combines the power of traditional vectorizers with the quality of data-driven layer index prediction, enabling state-of-the-art performance on both fronts.

\subsection{Text-to-Vector-Graphics Generation}
\label{sec:texttovector}
The creation of high-quality vector graphics remains a challenging problem. Direct generation techniques, such as those employing Score Distillation Sampling (SDS)~\cite{zhang_text--vector_2024, polaczek_neuralsvg_2025} or Large Language Models (LLMs)~\cite{rodriguez_starvector_2025, rodriguez_rendering-aware_2025, yang_omnisvg_2025}, have not yet matched the visual fidelity achieved by state-of-the-art text-to-image generative models. Here again, our illustrator's depth neural prediction can dramatically help in obtaining high-quality editable illustrations.

\paragraph{Pipeline}~Leveraging recent advances in high-quality image generation~\cite{labs_flux1_2025, google_gemini_2025}, we first generate vector-style raster images (prompts are detailed in the Supplementary Material). These raster images are subsequently transformed into structured, editable, and layered SVG using our specialized vectorization pipeline described in \cref{sec:vectorization}.

\begin{figure}[h]
    \centering
    \begin{subfigure}[b]{0.49\linewidth}
        \centering
        \includegraphics[width=\linewidth]{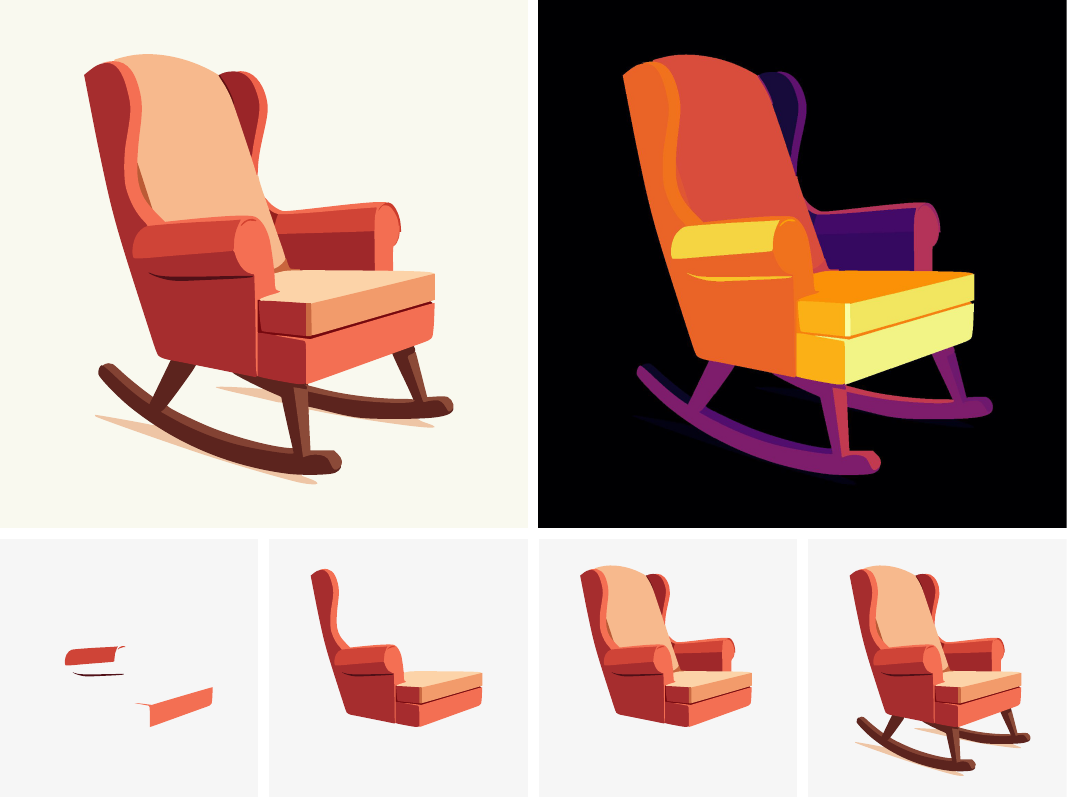}
    \end{subfigure}
    \hfill
    \begin{subfigure}[b]{0.49\linewidth}
        \centering
        \includegraphics[width=\linewidth]{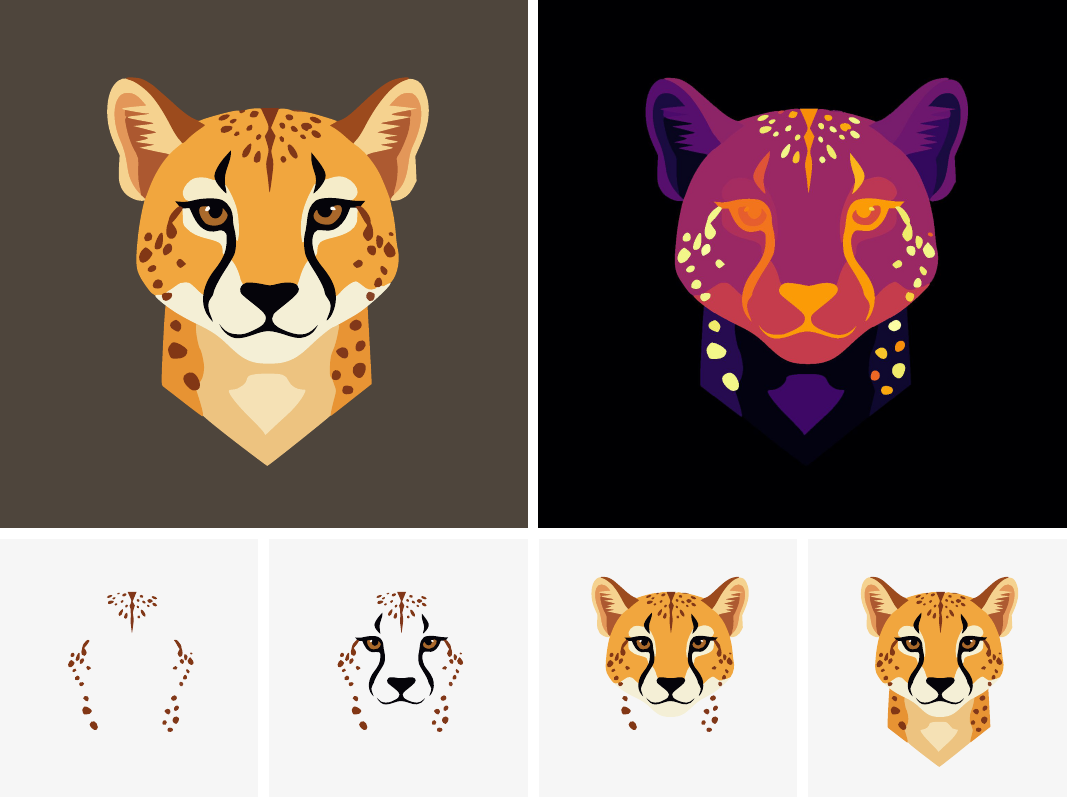}
    \end{subfigure}
    \caption{
    \textbf{Vector graphics generation.} 
    By augmenting text-to-image diffusion models like Flux~\cite{labs_flux1_2025} with illustrator’s depth, generated images can be automatically transformed into editable vector graphics. Layers (bottom, displayed from front to back) facilitate intuitive manipulation of individual elements.
    }
    

    \label{fig:t2v_comparison}
\end{figure}

\begin{figure}[!t]
    \centering
    \includegraphics[width=\linewidth]{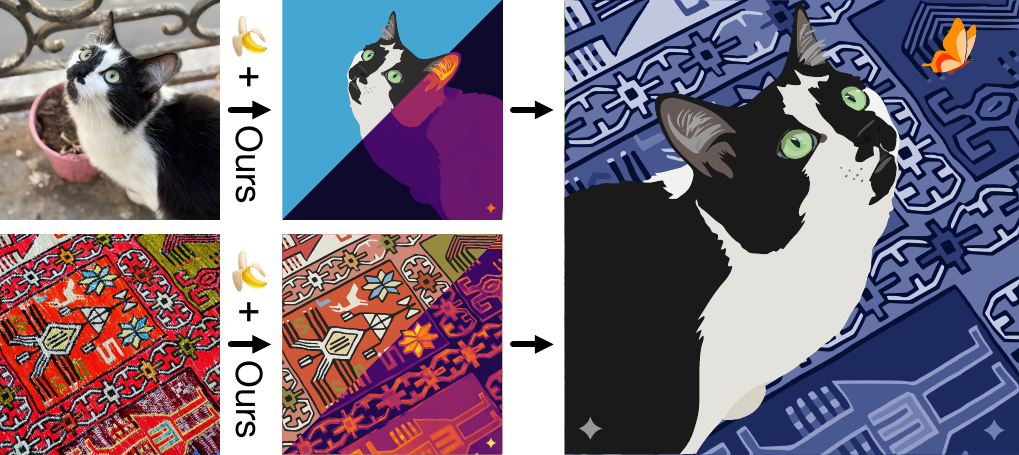}
    \caption{\textbf{Illustrator's depth in generative workflows.} 
    Starting from a cellphone photo and a rug texture (left), a pipeline based on Nano Banana~\cite{google_gemini_2025} and illustrator's depth synthesizes a vector-graphics illustration and converts it into a layered SVG, supporting depth-aware editing such as recoloring and object insertion.
    \vspace*{-2mm}
    }
    
    \label{fig:t2vec_workflow}
\end{figure}

\paragraph{Results}~ \cref{fig:t2v_comparison} presents examples generated via Flux~\cite{labs_flux1_2025} and postprocessed with illustrator's depth. The resulting SVG illustrations exhibit high visual complexity and coherent layer organization, facilitating the intuitive grouping and editing of individual elements (see supplementary video). Our vectorization can be similarly integrated to Nano Banana~\cite{google_gemini_2025} to offer a more advanced, multi-stage generative workflow as illustrated in \cref{fig:t2vec_workflow}. 
We also show comparisons with Neural Path Representation~\cite{zhang_text--vector_2024}, NeuralSVG~\cite{polaczek_neuralsvg_2025}, and LayerTracer~\cite{song_layertracer_2025}, in the Supplementary Material.

\begin{figure}[b] \vspace*{-3mm}
    \centering
    \begin{subfigure}[b]{0.325\linewidth}
        \centering
        \includegraphics[width=\linewidth]{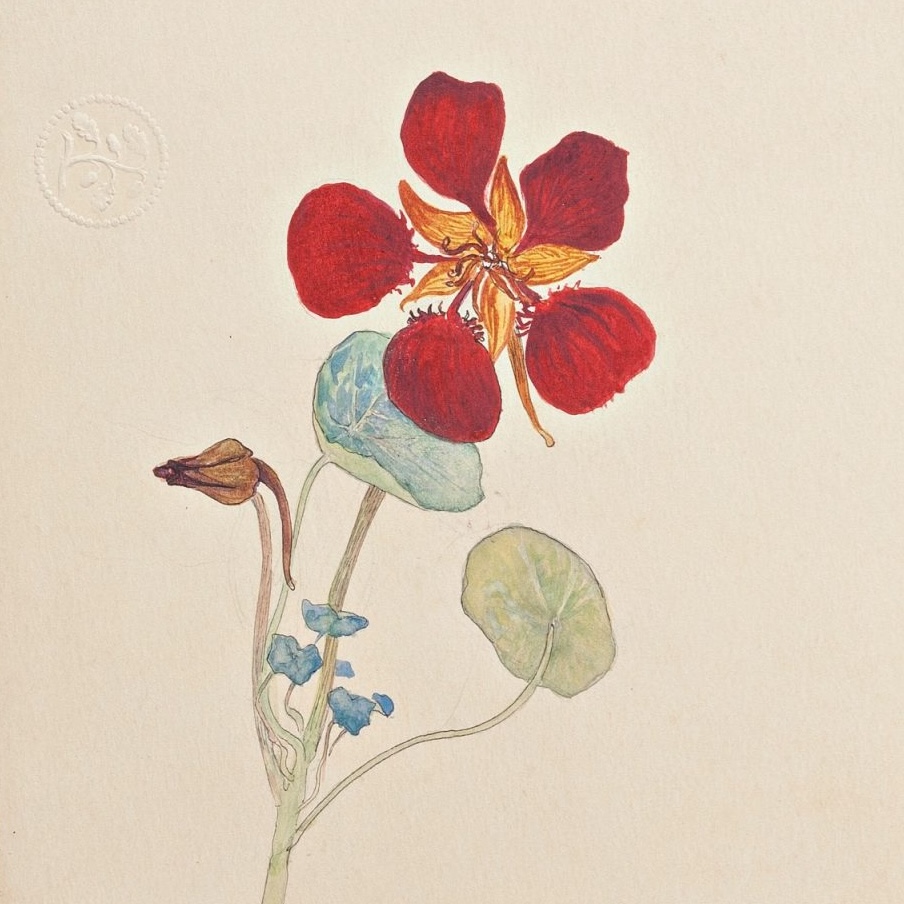}
    \end{subfigure}
    \hfill
    \begin{subfigure}[b]{0.325\linewidth}
        \centering
        \includegraphics[width=\linewidth]{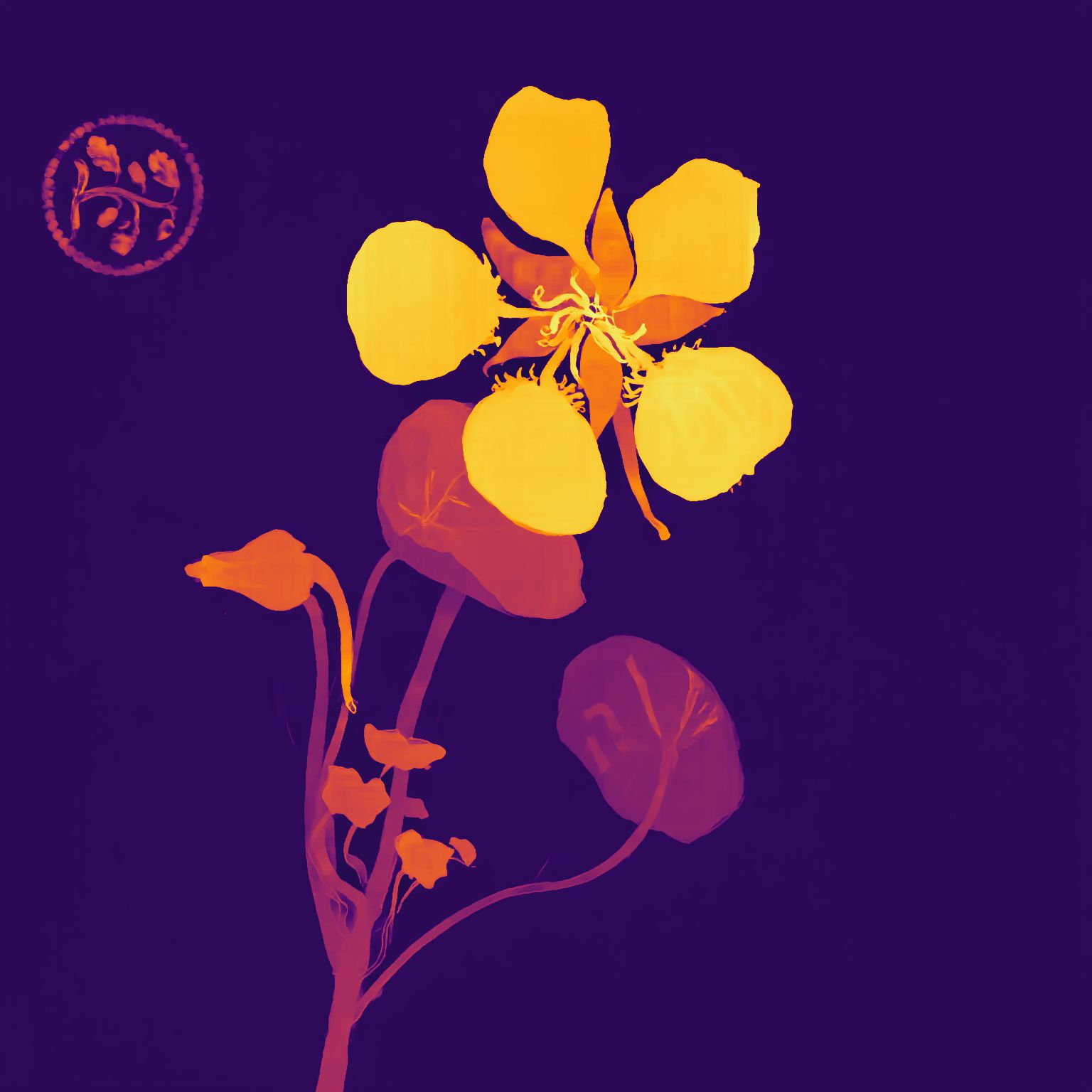}
    \end{subfigure}
    \hfill
    \begin{subfigure}[b]{0.325\linewidth}
        \centering
        \includegraphics[width=\linewidth]{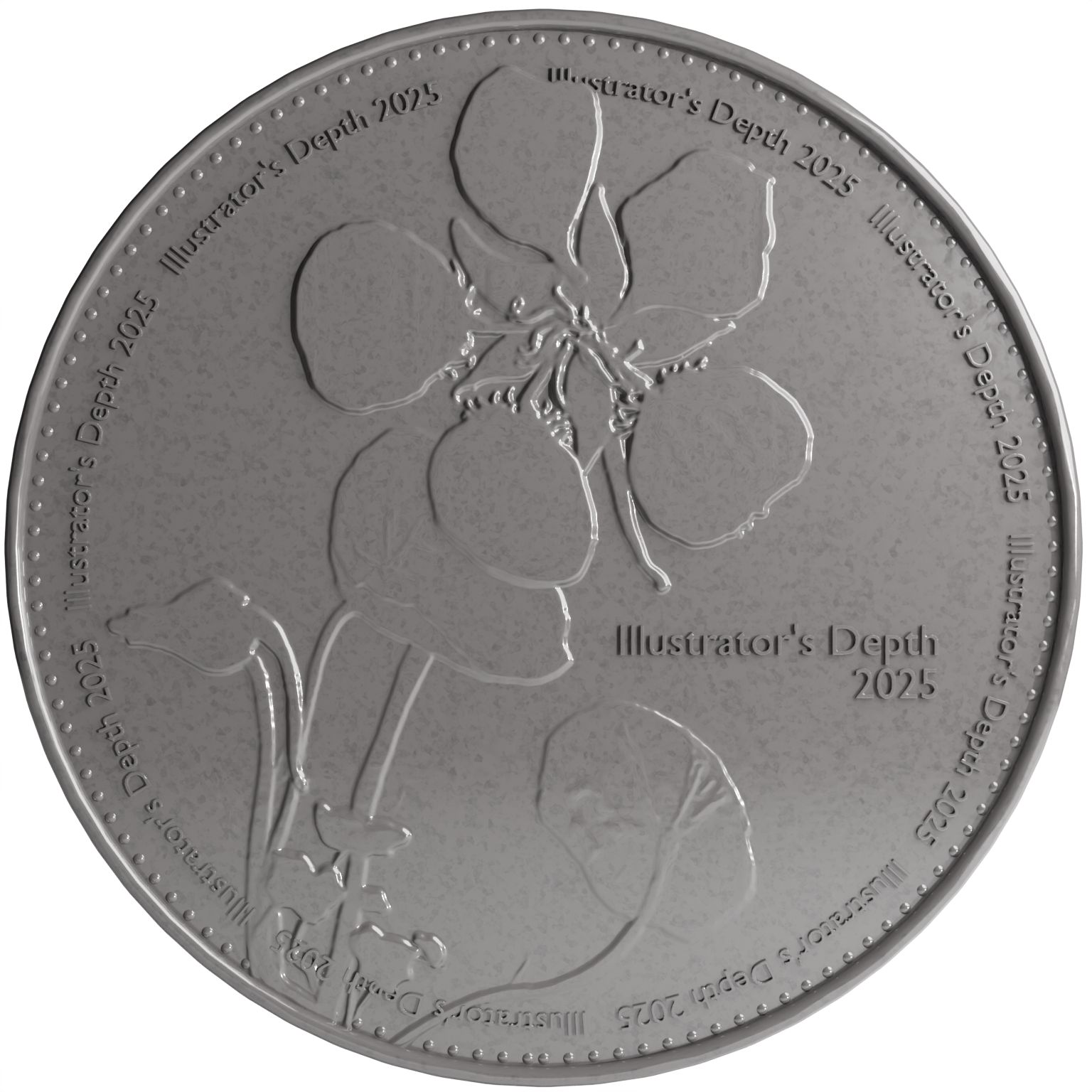}
    \end{subfigure}
    \centering
    \begin{subfigure}[b]{0.325\linewidth}
        \centering
        \includegraphics[width=\linewidth]{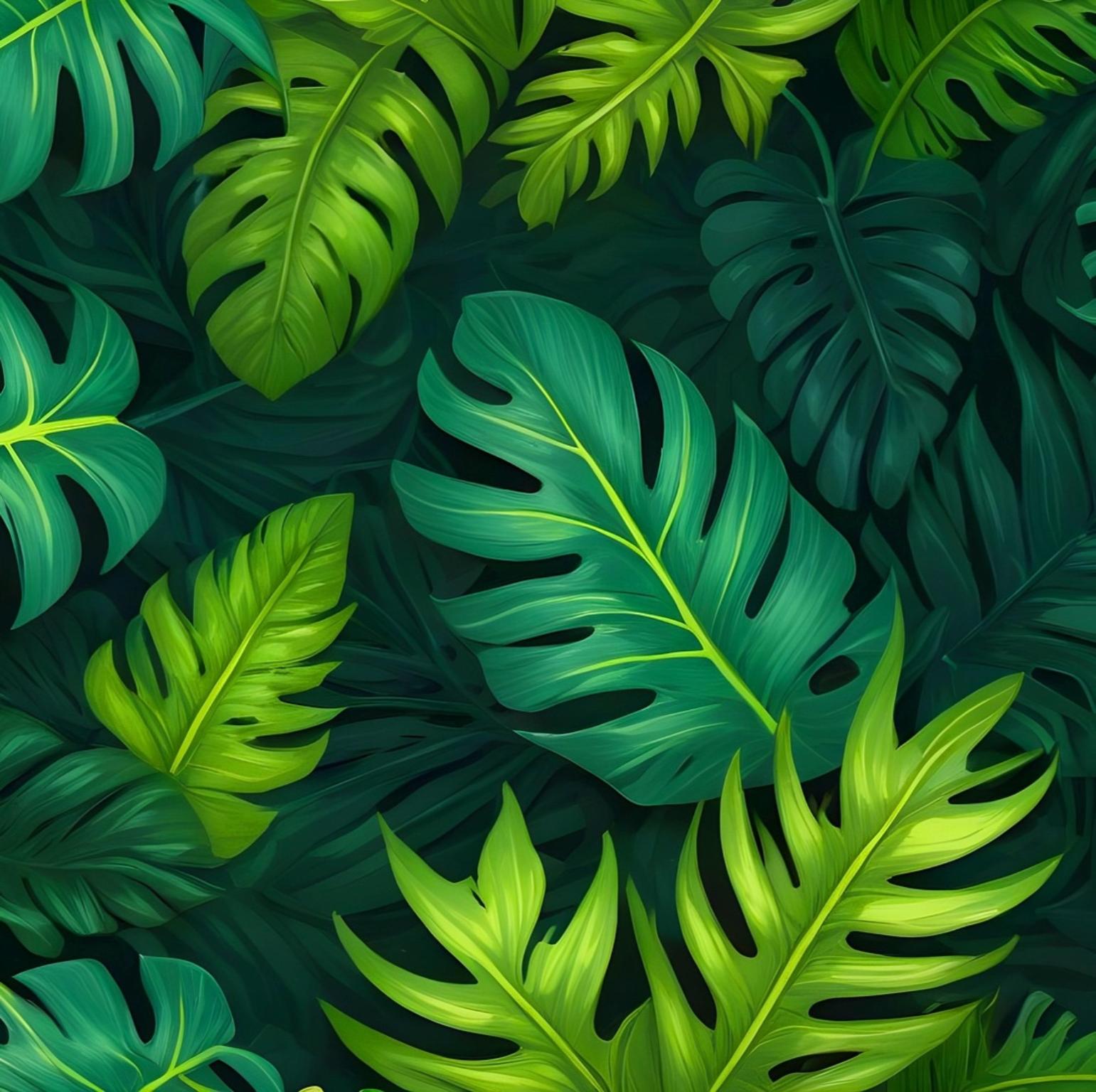}
    \end{subfigure}
    \hfill
    \begin{subfigure}[b]{0.325\linewidth}
        \centering
        \includegraphics[width=\linewidth]{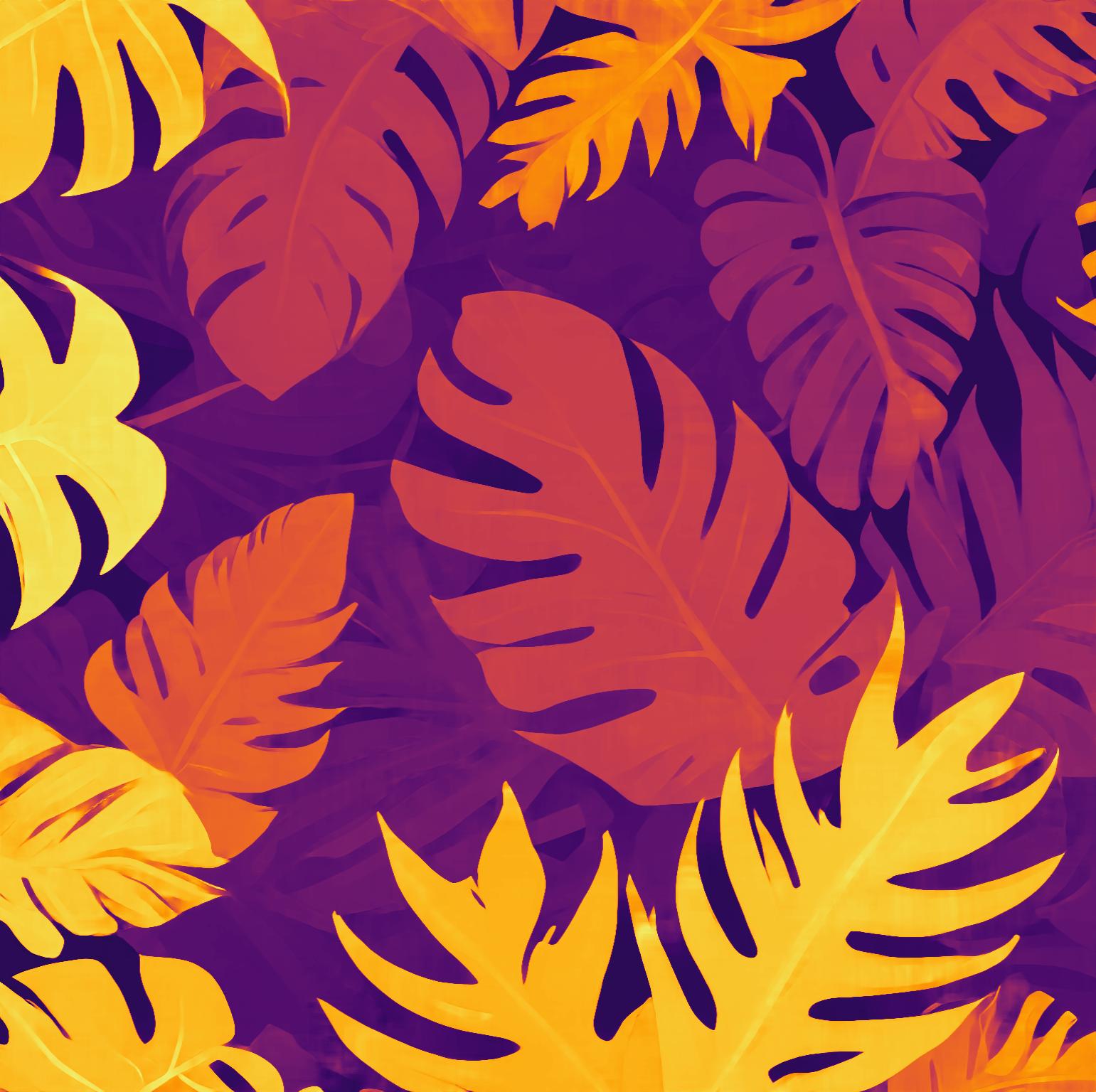}
    \end{subfigure}
    \hfill
    \begin{subfigure}[b]{0.325\linewidth}
        \centering
        \includegraphics[width=\linewidth]{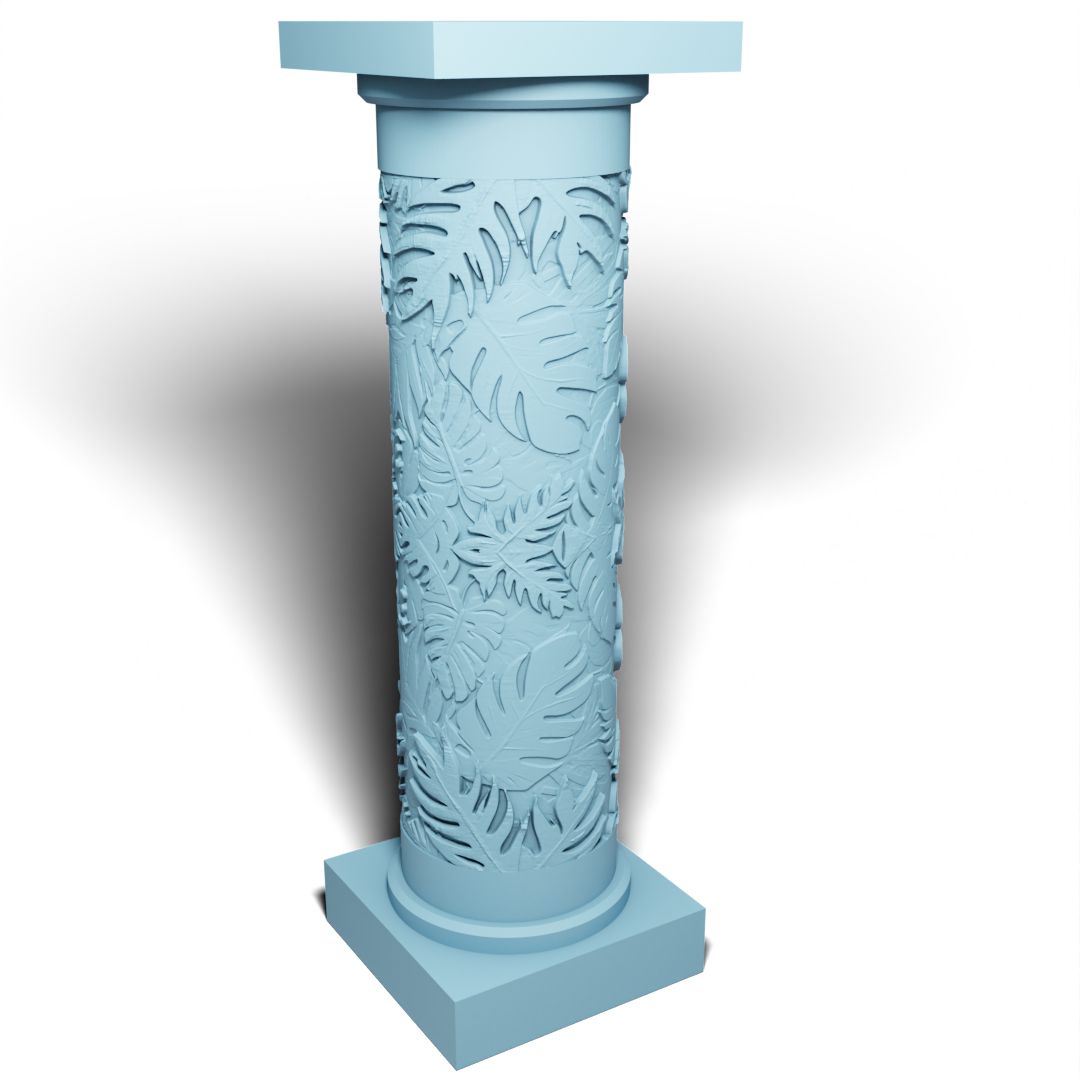}
    \end{subfigure}
    
    \centering
    \begin{subfigure}[b]{0.325\linewidth}
        \centering
        \includegraphics[width=\linewidth]{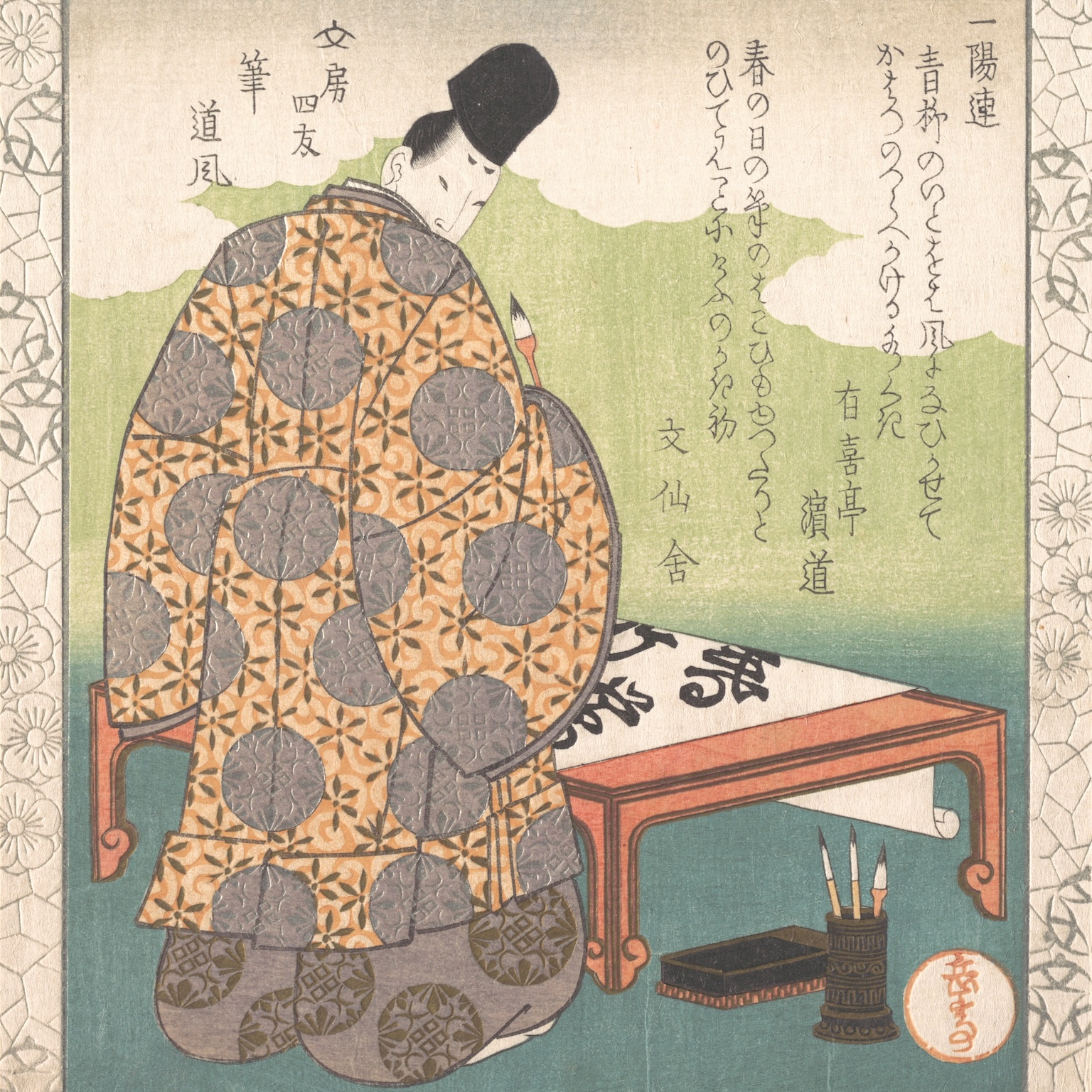}
        \caption{Input Image}
    \end{subfigure}
    \hfill
    \begin{subfigure}[b]{0.325\linewidth}
        \centering
        \includegraphics[width=\linewidth]{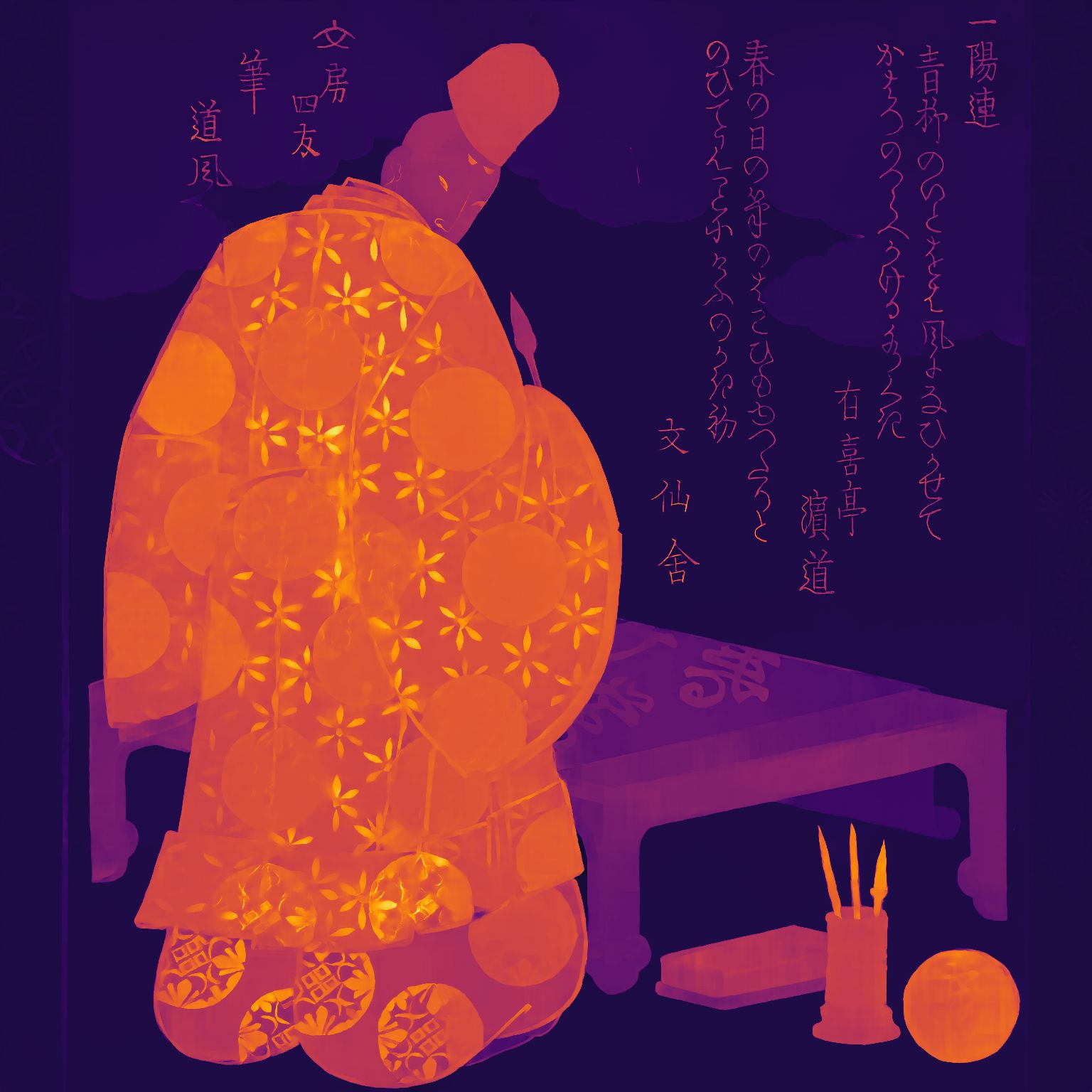}
        \caption{Illustrator's Depth}
    \end{subfigure}
    \hfill
    \begin{subfigure}[b]{0.325\linewidth}
        \centering
        \includegraphics[width=\linewidth]{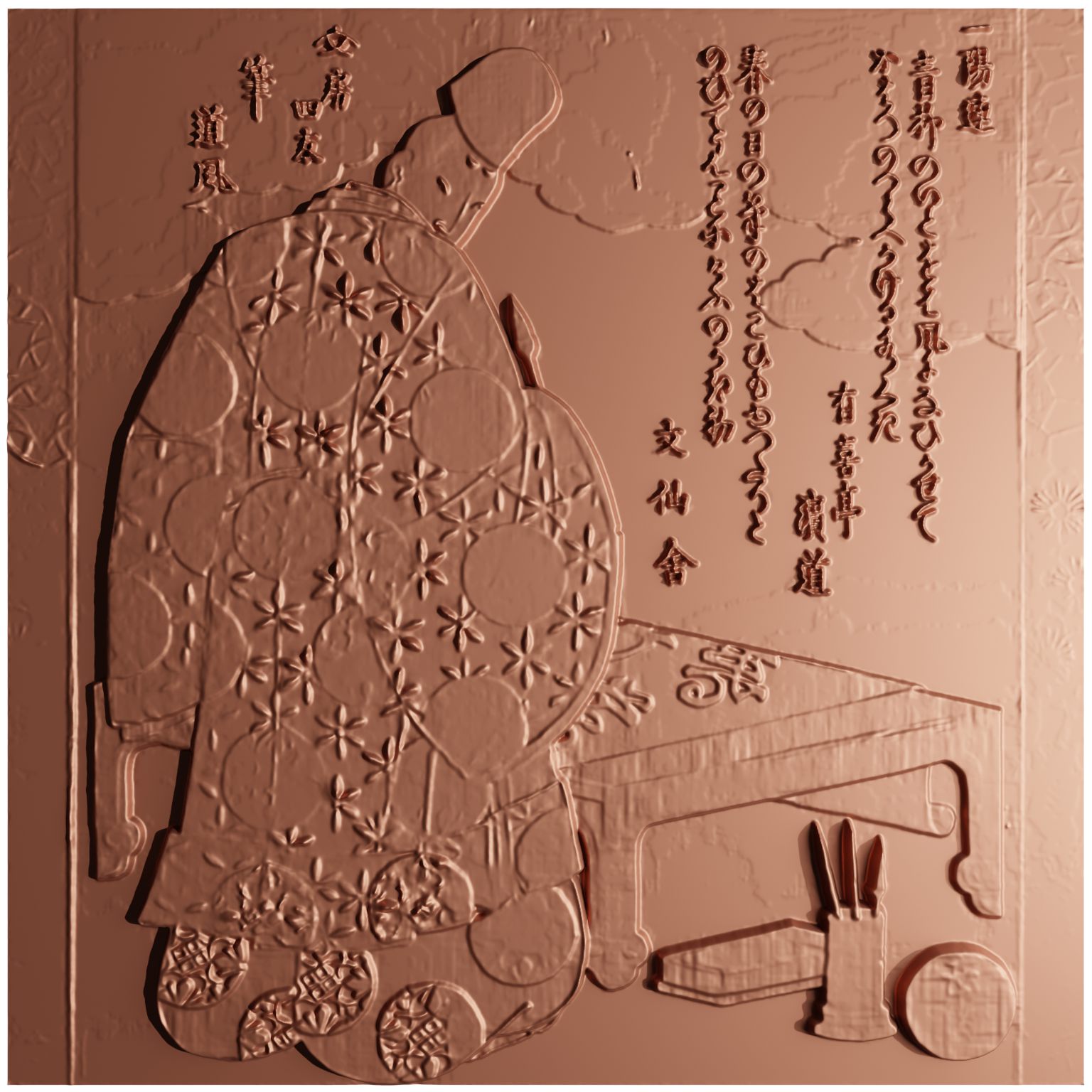}
        \caption{Relief, 3D rendering}
    \end{subfigure}
    \vspace*{-6mm}
    \caption{\textbf{Automatic relief generation from single images.} 
    With no manual intervention, illustrator’s depth (middle) can be converted into 3D surfaces by interpreting predicted depth as elevation. The resulting meshes, shown on the right, demonstrate how images can be transformed into tactile or printable reliefs.
    }
    \label{fig:tactile}
\end{figure}

\subsection{Beyond Vector Graphics}
\label{sec:beyond}

Despite being trained exclusively on depth data generated from simple SVG images, our model demonstrates a remarkable ability to generalize beyond this narrow scope. 
It successfully infers useful illustrator's depths across highly diverse inputs, ranging from simple illustrations to complex artistic renderings, most likely due to our use of pretrained priors~\cite{bochkovskii_depth_2025, oquab_dinov2_2024} learned from millions of images. While quantitative evaluation of this generalization is difficult in the absence of ground truth layerings (we provide an attempt in 
our Supplementary Material), we now showcase two practical applications leveraging this generalization ability.
Additional qualitative results and discussions of failure cases are provided in~Figs.\ref{fig:intro_overview} \& \ref{fig:depth_overview}, and in the Supplementary Material.

\subsubsection{Automatic Relief Generation From a Single Image}
\label{sec:tactile}

\paragraph{Task}~Relief is a sculptural method where elements remain attached to a solid background to give the impression that the sculpture has been raised above the background. Bas-relief, a shallow form of this technique, is widely applied, from coinage to architectural ornament~\cite{zhang_computer-assisted_2019}. Current methods for generating 3D reliefs from 2D images are fundamentally limited by their reliance on user-defined depth ordering~\cite{reichinger_high-quality_2011}. We eliminate this user interaction entirely by leveraging the fully automated output of our model.

\paragraph{Pipeline}~Given an input image, our system first generates a pixel-wise illustrator's depth map $d_\theta(i,j)$. This depth is then directly used to build a triangulated surface by transforming each pixel into a vertex with 3D coordinates $(i,j, d_\theta(i,j))$, and triangulating adjacent vertices.

\paragraph{Results}~The resulting mesh easily integrates into any 3D application as illustrated in~\cref{fig:tactile}. Crucially, our illustrator's depth transforms flat paintings into 3D objects without any manual annotation, offering an alternative, intuitive, and tangible interaction with works of art.

\subsubsection{Depth-Based Editing}
\label{sec:depth-based-editing}

\begin{figure}[b] \vspace*{-4mm}
    \centering
    \begin{subfigure}[b]{0.24\linewidth}
        \centering
        \includegraphics[width=\linewidth]{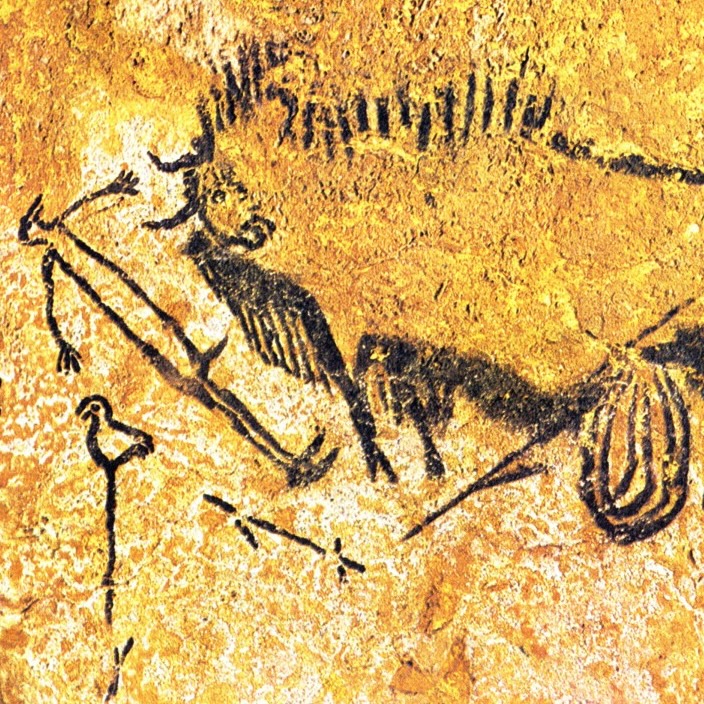}
    \end{subfigure}
    \hfill
    \begin{subfigure}[b]{0.24\linewidth}
        \centering
        \includegraphics[width=\linewidth]{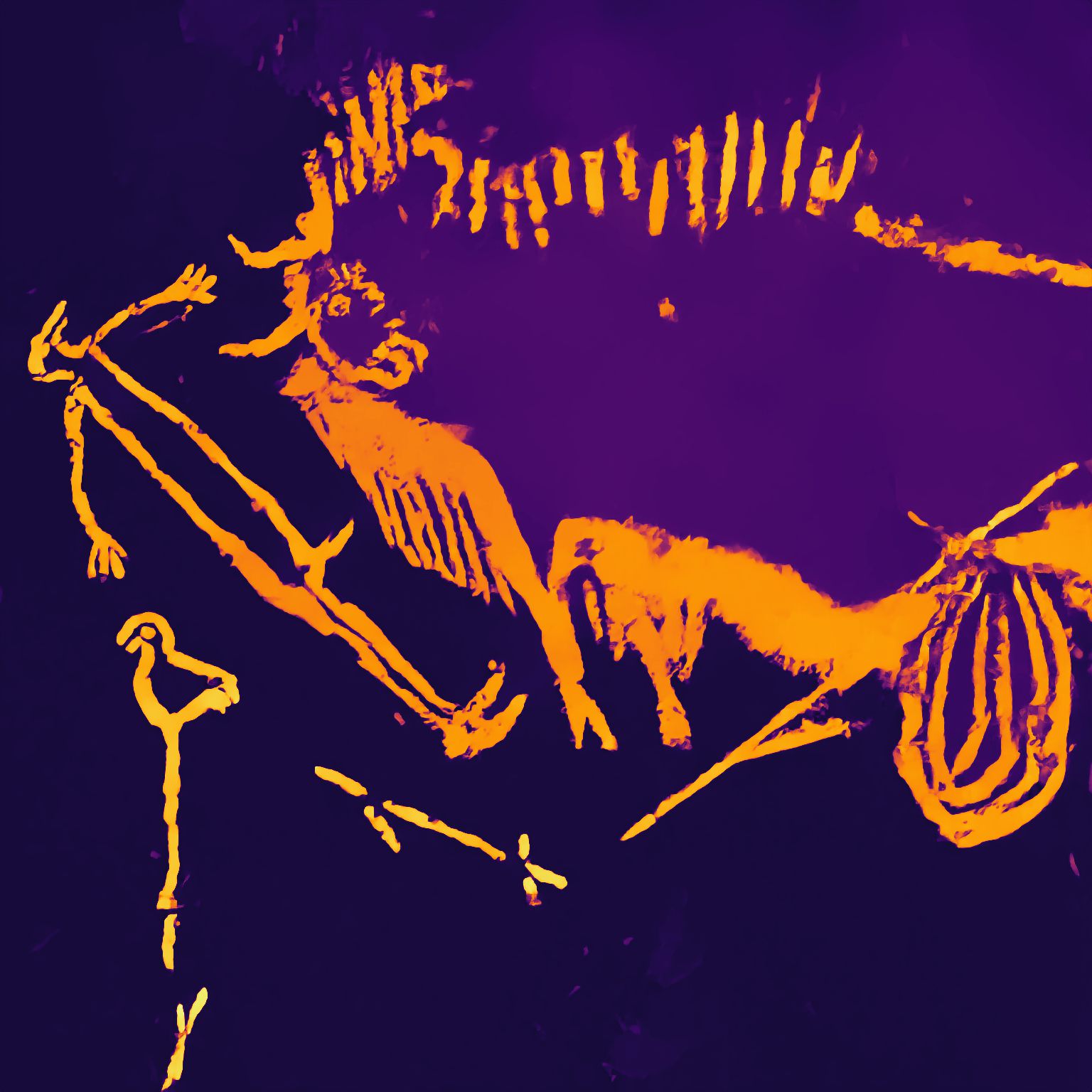}
    \end{subfigure}
    \hfill
    \begin{subfigure}[b]{0.24\linewidth}
        \centering
        \includegraphics[width=\linewidth]{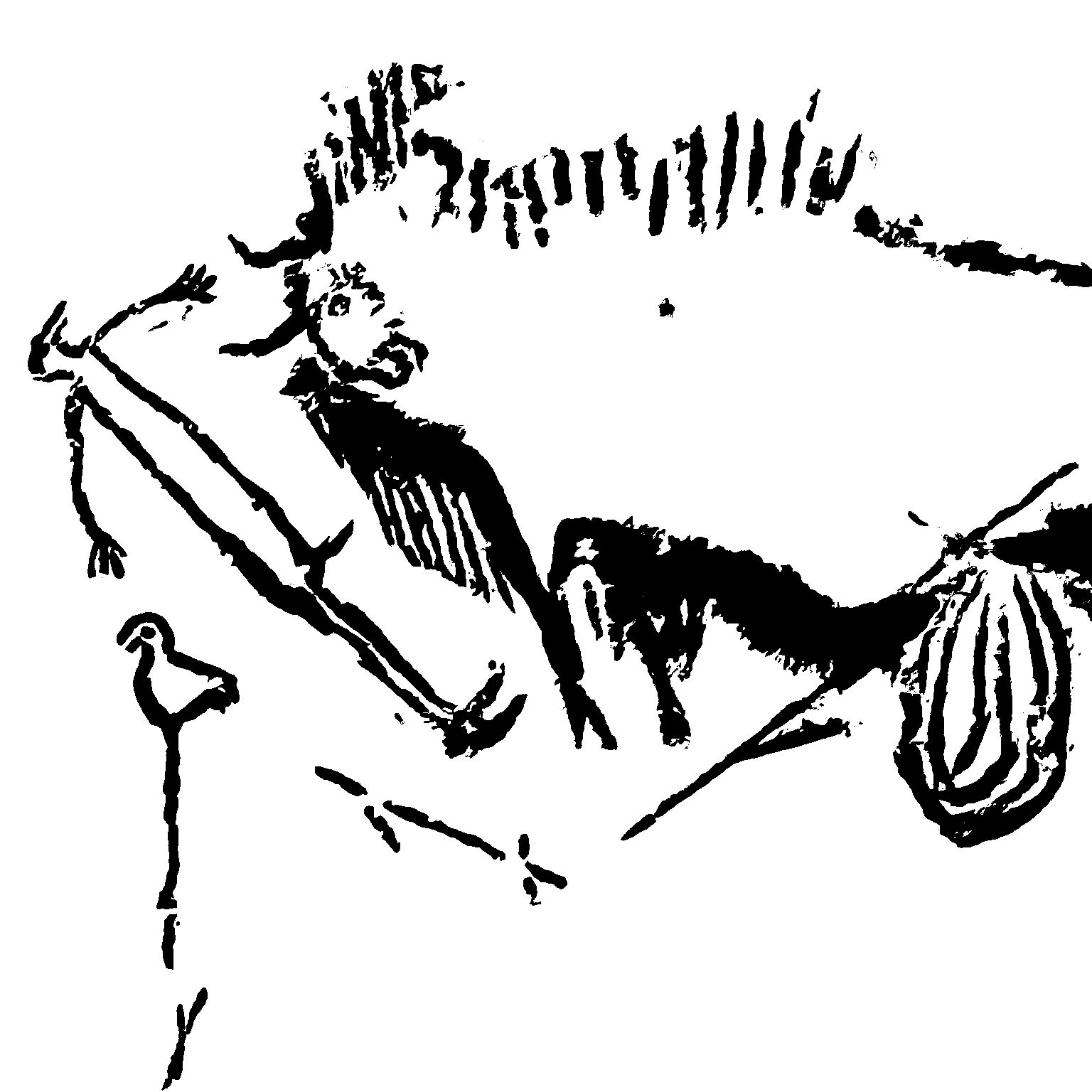}
    \end{subfigure}
    \hfill
    \begin{subfigure}[b]{0.24\linewidth}
        \centering
        \includegraphics[width=\linewidth]{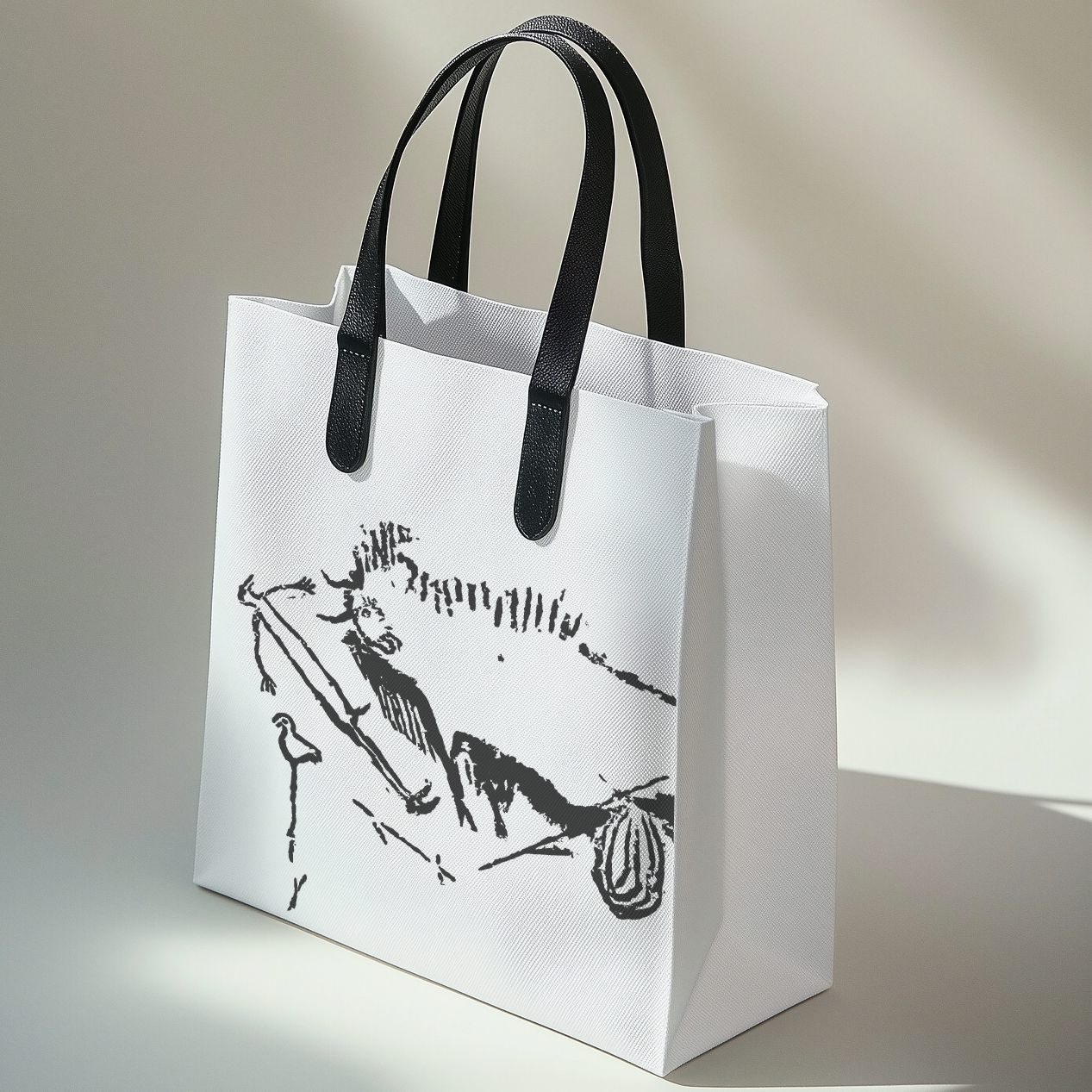}
    \end{subfigure}
    \centering
    \begin{subfigure}[b]{0.24\linewidth}
        \centering
        \includegraphics[width=\linewidth]{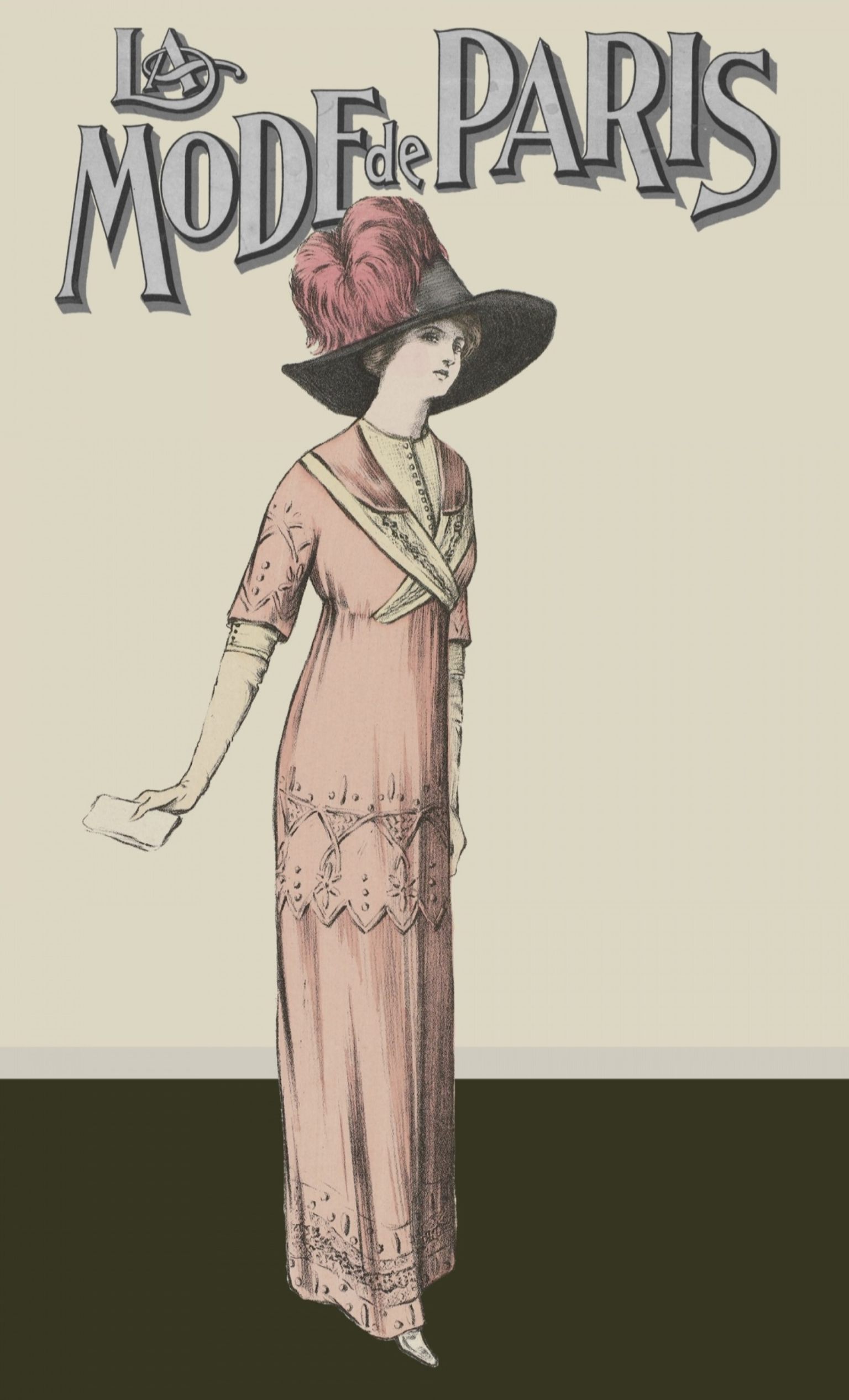}
        \caption{Input Image}
    \end{subfigure}
    \hfill
    \begin{subfigure}[b]{0.24\linewidth}
        \centering
        \includegraphics[width=\linewidth]{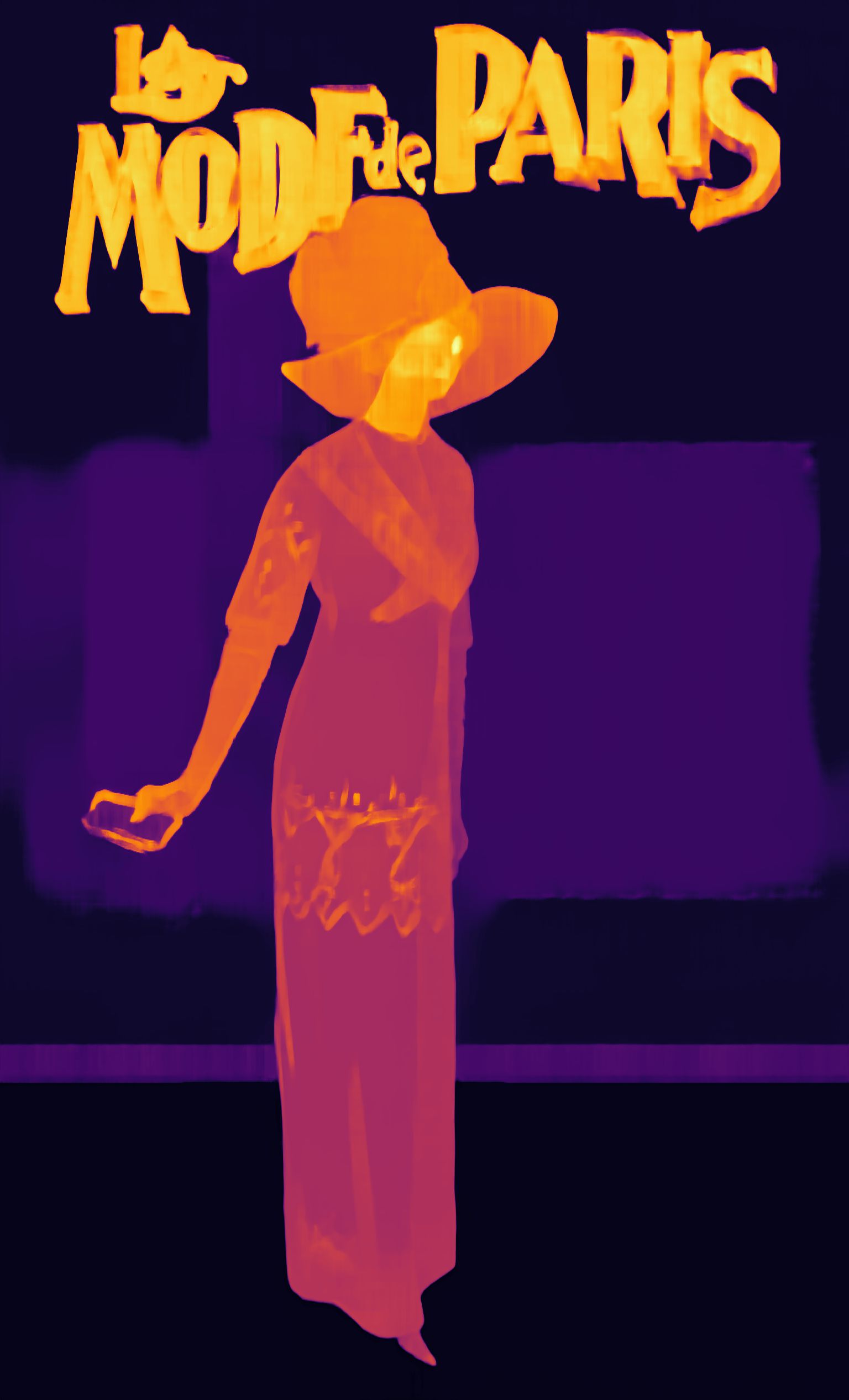}
        \caption{I. D.}
    \end{subfigure}
    \hfill
    \begin{subfigure}[b]{0.24\linewidth}
        \centering
        \includegraphics[width=\linewidth]{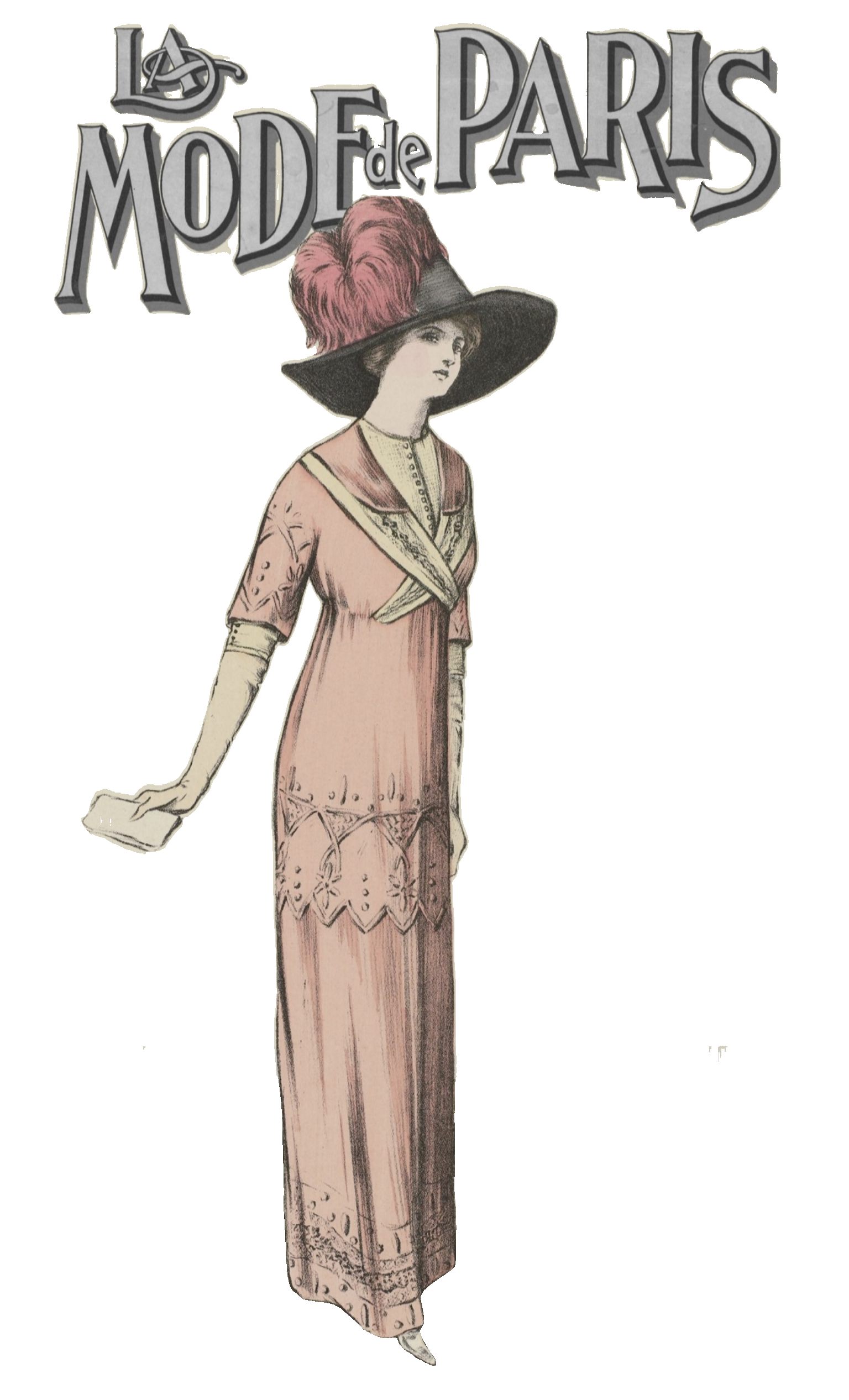}
        \caption{Foreground}
    \end{subfigure}
    \hfill
    \begin{subfigure}[b]{0.24\linewidth}
        \centering
        \includegraphics[width=\linewidth]{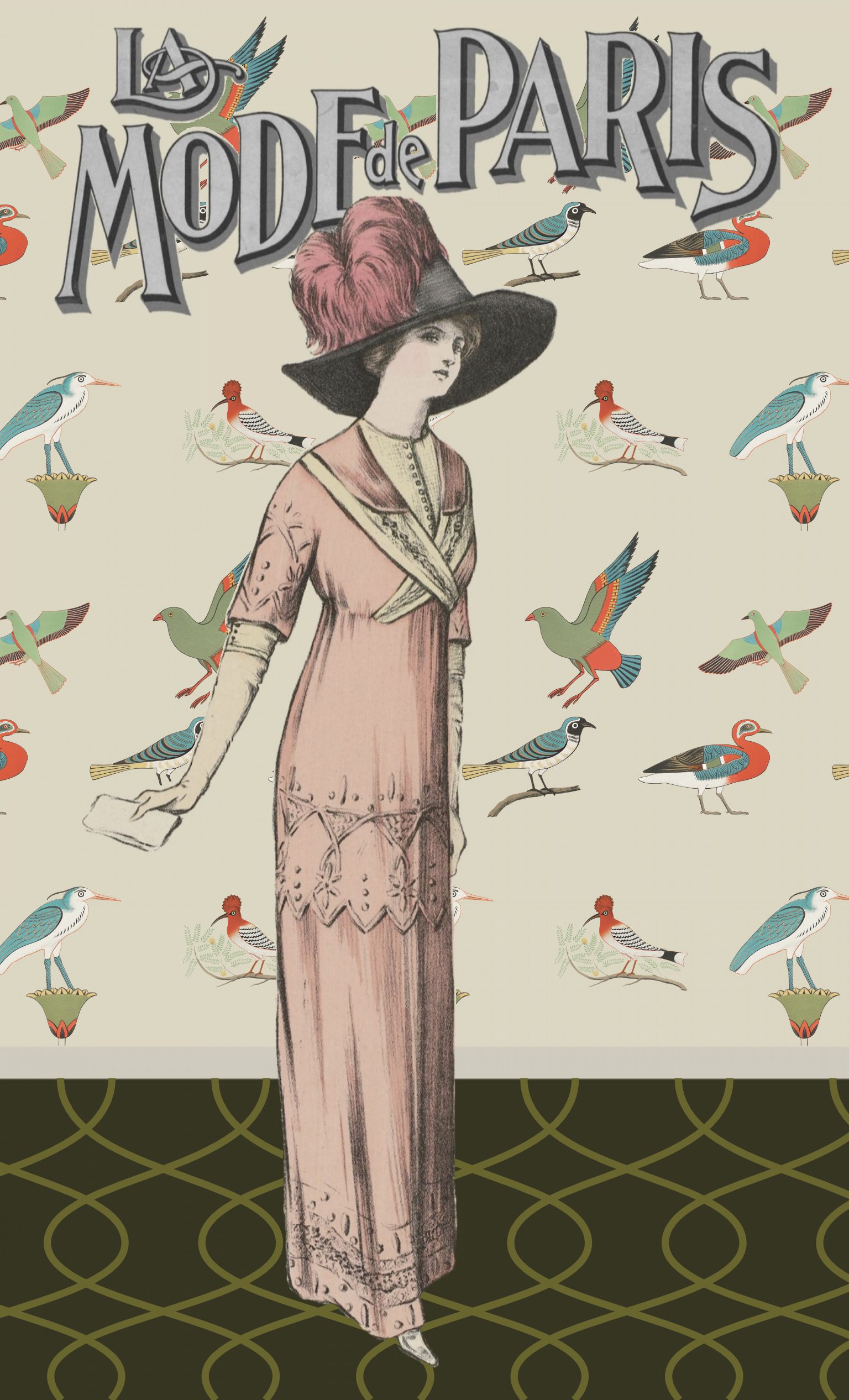}
        \caption{Editing}
    \end{subfigure}
    \vspace*{-3mm}
    \caption{\textbf{Depth-Aware Image Editing.} 
    From an input image (a), illustrator’s depth (b) enables selective separation of key image regions (c) for seamless compositing or depth-aware insertion (d). 
    }
    \label{fig:raster_editing}
\end{figure}

\paragraph{Task}~Raster image editing relies on the composition of multiple layers. Existing segmentation tools
often fall short when handling ambiguous requests: if a user clicks on a face, do they mean the face, the character, or the entire foreground? 
Our work can resolve this ambiguity by providing a depth-based way to segment an input illustration.

\paragraph{Pipeline}~Illustrator's depth is easily leveraged to inform segmentation: 
based on a user-defined threshold value $t$ adjustable in realtime via a slider, an image can be split into two layers, one (foreground) defined as illustrator's depths satisfying $D[i,j]\!>\!t$ and one (background) for all others. 
More generally, any binning strategy into $N$ layers, found through a quick analysis of the entire map $D$ or derived manually, provides a decomposition into layers by ranges of illustrator's depths, which can be directly uploaded in raster graphic editors to allow for direct editing. 

\begin{figure}[!t]
    \centering
    \includegraphics[width=\linewidth]{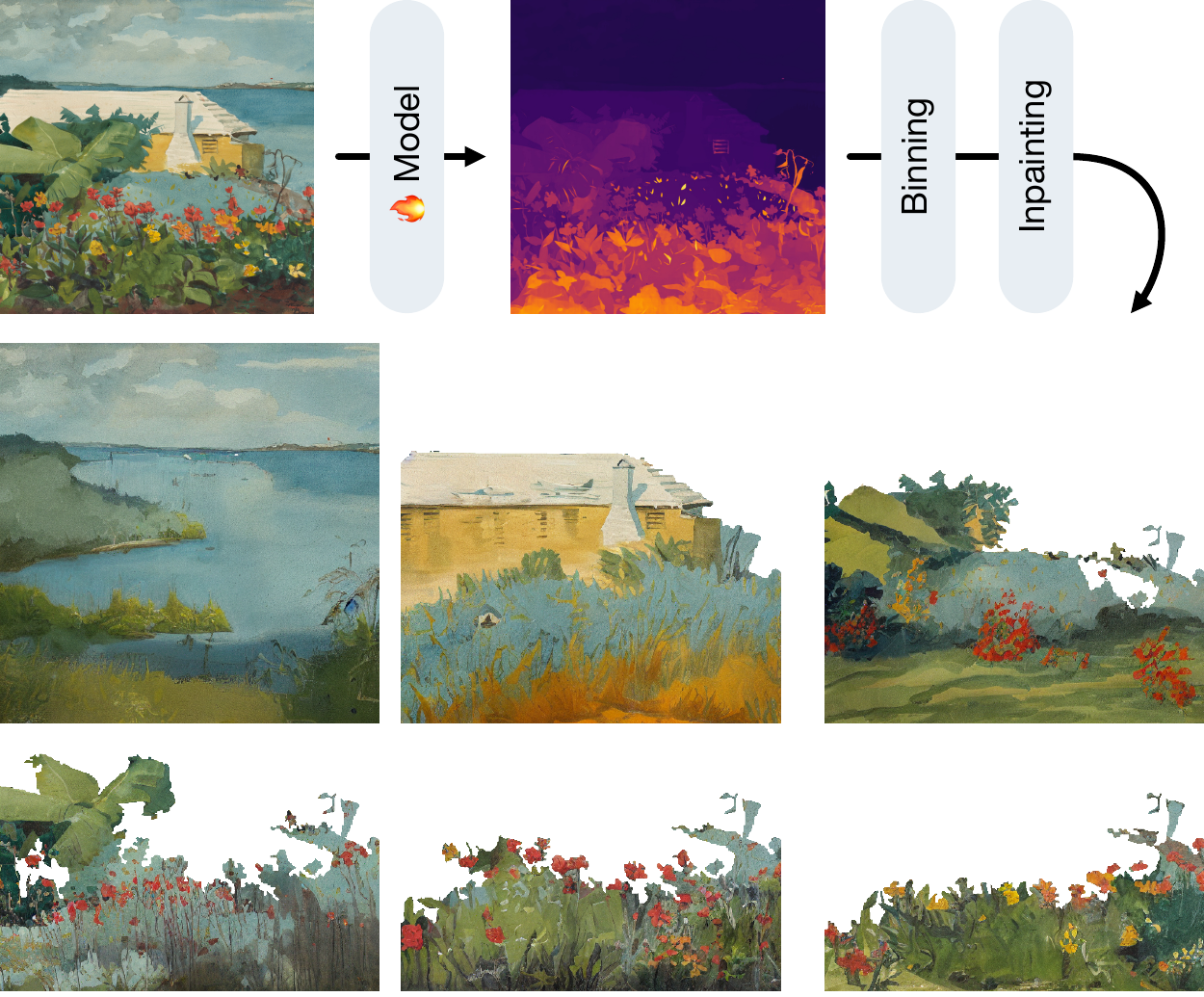}
    \vspace*{-6.5mm}     \caption{\textbf{Illustrator's depth with an inpainting model.} Given our illustrator's depth (top), we can bin the values into several layers and inpaint each occluded regions with Stable Diffusion~\cite{rombach_high-resolution_2022}. The resulting set of overlapping layers can be directly used for 3D parallax effects as demonstrated in our supplemental video. 
    \vspace*{-4mm}
    }
    \label{fig:inpainting}
\end{figure}

\paragraph{Results}~While not designed to rival with modern segmentation models, illustrator's depth provides an intuitive mechanism for selective element isolation within the context of raster image editing, as demonstrated in \cref{fig:raster_editing}. 
Paired with any inpainting model such as~\cite{rombach_high-resolution_2022}, our method can produce $N$ overlapping layers to allow for parallax effects for instance, see~\cref{fig:inpainting}. Additional examples can be found in the accompanying Supplementary Material and video.

\section{Conclusion and Future Work}



We introduced Illustrator's Depth, a novel concept that augments image pixels with additional layer indices, enabling straightforward decomposition into an edit-ready stack. Trained on a curated dataset of SVG files, our network can infer illustrator’s depth across a wide range of inputs, ranging from simple icons to complex raster graphics.
We demonstrated that our method achieves SOTA performance in image vectorization and facilitates a number of downstream tasks beyond vector graphics, such as text-to-vector generation, interactive editing, and relief generation.\smallskip

Although our current model is trained specifically for this task with a curated dataset of SVGs, the rapid advancement of vision models toward one-shot and zero-shot generalization~\cite{wiedemer_video_2025} suggests a near-future where illustrator’s depth could be inferred directly from natural prompts, without explicit training. Beyond its current technical form, we believe that the underlying concept of illustrator’s depth will remain relevant across a variety of creative domains: by shifting the notion of \emph{depth} from a physical metric to a layer-based ready-to-edit abstraction, our work introduces a new paradigm for intelligent creative tools to better assist the artistic process. Illustrator’s depth transforms image decomposition from a mere technical challenge into a creative and assistive foundation for the next generations of computational art and design systems.

\section{Acknowledgments}
This work was supported by the French government through the 3IA Cote d’Azur Investments in the project managed by the National Research Agency (ANR-23-IACL-0001), Ansys, and a Choose France Inria chair.


{
    \small
    
    \bibliographystyle{ieeenat_fullname}
    \bibliography{main}
}

 \clearpage
\setcounter{page}{1} 
\maketitlesupplementary

This supplementary material provides additional details, results, and comparisons to complement our CVPR paper on \emph{Illustrator's Depth}. 
The reader is also encouraged to browse a number of additional files that we provide with our submission: an interactive layer visualization tool (\emph{interactive.html}) to compare various vectorization methods on the two-cherry example from ~\cref{fig:layer_decomp},
an \emph{mmsvg-100.html} file containing all the results of our approach and previous methods on the MMSVG-100 dataset, a \emph{showcase.html} file focusing on 31 other complex examples of various natures (paintings, illustrations, natural images) demonstrating the strength of our approach, as well as a video summarizing our approach and its benefits.

\section{Evaluation on SVGX}

\begin{figure}[ht]
    \centering
      \begin{subfigure}[b]{0.325\linewidth}
        \centering
        \includegraphics[width=\linewidth]{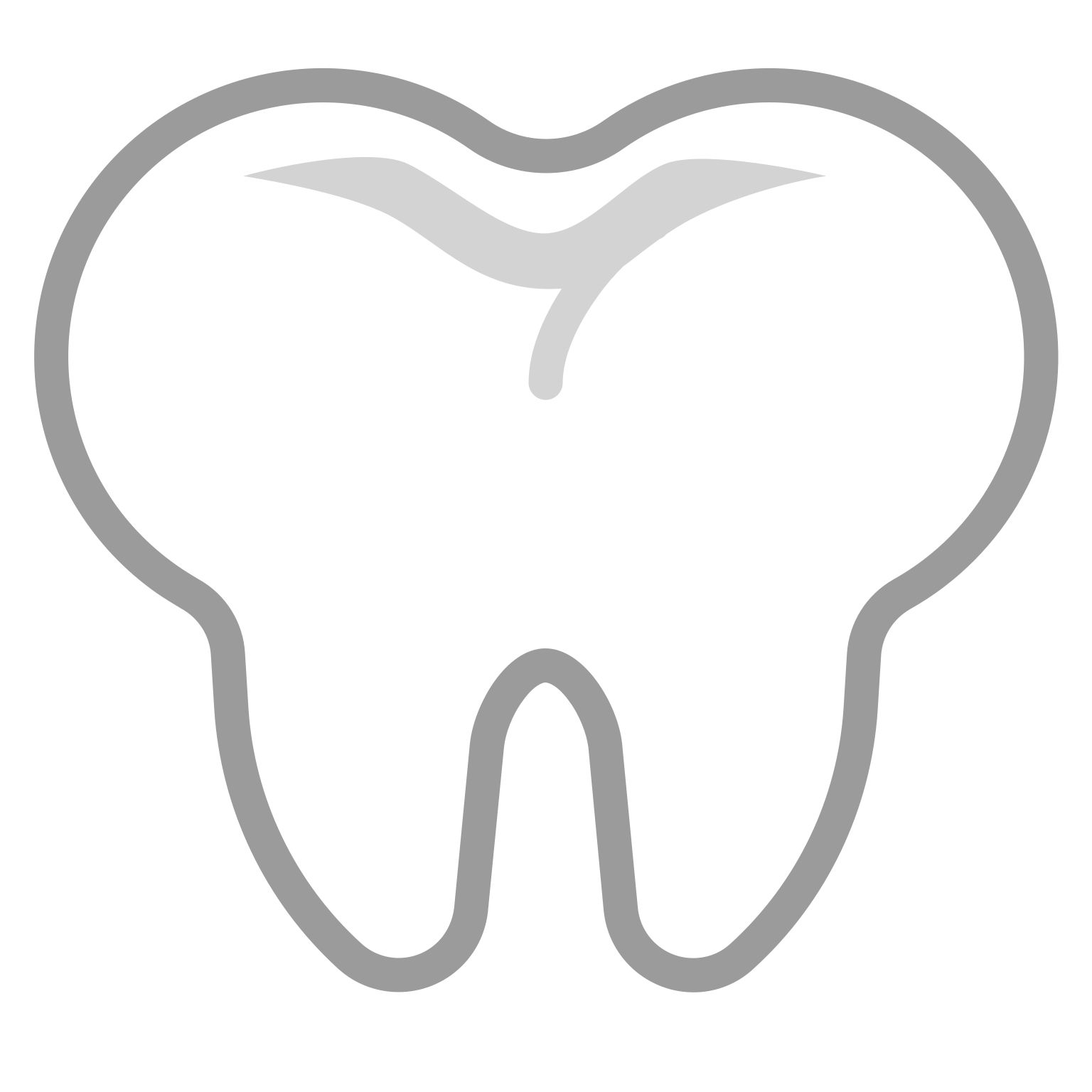}
    \end{subfigure}
    \hfill
    \begin{subfigure}[b]{0.325\linewidth}
        \centering
        \includegraphics[width=\linewidth]{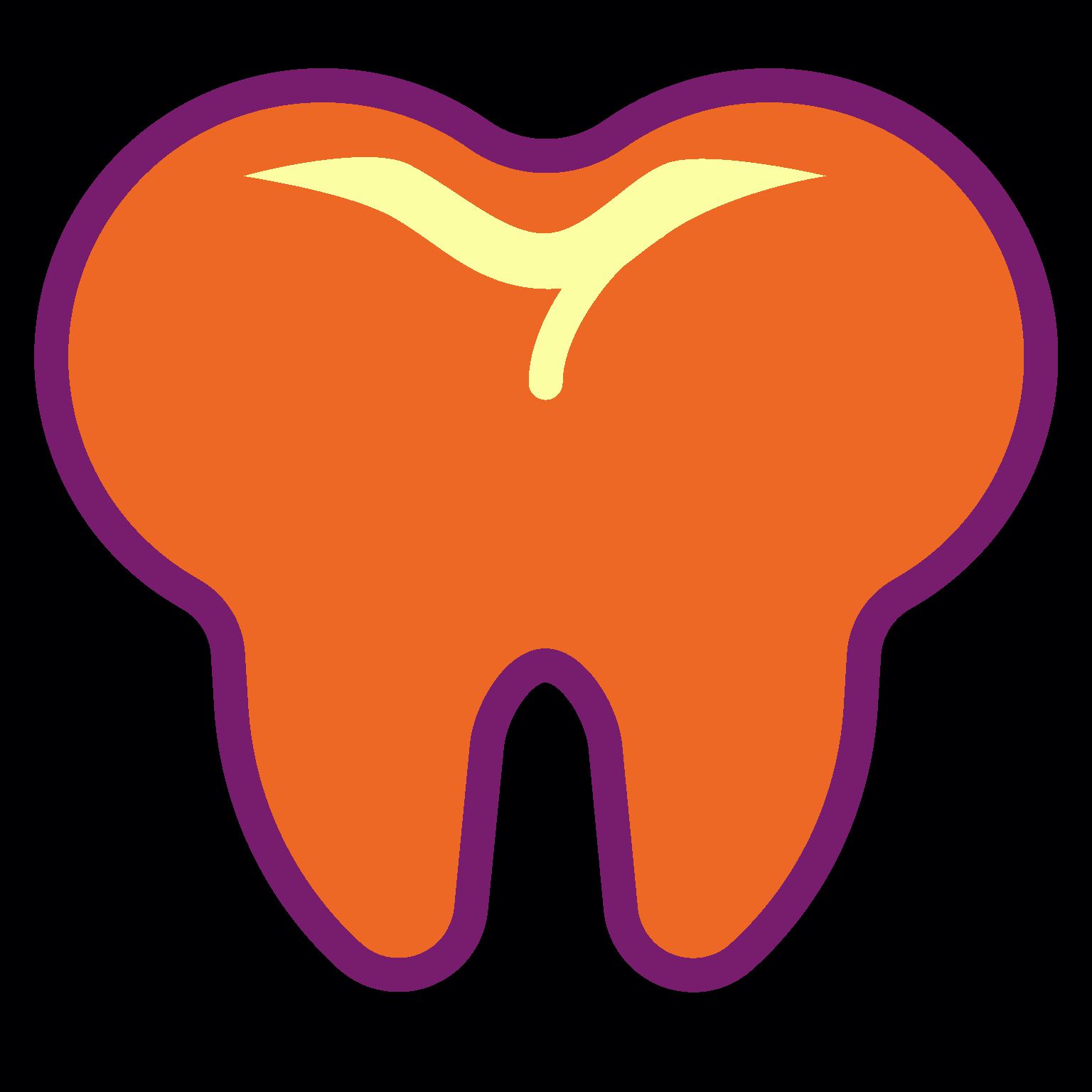}
    \end{subfigure}
    \hfill
    \begin{subfigure}[b]{0.325\linewidth}
        \centering
        \includegraphics[width=\linewidth]{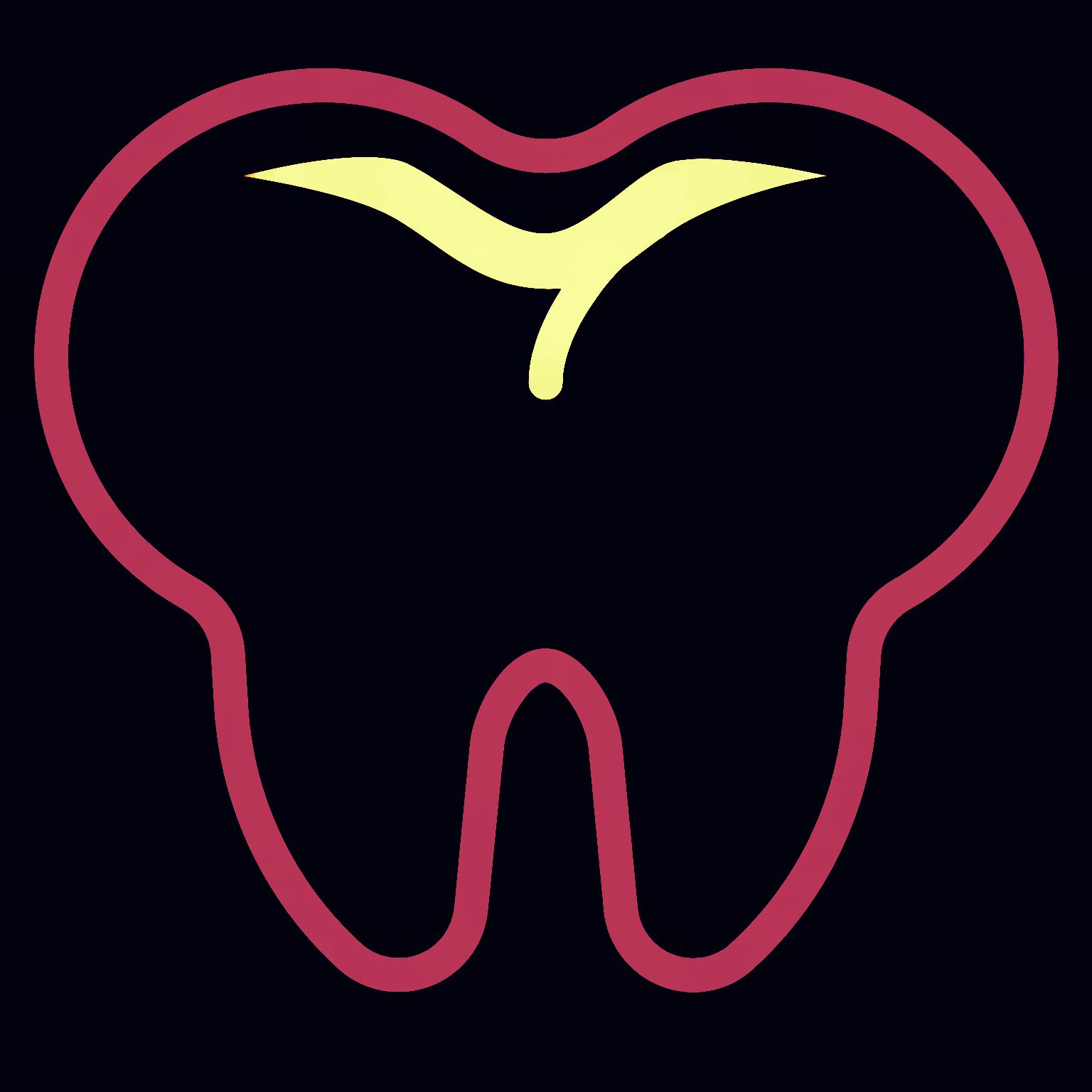}
    \end{subfigure}

    \begin{subfigure}[b]{0.325\linewidth}
        \centering
        \includegraphics[width=\linewidth]{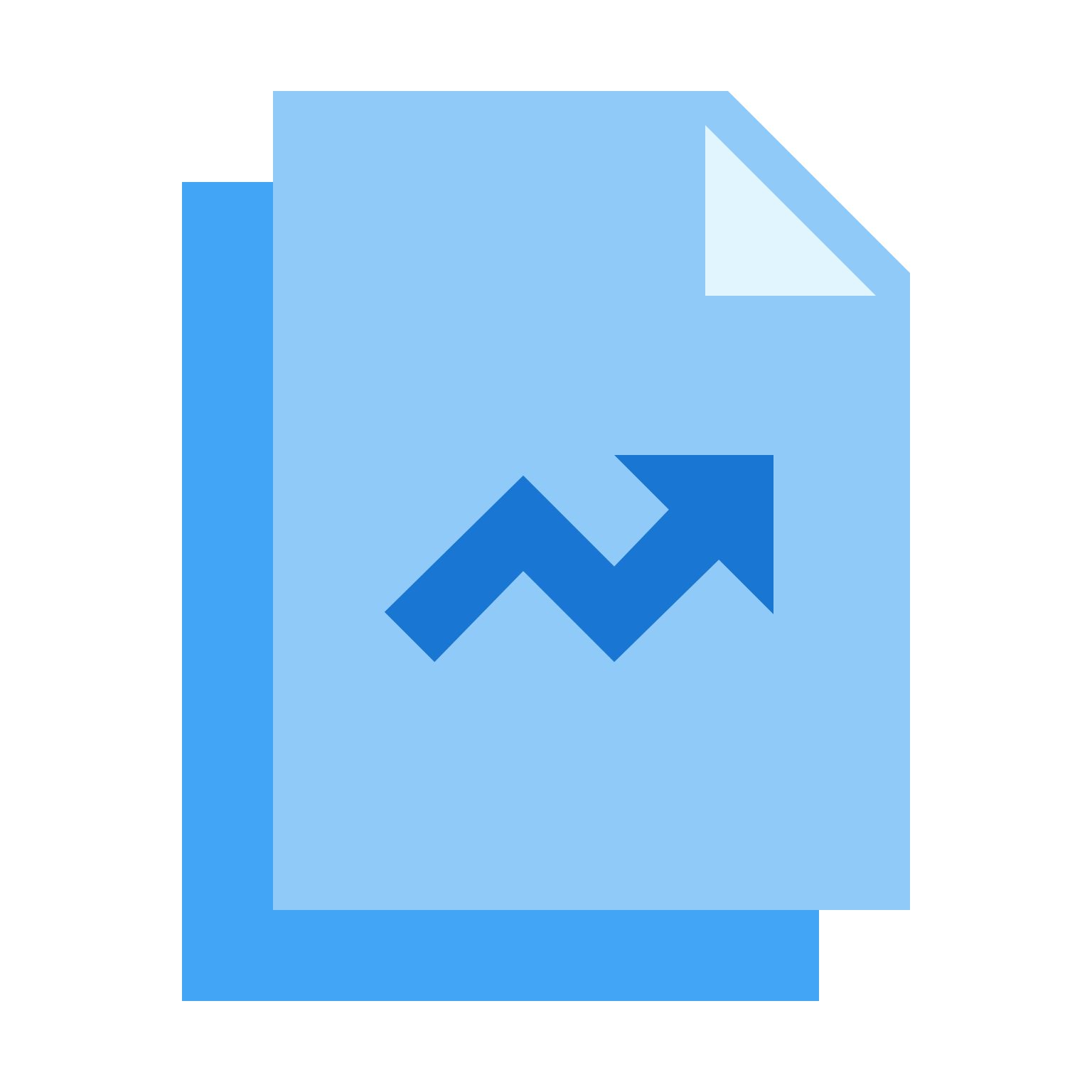}
    \end{subfigure}
    \hfill
    \begin{subfigure}[b]{0.325\linewidth}
        \centering
        \includegraphics[width=\linewidth]{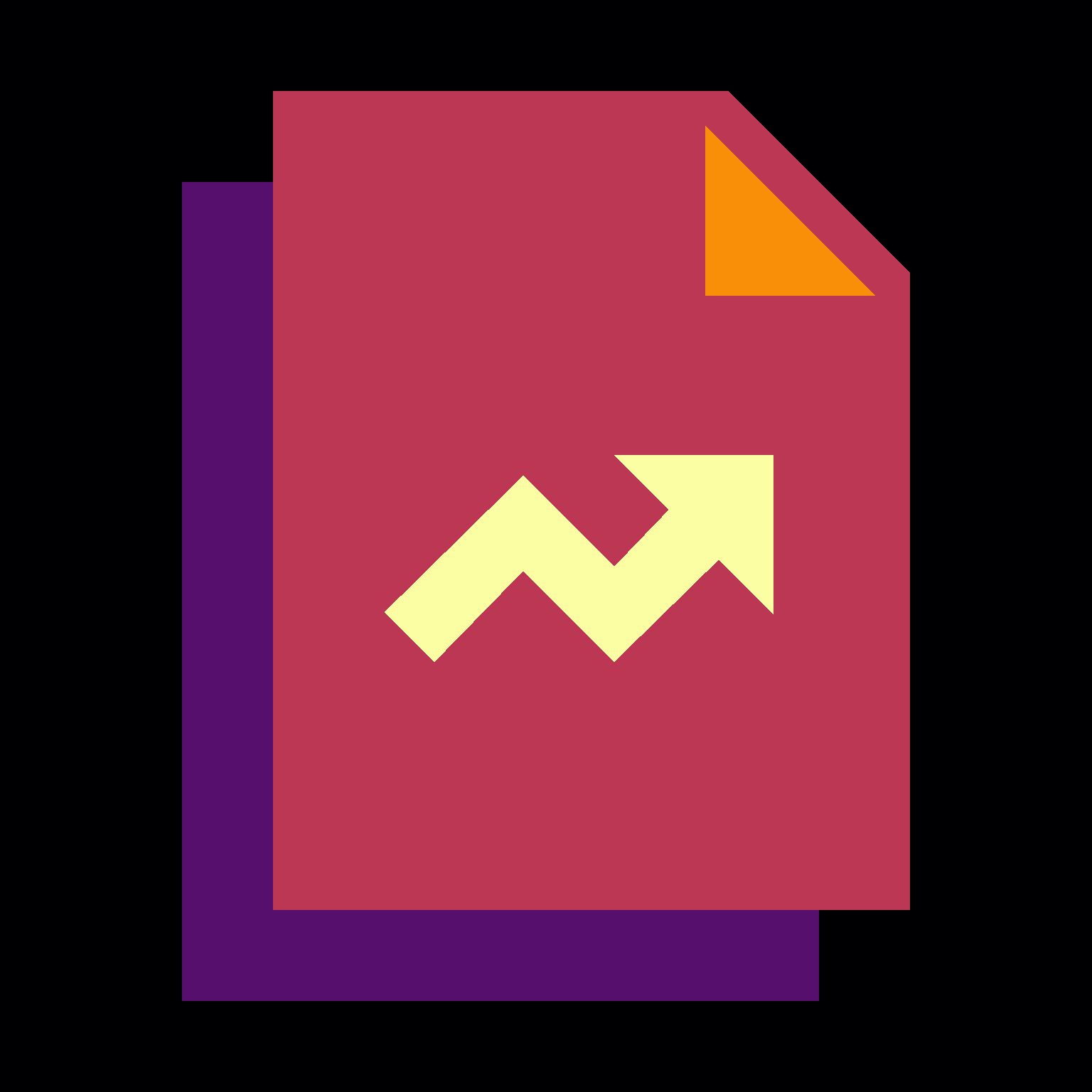}
    \end{subfigure}
    \hfill
    \begin{subfigure}[b]{0.325\linewidth}
        \centering
        \includegraphics[width=\linewidth]{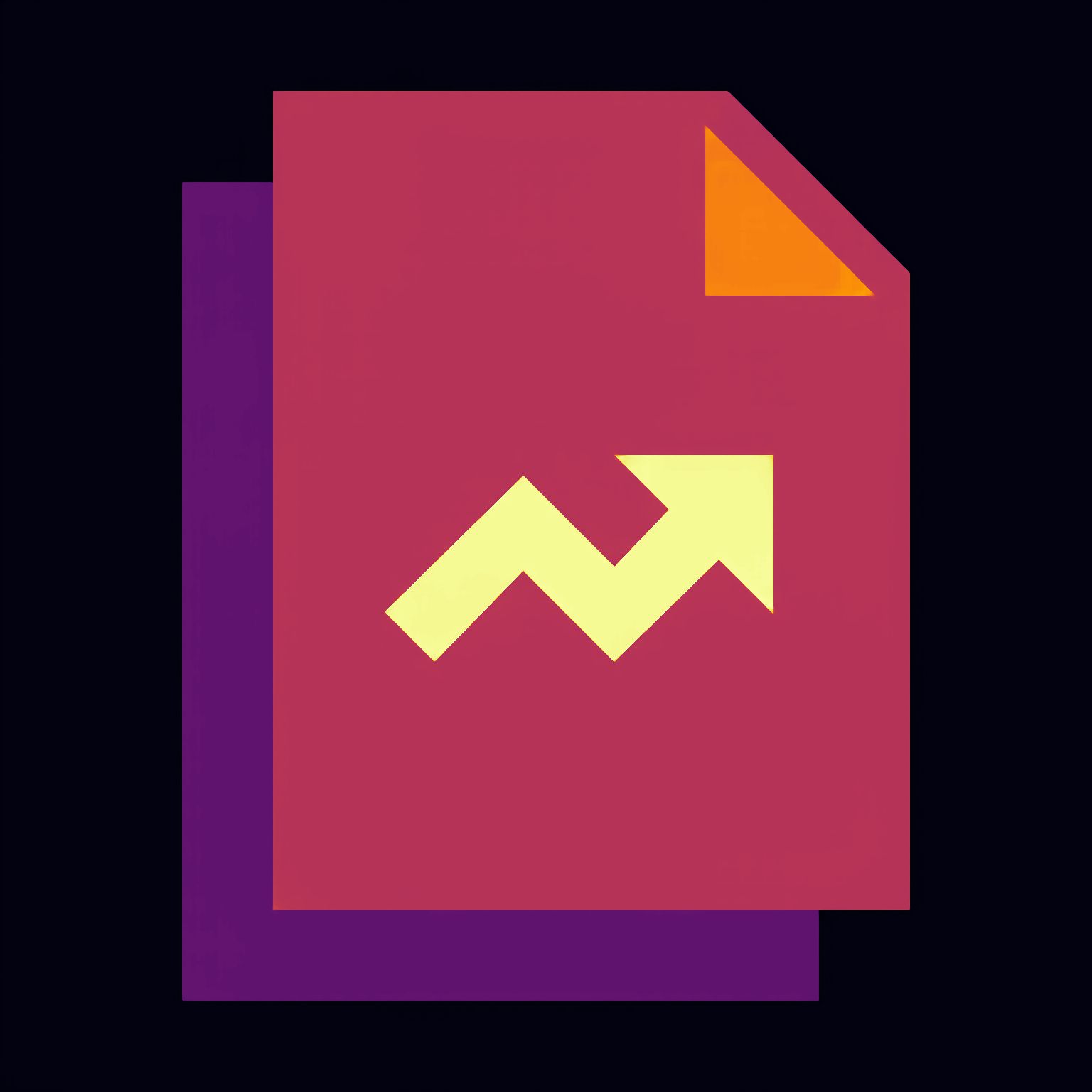}
    \end{subfigure}    

    \begin{subfigure}[b]{0.325\linewidth}
        \centering
        \includegraphics[width=\linewidth]{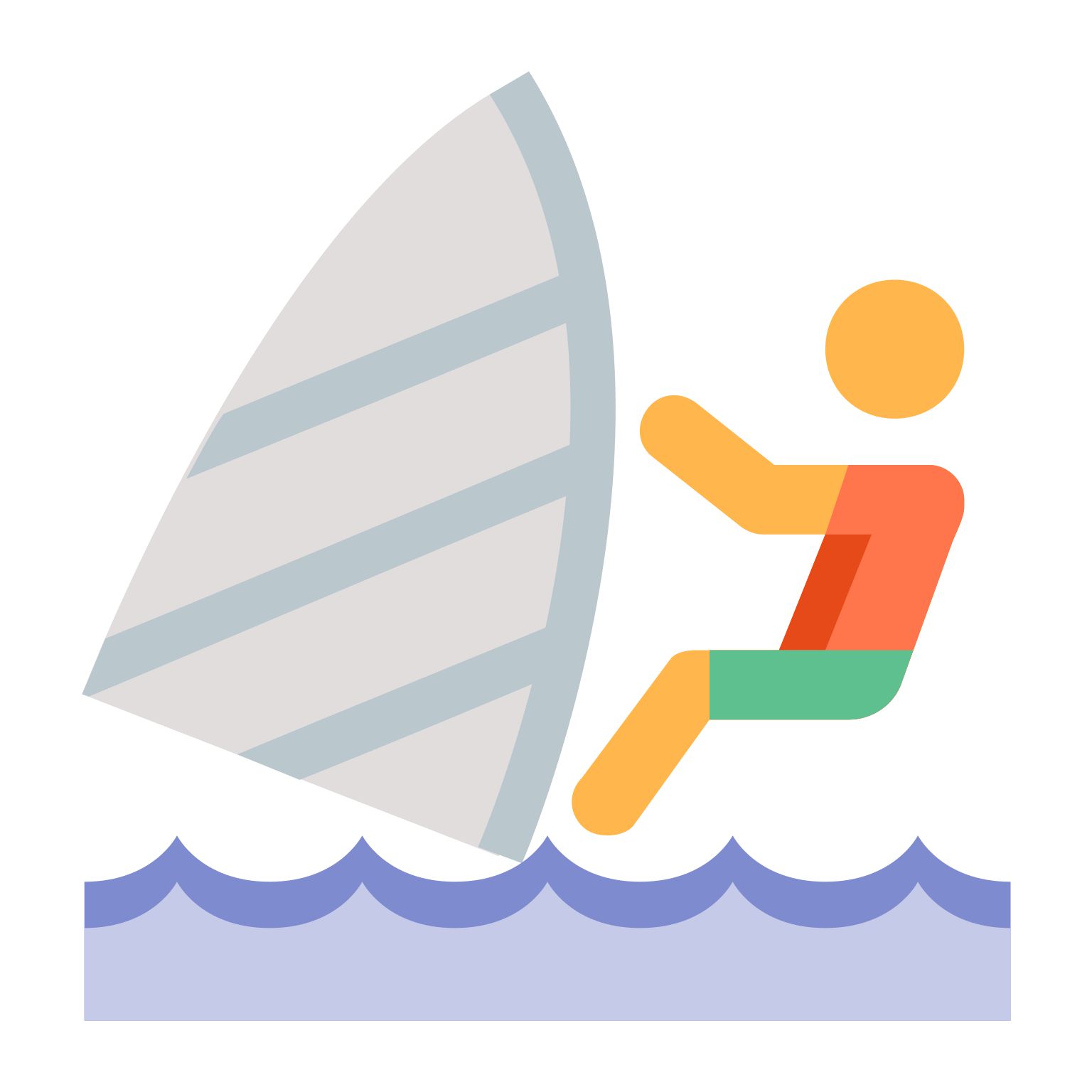}
    \end{subfigure}
    \hfill
    \begin{subfigure}[b]{0.325\linewidth}
        \centering
        \includegraphics[width=\linewidth]{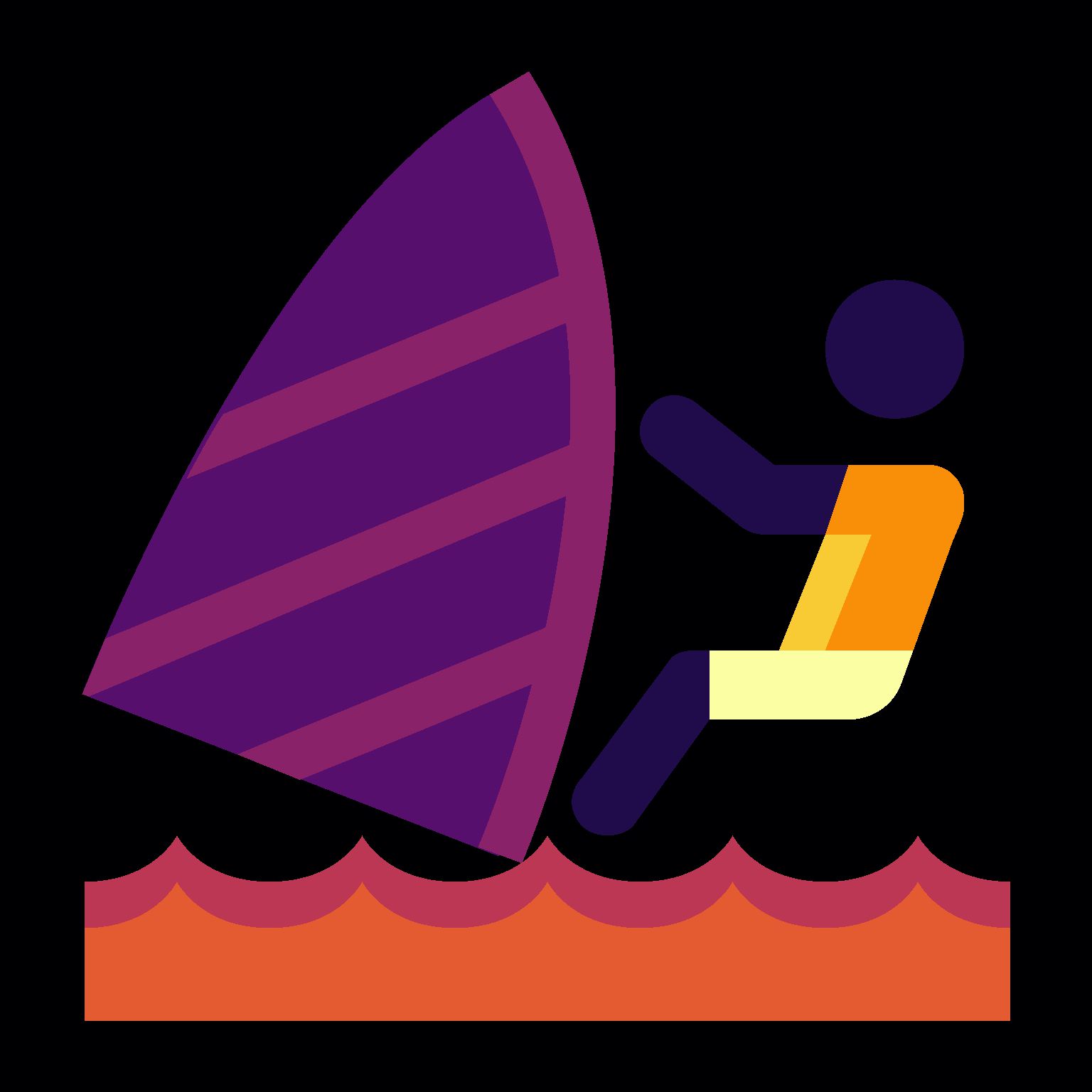}
    \end{subfigure}
    \hfill
    \begin{subfigure}[b]{0.325\linewidth}
        \centering
        \includegraphics[width=\linewidth]{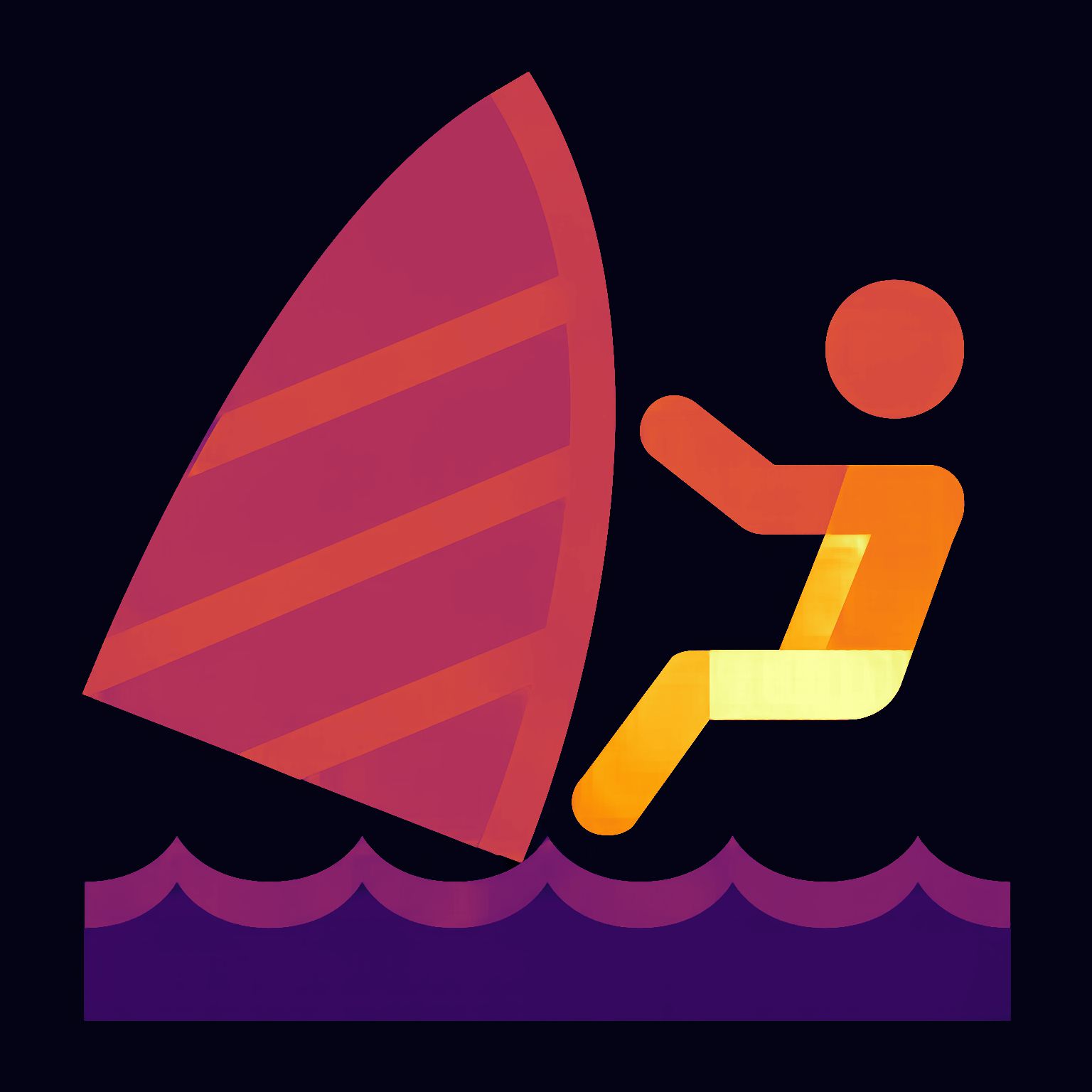}
    \end{subfigure}

    \begin{subfigure}[b]{0.325\linewidth}
        \centering
        \includegraphics[width=\linewidth]{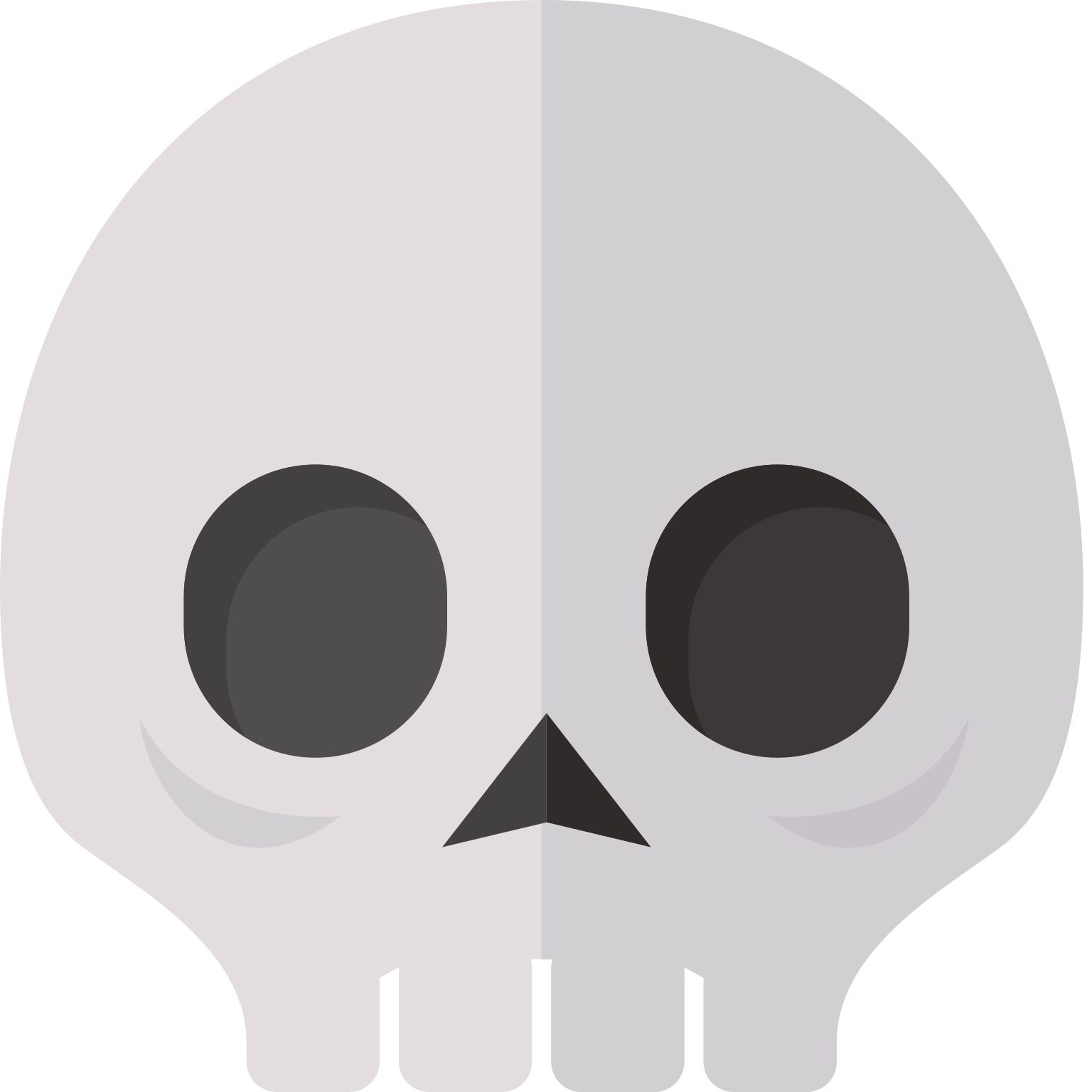}
        \caption{Input Image}
    \end{subfigure}
    \hfill
    \begin{subfigure}[b]{0.325\linewidth}
        \centering
        \includegraphics[width=\linewidth]{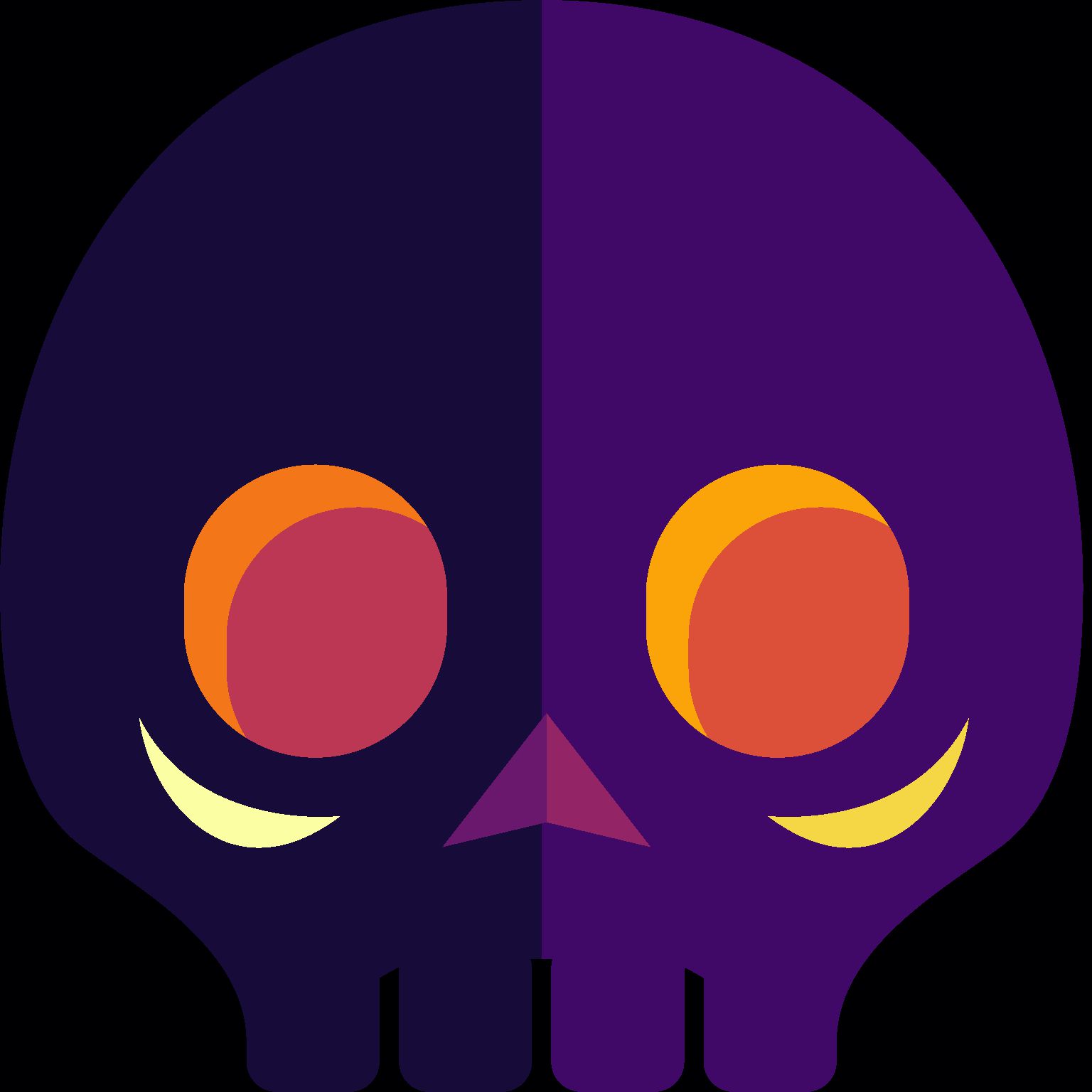}
        \caption{Ground Truth}
    \end{subfigure}
    \hfill
    \begin{subfigure}[b]{0.325\linewidth}
        \centering
        \includegraphics[width=\linewidth]{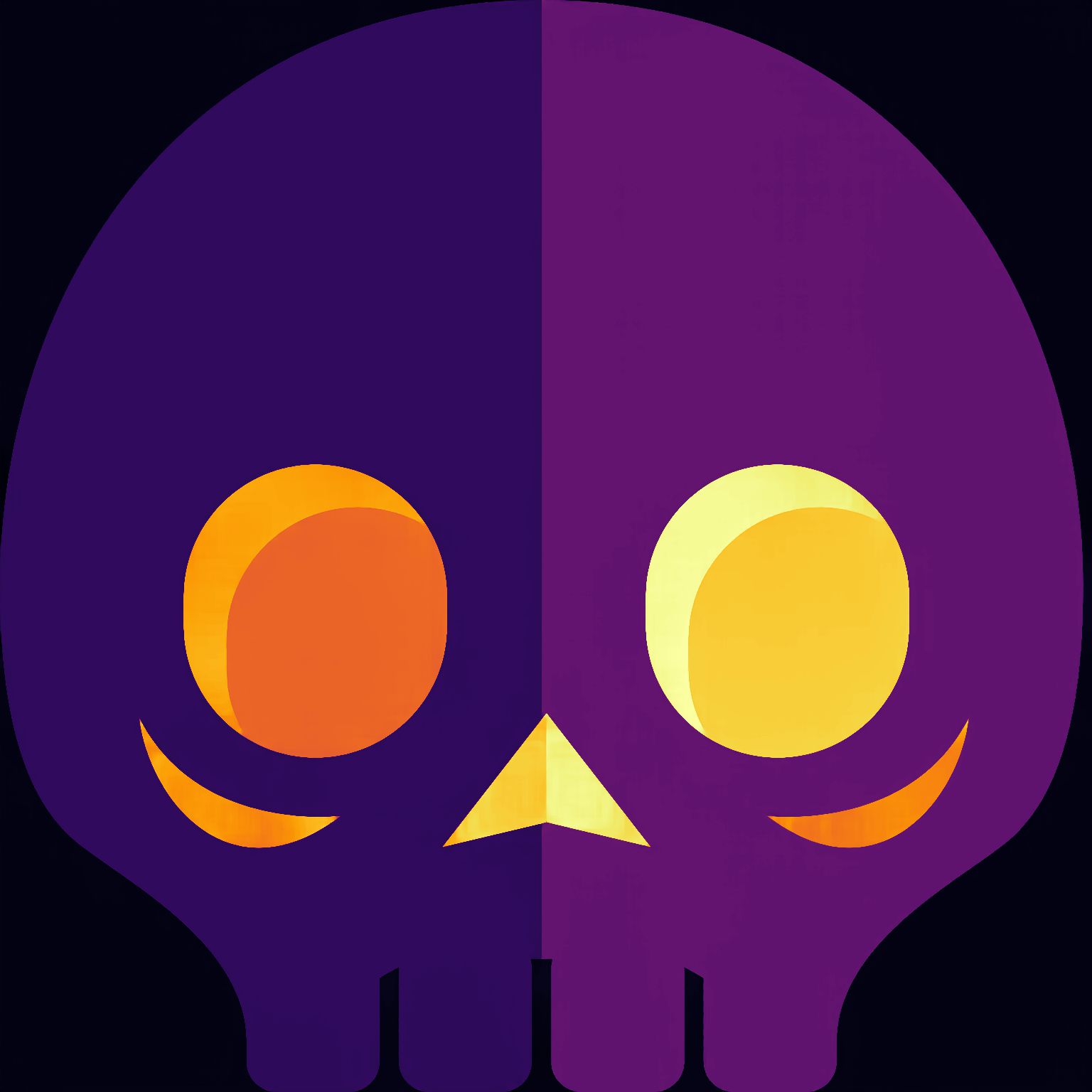}
        \caption{Ours}
    \end{subfigure}

    \caption{\textbf{Evaluation of the inferred layer indices.} When evaluated on the SVGX-Core-250k dataset, our method predicts a satisfactory illustrator's depth even if some conventions are different from in our training dataset (for instance, the outline of the tooth is placed \emph{below} the filled-in shape).}
    \label{fig:suppl_svgx}
\end{figure}

While we trained our model on (a curated subset of) the MMSVG-Illustration dataset~\cite{yang_omnisvg_2025}, we also evaluated our layer index predictions on the SVGX-Core-250k dataset curated by~\cite{xing_empowering_2025} for completeness. Similar to MMSVG, we randomly select 100 images for quantitative analysis. As shown in \cref{tab:suppl_svgx} and \cref{fig:suppl_svgx}, our model demonstrates strong generalization and maintains excellent performance.

\begin{table}[h]
    \centering
    \caption{\textbf{Evaluation of our method on different datasets predicted depth.} Raw outputs of the network.}
    \label{tab:suppl_svgx}
    \resizebox{0.99\linewidth}{!}{
    \begin{tabular}{lccc}
    \toprule
       & Order $\uparrow$ & MAE $\downarrow$ & MSE $\downarrow$\\
     \midrule
    MMSVG 
    & 0.987 & 0.12 &  0.26 
    \\
    SVGX-Core-250k 
    &  0.984 &  0.16 & 0.53
    \\
     \bottomrule
    \end{tabular}}
\end{table}
 
\section{Ablation Studies}

\begin{figure}[h]
    \centering
      \begin{subfigure}[b]{0.325\linewidth}
        \centering
        \includegraphics[width=\linewidth]{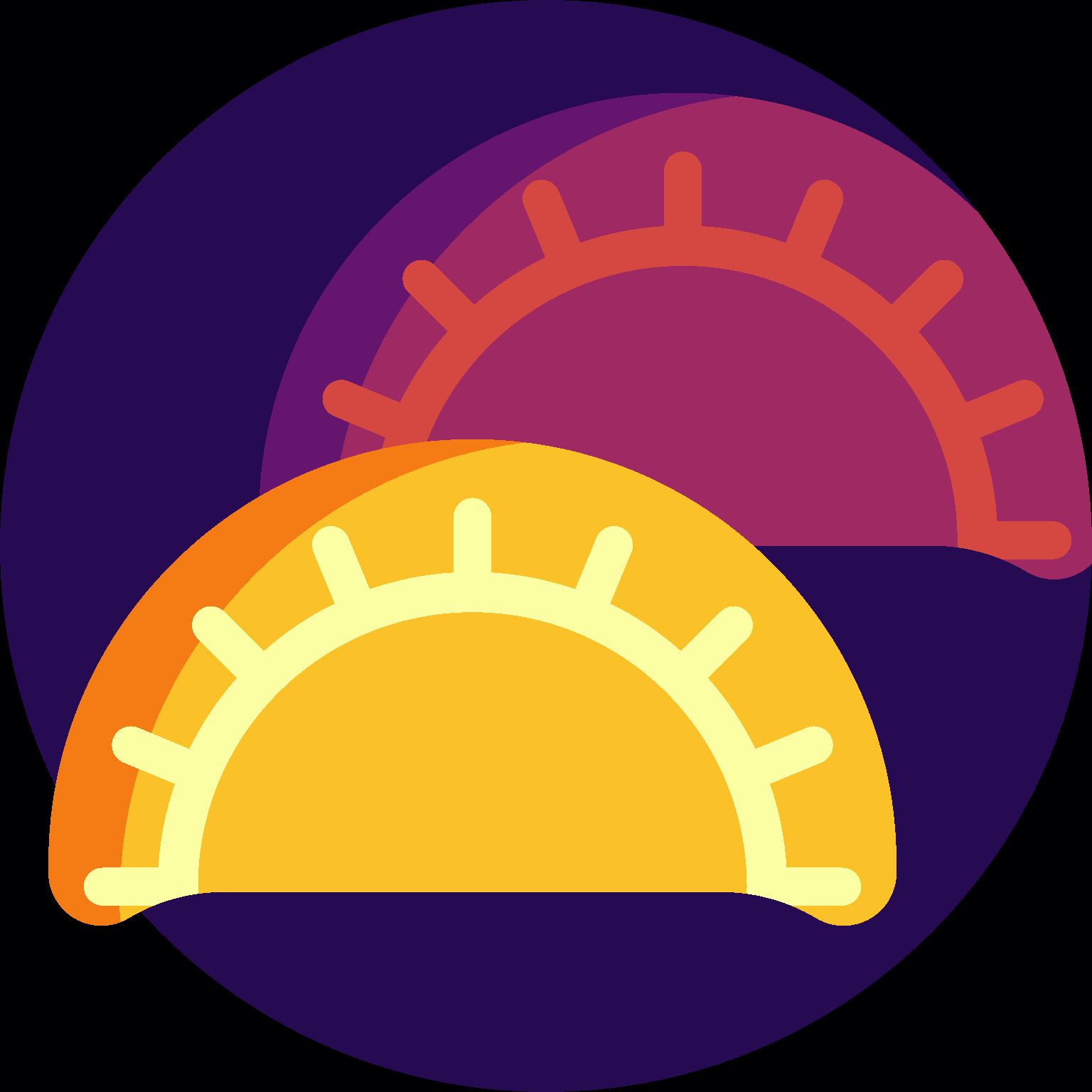}
        \caption{GT}
    \end{subfigure}
    \hfill
      \begin{subfigure}[b]{0.325\linewidth}
        \centering
        \includegraphics[width=\linewidth]{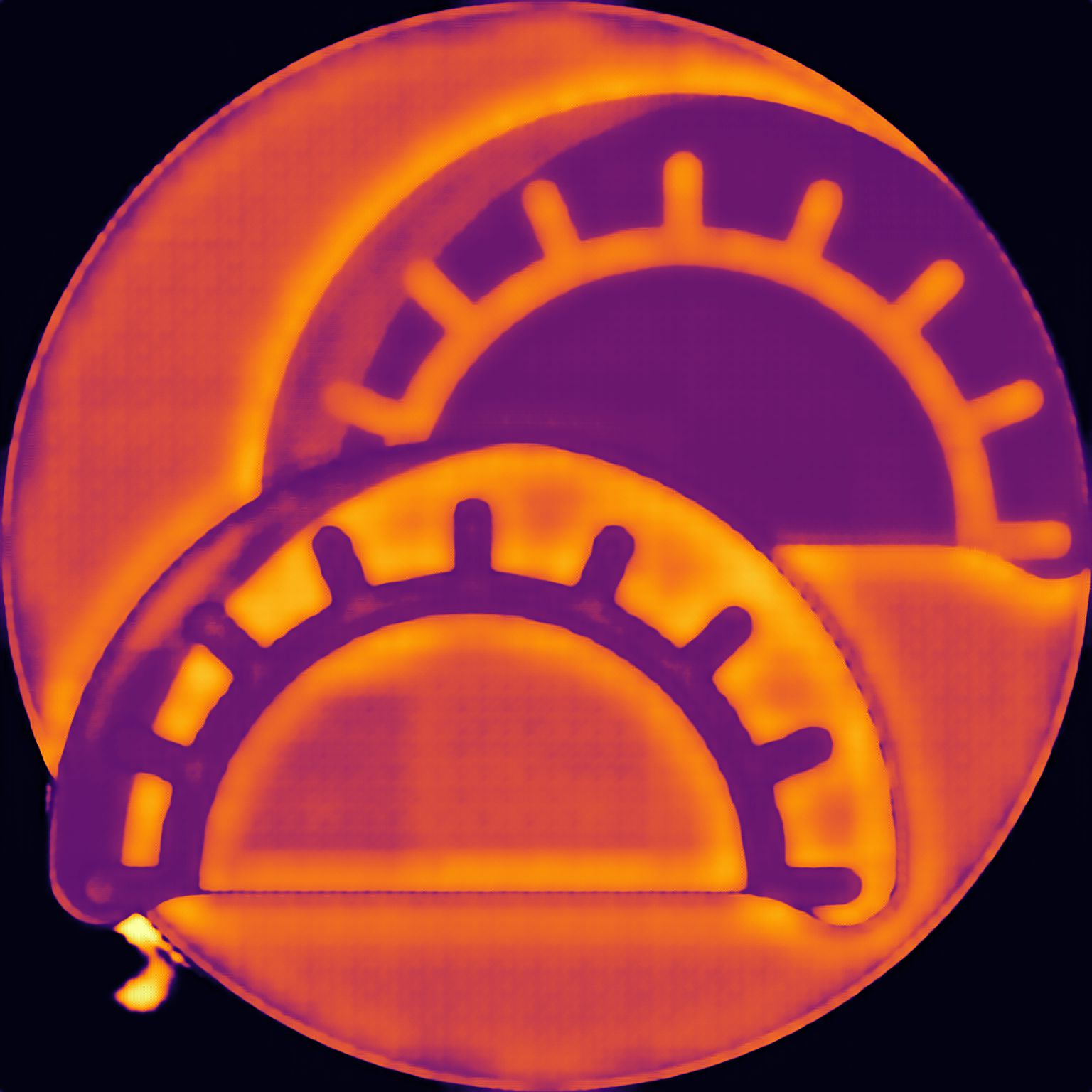}
        \caption{W/o Data Cleaning}
    \end{subfigure}
    \hfill
      \begin{subfigure}[b]{0.325\linewidth}
        \centering
        \includegraphics[width=\linewidth]{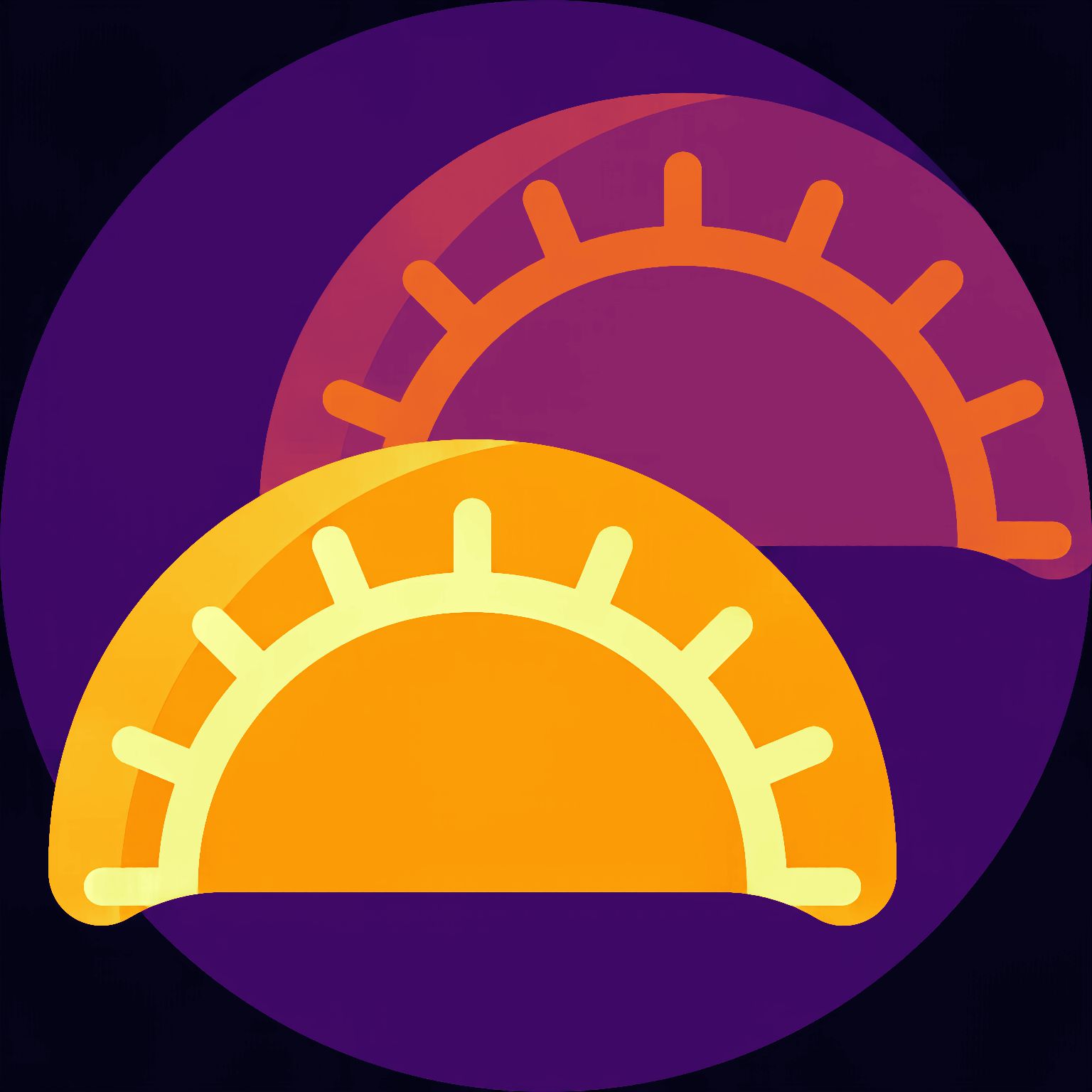}
        \caption{Ours}
    \end{subfigure}

    \centering
      \begin{subfigure}[b]{0.325\linewidth}
        \centering
        \includegraphics[width=\linewidth]{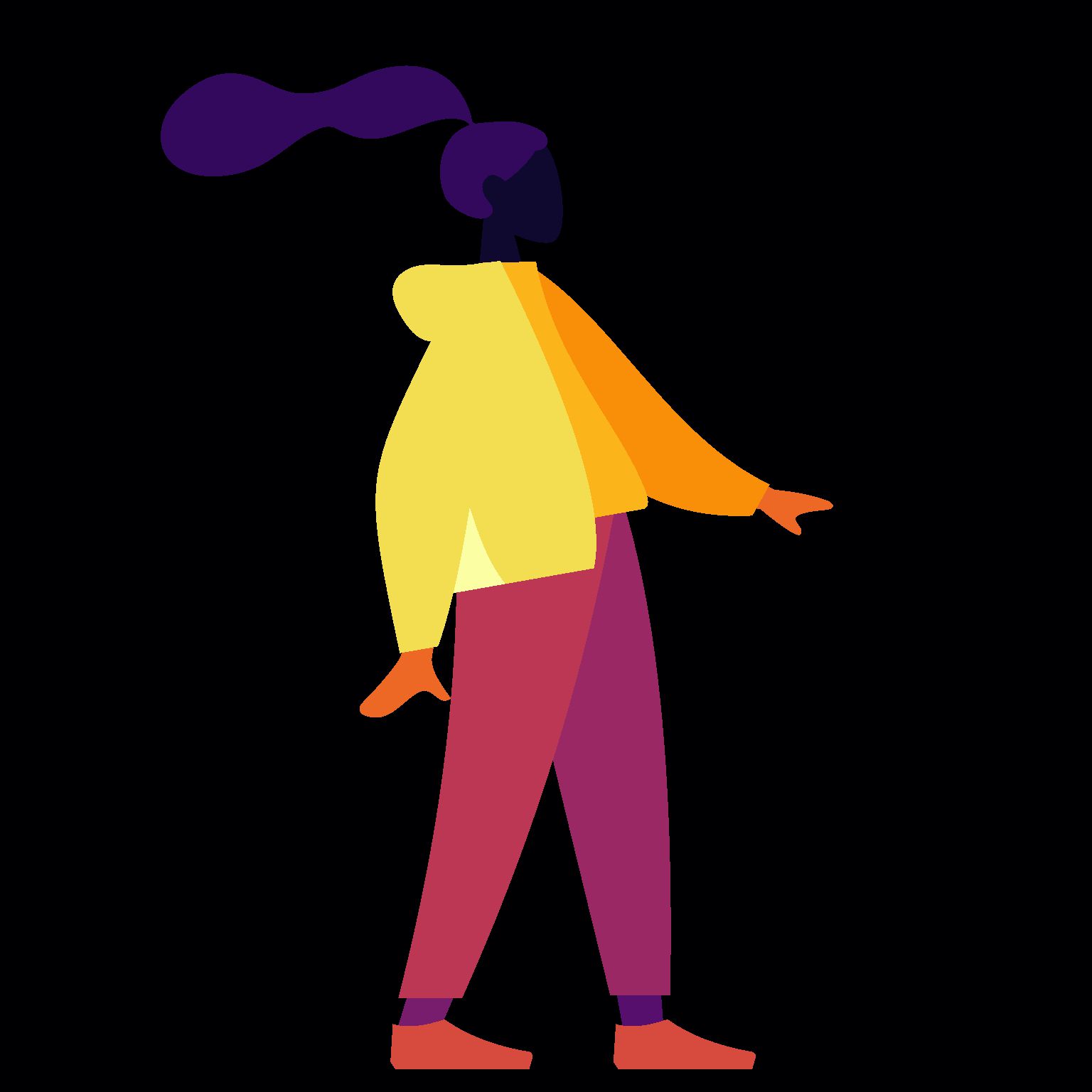}
        \caption{GT}
    \end{subfigure}
    \hfill
      \begin{subfigure}[b]{0.325\linewidth}
        \centering
        \includegraphics[width=\linewidth]{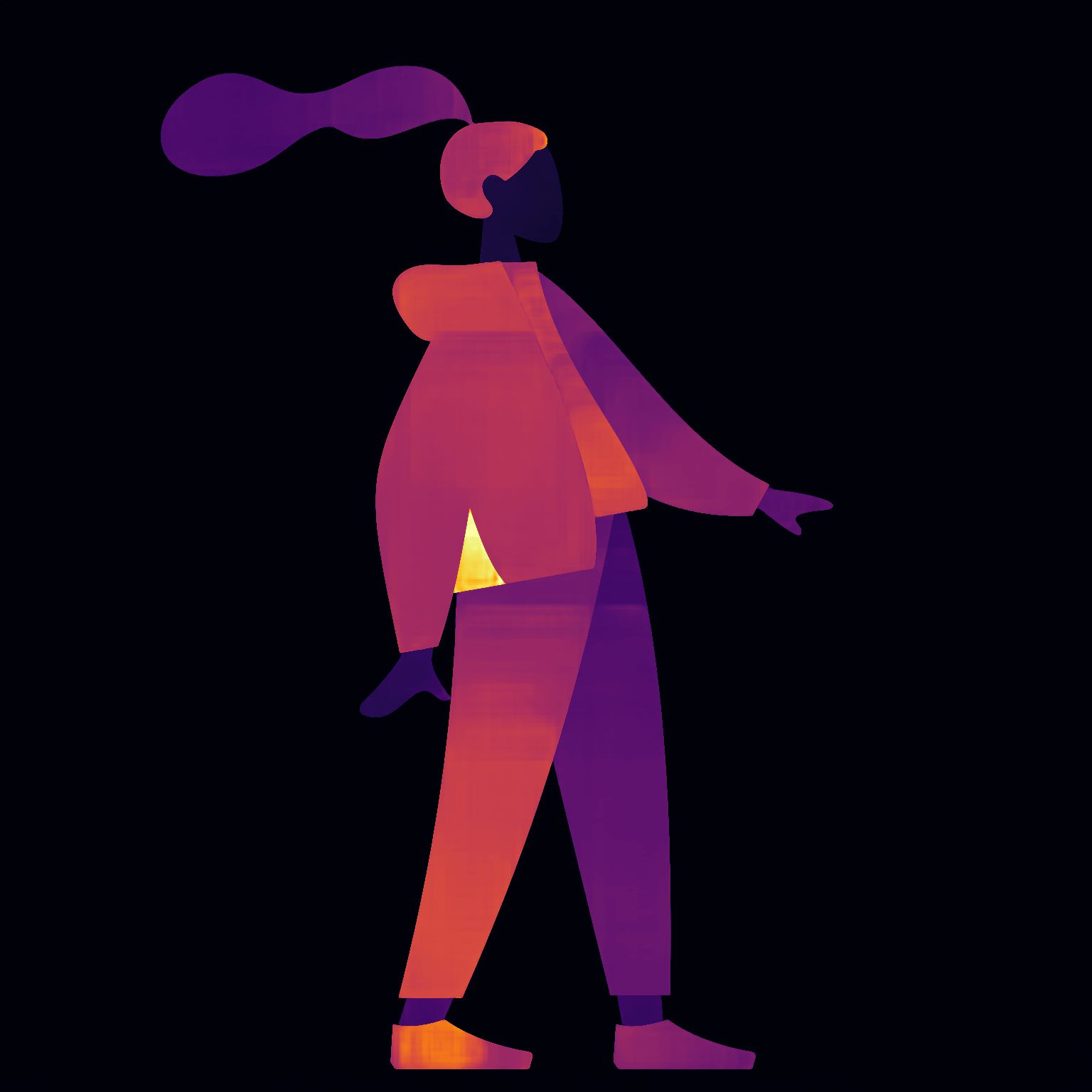}
        \caption{W/ Direct Index}
    \end{subfigure}
    \hfill
      \begin{subfigure}[b]{0.325\linewidth}
        \centering
        \includegraphics[width=\linewidth]{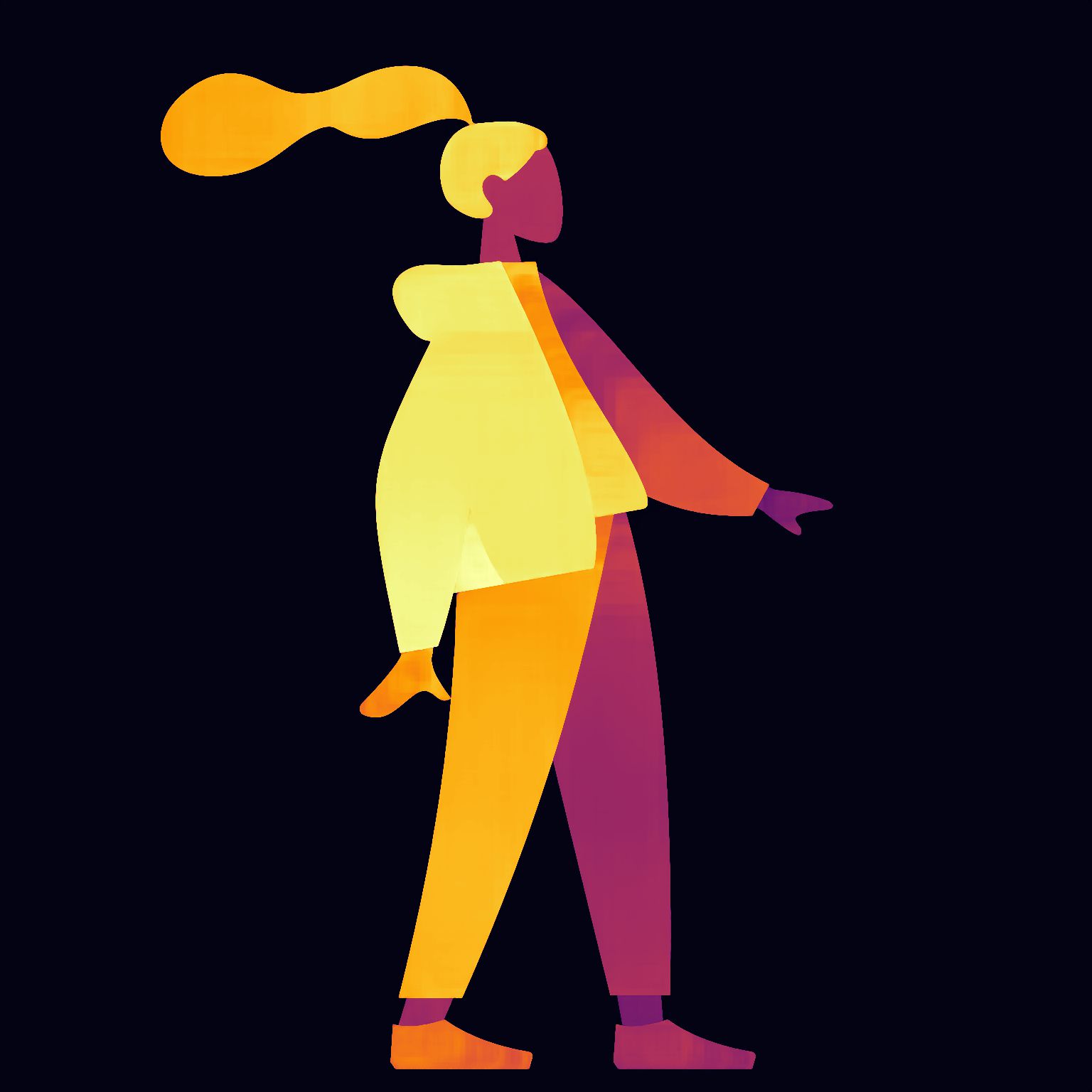}
        \caption{Ours}
    \end{subfigure}
    
    \centering
      \begin{subfigure}[b]{0.325\linewidth}
        \centering
        \includegraphics[width=\linewidth]{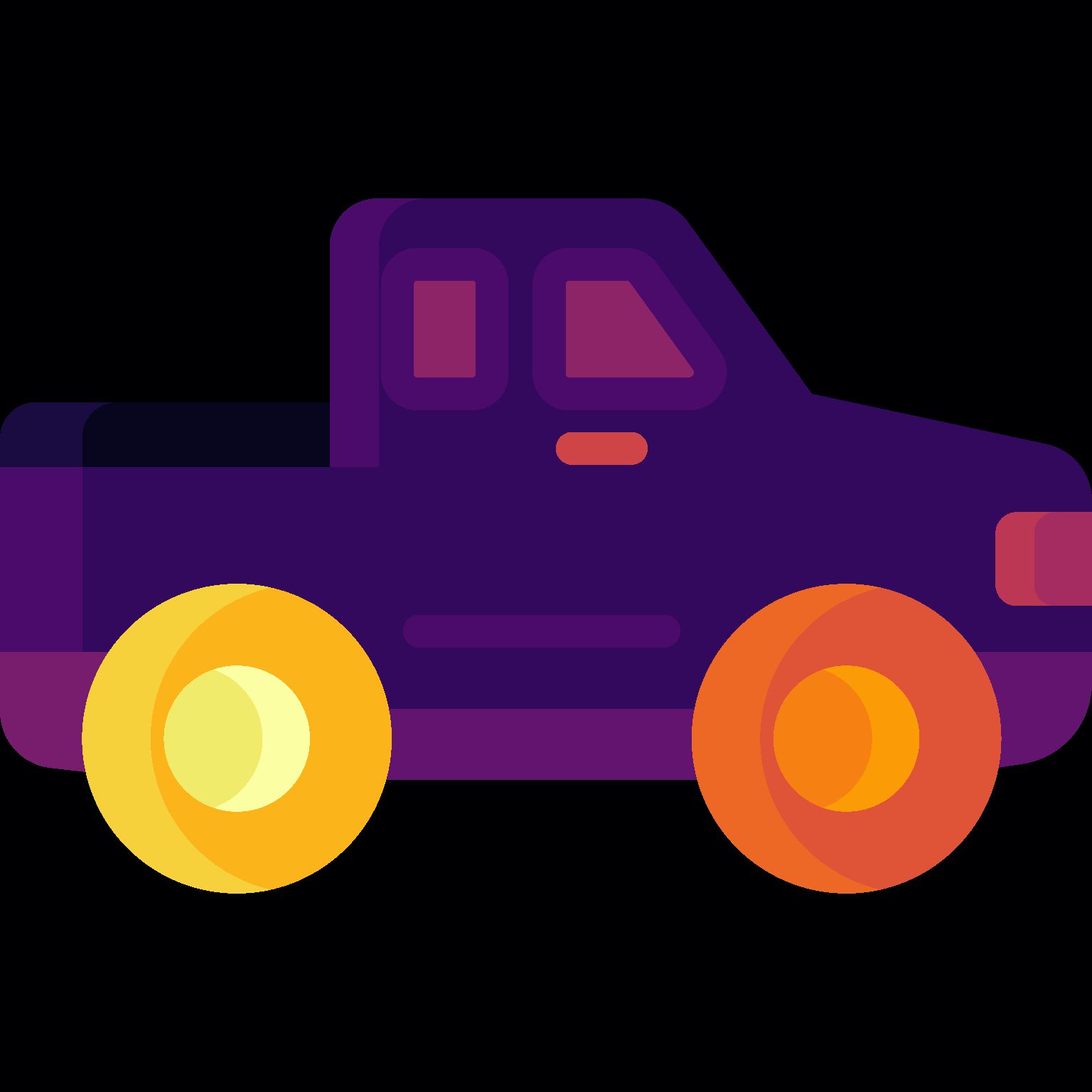}
        \caption{GT}
    \end{subfigure}
    \hfill
      \begin{subfigure}[b]{0.325\linewidth}
        \centering
        \includegraphics[width=\linewidth]{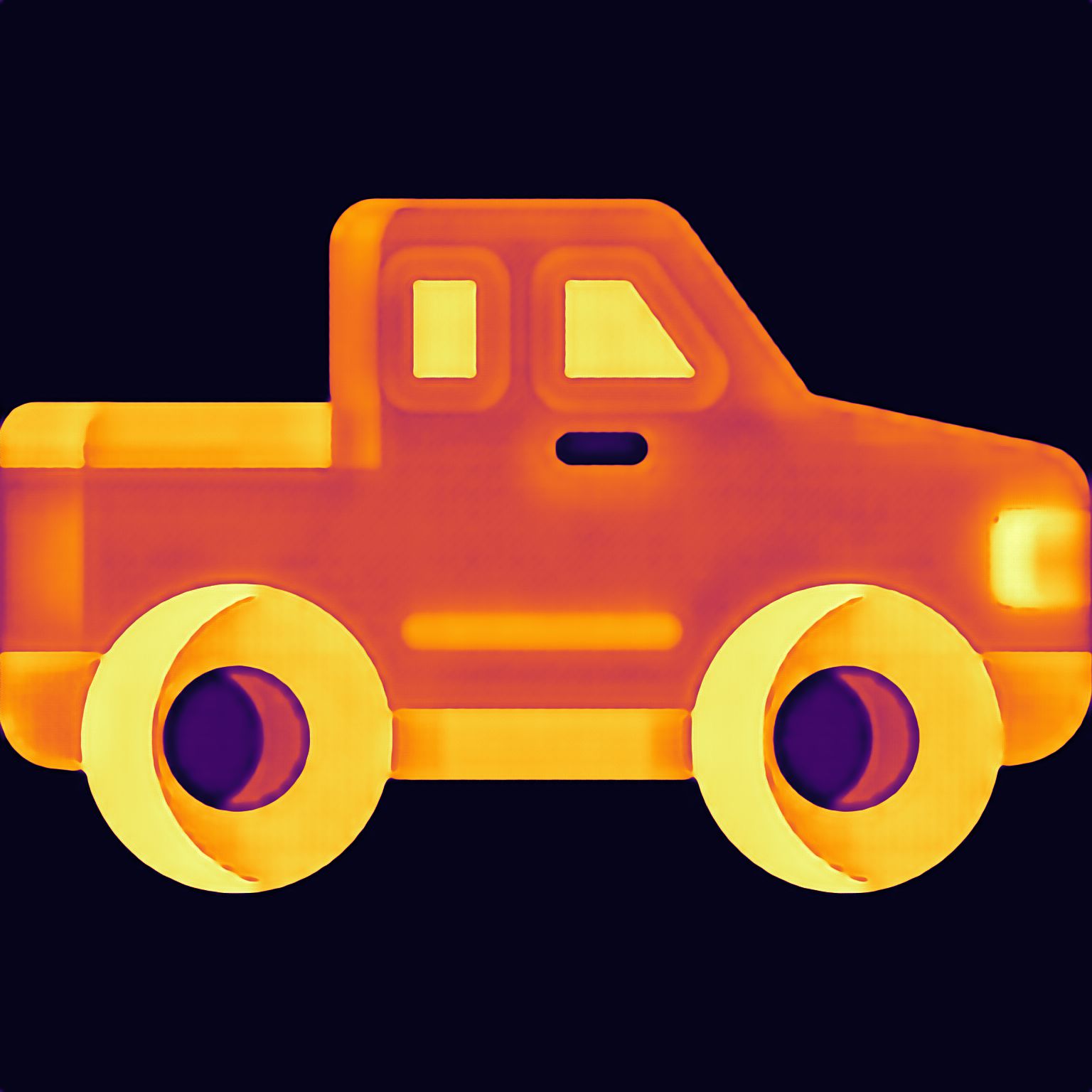}
        \caption{W/o Depth Prior}
    \end{subfigure}
    \hfill
      \begin{subfigure}[b]{0.325\linewidth}
        \centering
        \includegraphics[width=\linewidth]{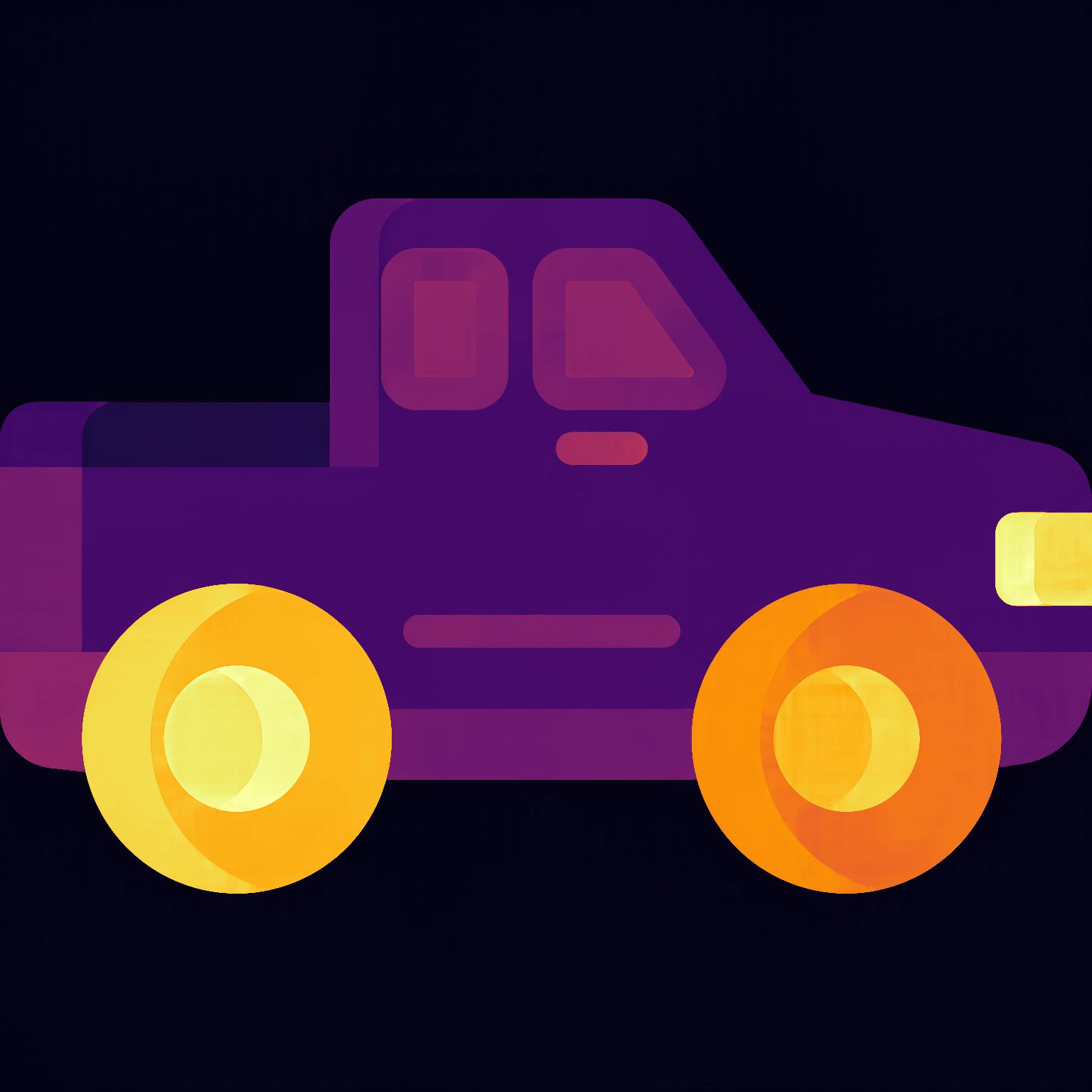}
        \caption{Ours}
    \end{subfigure}
    
    \caption{\textbf{Ablation studies.} Data cleaning (top row) and direct indexing (middle row) ease the burden of the model, resulting in cleaner predictions, while using a depth prior initialization (bottom row) significantly improves our model's performance.}
    \label{fig:suppl_ablation}
\end{figure}

We present additional qualitative results in~\cref{fig:suppl_ablation} to complement the quantitative findings reported in Table~2 of the main paper. While data cleaning and the use of depth priors lead to pronounced improvements, the choice of layer indices vs. disparity space ($d$ vs. $1/d$) yields more subtle effects, yet still provides noticeable gains in these examples. 

\section{Details on our vectorization pipeline}

\begin{figure}[h]
    \centering
      \begin{subfigure}[b]{0.24\linewidth}
        \centering
        \includegraphics[width=\linewidth]{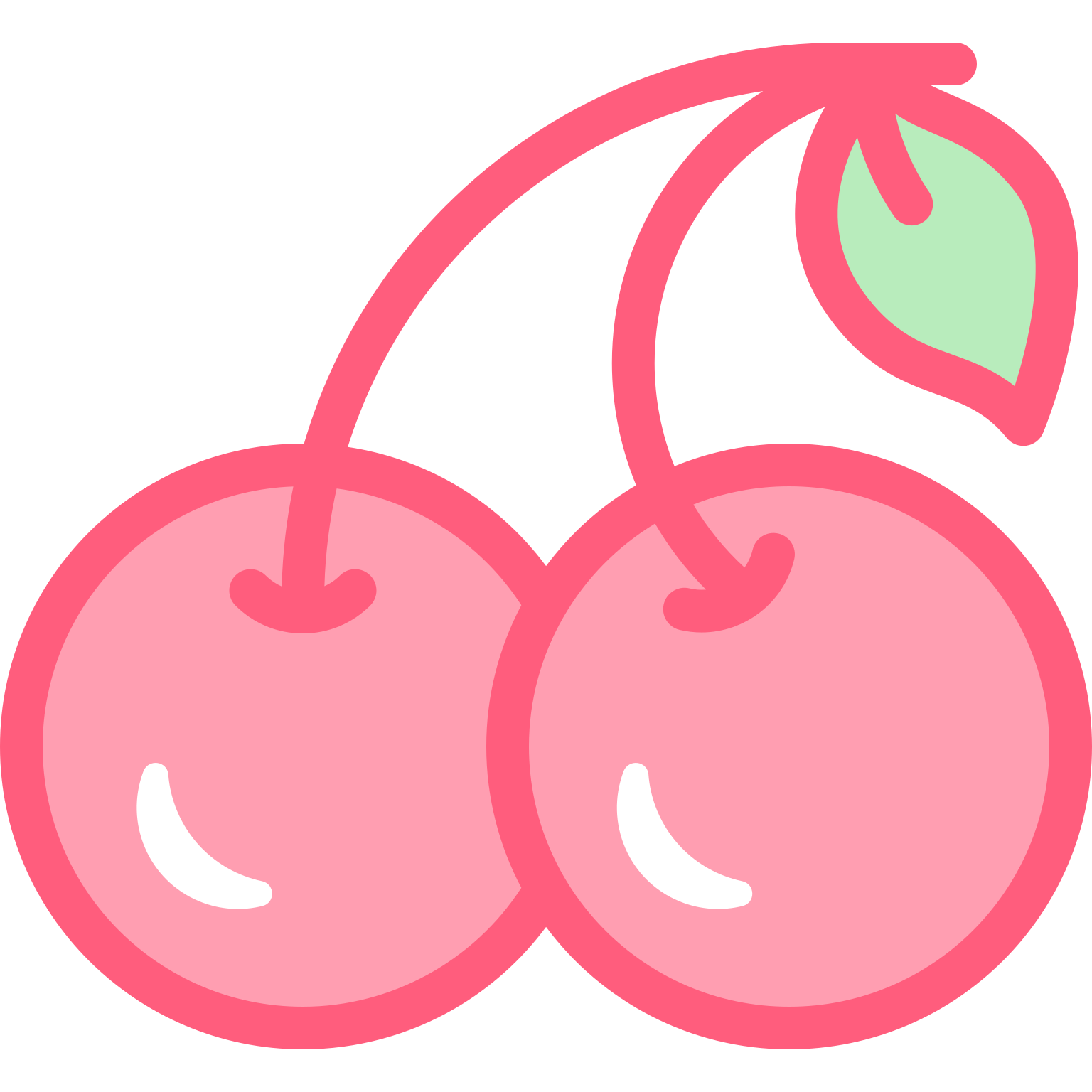}
        \caption{GT}
    \end{subfigure}
    \hfill
    \begin{subfigure}[b]{0.24\linewidth}
        \centering
        \includegraphics[width=\linewidth]{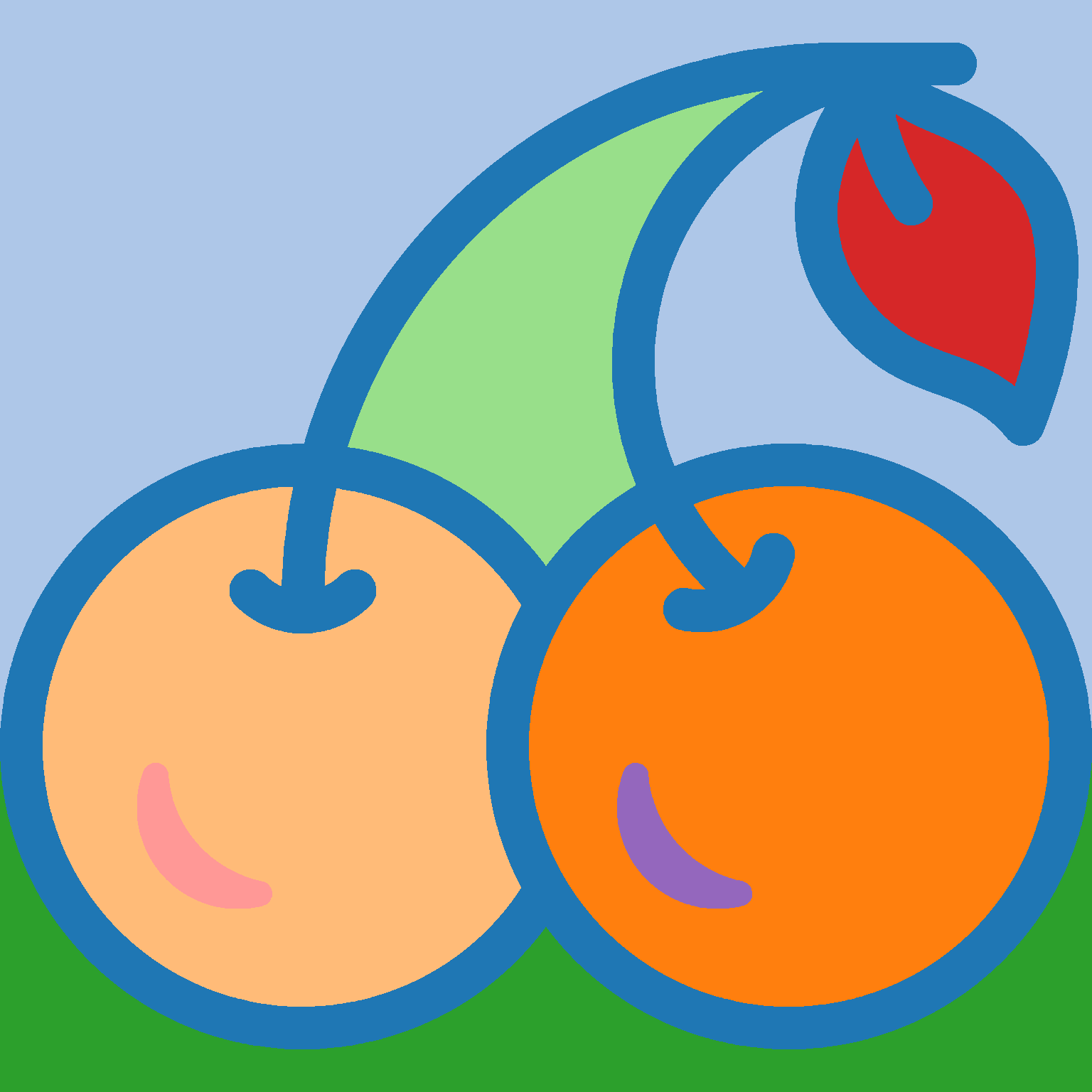}
        \caption{9 clusters}
        \label{fig:supp_vectorization_cluster}
    \end{subfigure}
    \hfill
      \begin{subfigure}[b]{0.24\linewidth}
        \centering
        \includegraphics[width=\linewidth]{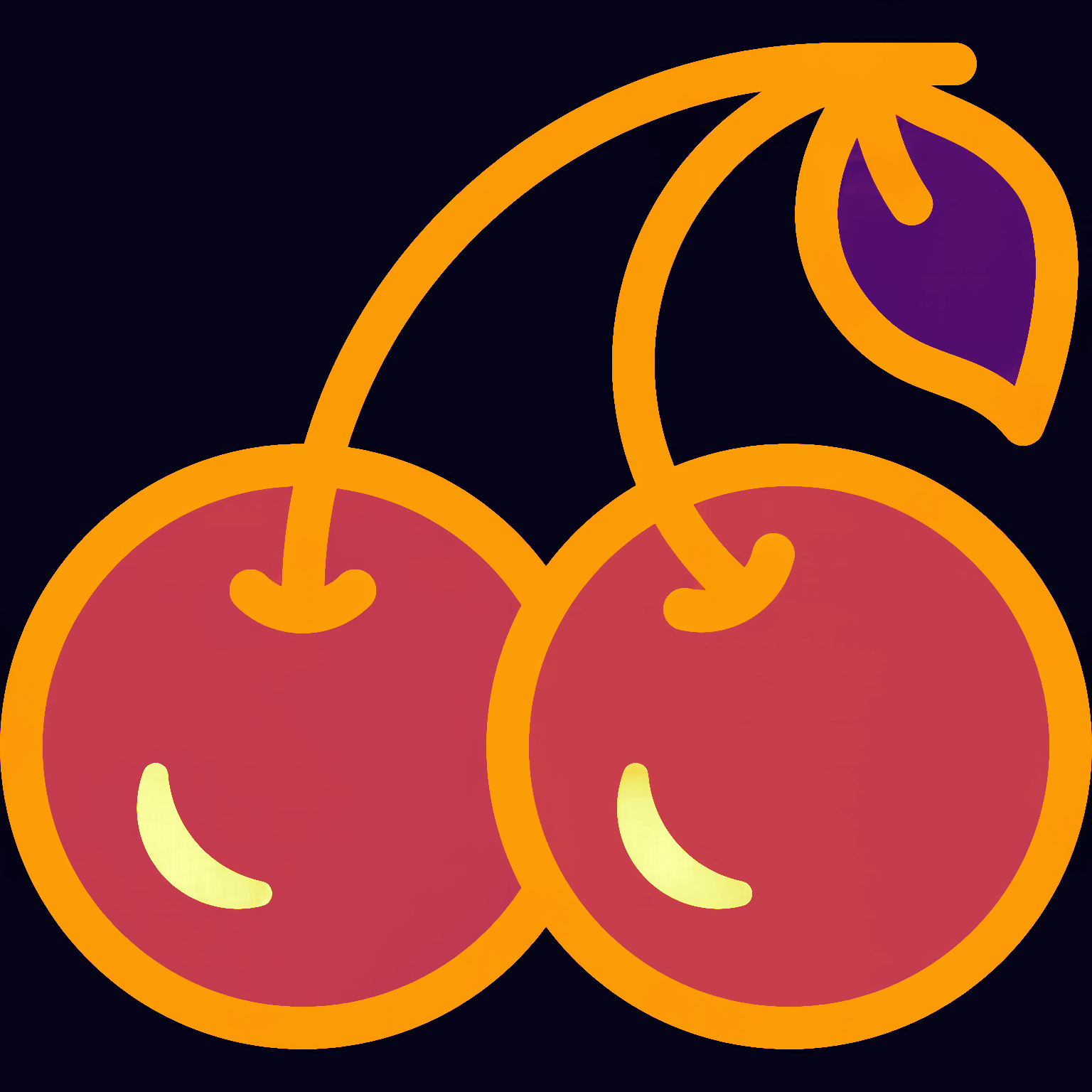}
        \caption{Ours}
        \label{fig:supp_vectorization_prediction}
    \end{subfigure}
    \hfill
    \centering
      \begin{subfigure}[b]{0.24\linewidth}
        \centering
        \includegraphics[width=\linewidth]{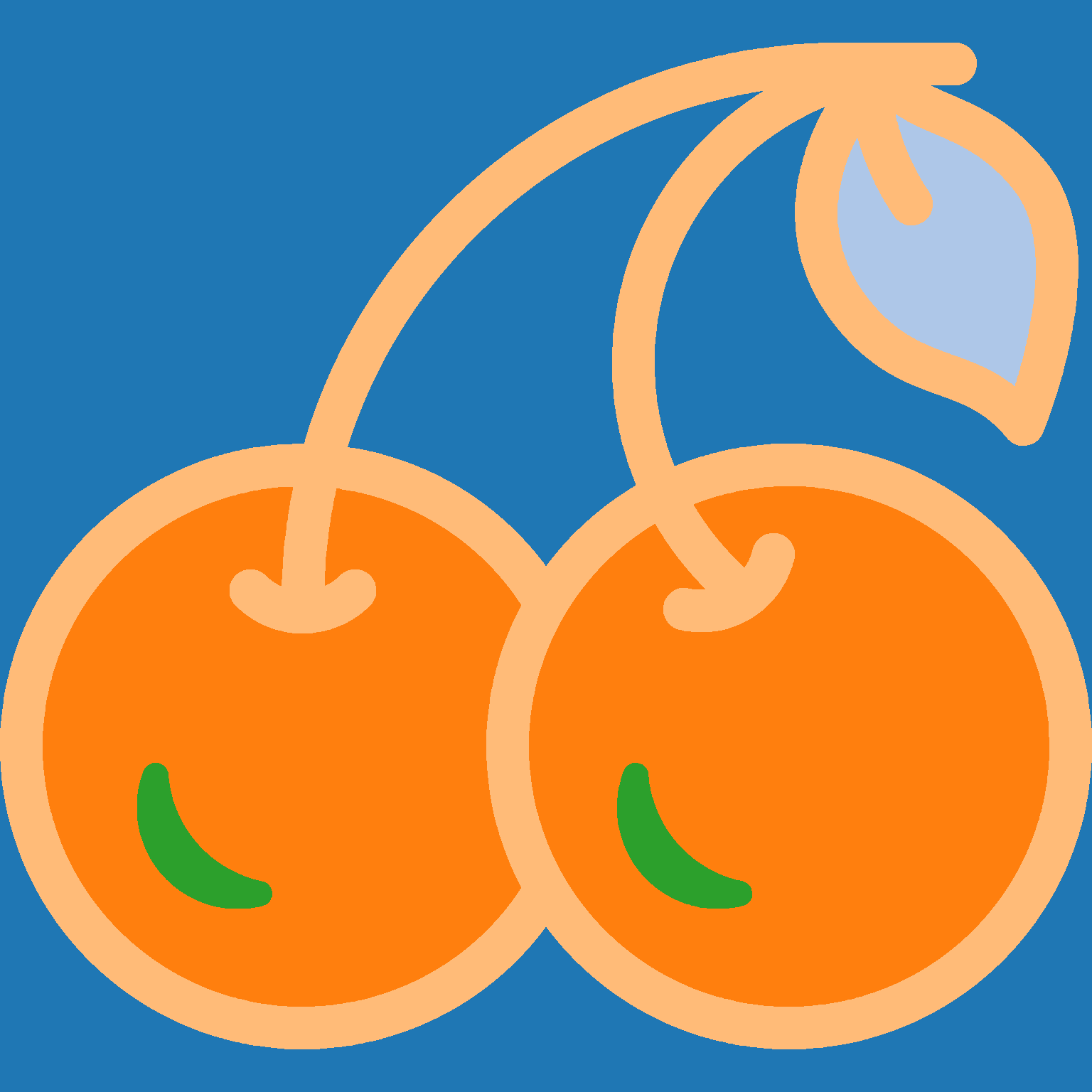}
        \caption{5 Clusters}
        \label{fig:supp_vectorization_simplification}
    \end{subfigure}
    \vspace*{-2mm}
    \caption{\textbf{Depth-aware clustering.} Given an input image (a), VTracer~\cite{pun_vtracer_2025} provides a list of color-constant clusters (b). We order these clusters based on our predicted depth map (c) and merge them to form the final decomposition (d).}
    \label{fig:supp_vectorization}
\end{figure}
Our vectorization tests were performed on a pipeline combining VTracer~\cite{pun_vtracer_2025} and Potrace~\cite{selinger_potrace_2003} with our contributions. Specifically, illustrator's depth based vectorization is achieved as follows: 
\begin{enumerate}
    \item We find color-constant clusters using VTracer (\cref{fig:supp_vectorization_cluster}). The values of several hyper-parameters are important, such as \emph{filter\_speckle} to suppress noise, \emph{color\_precision} and \emph{layer\_difference} to accurately split the image in distinct regions. All of them are provided in our code. 
    \item Instead of relying on VTracer's heuristics to sort the clusters, we leverage our predicted illustrator's depth (\cref{fig:supp_vectorization_prediction}) for layering, assigning the cluster's depth order to the median of the predicted depth for each cluster. 
    \item Cluster grouping is important to ensure a well-layered, compact output. After sorting, we further merge layers with neighboring indices if their RGB colors are within a certain threshold $\tau \!=\! 0.05$ in the $L^2$ norm. This results in an ordered clustering image $C \!\in\! [1, ...N]^{\scriptscriptstyle H\times W}$ (\cref{fig:supp_vectorization_simplification}). 
    \item While it doesn't affect the final rendering, filling holes and bridging gaps yields simpler, overlapping layers that are compact and easy to edit. Given a cluster with index $n$, we create a binary mask $\mathbf{1}_{C{[i,j]} > n}$, and inpaint the missing regions of $\mathbf{1}_{C{[i,j]}==n}$ using off-the-shelf algorithms (see~\cref{sec:supp_inpainting} and~\cref{fig:supp_inpainting}). 
    \item This layer collection is then vectorized with Potrace~\cite{selinger_potrace_2003} and assembled to form the final vector graphics. 
\end{enumerate}

\section{Inpainting}
\label{sec:supp_inpainting}
\begin{figure}[h]
    \centering
    \begin{subfigure}[b]{.49\linewidth}
        \centering
        \includegraphics[width=\linewidth]{figures/mde_comp/15.jpeg}
        \caption{GT}
    \end{subfigure}
    \hfill
    \begin{subfigure}[b]{.49\linewidth}
        \centering
        \includegraphics[width=\linewidth]{figures/mde_comp/15_ours.jpeg}
        \caption{Illustrator's Depth}
    \end{subfigure}
    
    \begin{subfigure}[b]{.49\linewidth}
        \centering
        \includegraphics[width=\linewidth]{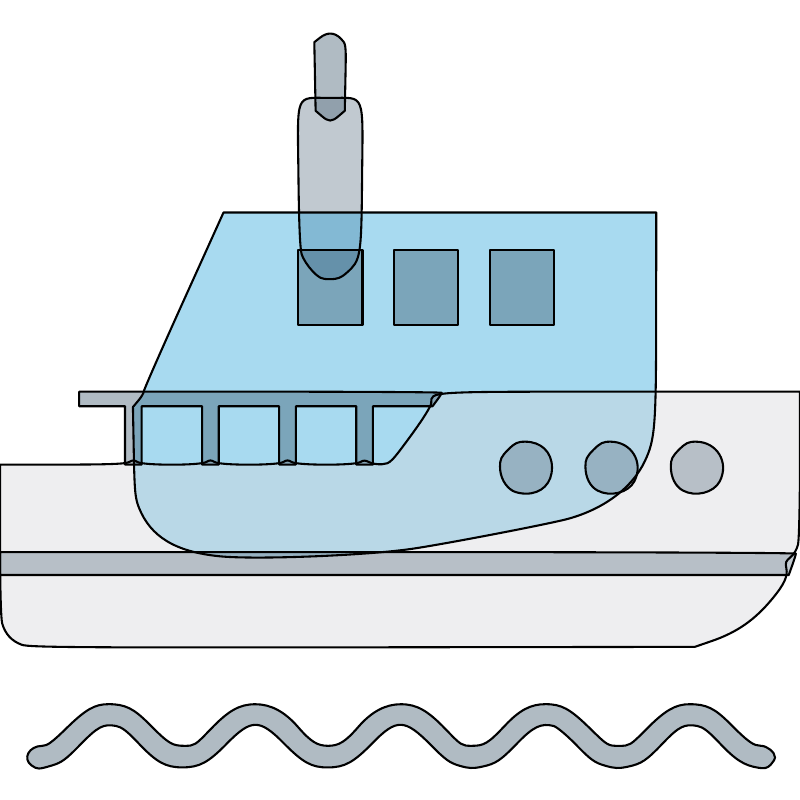}
        \caption{Scikit Biharmonic Inpainting}
    \end{subfigure}
    \hfill
    \begin{subfigure}[b]{.49\linewidth}
        \centering
        \includegraphics[width=\linewidth]{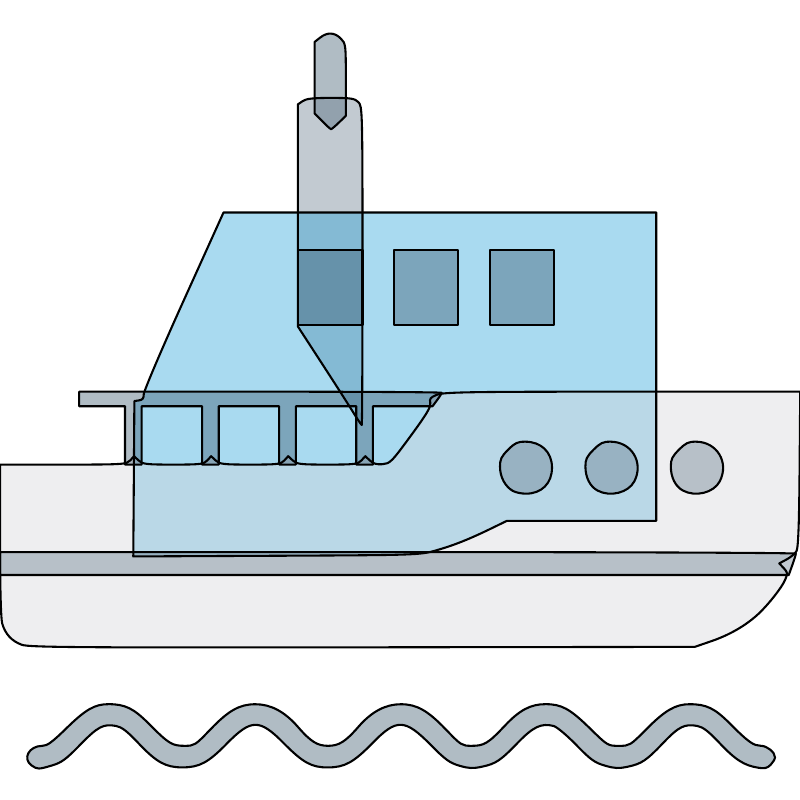}
        \caption{Scikit Closest Point Inpainting}
    \end{subfigure}
    \caption{\textbf{Inpainting with Scikit~\protect\cite{van2014scikit}.}~Using the boat example (a) from~\cref{fig:mde_comp}, we display two examples of the same layer-wise decomposition produced by our method (b), where inpainting is done via a biharmonic (c) or closest point (d) variant.}
    \label{fig:supp_inpainting}
\end{figure}

While not part of our contributions, we also show examples of inpainting strategies once our layer index prediction has been generated, see~\cref{fig:supp_inpainting}. For vector graphics, we rely on fast, off-the-shelf algorithms provided by Scikit-image~\cite{van2014scikit}. We experimented with two variants in order to fill the missing regions: one that interpolates using the nearest unmasked point, and another based on biharmonic interpolation. Depending on the application, users may prefer one approach over the other: the biharmonic method produces smoother curves, whereas the closest-point interpolation yields sharper, crisper boundaries (see~\cref{fig:supp_inpainting}). We generally use the latter in our code due to its faster computational time. 
While this simple hole-filling approach is sufficient for most vector graphics, data-driven inpainting may be desired for more involved applications, including raster image editing: here again, leveraging off-the-shelf inpainting models (see~\cref{fig:layer_decomp}) offers a solution that doesn't require any additional training.

\section{Prompts for vector-styled images}
\label{sec:supp_prompts}

The main paper shows two text-to-image examples, one using FLUX~\cite{labs_flux1_2025} and one using Nano Banana~\cite{google_gemini_2025}.
For FLUX (\cref{fig:t2v_comparison} in the main paper), we found that, given a desired object to be drawn (underlined below), a mix of positive and negative prompts provides clean vector-styled raster images that are easy to process with our pipeline; for instance,
\begin{lstlisting}
{  "prompt": "Vector graphics of a simple (*@\underline{cheetah head}@*).",
  "prompt_2": "Vector graphics of a simple (*@\underline{cheetah head}@*). SVG file. Filled shapes, minimalist design. Abstract.",
  "negative_prompt": "Gradient, 3D. Small details. Fineline details.",
  "negative_prompt_2": "Gradient, 3D. Small details. Fineline details.",
  "num_inference_steps": 28,
  "num_images_per_prompt": 1
}
\end{lstlisting}
For Nano Banana (\cref{fig:t2vec_workflow} in the main paper), we simply prompt the model through:
\begin{lstlisting}
{  "prompt": "Vector graphic illustration of a (*@\underline{cat}@*). SVG style, (*@\underline{blue}@*) background. Smooth, flowy shapes."
}
\end{lstlisting}

\section{Comparison with Text2Vector Generators}

\begin{figure}[t]
    \centering
    \begin{subfigure}[b]{0.49\linewidth}
        \centering
        \includegraphics[width=\linewidth]{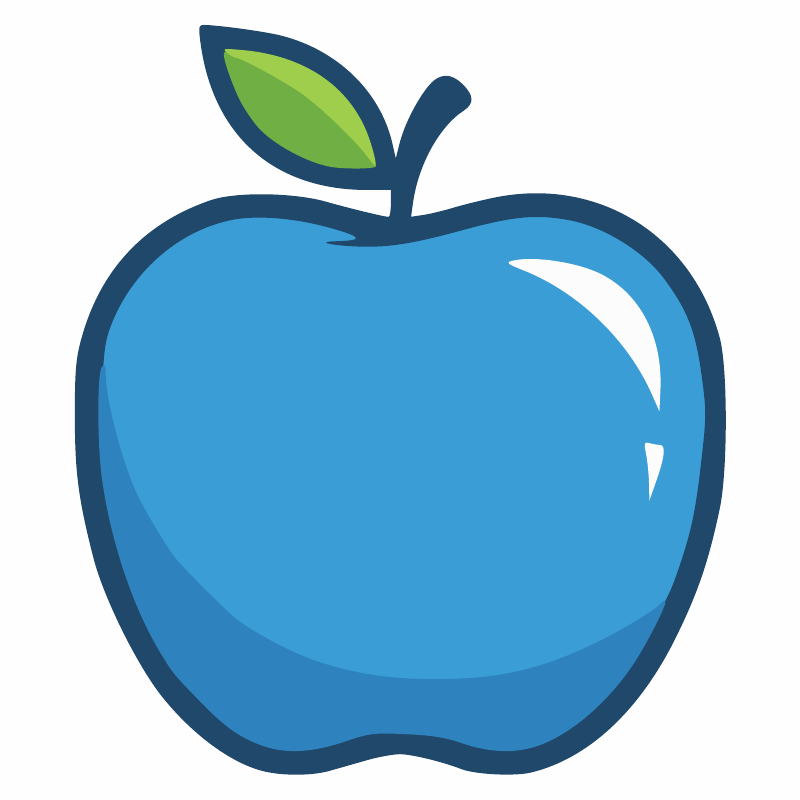}
        \caption{Ours + \cite{google_gemini_2025}}
    \end{subfigure}
    \hfill
    \begin{subfigure}[b]{0.49\linewidth}
        \centering
        \includegraphics[width=\linewidth]{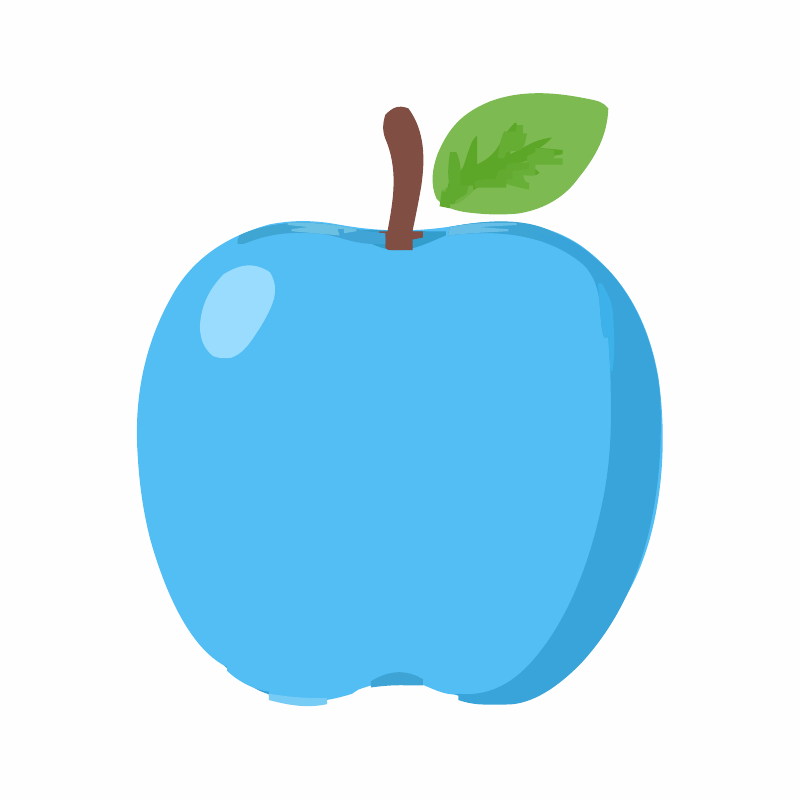}
        \caption{LayerTracer~\cite{song_layertracer_2025}}
    \end{subfigure}
    \hfill

    \begin{subfigure}[b]{0.49\linewidth}
        \centering
        \includegraphics[width=\linewidth]{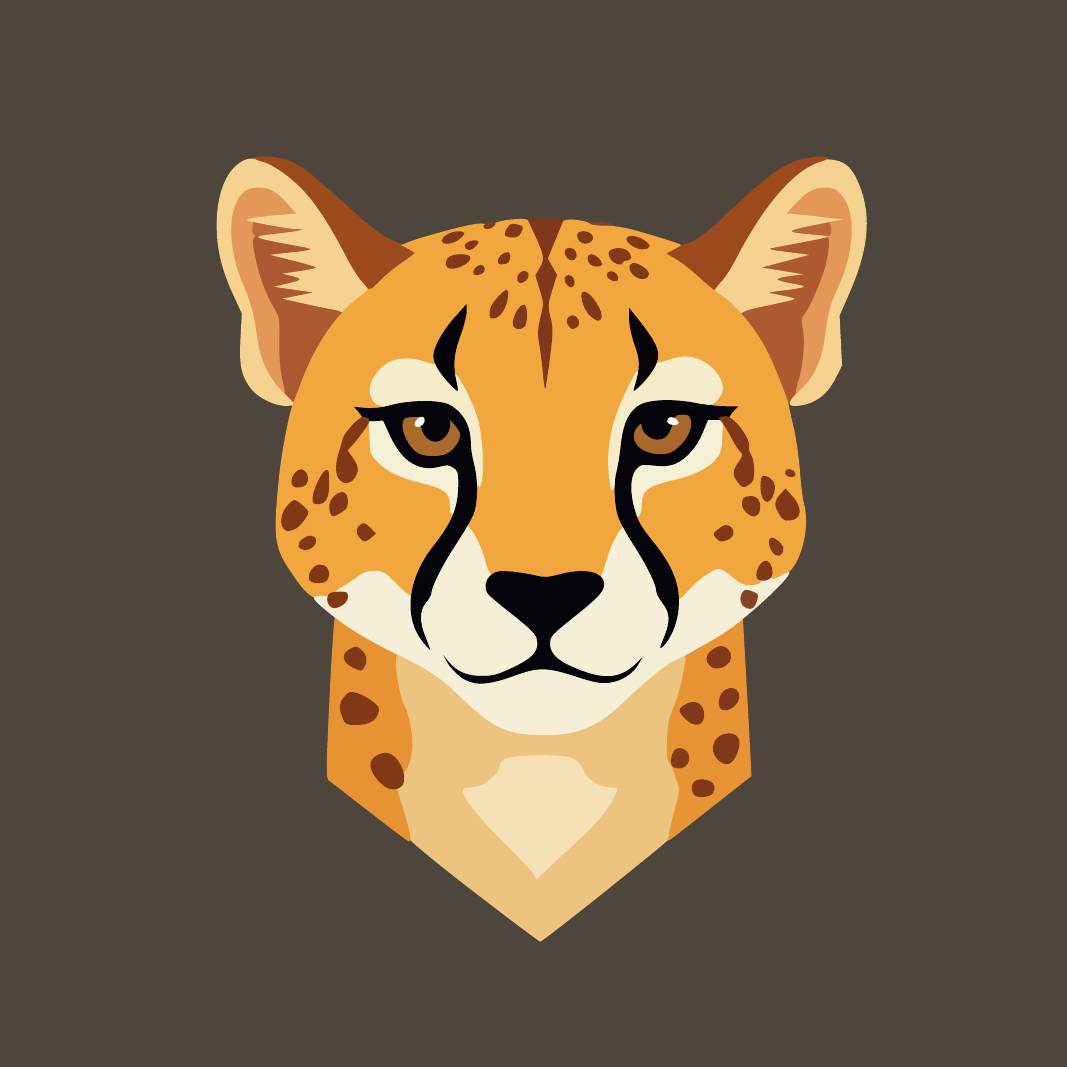}
        \caption{Ours + \cite{labs_flux1_2025}}
    \end{subfigure}
    \hfill
    \begin{subfigure}[b]{0.49\linewidth}
        \centering
        \includegraphics[width=\linewidth]{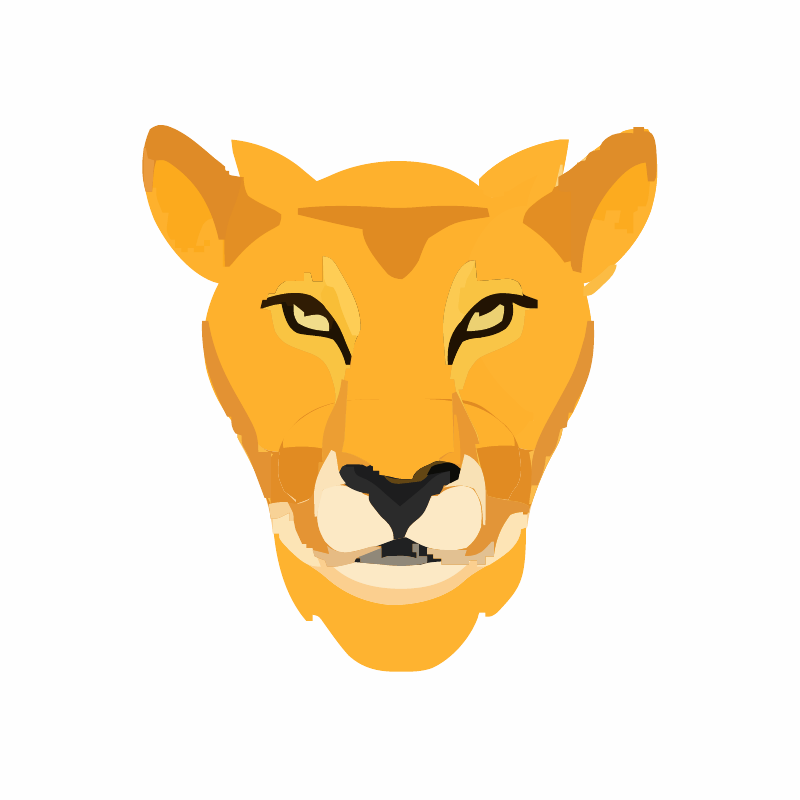}
        \caption{LayerTracer~\cite{song_layertracer_2025}}
    \end{subfigure}
    \hfill
    
    \begin{subfigure}[b]{0.49\linewidth}
        \centering
        \includegraphics[width=\linewidth]{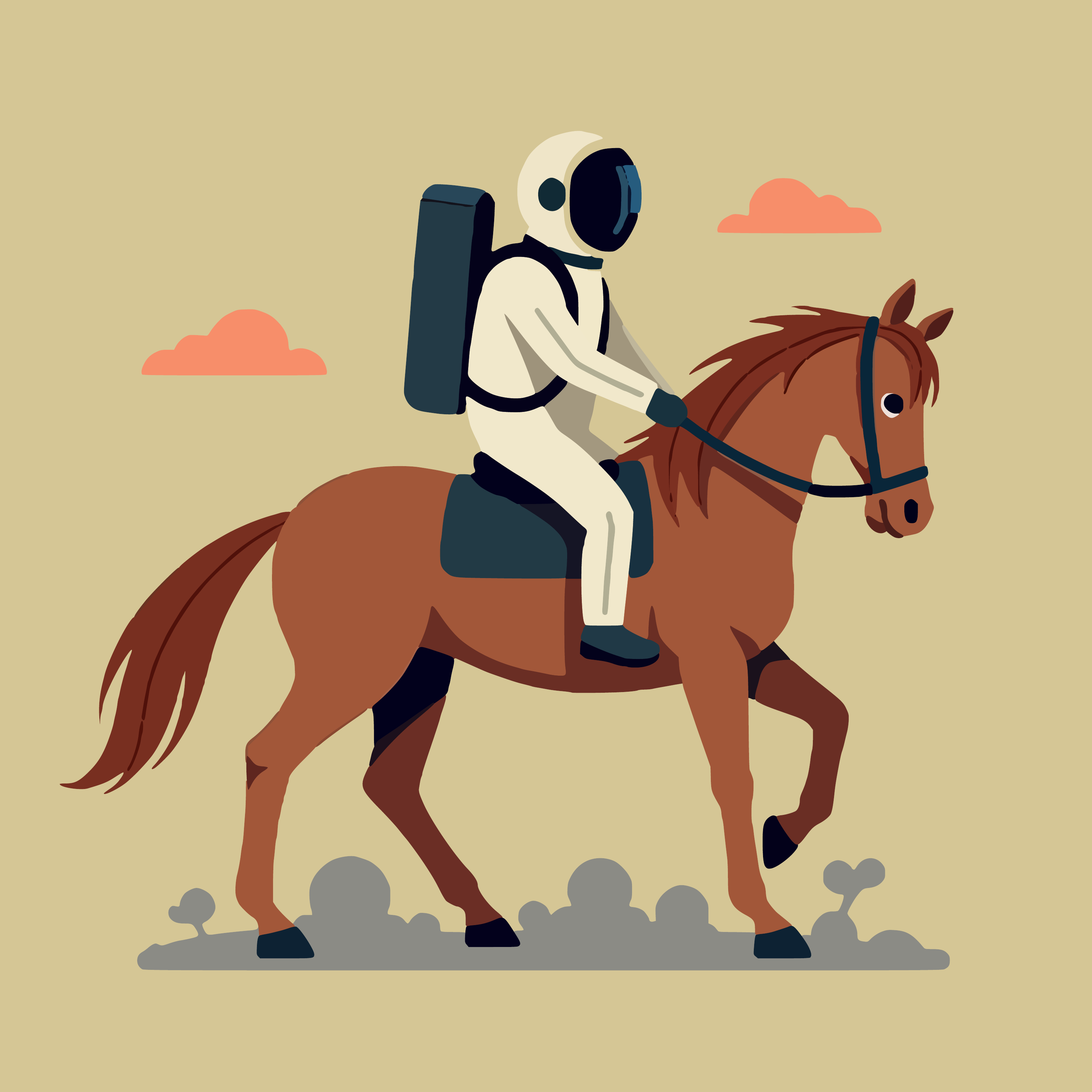}
        \caption{Ours + \cite{labs_flux1_2025}}
    \end{subfigure}
    \hfill
    \begin{subfigure}[b]{0.49\linewidth}
        \centering
        \includegraphics[width=\linewidth]{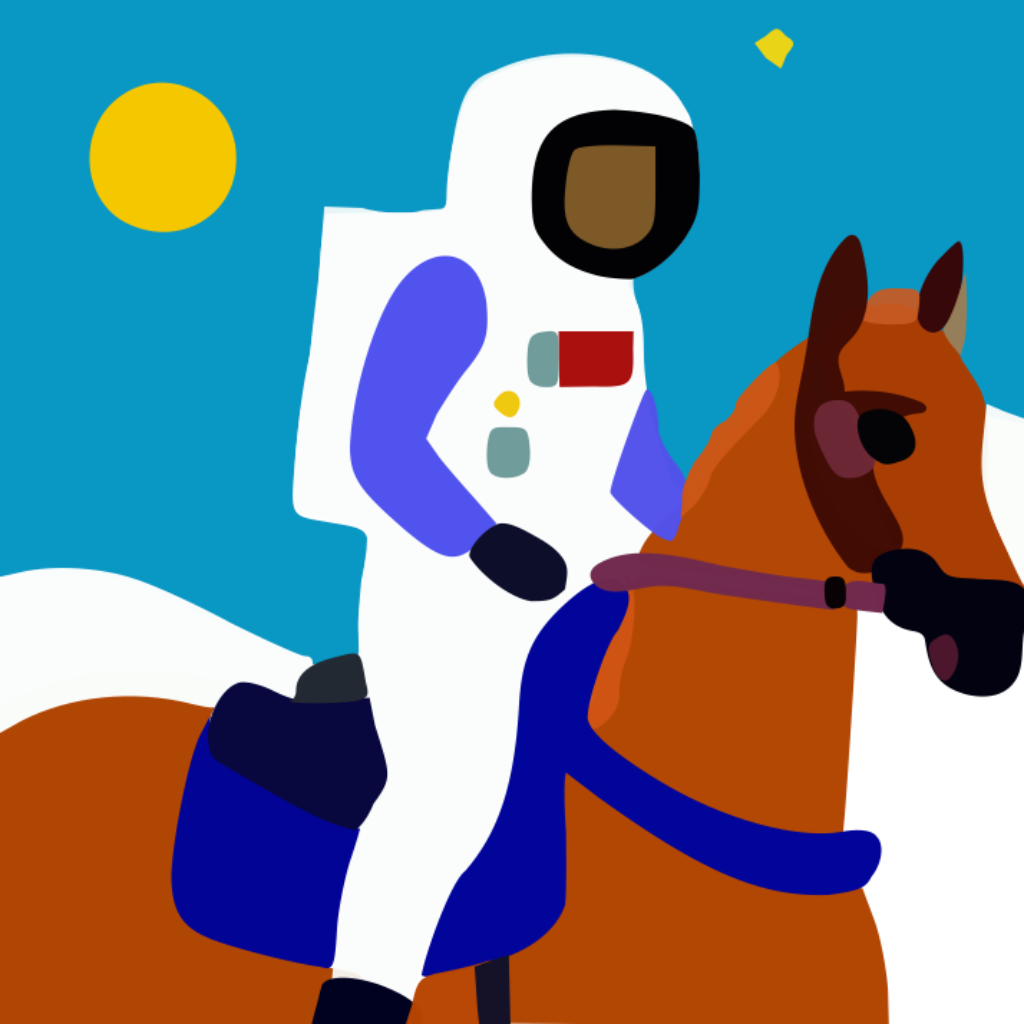}
        \caption{Neural Paths\cite{zhang_text--vector_2024}}
    \end{subfigure}

    \begin{subfigure}[b]{0.49\linewidth}
        \centering
        \includegraphics[width=\linewidth]{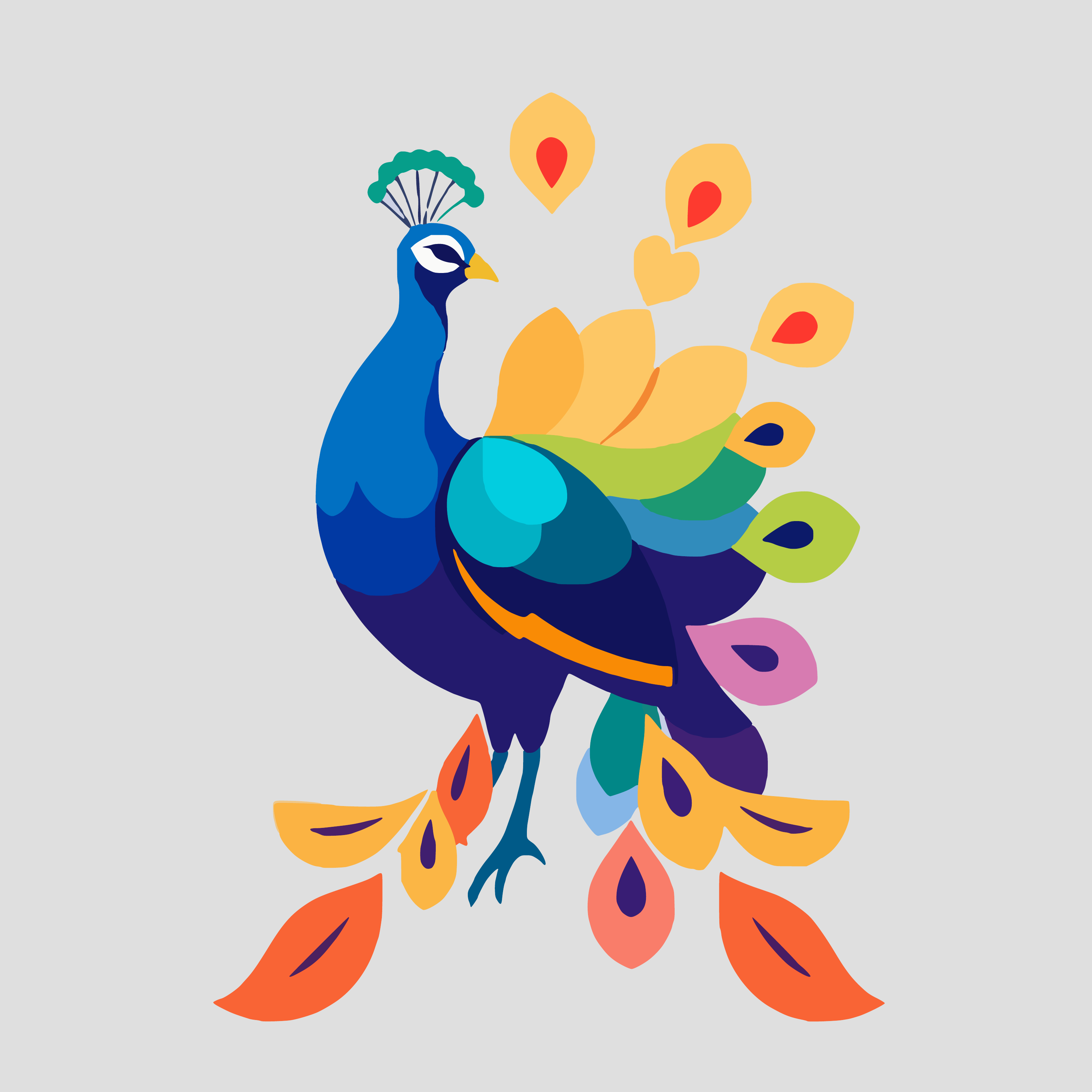}
        \caption{Ours + \cite{labs_flux1_2025}}
    \end{subfigure}
    \hfill
    \begin{subfigure}[b]{0.49\linewidth}
        \centering
        \includegraphics[width=\linewidth]{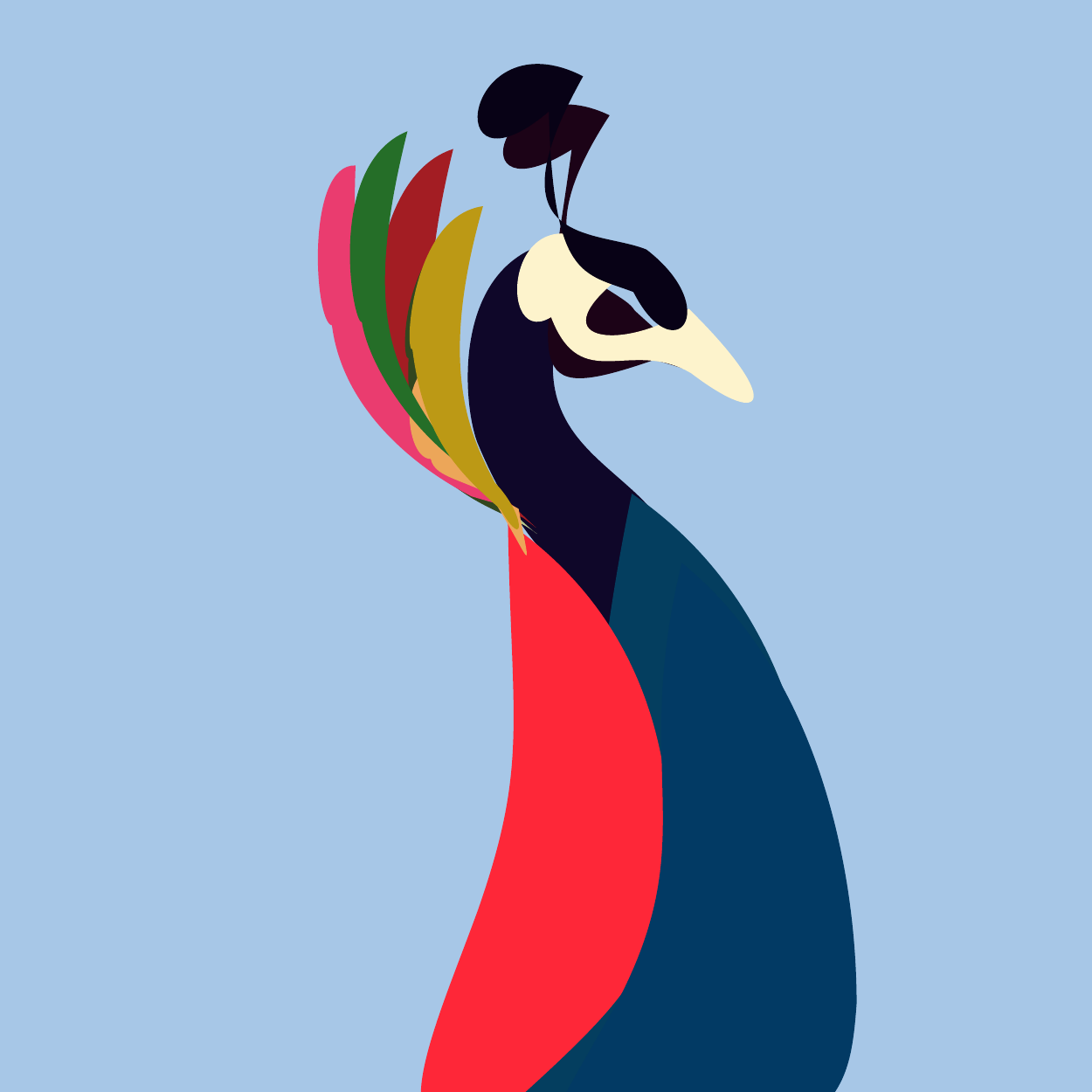}
        \caption{NeuralSVG~\cite{polaczek_neuralsvg_2025}}
    \end{subfigure}
    \caption{\textbf{Text2Vector models.}~Pairing Illustrator's Depth with powerful image generative models produces more complex and detailed illustrations than current text-to-vector diffusion models. In our results (left column), the models are prompted as described in~\cref{sec:supp_prompts} using ``a blue apple", ``a cheetah head", ``an astronaut riding a horse", and ``a colorful peacock". \vspace*{-3mm}}
    \label{fig:supp_t2vec}
\end{figure}

In addition to the results discussed in 
\cref{sec:texttovector}
of the main paper, more examples of text-to-vector-graphics generations are given in~\cref{fig:supp_t2vec}. Both text-to-vector generations using Neural Path Representations~\cite{zhang_text--vector_2024} and NeuralSVG~\cite{polaczek_neuralsvg_2025} are based on Score Distillation Sampling (SDS) that relies on a pretrained diffusion model to backpropagate gradients to B\'ezier curve parameters. Consequently, their generated illustrations  are relatively simple and lack fine details (we reproduce the images provided in their articles in \cref{fig:supp_t2vec}). Although LayerTracer~\cite{song_layertracer_2025} employs its own custom diffusion model, it exhibits similar limitations, producing simple emoji-like graphics; note that we used the prompting setup provided in their public repository. In contrast, our method can decompose any output into layered SVG representations, effectively decoupling generation from vectorization --- and thus fully leveraging the capabilities of modern generative models. Our modular pipeline, compatible with both Flux~\cite{labs_flux1_2025} and Nano Banana~\cite{google_gemini_2025}, produces detail-rich vector illustrations within seconds (see \cref{sec:supp_prompts} for the full prompt configurations).

\section{Failure cases}

\begin{figure}[h]
    \centering
    \begin{subfigure}[b]{0.495\linewidth}
        \centering
        \includegraphics[width=\linewidth]{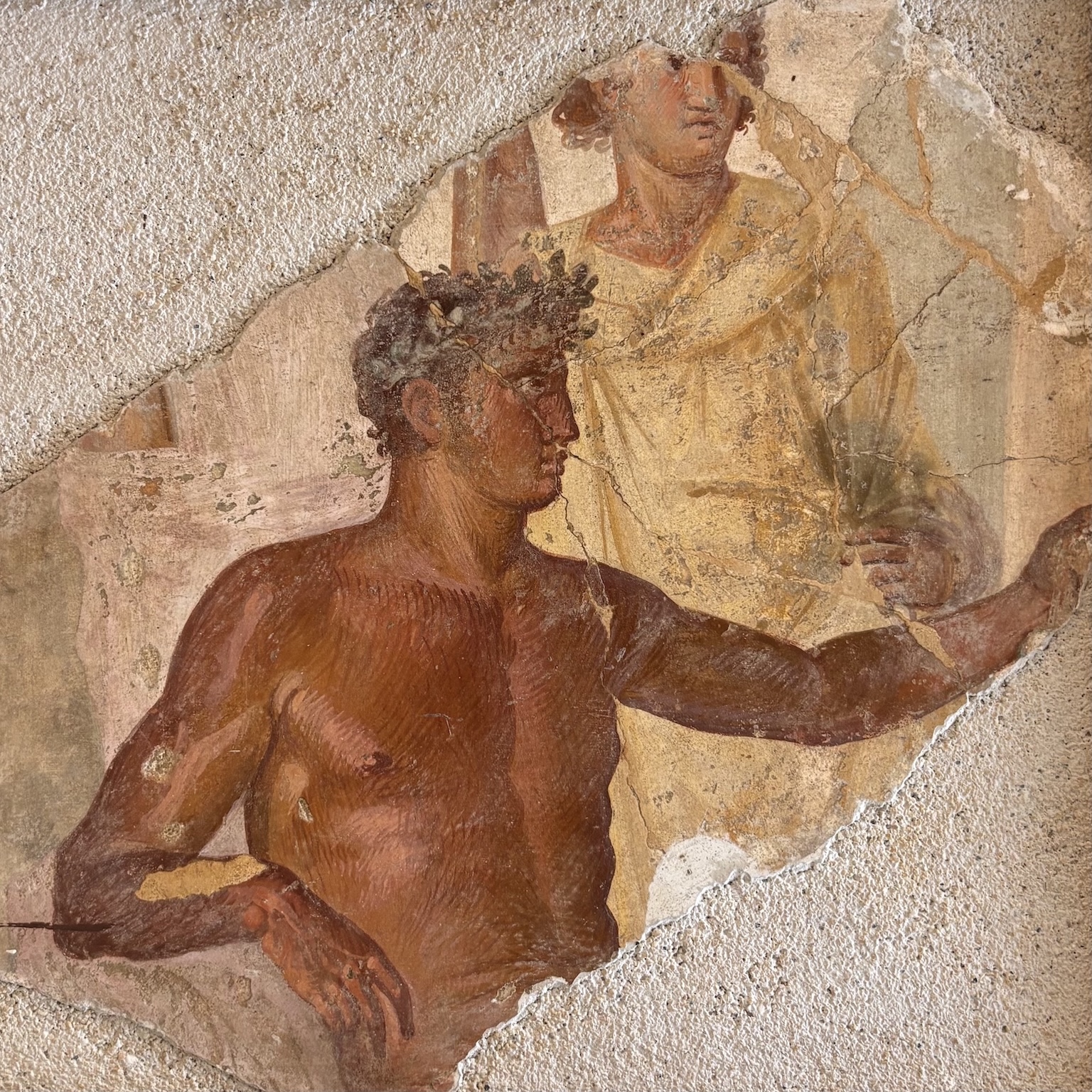}
        \caption{Input Image}
    \end{subfigure}
    \hfill
    \begin{subfigure}[b]{0.495\linewidth}
        \centering
        \includegraphics[width=\linewidth]{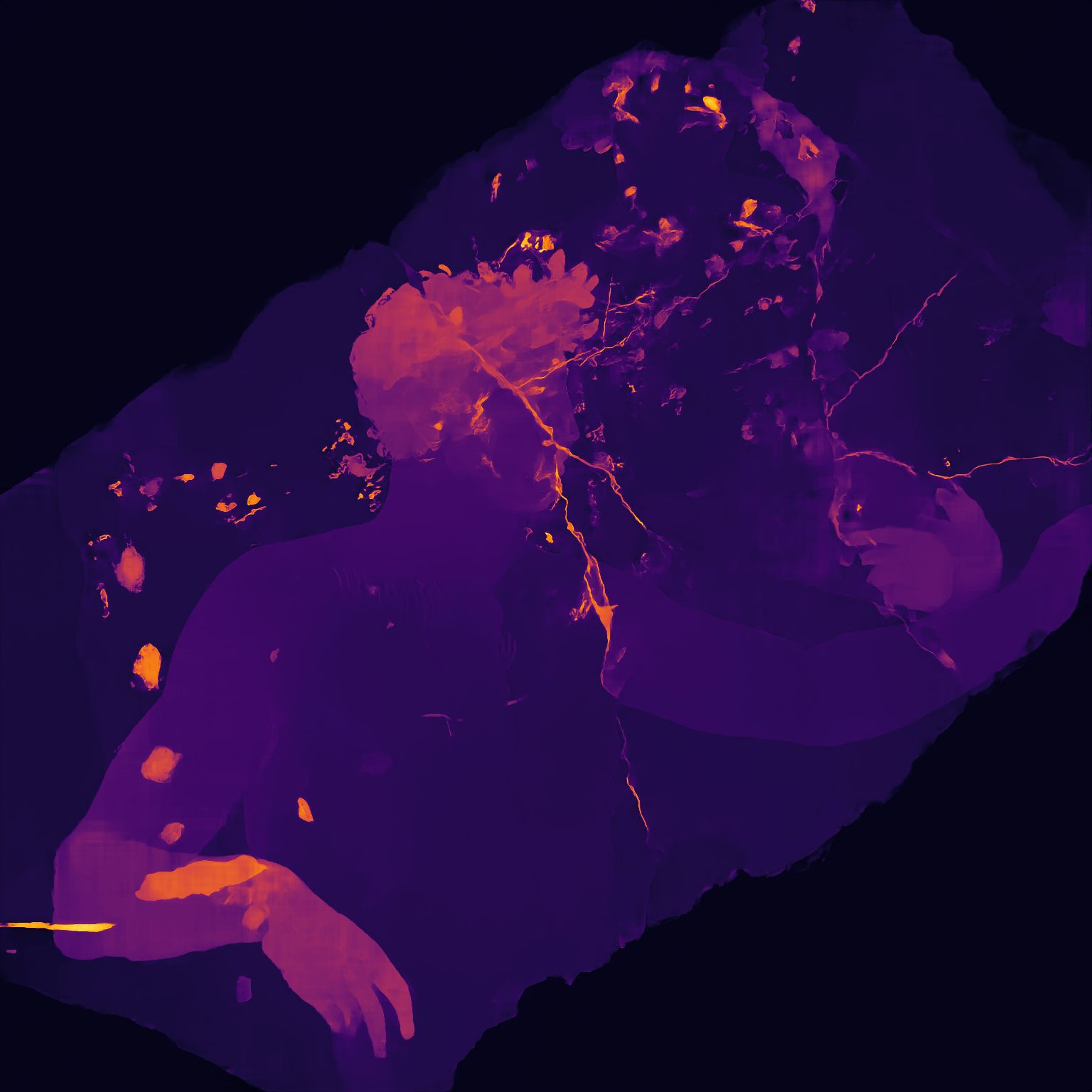}
        \caption{Illustrator's Depth}
    \end{subfigure}
    \hfill
    \begin{subfigure}[b]{0.495\linewidth}
        \centering
        \includegraphics[width=\linewidth]{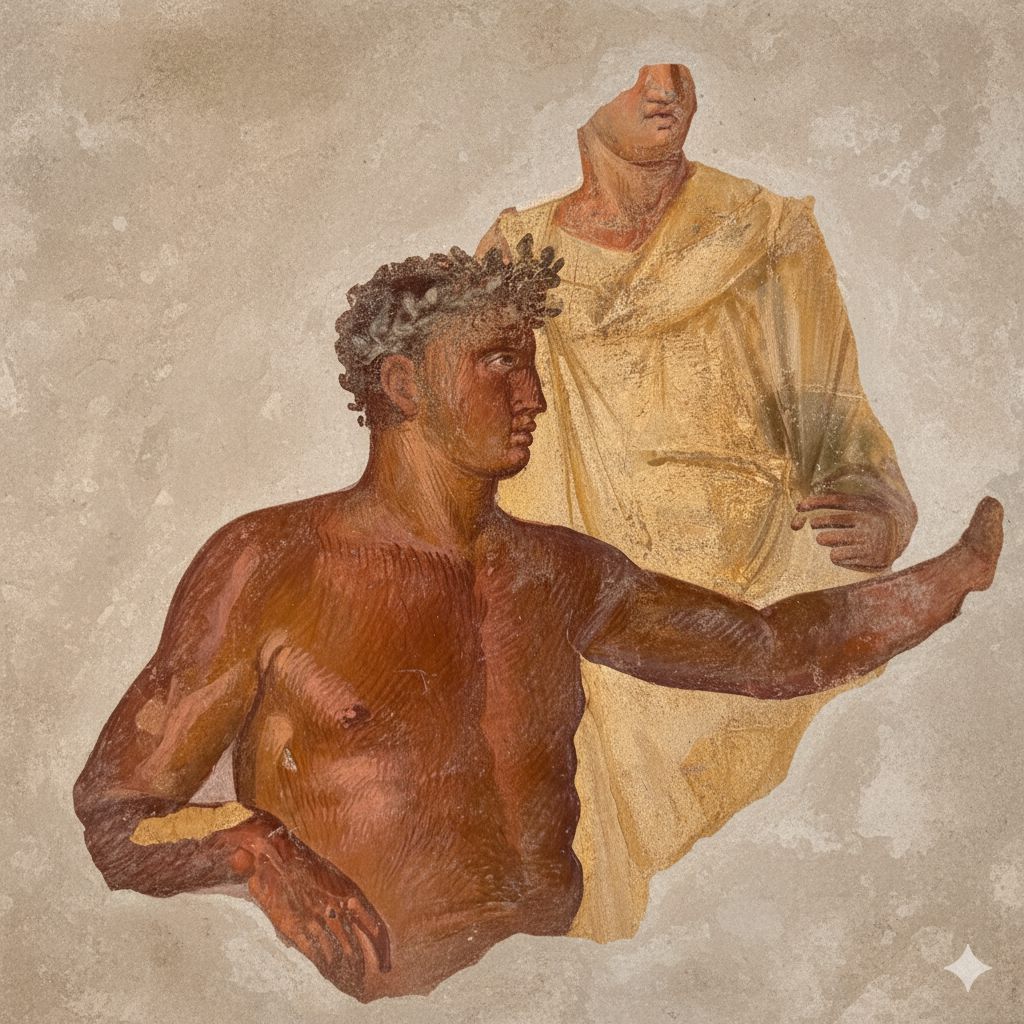}
        \caption{Input Image (cleaned)}
    \end{subfigure}
    \hfill
    \begin{subfigure}[b]{0.495\linewidth}
        \centering
        \includegraphics[width=\linewidth]{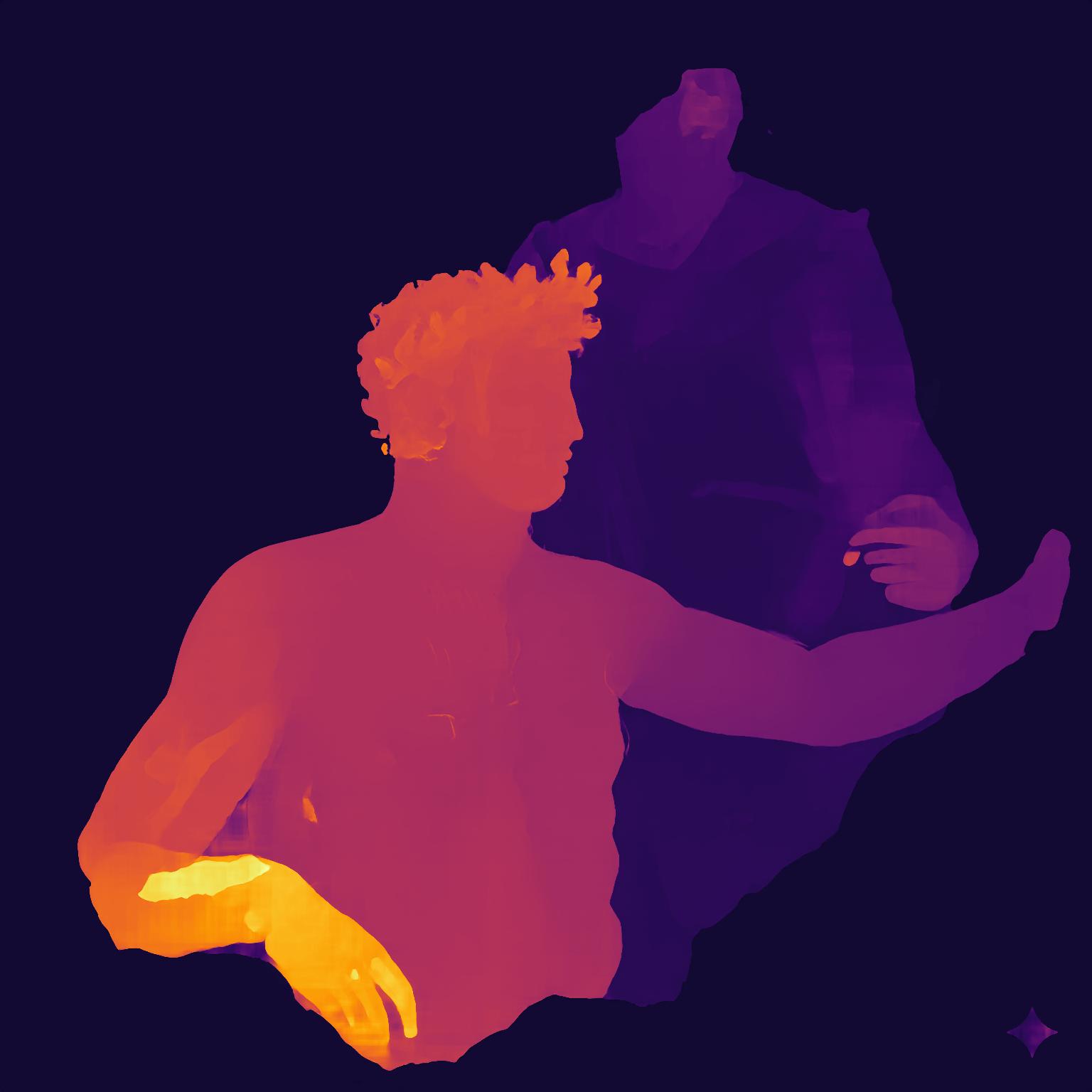}
        \caption{Illustrator's Depth}
    \end{subfigure}

    \caption{\textbf{Sensitivity to texture.}~The fresco (top left) contains several missing regions and cracks, which our model identifies as foreground elements (top right). If these artifacts are undesired, one can first use Nano Banana~\cite{google_gemini_2025} to remove defects, then reapply our model to obtain a cleaner result.}
    \label{fig:suppl_failure}
\end{figure}

\paragraph{Texture artifacts}~Since our model is trained on clean SVG data, a failure case arises when the input image contains canvas textures or defects. These issues can be easily mitigated by using a generative model (e.g., Nano Banana~\cite{google_gemini_2025}) to clean the image before applying our method (see~\cref{fig:suppl_failure}).

\begin{figure}[h]
    \centering
    \begin{subfigure}[b]{0.495\linewidth}
        \centering
        \includegraphics[width=\linewidth]{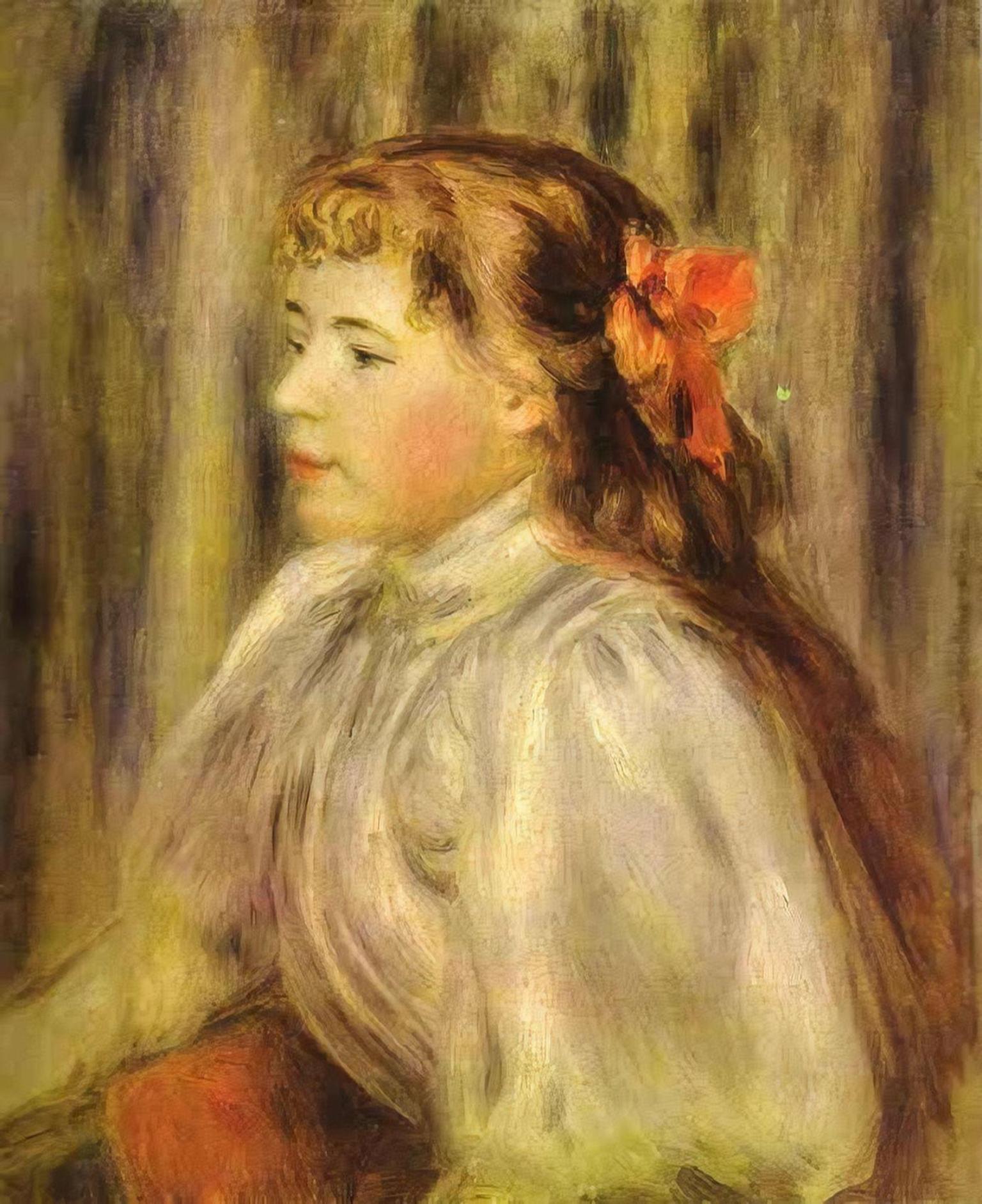}
    \end{subfigure}
    \hfill
    \begin{subfigure}[b]{0.495\linewidth}
        \centering
        \includegraphics[width=\linewidth]{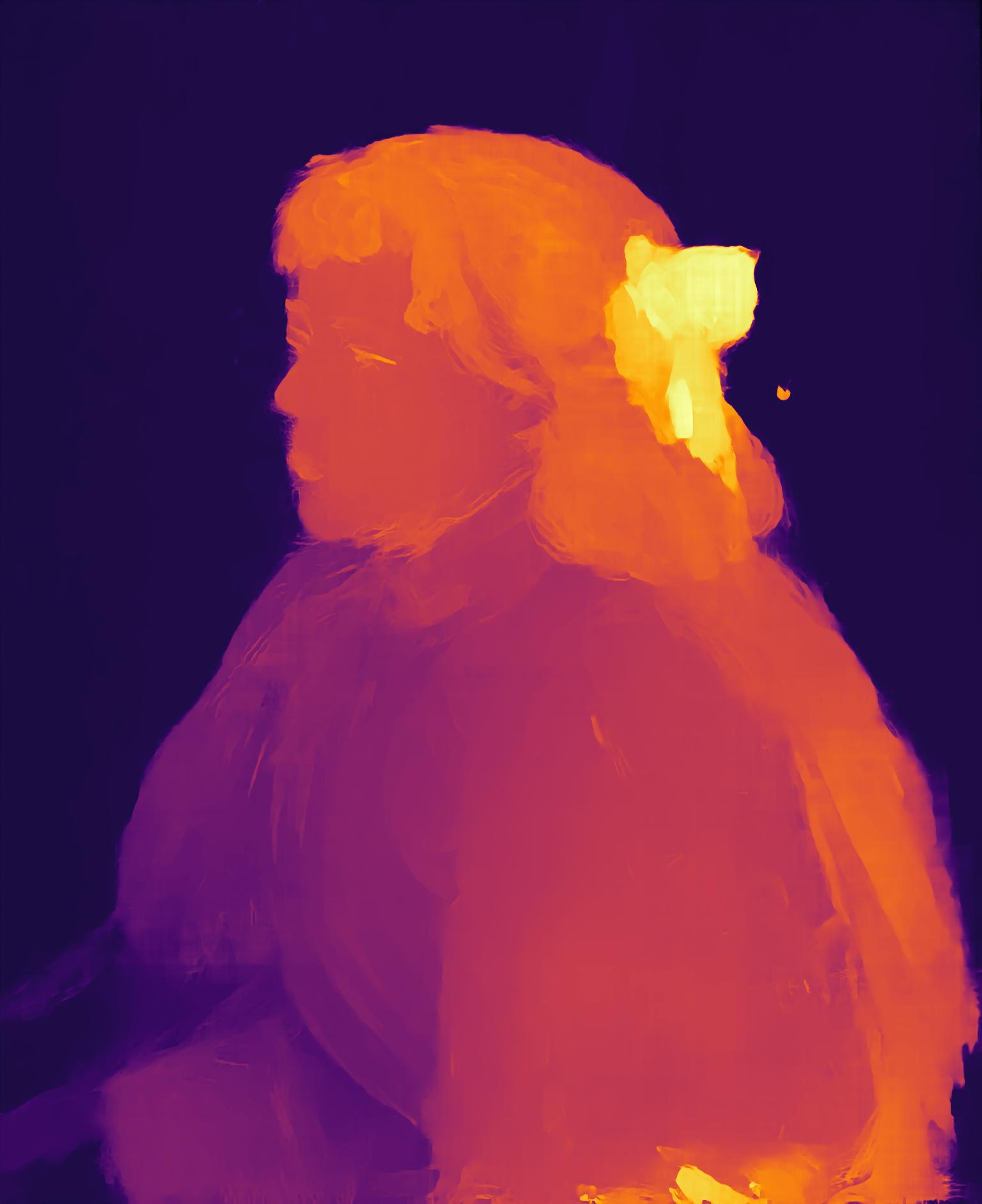}
    \end{subfigure}
    \hfill
    \begin{subfigure}[b]{0.495\linewidth}
        \centering
        \includegraphics[width=\linewidth]{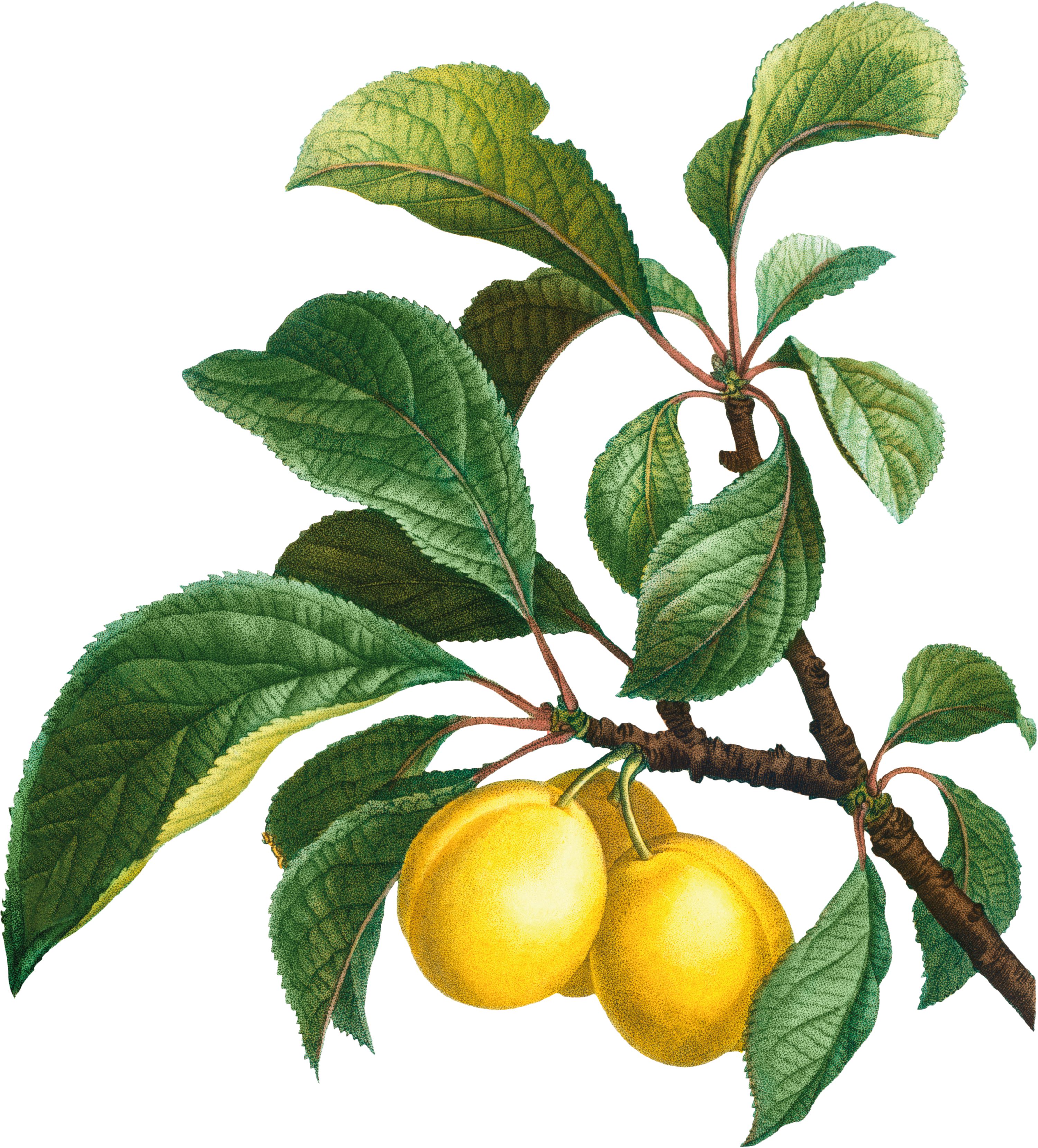}
        \caption{Input Image}
    \end{subfigure}
    \hfill
    \begin{subfigure}[b]{0.495\linewidth}
        \centering
        \includegraphics[width=\linewidth]{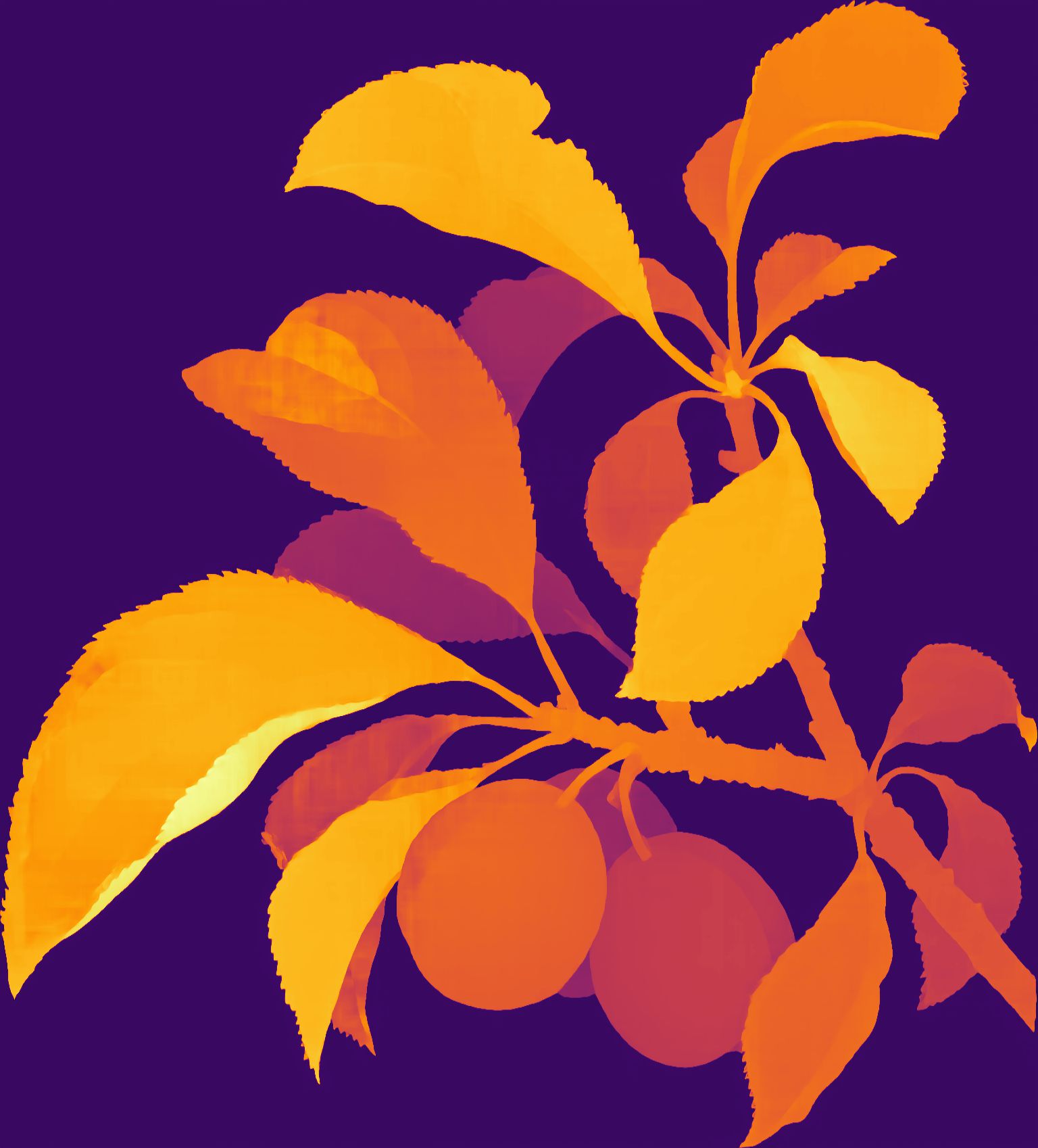}
        \caption{Illustrator's Depth}
    \end{subfigure}

    \caption{\textbf{Failure cases.}~Our models can ignore the background elements, such as the stripes (top row), or incorrectly predict the illustrator's depth: in the bottom row, the leftmost plum should be on top of the leaf rather than behind.}
    \label{fig:suppl_failure2}
\end{figure}

\paragraph{Incorrect ordering}~Like any machine learning model, ours can occasionally make mistakes (see~\cref{fig:suppl_failure2}, bottom row). Quantitatively, such errors are rare: as shown in~\cref{tab:mde_comp}, over $98\%$ of randomly sampled pixel pairs are correctly ordered in our experiment with MMSVG.

\paragraph{Foreground Focus}~Our training set primarily contains single objects over white backgrounds. Consequently, the model sometimes neglects background elements, which may be undesirable in certain scenarios (see~\cref{fig:suppl_failure2}, top row). Future work could address this limitation by training on more complex or synthetic SVG datasets that include background elements.


\section{Depth ordering consistency metric}

To compute the depth ordering consistency from a ground-truth illustrator's depth map $D$ and a predicted map $D_{\theta}$, we adapt the approach of~\cite{zhang_monocular_2015} and proceed as follows:

\begin{enumerate}[labelsep=0.4ex]
\item we uniformly sample $\frac{H \times W}{50}$ random pairs of pixel locations $(i,j)$ and $(k,l)$ and keep only those corresponding to two different layers in $D$, i.e., such that $D[i,j] \!\neq\! D[k,l]$; 
\item we then check whether the relative ordering is preserved by comparing the signs of $(D[i,j] \!-\! D[k, l])$ and $(D_{\theta}[i,j]\!-\!D_{\theta}[k,l])$.
\item Finally, we compute the average consistency score $\bar{s}$ over all pairs by the ratio of preserved ordering over total number of pixel pairs.   
\end{enumerate}

\begin{figure}[!h]
    \centering     \includegraphics[width=1.0\linewidth]{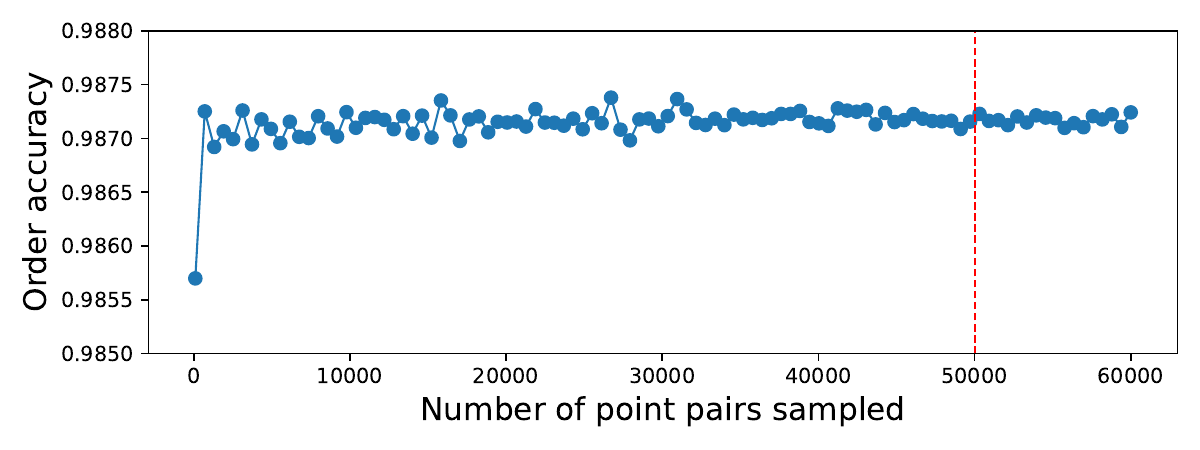}
    \vspace*{-6mm}
\caption{\textbf{Stability of order metric.} We plot the order metric when sampling one of our results from $100$ to $60$K points.}
    \label{fig:suppl_order}
\end{figure}

This formulation quantifies how effectively the predicted illustrator's depth maintains correct relative depths, independent of absolute scale. This metric, inherently stochastic as it relies on randomly sampled pixel pairs from the image, exhibits strong stability: sampling $50,000$ pairs on $1536 \times 1536$ images yielded no significant variations in our experiments (see~\cref{fig:suppl_order}). And as Tabs.~\ref{tab:mde_comp}-\ref{tab:svg_depth} from the original paper demonstrate, it offers a complementary measure of layering quality.

\section{Quantitative Evaluation on Natural Images} 
\label{sec:sup_quantitative_paintings}
\begin{figure}[h]
    \centering
    \captionsetup[subfigure]{justification=centering, singlelinecheck=false,aboveskip=1pt}
    \begin{subfigure}[b]{0.235\linewidth}
        \centering
        \includegraphics[width=\linewidth]{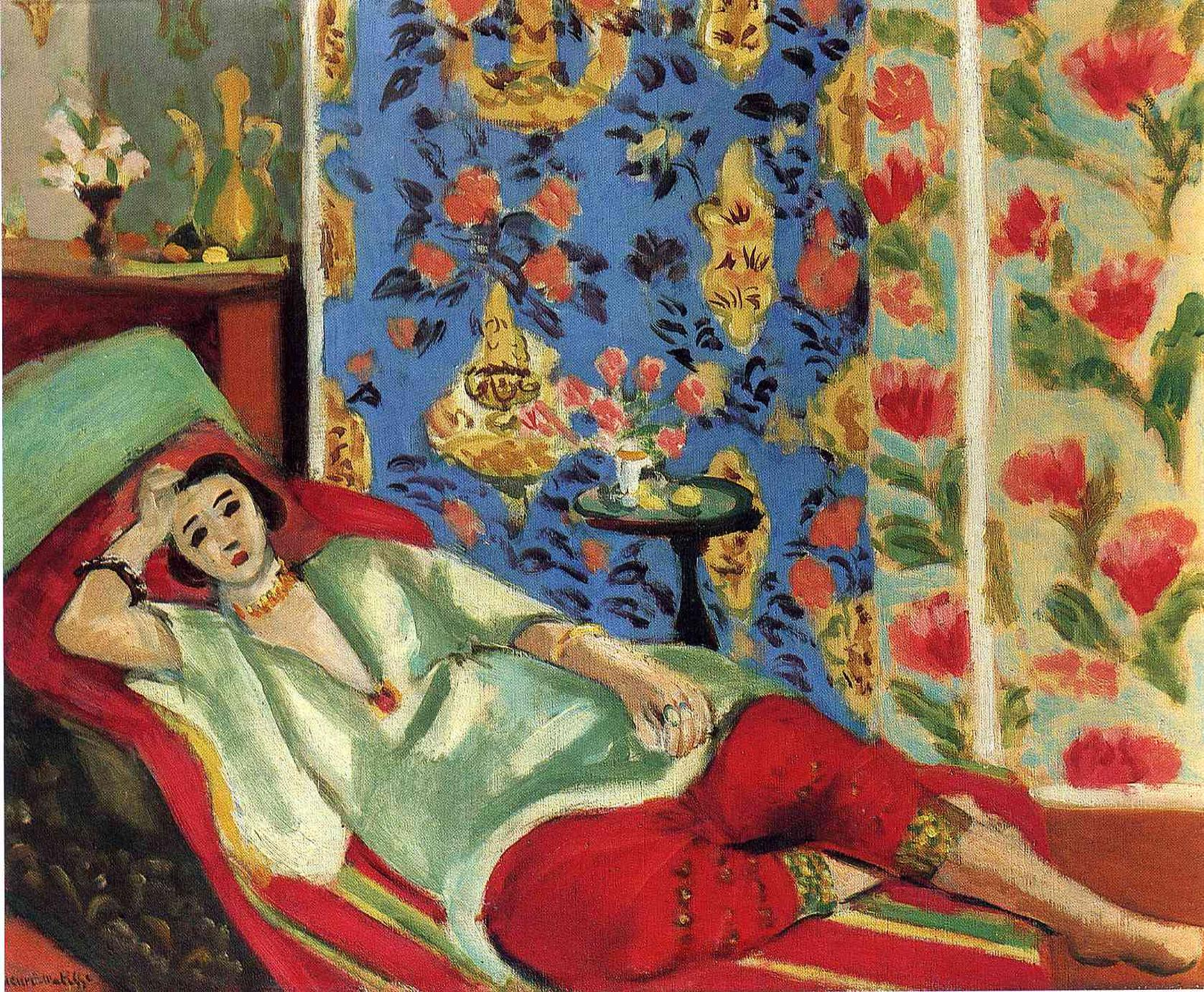} 
        \caption{Input}
    \end{subfigure}
    \hfill
    \begin{subfigure}[b]{0.235\linewidth}
        \centering
        \includegraphics[width=\linewidth]{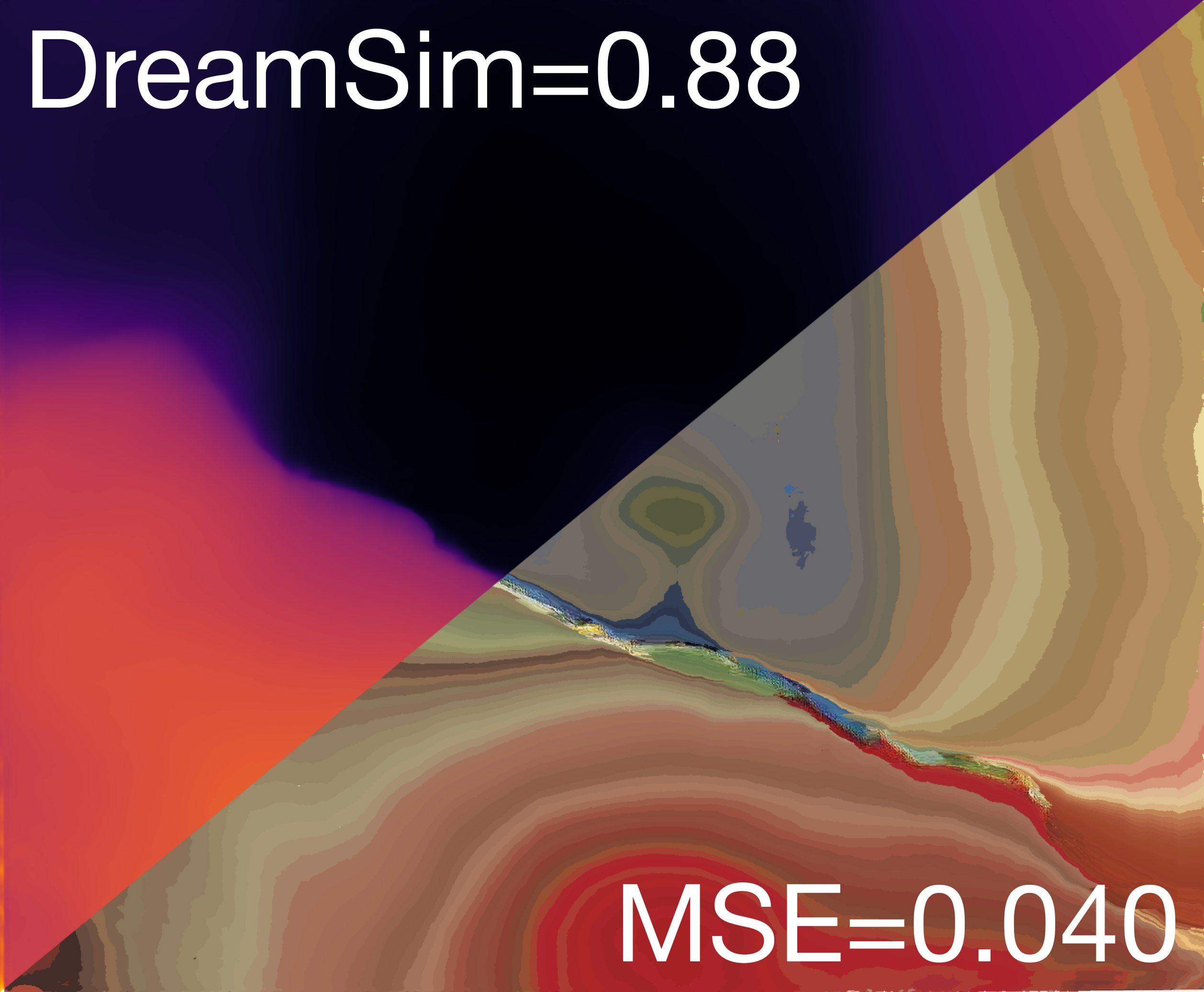}
        \caption{Depth Pro}
    \end{subfigure}
    \hfill
     \begin{subfigure}[b]{0.235\linewidth}
        \centering
        \includegraphics[width=\linewidth]{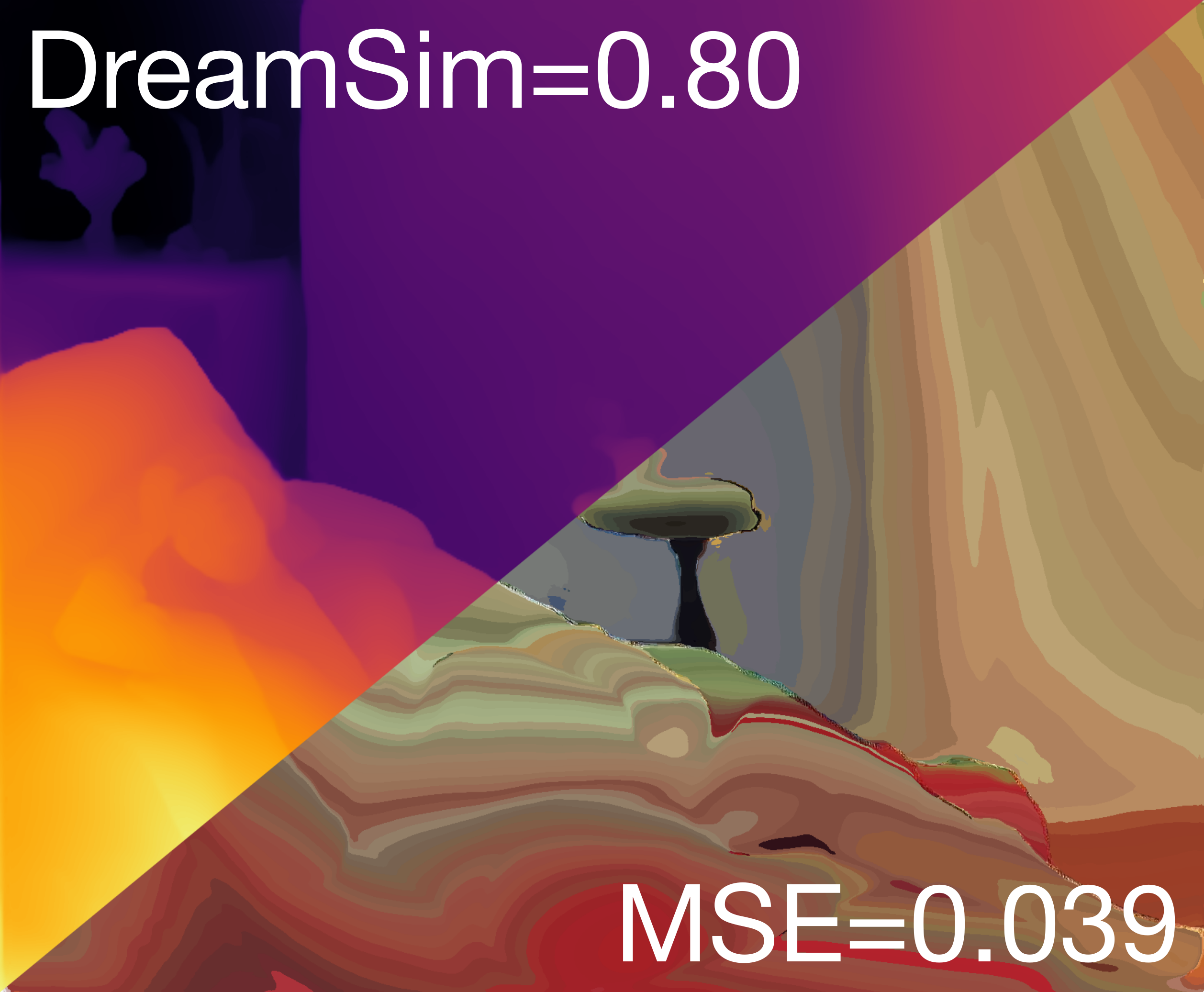}
        \caption{Depth A.-v2}
    \end{subfigure}
    \hfill
     \begin{subfigure}[b]{0.235\linewidth}
        \centering
        \includegraphics[width=\linewidth]{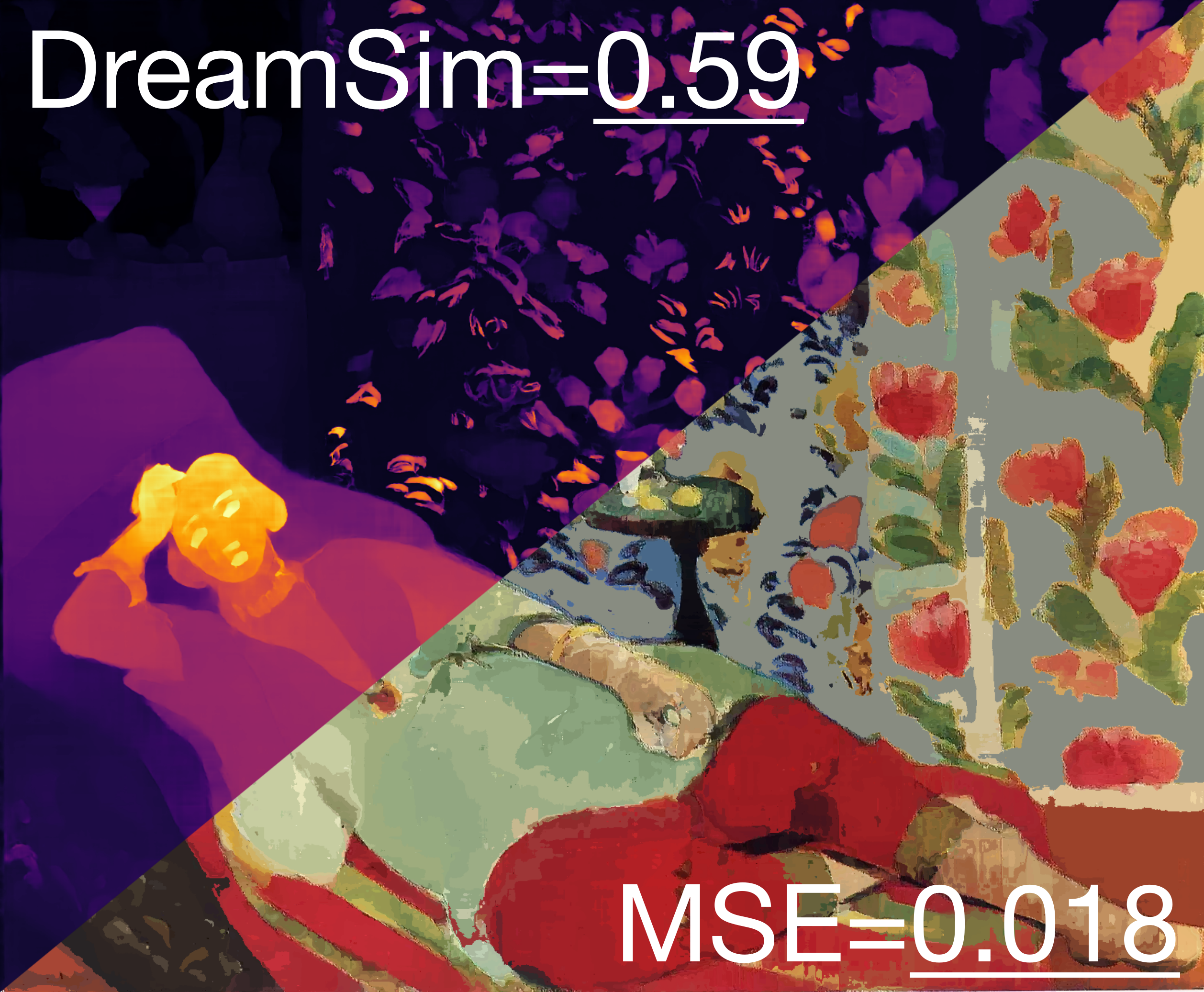}
        \caption{Ours}
    \end{subfigure}
    \caption{\textbf{Evaluation on Paintings.} Compared to traditional Monocular Depth Estimation models, our Illustrator's Depth model better aligns with color-constant regions typical of illustrations.}
    \label{fig:quantpaint}
\end{figure}

Quantitative evaluation of Illustrator’s Depth in the \emph{absence of ground-truth} (e.g., for bas-relief generation) is challenging. To further quantify how our method generalizes to real paintings, we propose here to (1) compare the predicted depth maps with the input using the human-perception-aligned CLIP Image Score~\cite{hessel2021clipscore} and DreamSim~\cite{fu2023dreamsim}, and (2) evaluate piecewise-constant RGB reconstructions after binning into $100$ layers with standard metrics. While these tests cannot possibly assess layer index fidelity, they demonstrate that solely training on SVGs improves perceptual scores for artwork images compared to our initialization Depth Pro (see \cref{table:quantpaint}), which can also be confirmed visually by comparing the different piecewise constant reconstructions on a painting as shown in \cref{fig:quantpaint}.

\begin{table}[htpb]
    \centering
    \caption{\textbf{Quantitative Evaluation on 1,000 Random Artworks from WikiArt.}}
    \label{table:quantpaint}
    \resizebox{1\linewidth}{!}{
\begin{tabular}{lccc||cc}
    \toprule
      WikiArt1000 &  MAE ($10^{-2}$)$\downarrow$ &MSE ($10^{-2}$)$\downarrow$  & SSIM$\uparrow$ & CLIP-I$\uparrow$ & DreamSim$\downarrow$\\
        \midrule
Depth Pro~[\textcolor{cvprblue}{2}] & 9.716 & 2.078 & 0.387 & 58.7 & 0.771\\
Depth A.-v2~[\textcolor{cvprblue}{53}]  & 8.238 & 1.574 & 0.427 & 64.1 & 0.650 \\
Ours &  \textbf{4.672} & \textbf{0.629} & \textbf{0.640}  &  \textbf{72.3} &  \textbf{0.507} \\
 \bottomrule 
    \end{tabular}}
\end{table}

\noindent 

\section{Nearest Neighbors}
\label{sec:sup_nn}

\begin{figure}[h] \vspace*{-2.5mm}
    \centering
    \captionsetup[subfigure]{justification=centering, singlelinecheck=false,aboveskip=1pt}
    \begin{subfigure}[b]{0.155\linewidth}
        \centering
        \includegraphics[width=\linewidth]{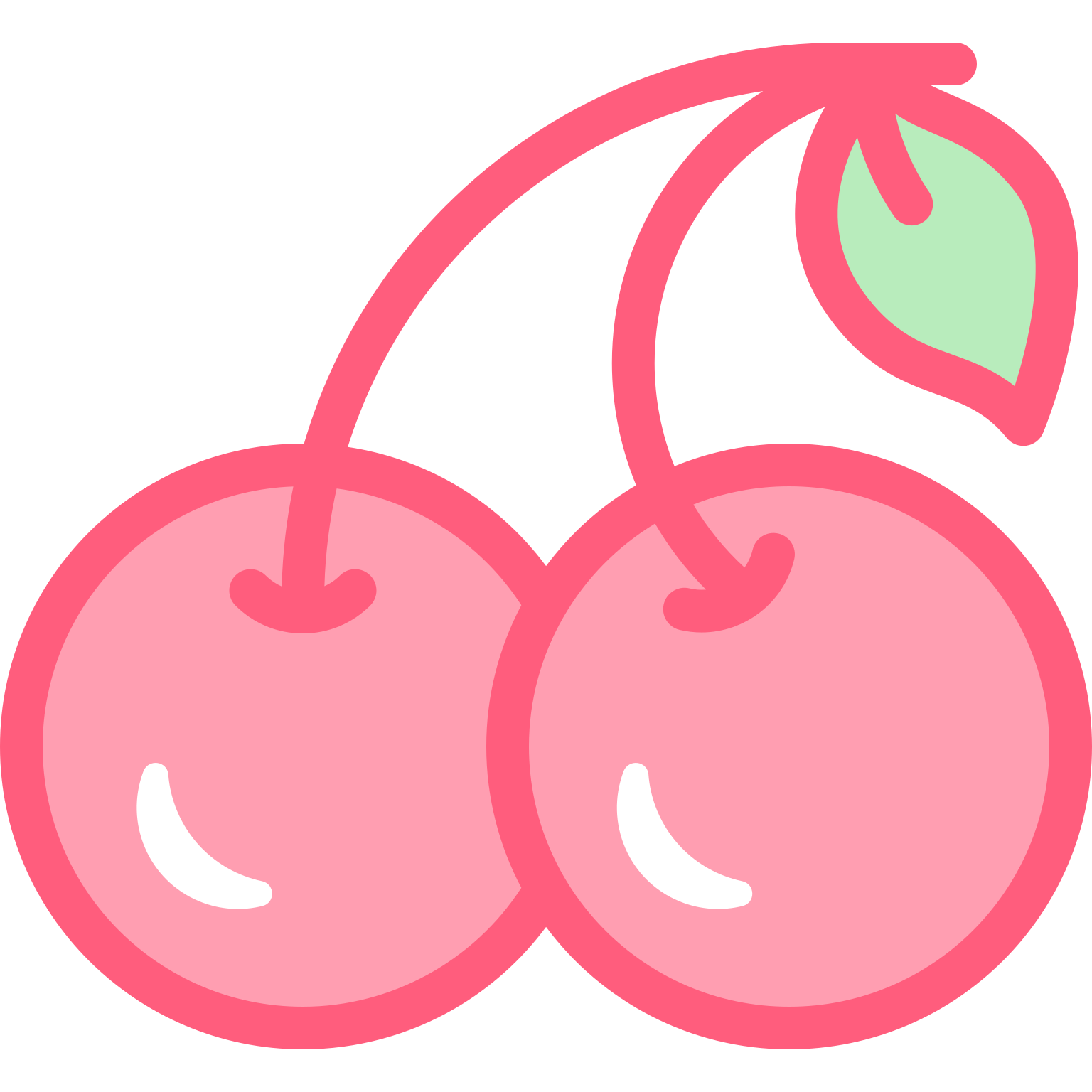}
        \caption{Test Image}
    \end{subfigure}
    \hfill
        \begin{subfigure}[b]{0.155\linewidth}
        \centering
        \includegraphics[width=\linewidth]{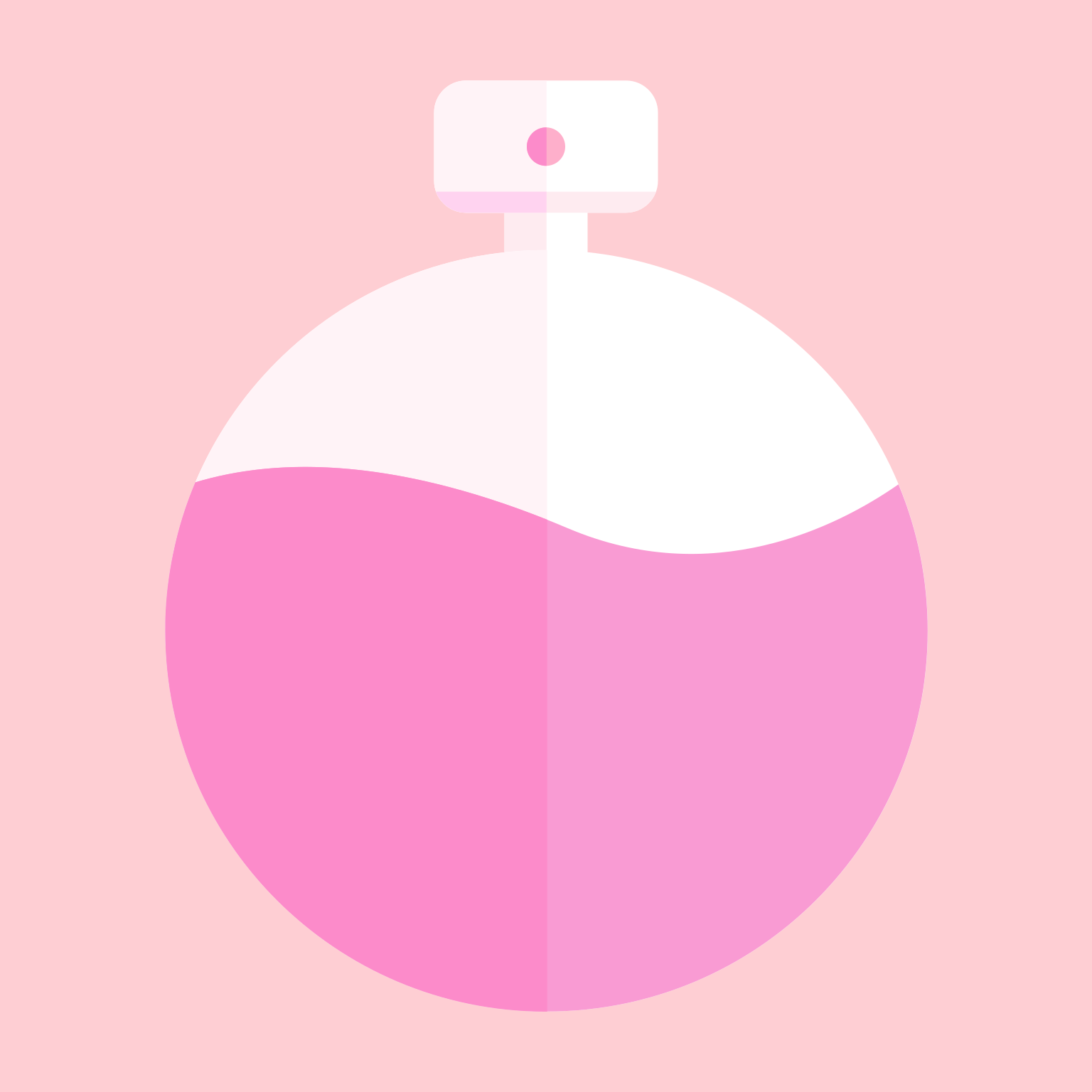}
         \caption{NN (RGB)}
    \end{subfigure}
    \hfill
    \begin{subfigure}[b]{0.155\linewidth}
        \centering
        \includegraphics[width=\linewidth]{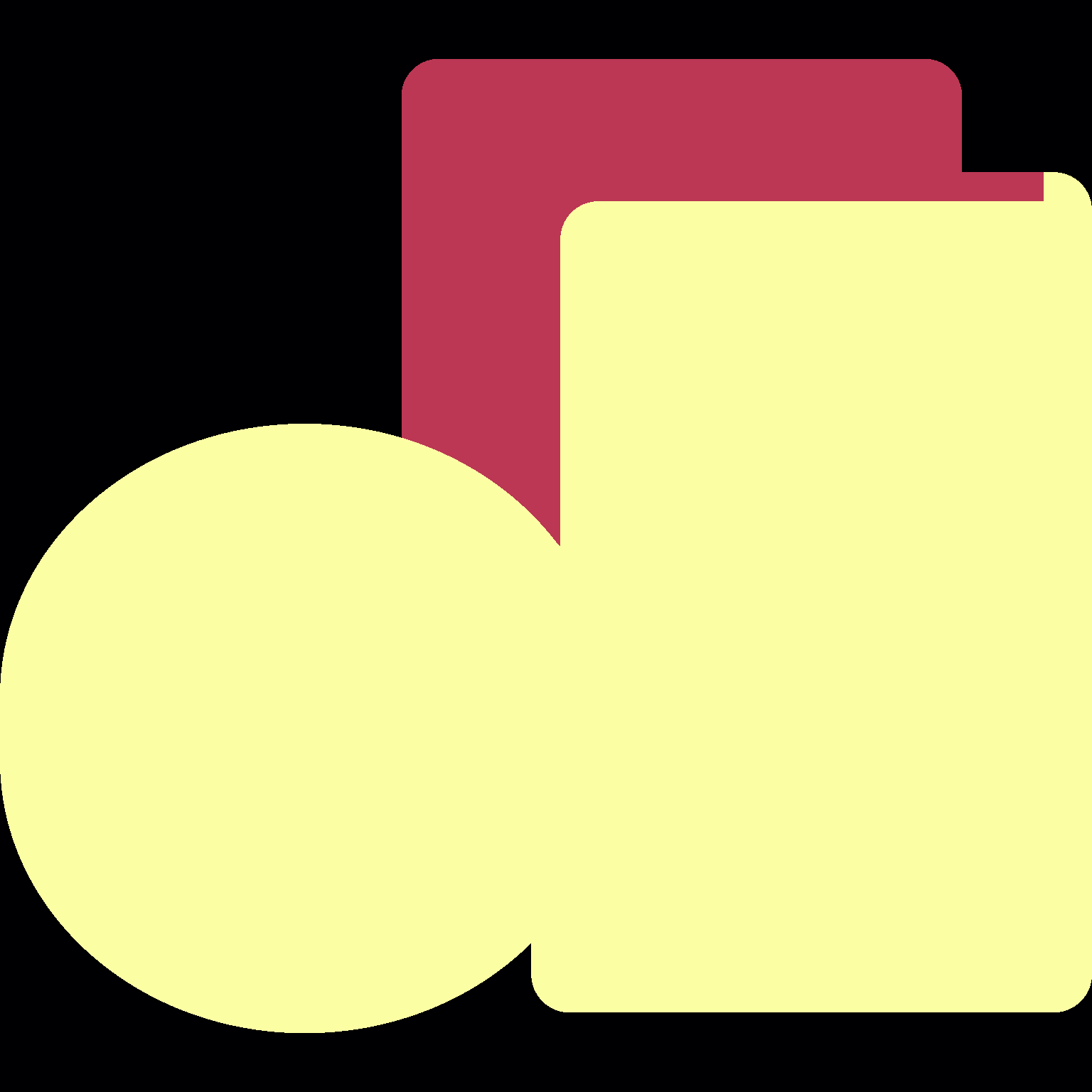}
        \caption{NN (Depth)} 
    \end{subfigure}
    \hfill
    \begin{subfigure}[b]{0.155\linewidth}
        \centering
        \includegraphics[width=\linewidth]{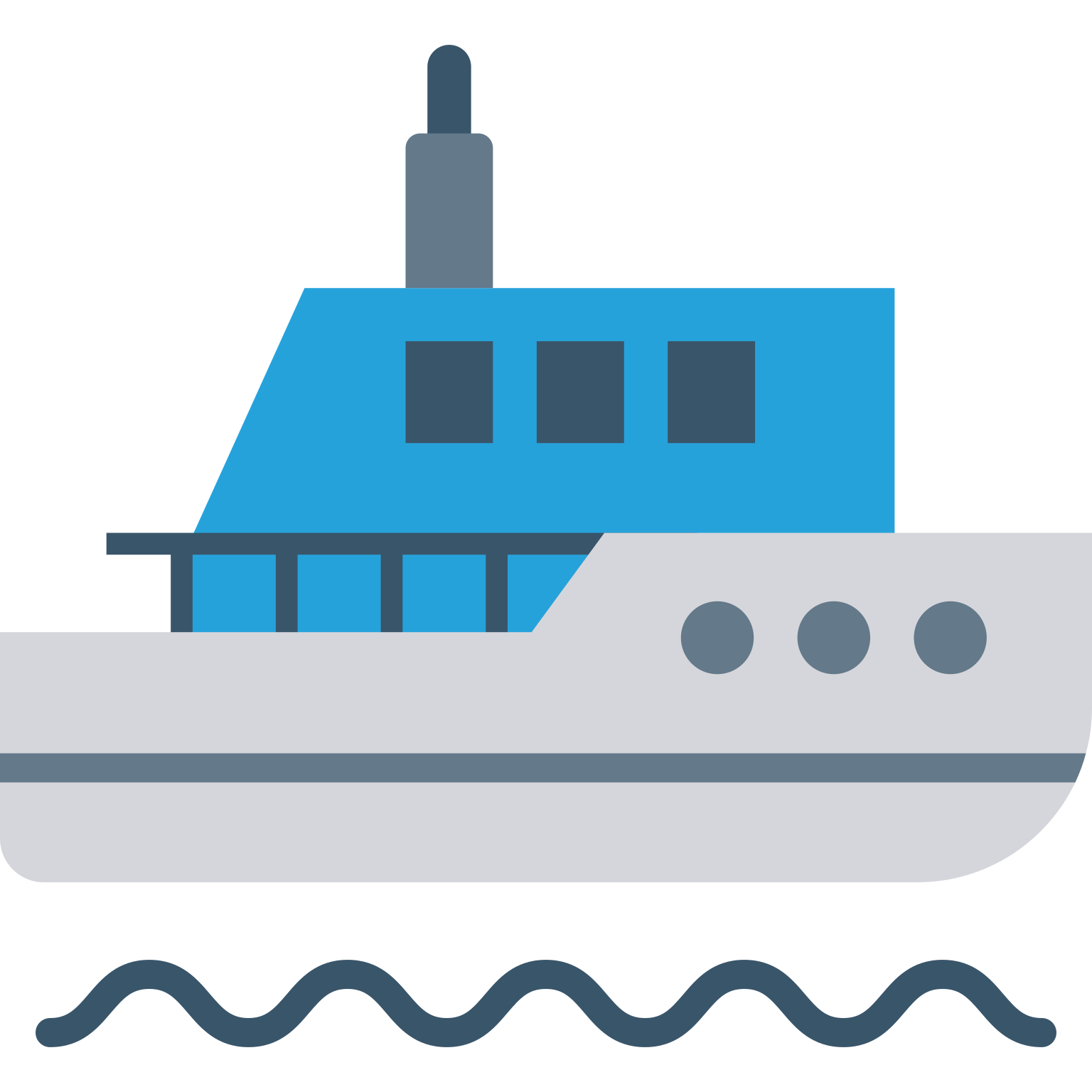}
        \caption{Test Image}
    \end{subfigure}
    \hfill
        \begin{subfigure}[b]{0.155\linewidth}
        \centering
        \includegraphics[width=\linewidth]{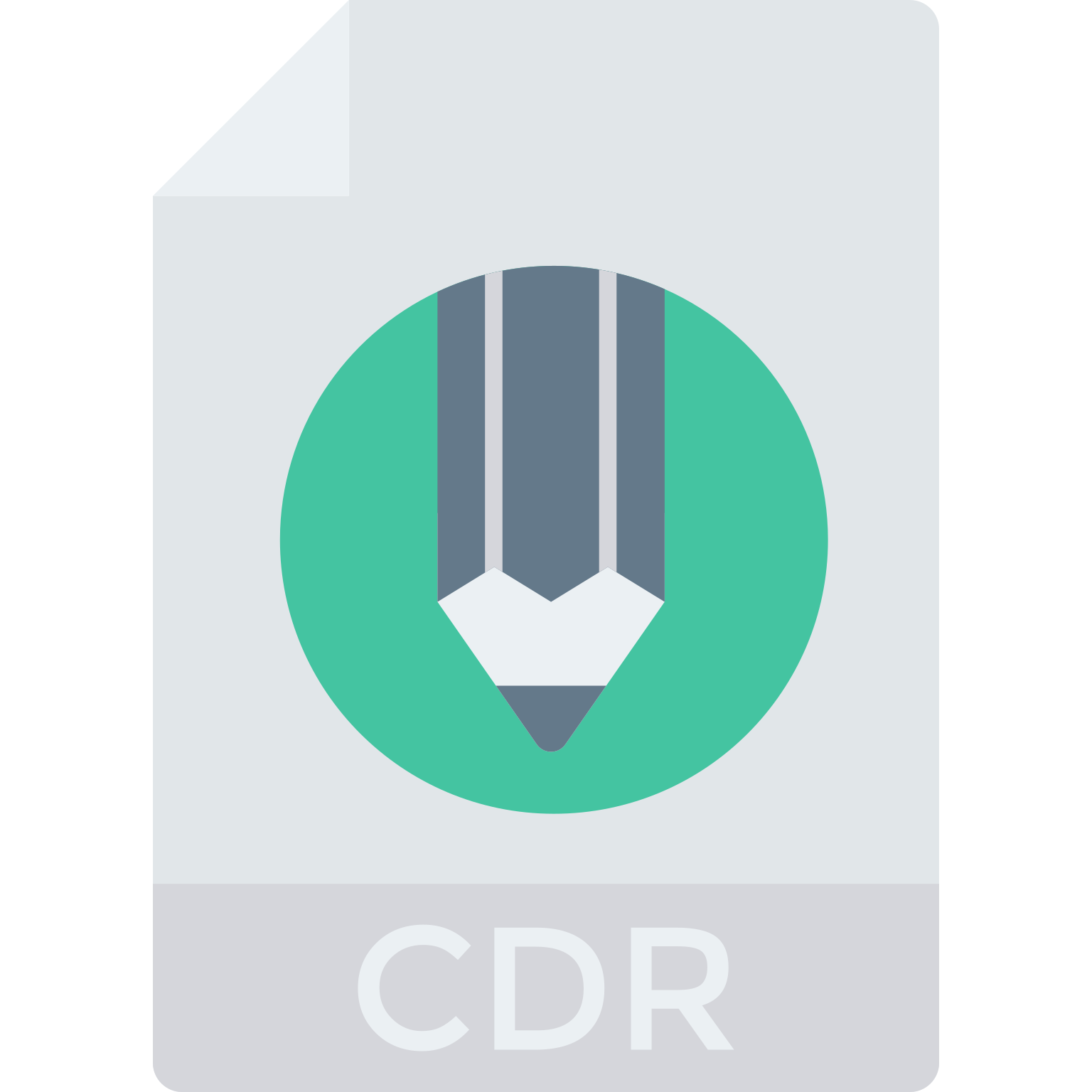}
         \caption{NN (RGB)}
    \end{subfigure}
    \hfill
    \begin{subfigure}[b]{0.155\linewidth}
        \centering
        \includegraphics[width=\linewidth]{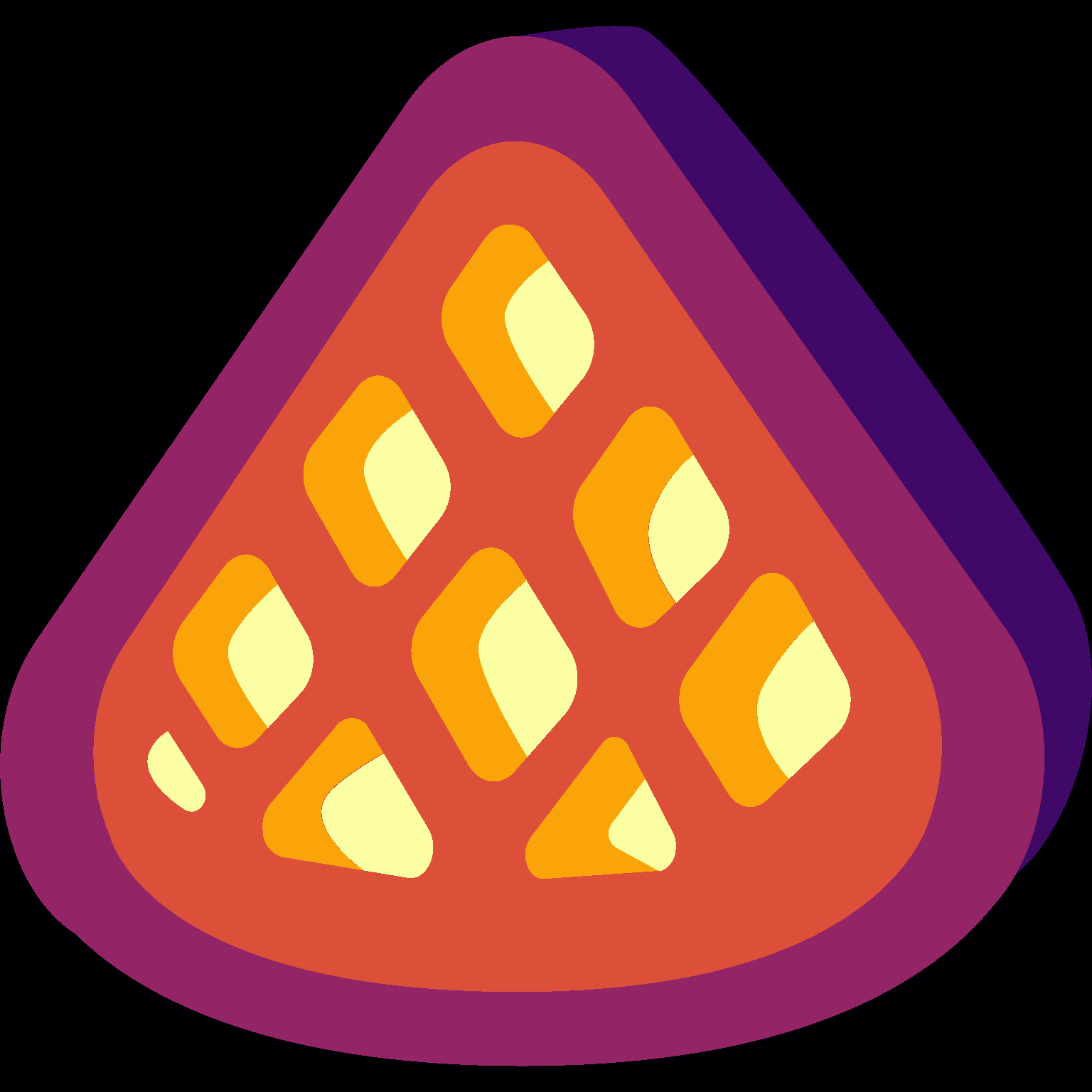}
        \caption{NN (Depth)} 
    \end{subfigure}
    \caption{\textbf{Nearest Neighbor.} For two test images (a) and (d), we show their nearest neighbors in the training set based on pixel-wise RGB and Illustrator’s Depth. This confirms their novelty for the network.}
    \label{fig:sup_nn}
\end{figure}

While our test set is relatively small, nearest neighbor analysis proves that our input images from the test set do not appear in the training set, as demonstrated in \cref{fig:sup_nn}.

\section{Additional results}

\begin{figure}[!h]
    \centering
    \includegraphics[width=1.0\linewidth]{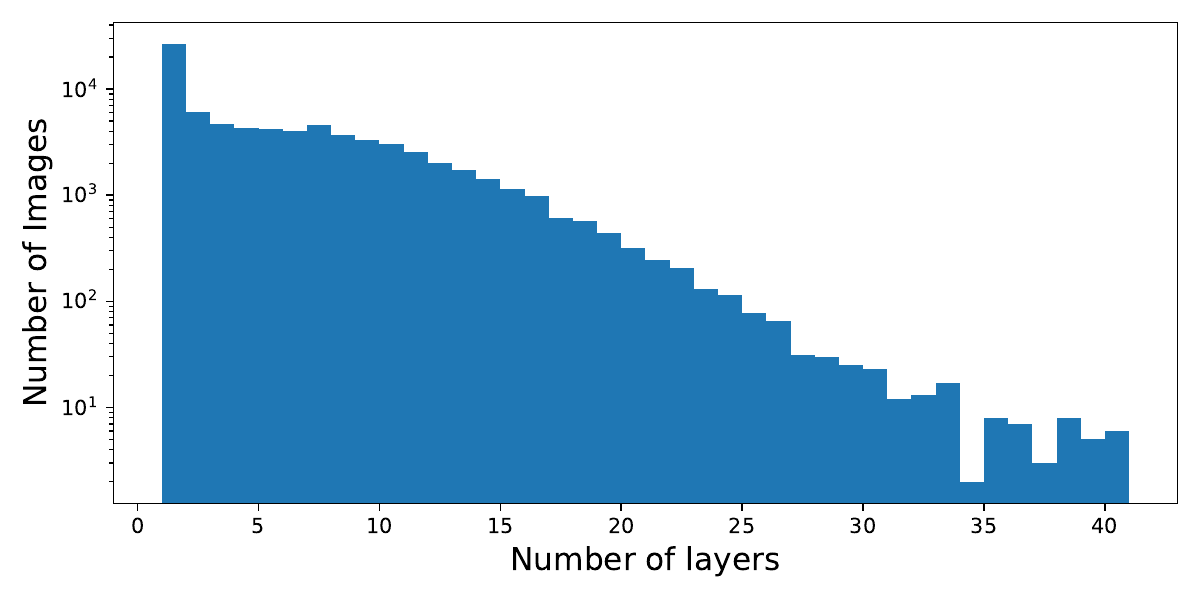}
    \vspace*{-7mm}
    \caption{\textbf{Number of layers in MMSVG training set.} We plot the histogram of the number of layers in our training dataset. Note that each layer may have many connected components, resulting in a large number of paths.}
    \label{fig:suppl_layers}
\end{figure}

\begin{figure}[h]
    \centering

    \includegraphics[width=\linewidth]{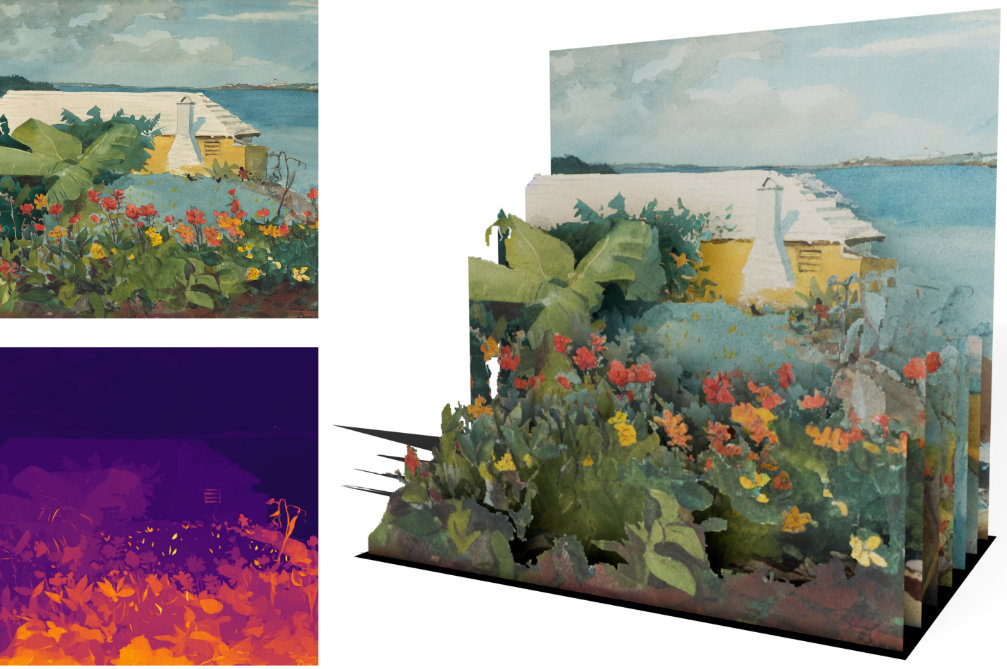}
    \caption{\textbf{Automatic pop-up card generation.} From an image (top left) and our predicted illustrator's depth (bottom left), a multi-layered pop-up card can easily be created using our method --- see video for animation.}
    \label{fig:supp_cutout}
\end{figure}

For completeness, we also provide a histogram of the number of layers present in our curated training dataset in~\cref{fig:suppl_layers}, as well as a figure demonstrating another potential use of our illustrator's depth 
in~\cref{fig:supp_cutout}, where a painting is automatically turned into a multi-layered pop-up card. 

\end{document}